%% file: main.tex
\documentclass[sigconf]{aamas} 

\usepackage{balance} % for balancing columns on the final page
\usepackage{amsmath}
\usepackage{dsfont}
\usepackage{graphicx}
\usepackage{float}
\usepackage{caption}
\usepackage{subcaption}
\usepackage{cleveref}
\usepackage{sidecap}
\sidecaptionvpos{figure}{t}
\usepackage[ruled,longend]{algorithm2e}
\setcopyright{ifaamas}
\acmConference[AAMAS '25]{Full version of the paper appearing in Proc.\@ of the 24th International Conference
on Autonomous Agents and Multiagent Systems (AAMAS 2025)}{May 19 -- 23, 2025}
{Detroit, Michigan, USA}{Y.~Vorobeychik, S.~Das, A.~Nowe (eds.)}
\copyrightyear{2025}
\acmYear{2025}
\acmDOI{}
\acmPrice{}
\acmISBN{}

\acmSubmissionID{667}

\title[AAMAS-2025 Formatting Instructions]{Human-Aligned Skill Discovery: Balancing Behaviour Exploration and Alignment}

\author{Maxence Hussonnois}
\affiliation{
  \institution{$A^2 I^2$, Deakin University}
  \city{Geelong}
  \country{Australia}}
\email{m.hussonnois@deakin.edu.au}

\author{Thommen George Karimpanal}
\affiliation{
  \institution{School of IT, Deakin University}
  \city{Geelong}
  \country{Australia}}
\email{thommen.karimpanalgeorge@deakin.edu.au}

\author{Santu Rana}
\affiliation{
  \institution{$A^2 I^2$, Deakin University}
  \city{Geelong}
  \country{Australia}}
\email{santu.rana@deakin.edu.au}
\begin{abstract}

\input{paper/sections/abstract}
\end{abstract}


\newcommand{\BibTeX}{\rm B\kern-.05em{\sc i\kern-.025em b}\kern-.08em\TeX}

\begin{document}

%

%%% The following commands remove the headers in your paper. For final 
%%% papers, these will be inserted during the pagination process.

\pagestyle{fancy}
\fancyhead{}

%%% The next command prints the information defined in the preamble.

\maketitle 

%%%%%%%%%%%%%%%%%%%%%%%%%%%%%%%%%%%%%%%%%%%%%%%%%%%%%%%%%%%%%%%%%%%%%%%%

\section{Introduction}
\input{paper/sections/introduction}

\section{Related Work}
\input{paper/sections/related_work}

\section{Approach}
\input{paper/sections/preliminaries}
\input{paper/sections/methods}

\input{paper/experiments/sections/introduction}
\section{Discussion and Conclusion}

\input{paper/sections/conclusion}

%%%%%%%%%%%%%%%%%%%%%%%%%%%%%%%%%%%%%%%%%%%%%%%%%%%%%%%%%%%%%%%%%%%%%%%%

%%% The acknowledgments section is defined using the "acks" environment
%%% (rather than an unnumbered section). The use of this environment 
%%% ensures the proper identification of the section in the article 
%%% metadata as well as the consistent spelling of the heading.

% \begin{acks}
% If you wish to include any acknowledgments in your paper (e.g., to 
% people or funding agencies), please do so using the `\texttt{acks}' 
% environment. Note that the text of your acknowledgments will be omitted
% if you compile your document with the `\texttt{anonymous}' option.
% \end{acks}

%%%%%%%%%%%%%%%%%%%%%%%%%%%%%%%%%%%%%%%%%%%%%%%%%%%%%%%%%%%%%%%%%%%%%%%%

%%% The next two lines define, first, the bibliography style to be 
%%% applied, and, second, the bibliography file to be used.
\balance
\bibliographystyle{ACM-Reference-Format} 
\bibliography{main}

%%%%%%%%%%%%%%%%%%%%%%%%%%%%%%%%%%%%%%%%%%%%%%%%%%%%%%%%%%%%%%%%%%%%%%%%
\newpage
\onecolumn
\input{paper/appendix/sections/appendix}
\end{document}

%% file: paper/sections/abstract.tex
Unsupervised skill discovery in Reinforcement Learning aims to mimic humans’ ability to autonomously discover diverse behaviors. However, existing methods are often unconstrained, making it difficult to find useful skills, especially in complex environments, where discovered skills are frequently unsafe or impractical. We address this issue by proposing \textit{Human-aligned Skill Discovery - HaSD} 
\footnote{See code here: \href{https://github.com/HussonnoisMaxence/HaSD-AAMAS}{link to github}}, a framework that incorporates human feedback to discover safer, more aligned skills. HaSD simultaneously optimises skill diversity and alignment with human values. This approach ensures that alignment is maintained throughout the skill discovery process, eliminating the inefficiencies associated with exploring unaligned skills. We demonstrate its effectiveness in both 2D navigation and SafetyGymnasium environments, showing that HaSD discovers diverse, human-aligned skills that are safe and useful for downstream tasks. Finally, we extend HaSD by learning a range of configurable skills with varying degrees of diversity-alignment trade-offs that could be useful in practical scenarios.

%% file: paper/sections/introduction.tex
Deep reinforcement learning \citep{Mnih2015} aims to solve sequential decision-making problems by maximising pre-specified rewards over time. Despite its proven success in a number of applications ranging from Atari games to robotics \citep{Mnih2015,Lillicrap2016}, the framework is typically task-specific, resulting in agents with poor generalisation abilities.

Humans, on the other hand, can autonomously discover diverse and complex skills that can be combined later for better generalisation. In line with this objective, Unsupervised Skill Discovery (USD) methods \citep{eysenbach2018, Sharma2020Dynamics-Aware, Campos2020ExploreDA, park2021lipschitz} aim to learn a library of policies (skills) driven by an intrinsic reward. In addition to locomotion and manipulation tasks \citep{park2023controllability}, these methods have also demonstrated promising results with pixel-based agents \citep{park2023metra}.

\begin{figure}{}
    \centering
    \includegraphics[width=0.4\textwidth]{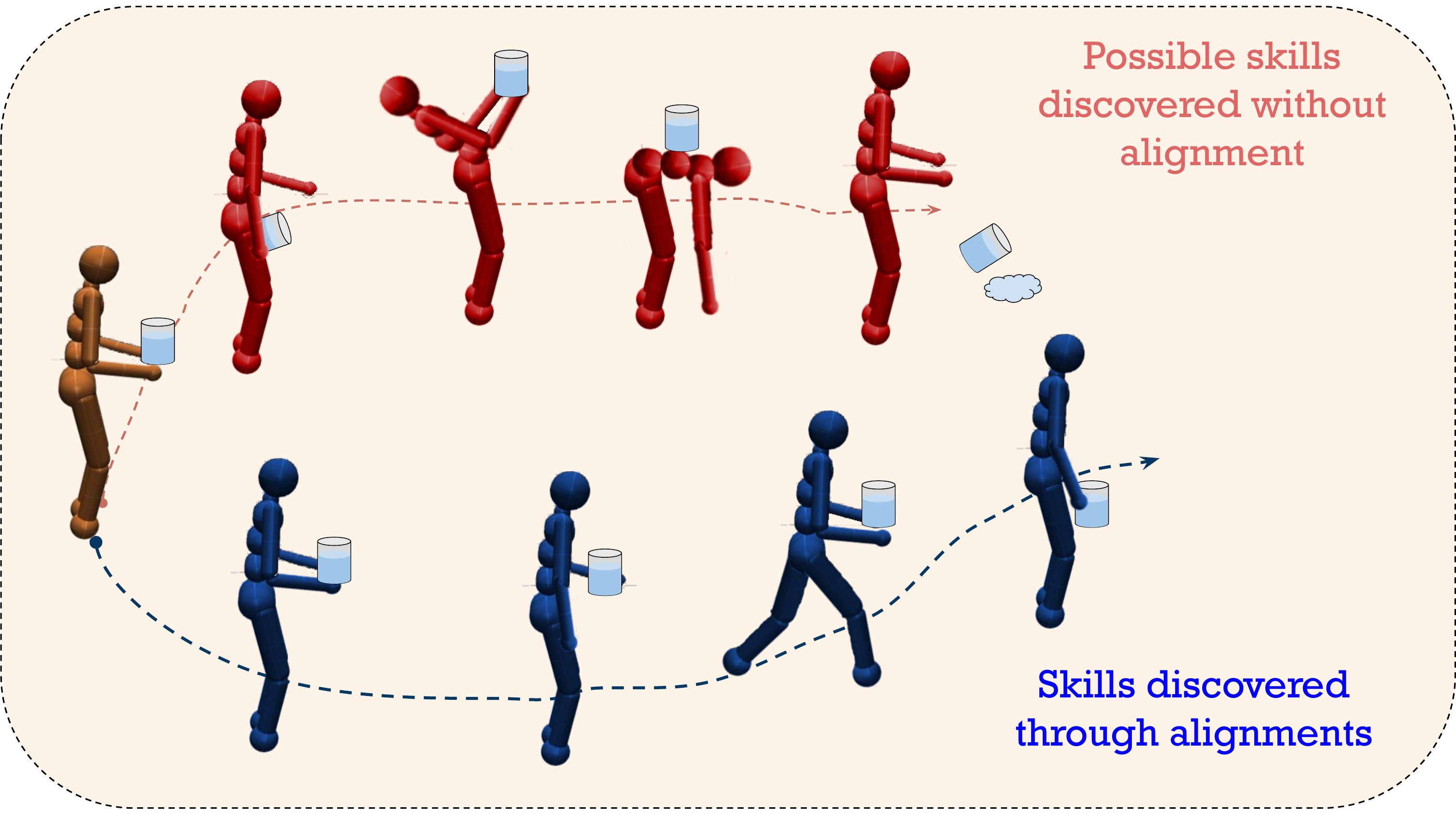} 
    \caption{Without an alignment signal, discovering desirable skills in complex environments is like searching for a needle in a haystack, often leading to skills that achieve tasks in undesirable ways, such as carrying a glass of water awkwardly (red robots). Aligning skills during discovery ensures they meet human preferences (blue robots).}
    % Without an alignment signal, discovering desirable skills in complex environments is like searching for a needle in a haystack. Agents may learn skills that achieve a task in a way undesirable to humans, for example carrying a glass of water in peculiar positions. In contrast, by aligning skills during discovery, we can ensure their desirability.}
    \label{fig: introduction|example}
    \Description{Schematic describing the HaSD framework. At the top, equations describe the passage from collecting preferences feedback to maximising a alignement reward. At the bottom, equations describes the passage from collecting data to maximising a skill discovery reward.}
\end{figure}

However, to discover dynamic behaviours these methods rely on correlating skills with changes in the environment regardless of the underlying safety or desirability of these changes. As illustrated in Figure \ref{fig: introduction|example}, the same skill, that of delivering a glass of water, can have very different degrees of desirability. Without an alignment objective, although this task may be successfully completed, say, by the humanoid carrying the glass on its back, such skills are in general, not desirable, nor expected. This suggests a need to constrain skill discovery with alignment signals from human.

Recent works \citep{klemsdal2021learning} have suggested relying on expert demonstrations to guide the agent. Such demonstrations are generally expensive and thus not available in large quantities, negatively impacting skill diversity. More generally, \citet{kim2023safeskill} introduces the problem of safety-aware skill discovery (SASD) , which aims to discover diverse skills that satisfy user-predefined safety constraints. However, these methods are limited to a user's domain knowledge. Therefore, they cannot adapt to unforeseen unsafe scenarios that the agent may discover. To enable such adaptation, we advocate for the development of an online approach, where a human oversees training while providing necessary feedback.

In this regard, Human-in-the-loop reinforcement learning \citep{zhang2021recent}, in which humans are integrated with the agents' training to provide meaningful feedback, has been able to tackle problems such as safe reinforcement learning \citep{saunders2018trial} and reward engineering \citep{Christiano2017} while being practical, scalable and efficient \citep{2021pebble}. However, its incorporation into unsupervised skill discovery methods remains mostly under-explored. In this context, \citet{hussonnois2023controlled} framed the problem of controlling skill diversity as restraining goal-based skills to a certain region of the environment deemed more desirable by human preference. However, an agent may reach a desired goal in an unaligned manner as illustrated in Figure \ref{fig: introduction|example}. This can be remedied only by aligning the entire behaviour.
To address this issue, we formulate the problem of discovering diverse skills with aligned behaviour as a multi-objective problem, where the primary components are a skill discovery objective and an alignment objective. We call this framework Human-aligned Skill Discovery (HaSD). Additionally, we propose a method to optimise this objective that maximises the combination of a skill discovery reward and an alignment reward learned with the Preference-Based RL \citep{Christiano2017} framework. As human preferences evolve in response to the discovery of more diverse skills, it enables the discovery of increasingly more complex and aligned skills. Furthermore, we propose \begin{math} \alpha \end{math}-HaSD, a more general framework that addresses the problem of balancing both rewards by conditioning the skills on the diversity-alignment trade-off variable \begin{math} \alpha \end{math}. By doing so, \begin{math}\alpha\end{math}-HaSD can produce a range of skills corresponding to varying degrees of diversity-alignment trade-offs. 
In summary, the main contributions of this work are:

\begin{itemize}

  \item \emph{Human-aligned Skill Discovery} (HaSD), a novel framework to discover diverse and aligned skills.

  \item \emph{Configurable Human-Aligned Skills} (\begin{math}\alpha\end{math}-HaSD), to discover diverse and aligned skills that are conditioned on the diversity-alignment trade-off.

  \item Qualitative and quantitative evaluation of the proposed methods, with suitable comparisons with existing baselines for learning diverse and aligned skills.

\end{itemize}

%% file: paper/sections/related_work.tex
\paragraph{\textbf{Unsupervised Skill Discovery}} In unsupervised skill discovery approaches, the goal is to explore the space of learnable behaviours to acquire a set of useful and diversified temporally extended actions or skills \citep{SUTTON1999181} without a reward function. DIAYN \citep{eysenbach2018}, VIC \citep{Gregor2016}, VALOR \citep{achiam2018} and DADS \citep{Sharma2020Dynamics-Aware} suggested to maximise mutual information between states and skills. By maximising skill distinguishability, they learned diverse skills in locomotion environments. On the other hand, these methods learn to maximise the mutual information between state and skills with only small state variations, resulting in mostly static skills. Thus \citet{park2021lipschitz} proposed to maximise state variations with novel distance-maximising skill discovery objectives that learn more dynamic skills in locomotion tasks \citep{park2021lipschitz}, manipulation tasks \citep{park2023controllability} and pixel-based environments \citep{park2023metra}. However, skill discovery methods do not take into account the underlying context of the environment. This results in an inefficient discovery process that leads to unsafe and unusable skills. Our proposed method addresses this issue by integrating Unsupervised Skill Discovery methods with alignment techniques such as preference-based RL \citep{Christiano2017}. % \textcolor{orange}{integrating USD methods with preference based RL you mean}.
In recent work, \citet{kim2023safeskill} examined safety-aware skill discovery, which focused on finding inherently safe skills. Specifically, they proposed regularising skill discovery using a safety critic \citep{srinivasan2020learning} that learns from any user-defined safety constraints. In contrast, our work focuses on discovering skills that align with human values learned during training.
%However EDL \citep{Campos2020ExploreDA} and \citep{liu2023comsd} showed that such skill discovery methods suffer from poor exploration and proposed to integrate exploration techniques such as SMM \citep{smm2019}, or particle-based state entropy estimation \citep{Liu2021BehaviorFT}.

\paragraph{\textbf{Human-in-the-Loop and Preference Based RL}} Human in the loop reinforcement learning (HIL-RL) methods focus on learning through feedback from humans during training. In this regard, Preference-based RL\citep{Christiano2017} uses human preferences over agents' trajectories to infer a reward model and train an agent with it. Since human preferences are expensive, PEBBLE \citep{2021pebble} aimed to mitigate this via improved sample and feedback efficiency by leveraging exploration methods and off-policy learning. SURF \citep{park2021surf} and REED \citep{metcalf2022rewards} further improved feedback efficiency by using supervised learning techniques and self-supervised representation learning. We follow this line of work to align skills with human preference. However, the settings differ in that the agent also optimises a skill discovery reward, which reduces the need for pre-training phases. 

\paragraph{\textbf{Unsupervised Skills Discovery with Preferences}} Unsupervised Skill Discovery with Preferences aims to learn more desirable skills which is still an under-explored area of research. In this regard, Skill Preferences (SkiP) \citep{wang2022skill} used preferences over an offline dataset to extract relevant human-aligned skills. Alternatively, CDP \citep{hussonnois2023controlled} uses preferences in an online setting to first identify rewarding regions of the state space and then learn goal-based skills within those regions. However aligning goals of goal-based skills do not ensure that the entire trajectory of the skills is aligned. To overcome this limitation, we align the entire skill's trajectory by optimising both skill discovery and alignment (via preferences) objectives.

\paragraph{\textbf{Quality-Diversity Policy Optimisation}} Quality-Diversity Policy Optimisation \citep{kumar2020one, chen2024dgpo, zahavydiscovering} focuses on discovering multiple strategies for solving a given task. Generally, these methods are based on a sophisticated combination of diversity reward and task reward from the environment. In SMERL \citep{kumar2020one} the optimal policies are diversified by adding the DIAYN diversity reward to transitions along trajectories that produce a known optimal return. Alternatively, DGPO \citep{chen2024dgpo} and DOMINO \citep{zahavydiscovering}  achieve better results by alternately constraining the diversity of the strategies while simultaneously constraining the extrinsic reward. Similar to our work, these works combine an extrinsic reward with a diversity reward. However, their motivation and approach differ fundamentally from ours, as they first seek to optimise performance over task reward, then diversify those performances, whereas we first diversify policies by discovering skills, then adjust them to fit desirability through human preference. Our approach is set up as such, as unlike the former methods, we do not assume direct access to the task rewards. We thus tackle a more challenging setting where an agent must simultaneously infer this knowledge from human feedback.%This approach is only possible because of direct access to task rewards. In contrast, in our setting, we do not assume this information is available and must collect it from humans. Thus, we first diversify policies by discovering skills, then adjust them to fit desirability through human preference.

%% file: paper/sections/preliminaries.tex
\begin{figure*}[htbp]
  \centering
  \includegraphics[width=1.0\linewidth]{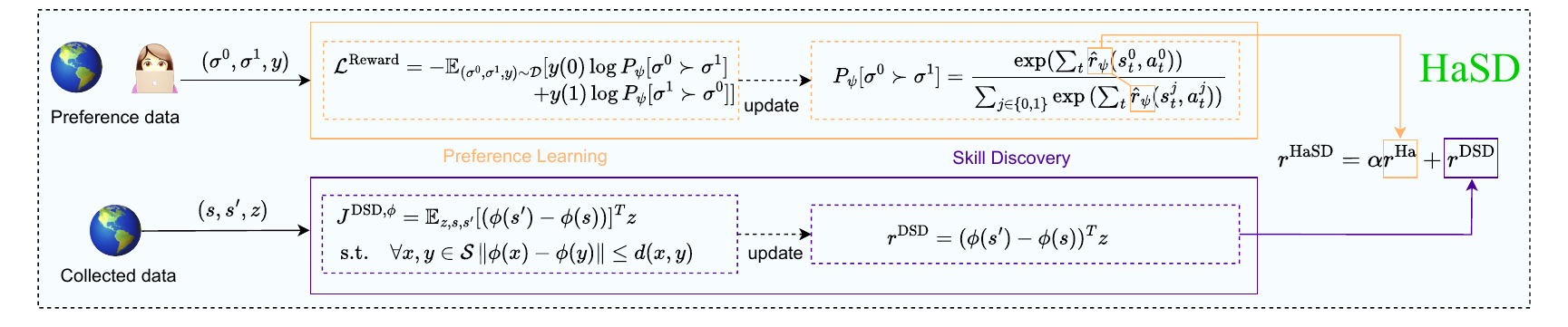} 
  \caption{Illustration of the HaSD reward components. Skill Discovery rewards are computed using the Distance-Maximising Skill Discovery (DSD) objective and data collected from interaction with the environment. The reward encourages skills to be more dynamic and diverse. Then, we add a \begin{math} r_{Ha} \end{math} human-aligned reward learned with preference learning through data collected from interaction with the environment and human preferences. This reward encourages skills to align with human preferences.}
  \label{fig:diagramme}
  \Description{Schematic describing the HaSD framework. At the top, equations describe the passage from collecting preferences feedback to maximising an alignment reward. At the bottom, equations describe the passage from collecting data to maximising a skill discovery reward.}
\end{figure*}
To discover diverse and complex skills that are more aligned with human values, agents must be able to adjust the skill discovery process based on human feedback. To this end, we consider the problem of discovering diverse skills that satisfy humans' preferences. We first detail the problem settings and the resulting Human-aligned Skill Discovery (HaSD) Objective. We then present the details of the optimisation in Section \ref{sec: hasd}. Furthermore, in Section \ref{sec: chasd} we propose an extension of HaSD by learning configurable skills that can produce a range of diversity-alignment trade-offs.

\subsection{Problem Settings}
We consider a Markov Decision Process (MDP) without a reward function, defined as \begin{math}\mathcal M = (\mathcal S, \mathcal A, \mathcal P) \end{math}, where \begin{math}\mathcal S\end{math} and \begin{math}\mathcal A\end{math} respectively denote the state and action spaces, and \begin{math}\mathcal P\end{math} is the transition function governing the agent-environment interaction dynamics. 
Consistent with prior work \citep{Campos2020ExploreDA,park2023controllability}, we also consider Skills as policies \begin{math}\pi(a|s,z)\end{math} conditioned on latent vector \begin{math}z \in Z \end{math}.
Finally, we assume that there exists a human in the loop who has an opinion on the desirability of the agent's behaviour. The human communicates this opinion during training to the agent using a set of preferences \begin{math}\zeta\end{math}.
We aim to learn a set of diverse, useful behaviours \begin{math}\pi(a|s,z)\end{math} that align with human preferences $\zeta$. 

\subsection{Human-aligned Skill Discovery Objective}
First, for \begin{math}\pi(a|s,z)\end{math} to learn diverse skills, we follow previous skill discovery approaches that rely on maximising the mutual information between \begin{math}\mathcal S\end{math} and \begin{math}Z\end{math}, denoted as  \begin{math}I(\mathcal S;Z)\end{math} Then, to align skills with human values, we also want to maximise the probability \begin{math}Pr\end{math} of realising trajectories with the policies \begin{math}\pi(a|s,z)\end{math}  that satisfy the set of preferences \begin{math}\zeta\end{math}. Thus we propose the following novel multi-objective problem:
\begin{eqnarray}
\label{eq: hasd_objective}
\max_\pi (I(\mathcal S;Z), Pr(\zeta) )),
\end{eqnarray}
where the resulting policies will learn to discover diverse skills with a high chance of satisfying human preferences. In the following section, we describe how each objective will be addressed in our method.

\subsection{Distance-Maximising Skill Discovery}
\label{sec:DSD}
To maximise \begin{math} I(\mathcal S;Z)\end{math} with  \begin{math}\pi(a|s,z)\end{math}, we use the Distance-Maximising Skill Discovery (DSD) objectives proposed by \citet{park2021lipschitz, park2023controllability, park2023metra}. Although our approach is not limited to a specific skill discovery method, we chose the DSD objective since it achieves state-of-the-art performance in discovering dynamic and diverse skills. Specifically, instead of maximising \begin{math} I(\mathcal S;Z)\end{math}, DSD maximizes \begin{math} I_w(\mathcal S;Z)\end{math}, which measures the Wasserstein dependency between states and skills \citet{NEURIPS2019_f9209b78}. DSD objectives are based on two elements, the state-representation function \begin{math} \phi :S\rightarrow Z\end{math} as well as a non-negative, arbitrary distance function \begin{math} d \end{math} that can be learned or specified (such as the Euclidean distance as in LSD \citep{park2021lipschitz}). The function \begin{math} \phi \end{math} is trained to represent the displacement in the state space under the distance function $d$ while staying aligned with the skill variable $z$. Thus, \begin{math} \phi \end{math}'s objective is the following :
\begin{equation}
\begin{aligned}
\label{eq: dsd_objective}
   J^{\text{DSD},\phi} := \mathbb{E}_{z,s,s'}[(\phi(s') -\phi(s))]^T z \quad\textrm{s.t.}  \\ 
   \quad  \forall x,y \in \mathcal S && \left\lVert \phi(x) - \phi(y) \right\rVert \leq d(x,y) 
\end{aligned}
\end{equation}
This constrained objective can be optimised with dual gradient descent \citep{boyd2004convex} as described in \citet{park2023controllability}.

We then use \begin{math} \phi \end{math} to train the skills \begin{math}\pi(a|s,z)\end{math} in order to generate trajectories with high differences in the latent state space. The skill-discovery reward per step is then given by: 
%Mainly they improve mutual information based skill discovery for learning dynamic skills, by maximising the following objectives with dual gradient descent:
\begin{eqnarray}
    \label{eq: dsd_reward}
    r^{\text{DSD}} =(\phi(s') -\phi(s))^T z.
\end{eqnarray}
This reward allows us to discover a continuous set of diverse and dynamic skills.

\subsection{Reward Learning from Preferences}
\label{sec:PBRL}
To maximise  \begin{math}Pr(\zeta)\end{math} with  \begin{math}\pi(a|s,z)\end{math}, we follow prior works in preference-based RL \citep{wilson2012,Christiano2017,2021pebble}, where a human is presented with two trajectory segments (state-action sequences)  \begin{math}\sigma^1\end{math} and \begin{math}\sigma^2\end{math}and is asked to indicate their preference  \begin{math}\zeta_i\end{math} for one over the other.  For instance, a preference for the first segment over the second is denoted as  \begin{math}\zeta_i = \sigma^1 \succ \sigma^2\end{math} and would result in the label \begin{math} y=(1,0) \end{math} and be stored in a buffer  \begin{math}\mathcal D\end{math} as  \begin{math}(\sigma^1 ,\sigma^2, y)\end{math}.
Then we model the human's internal reward function  \begin{math} \hat{r}_\psi \end{math} responsible for the indicated preferences via the Bradley-Terry model \citep{Bradley1952RankAO} as follows:
\begin{eqnarray}
\label{eq :BradleyTerry}
P_\psi[\sigma^1 \succ \sigma^2] = \frac{\exp({\sum_t \hat{r}_\psi (s^1_t,a^1_t))}}   {\sum_{j \in \{1,2\} } \exp {(\sum_t \hat{r}_\psi (s^j_t,a^j_t))} } .
\end{eqnarray}
As in \citet{2021pebble}, we model the reward function as a neural network with parameters $ \psi $, which is updated by minimising the following loss:

\begin{eqnarray}
\label{eq: BradleyTerryLoss}
\mathcal{L}^\text{Reward} = - \mathbb{E}_{(\sigma^1, \sigma^2, y)\sim \mathcal D} [y(1) \log  P_\psi[\sigma^1 \succ \sigma^2]  \nonumber \\ 
+  y(2)\log P_\psi[\sigma^2 \succ \sigma^1]] .
\end{eqnarray}

Thus a policy \begin{math}\pi(a|s,z)\end{math} maximising the reward function \begin{math}\hat{r}_\psi\end{math} would also maximise \begin{math} Pr(\zeta)\end{math}.

%% file: paper/sections/methods.tex
\subsection{Human-Aligned Skill Discovery (HaSD)}
\label{sec: hasd}
\input{paper/figures/algo}
In this section, we present our novel Human-aligned Skill Discovery (HaSD) method that learns latent-conditioned policies \begin{math}\pi(a|s,z)\end{math} that maximise the Human-aligned Skill Discovery Objective in Equation \ref{eq: hasd_objective}. HaSD linearly combines the skill discovery reward and the preference reward as follows:
\begin{equation}
\begin{aligned}
\label{eq: hasd_reward}
   r^{\text{HaSD}} &= \alpha r^{\text{Ha}} + r^{\text{DSD}}
\end{aligned}
\end{equation}
where :
\begin{equation}
\begin{aligned}
\label{eq:r_detailed}
r^{\text{Ha}} = \hat{r}_{\psi}(s,a)  \quad\text{and}\quad   r^{\text{DSD}} =   (\phi(s') -\phi(s))^T z.
\end{aligned}
\end{equation}
\noindent \begin{math} \alpha \end{math} is a hyper-parameter that represents diversity-alignment trade-offs, \begin{math} \hat{r}_{\psi}\end{math} is the reward learned from human feedback as described in Section \ref{sec:PBRL} and \begin{math} \phi \end{math} is the state-representation function from the DSD objective presented in Section \ref{sec:DSD}. 
\paragraph{Intuition and selecting \begin{math} \alpha\end{math}.} Intuitively, Equation \ref{eq: hasd_objective} implies that agents should learn diverse and dynamic skills in the latent space while maximising some degree of alignment (\begin{math}\alpha\end{math}) with human values. In practice, \begin{math} \alpha \end{math} must not be negative, as it would lead in discovering skills contrary to human values. Generally, a higher \begin{math} \alpha \end{math} will result in more aligned skills but a less diverse skill set. A lower  \begin{math} \alpha \end{math} will have the opposite effect. In the following section, we address the issue of selecting a suitable \begin{math} \alpha \end{math} by learning multiple diversity-alignment trade-offs. 

\paragraph{Pre-training phase.}
At the start of training, state coverage or coherent behaviours 
are limited, resulting in non-informative queries \citep{2021pebble}. To mimic the pre-training phase of PEBBLE \citep{2021pebble} , we let the policy be trained only on the skill discovery reward $r^{\text{DSD}}$, which is equivalent to setting \begin{math}\alpha\end{math} dynamically as:
\begin{equation}
  \alpha =
    \begin{cases}
      0& \text{if \begin{math} t \leq \tau \end{math}}\\
      c(constant) & \text{otherwise}
    \end{cases}       
\end{equation}
 where \begin{math} \tau \end{math} is a hyper-parameter that indicates the time step from which when we start to elicit and learn from human feedback.
 
\subsection{Configurable Human-Aligned Skills ($\alpha$-HaSD)}
\label{sec: chasd}
Despite the consideration of alignment and skill discovery rewards \begin{math}r^{Ha}\end{math} and \begin{math}r^{DSD}\end{math}, we may not know what the Diversity-Alignment trade-offs across objectives should be. Performing a hyper-parameter search over \begin{math}\alpha\end{math} would require multiple trials, which would be impractical, considering the cost associated with collecting human feedback. Additionally, individual users may have different preferences and thus it may be useful to learn across the whole range of trade-offs and let an user choose a trade-off value at run-time. In this case, we can extend HaSD to learn a conditional policy on the trade-off value, thereby learning Configurable Human-aligned Skills (\begin{math}\alpha\end{math}-HaSD) \begin{math} \pi(a|s,z,\alpha) \end{math}. Then we could apply any search method over \begin{math}\alpha\end{math} to the trained conditional policy without the need for additional human feedback. We train this policy by augmenting states with a variety of trade-offs, corresponding to a range of \begin{math}\alpha\end{math} values and then optimising the \begin{math}\alpha\end{math}-HaSD objective. Our overall method is described in Algorithm \ref{alg: HaSD}.

%% file: paper/figures/algo.tex
\begin{algorithm}
\textbf{Initialise} \begin{math}\mathcal B\end{math} , \begin{math}\pi_z, r_\psi, \phi\end{math} and \begin{math}\mathrm{A}\end{math} \;
\For{each epoch}{
    // Collect data \;
    \For{each episode }{
        Sample \begin{math}z \sim p(z)\end{math} and  \begin{math}\alpha \sim \mathrm{A}\end{math}  \;
        Sample trajectory \begin{math}\tau\end{math} with \begin{math}\pi_\theta(a_t|s_t,z,\alpha)\end{math}\;
        Store trajectory \begin{math}\tau\end{math} in \begin{math} \mathcal{B}\end{math}
    }
    \If { it's time to update the preference} {   
        \For{ each query to instructor}{
            Sample \begin{math}(\sigma^0,\sigma^1) \sim \mathcal B\end{math} \;
            Collect preference from instructor \begin{math}y=\sigma^0 \succ \sigma^1\end{math} \;
            Store transitions \begin{math}\mathcal D \leftarrow \mathcal{D}\cup \{(\sigma^0,\sigma^1, y)\}\end{math}
                }  
        Update \begin{math}\hat{r}_\psi\end{math} with gradient descent on  \begin{math}\mathcal{L}^{\text{Reward}}\end{math} \eqref{eq: BradleyTerryLoss}\;    
        }
    Update \begin{math}\phi\end{math} with gradient ascent on $J^{\text{DSD}}$\;   
    Update \begin{math}\pi_\theta(a|s,z,\alpha)\end{math} using SAC \cite{pmlr-v80-haarnoja18b} and TQC \cite{kuznetsov2020controlling} with reward \begin{math}r^{\text{HaSD}}\end{math}\;    
}
\caption{Human-aligned Skill Discovery (\begin{math}\alpha\end{math}-HaSD)}
\label{alg: HaSD}
\end{algorithm}

%% file: paper/experiments/sections/introduction.tex
\section{Experiments}
\label{sec: experiments}
\input{paper/experiments/sections/nav2d/qualitative_results_nav2d}
In this section, we demonstrate that HaSD discovers diverse skills that align with human preferences and that $\alpha$-HaSD learns a wide range of diversity-alignment trade-offs. To this end, we first show how our method works in simple 2D navigation with safety costs in Section \ref{subsec: exp|nav2d|hasd}. Then we demonstrate in Safety-Gymnasium \citep{ji2023safety} the scalability of our methods across a variety of robots and environments in higher dimensions. We consider environments with different types of alignment objectives in Sections \ref{subsec: Experiment|HR1} and \ref{subsec: Experiment|HR3}. 

In one set of experiments, the alignment objective is safety, and in another set of experiments, the alignment objective involves specific interaction with an object. We note that as such, there are no inherent prerequisites for an alignment objective-just that a human should be able to indicate their preference in terms of this objective, given a pair of trajectories. For each experiment, the alignment objective is described as a sentence that could be given to a human annotator. Following previous work on preference-based RL \citep{2021pebble}, we simulate human preferences with a ground truth reward function which is assumed to reflect human annotator preferences. Unless explicitly states, we did not limit human feedback budget to ensure alignment. However, we examine the sensitivity of our method to the number of feedback samples in Section \ref{subsex: experiment|budget|nav2d}, and later, the sensitivity of our method to real human feedback in Section  \ref{subsec: experiment|Human|nav2d}. We include in Appendix \ref{subsec: apdx|conflicting_objectives} and \ref{subsec: apdx|a_generalisation} additional results showing that our methods can handle conflicting objectives and how \begin{math}\alpha\end{math}-HaSD generalises to unseen \begin{math}\alpha\end{math} values. Finally, we provide implementation details in Appendix \ref{subsec: apdx|implementation_details}.
%%% Experiments Structures
%4.1 Baselines
\input{paper/experiments/sections/baselines}
\input{paper/experiments/sections/nav2d/a_hasd_Q}
\input{paper/experiments/sections/environments}
\input{paper/experiments/sections/nav2d/QandQ}
\input{paper/experiments/sections/nav2d/feedback_sensitivity}

\input{paper/experiments/sections/nav2d/downstreamTask}
\input{paper/experiments/sections/safety_gymnasium/hazard}

\input{paper/experiments/sections/safety_gymnasium/push}
\input{paper/experiments/sections/Human_Feedback/hyp_sec}

%% file: paper/experiments/sections/nav2d/qualitative_results_nav2d.tex
\begin{figure*}[htbp]
    \centering
    \begin{subfigure}[b]{0.15\textwidth}
        \centering
        \includegraphics[width=\textwidth]{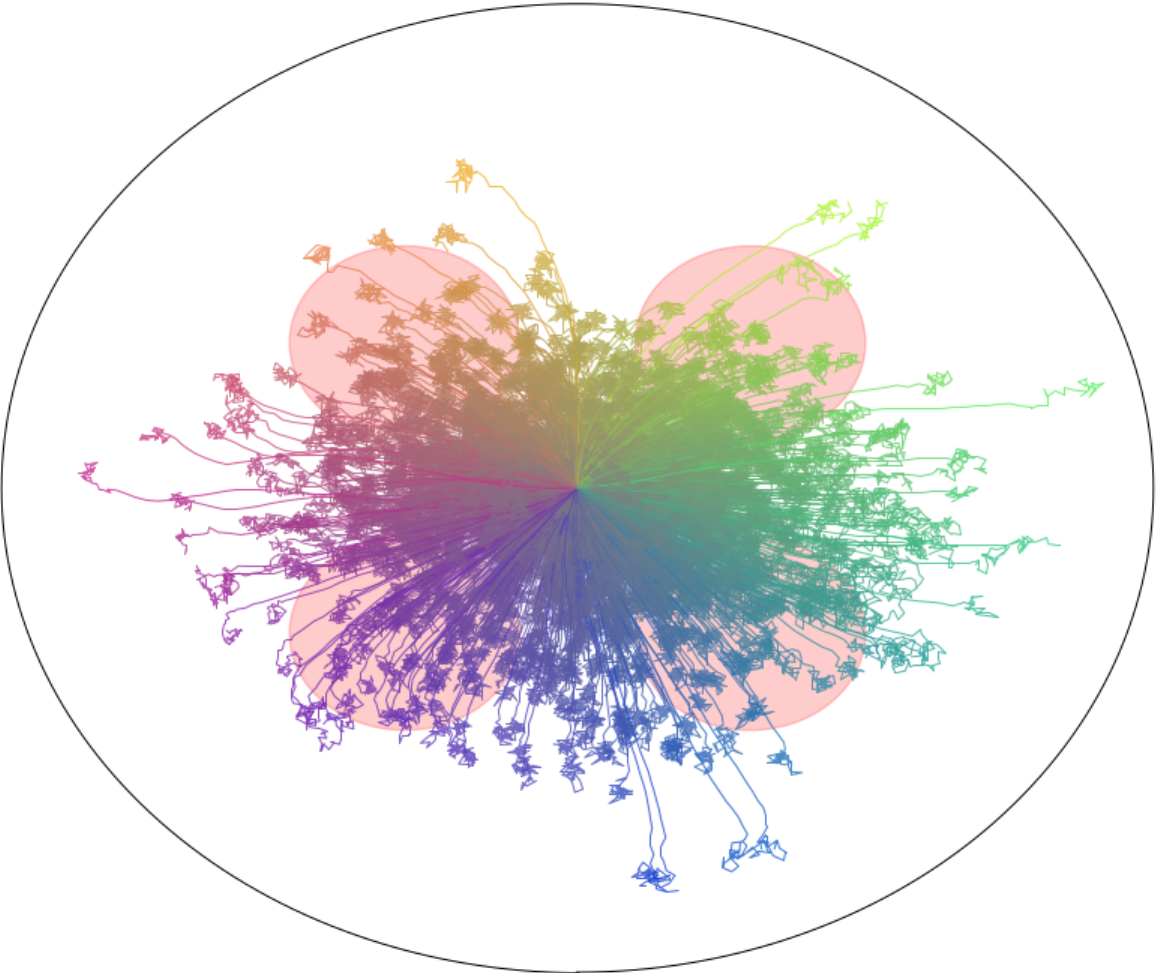} 
        \caption{DIAYN}
         \label{fig: diayn_skills_nav2d}
    \end{subfigure}
    \hfill
    \begin{subfigure}[b]{0.15\textwidth}
        \centering
        \includegraphics[width=\textwidth]{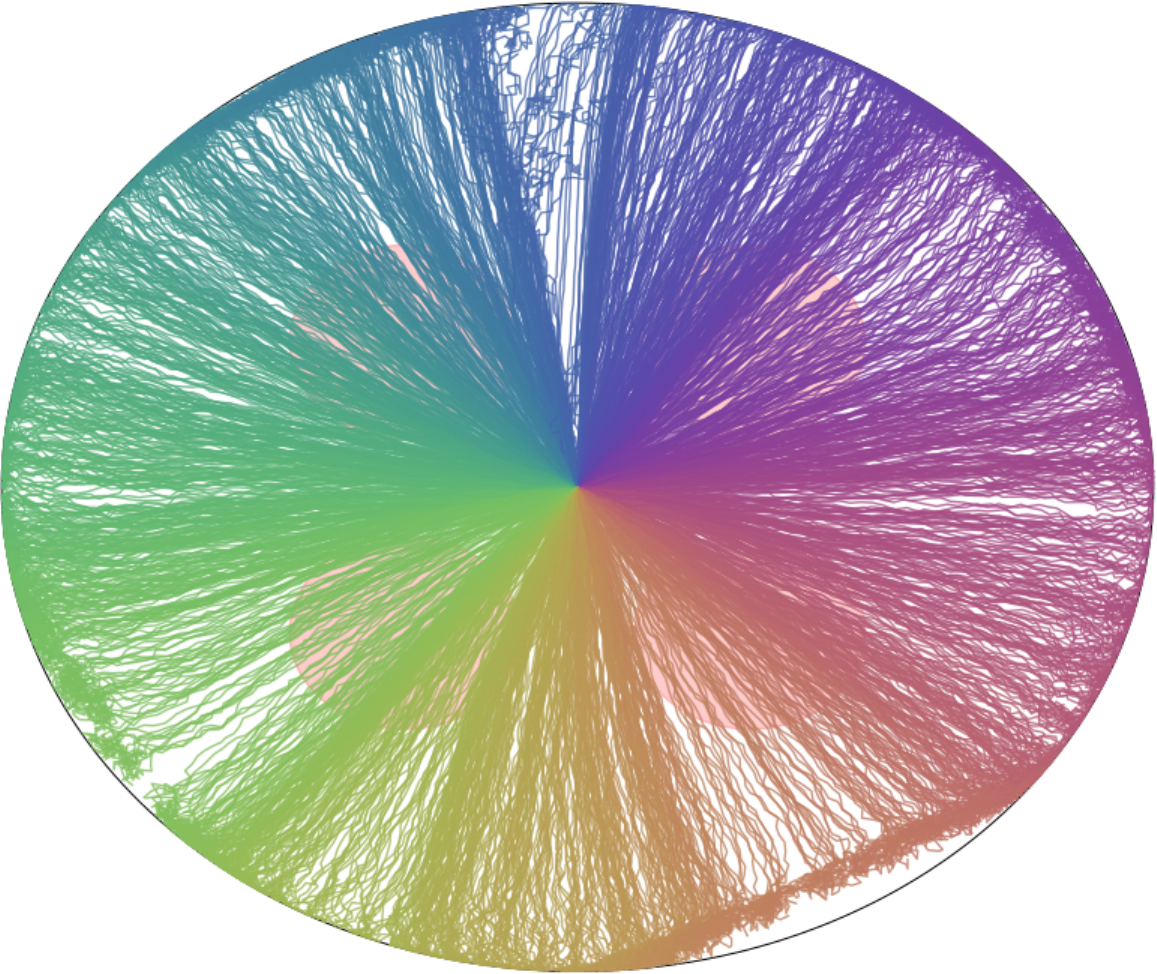} 
        \caption{LSD}
         \label{fig: lsd_skills_nav2d}
    \end{subfigure}
    \hfill
    \begin{subfigure}[b]{0.15\textwidth}
        \centering
        \includegraphics[width=\textwidth]{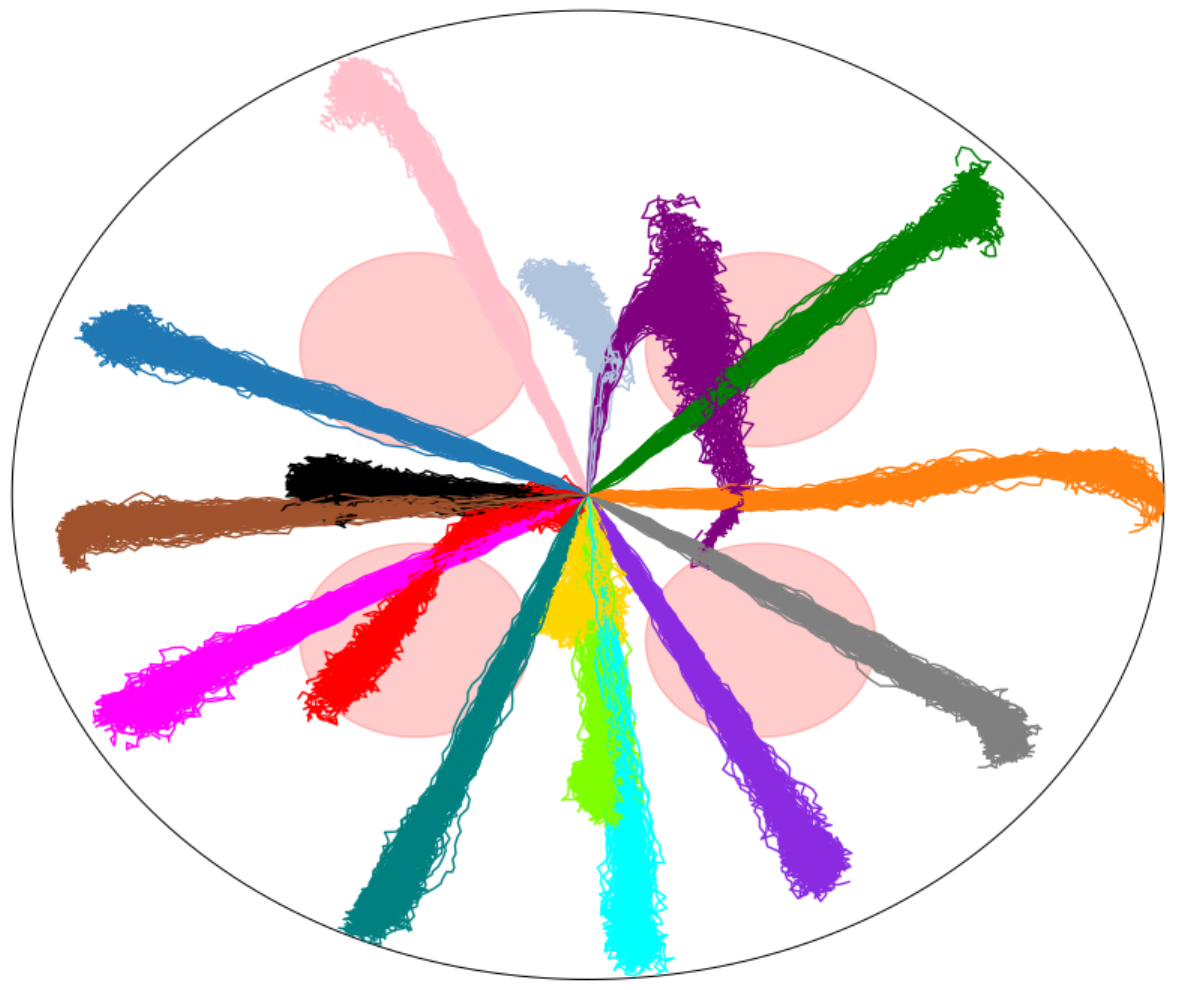} 
        \caption{CDP}
        \label{fig: cdp_skills_nav2d}
    \end{subfigure}
    \hfill
    \begin{subfigure}[b]{0.15\textwidth}
        \centering
        \includegraphics[width=\textwidth]{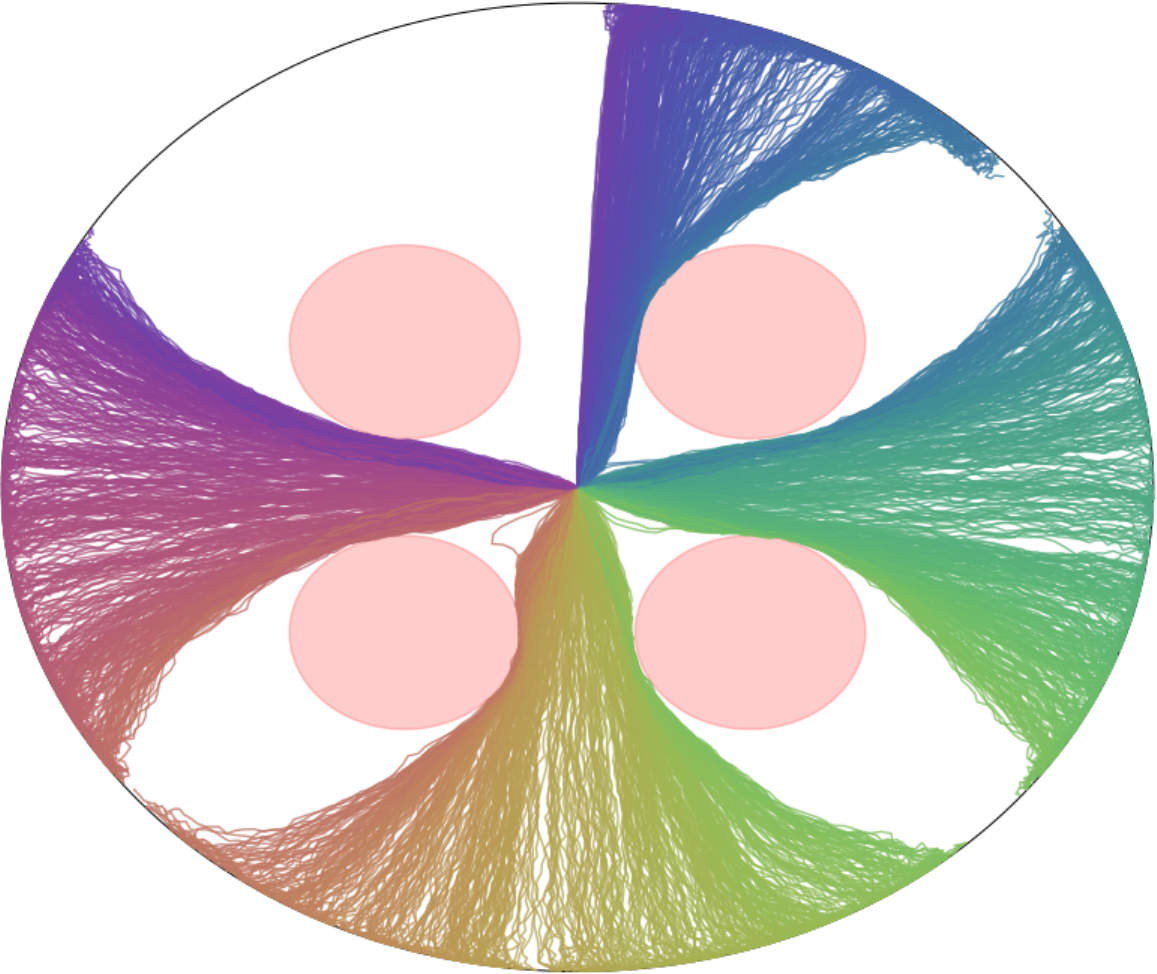} 
        \caption{SMERL}
        \label{fig: smerl_skills_nav2d}
    \end{subfigure} 
    \hfill
        \begin{subfigure}[b]{0.15\textwidth}
        \centering
        \includegraphics[width=\textwidth]{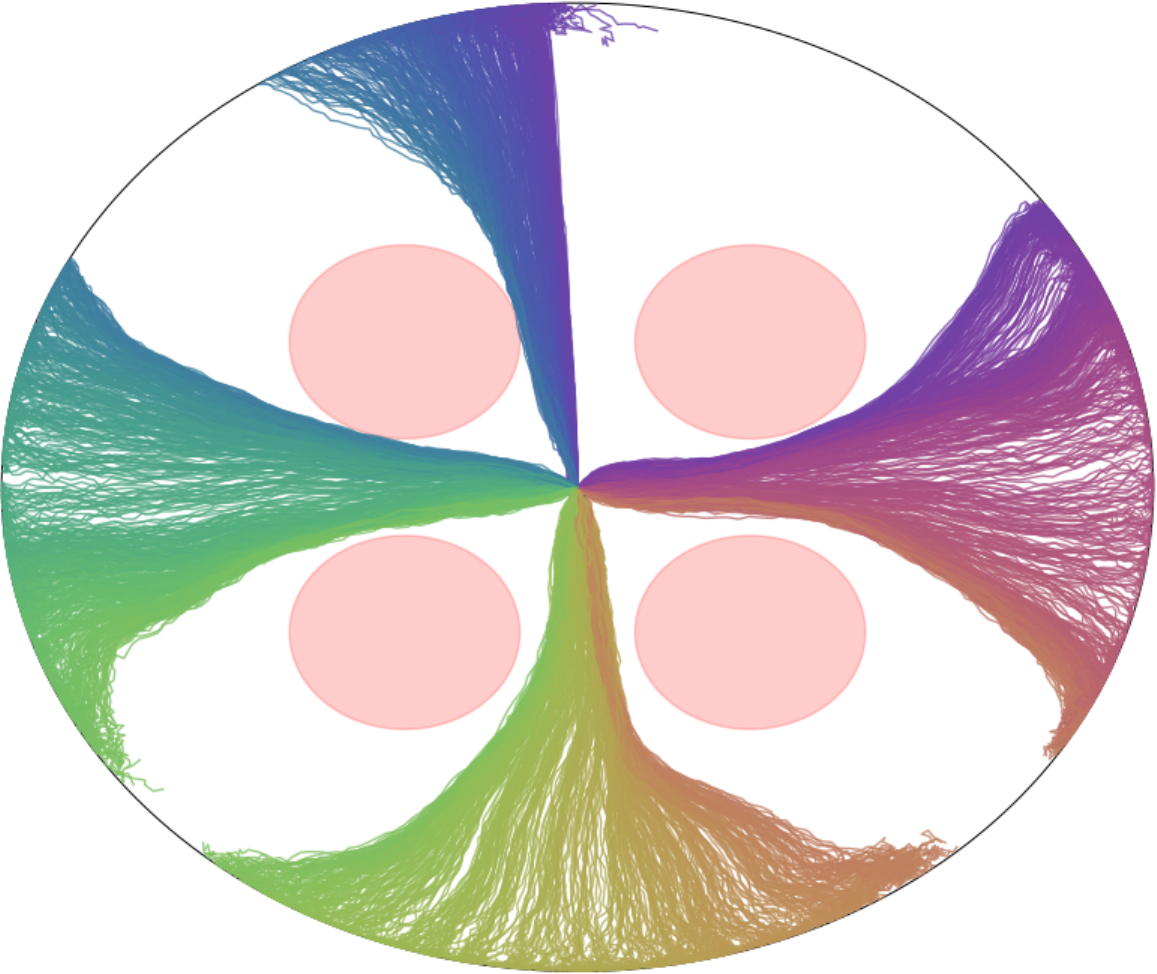} 
        \caption{SMERL+PbRL}
        \label{fig: smerl_pbrl_skills_nav2d}
    \end{subfigure} 
    \hfill
    \begin{subfigure}[b]{0.15\textwidth}
        \centering
        \includegraphics[width=\textwidth]{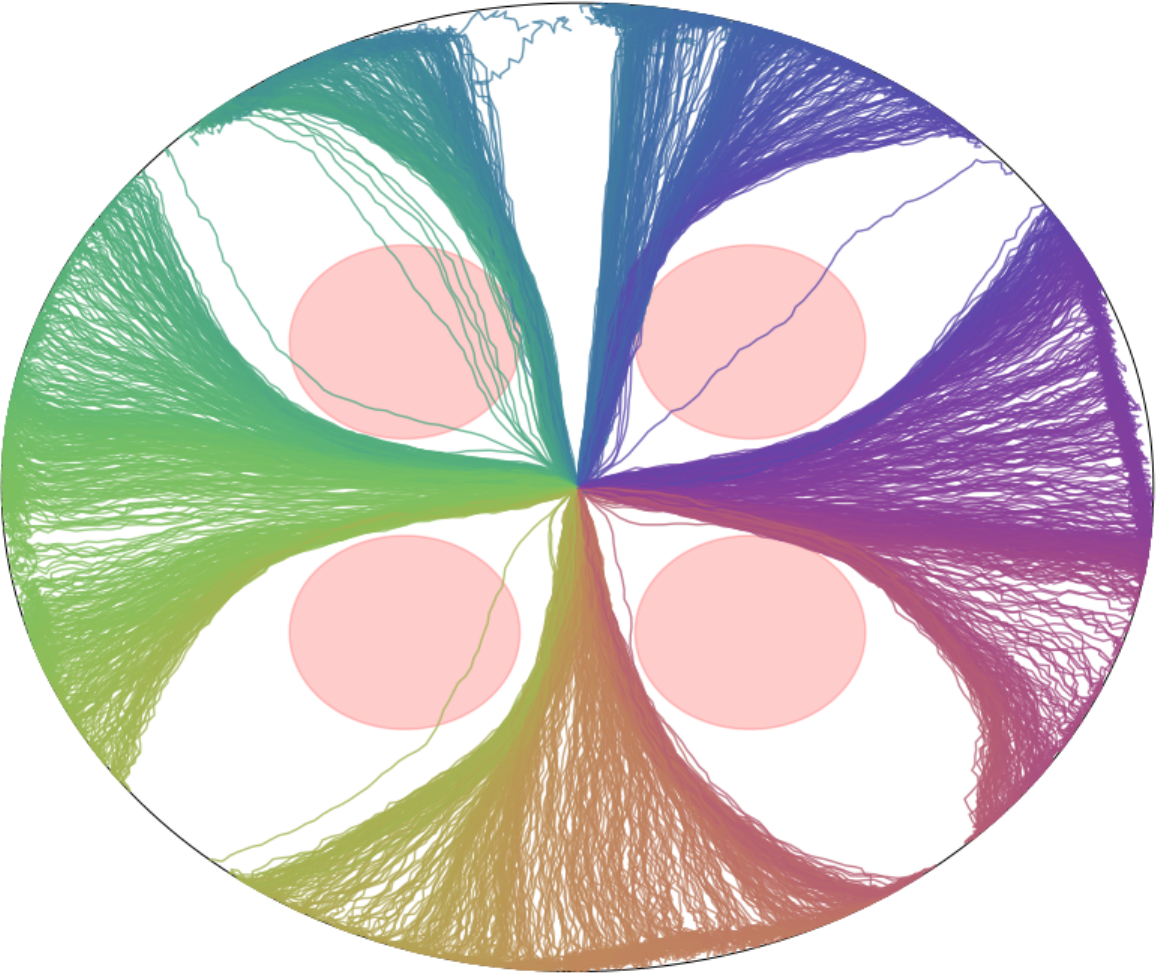} 
        \caption{HaSD (Ours)}
        \label{fig: hasd_skills_nav2d}
    \end{subfigure} 
    \caption{Skill sets learned from all baselines in the 2D navigation environment. LSD covers a larger region of the environment than DIAYN, while HaSD avoids hazardous areas while maintaining good coverage in safe places. SMERL methods are well aligned, but the coverage is not optimal.}
    \label{fig: Experiment|Nav2d|lsd_hasd_skills}
    \Description{A visual presentation of the skills learned with each method in a 2D navigation environment.}
\end{figure*}

%Skill sets learned from all baselines in the 2D navigation environment. LSD covers a larger region of the environment than DIAYN, while HaSD avoids hazardous areas while maintaining good coverage in safe places. SMERL methods are well aligned, but the coverage is not optimal. Finally, \begin{math}\alpha\end{math}-HaSD produce a range of aligned skills, where the higher the \begin{math}\alpha\end{math}, the lower the diversity and coverage is.
    %{Skill sets learned with DIAYN, LSD and HaSD in the 2D navigation environment. LSD covers a larger region of the environment than DIAYN, while HaSD avoids hazardous areas while maintaining good coverage in safe places.}

%% file: paper/experiments/sections/baselines.tex
\subsection{Baselines}
We compare HaSD with 5 baselines ranging from using no human information to using preference information, and finally to having direct access to human information. Wherever possible, we implement baselines over LSD\cite{park2021lipschitz}, the base for unsupervised skill discovery for our method. We overview the baselines below:
\begin{itemize}
  \item \textbf{Unaligned Skill discovery}: We train DIAYN\citep{eysenbach2018} and LSD\citep{park2021lipschitz} without any information about human preferences, serving as baselines for unaligned skill discovery.
  \item \textbf{Human-aligned Skill discovery with Complete Information}: We train original SMERL\citep{kumar2020one} which first maximises the ground truth rewards, and then diversifies skills, serving as a baseline with complete information.
  \item \textbf{Human-aligned Skill discovery with Preferences}: We train CDP\citep{hussonnois2023controlled}  and extend SMERL to incorporate human preferences (SMERL+PbRL), serving as a baseline using human preferences.
\end{itemize}

%% file: paper/experiments/sections/nav2d/a_hasd_Q.tex
\begin{figure*}[htbp]
    \centering
    \begin{subfigure}[b]{0.15\textwidth}
        \centering
        \includegraphics[width=\textwidth]{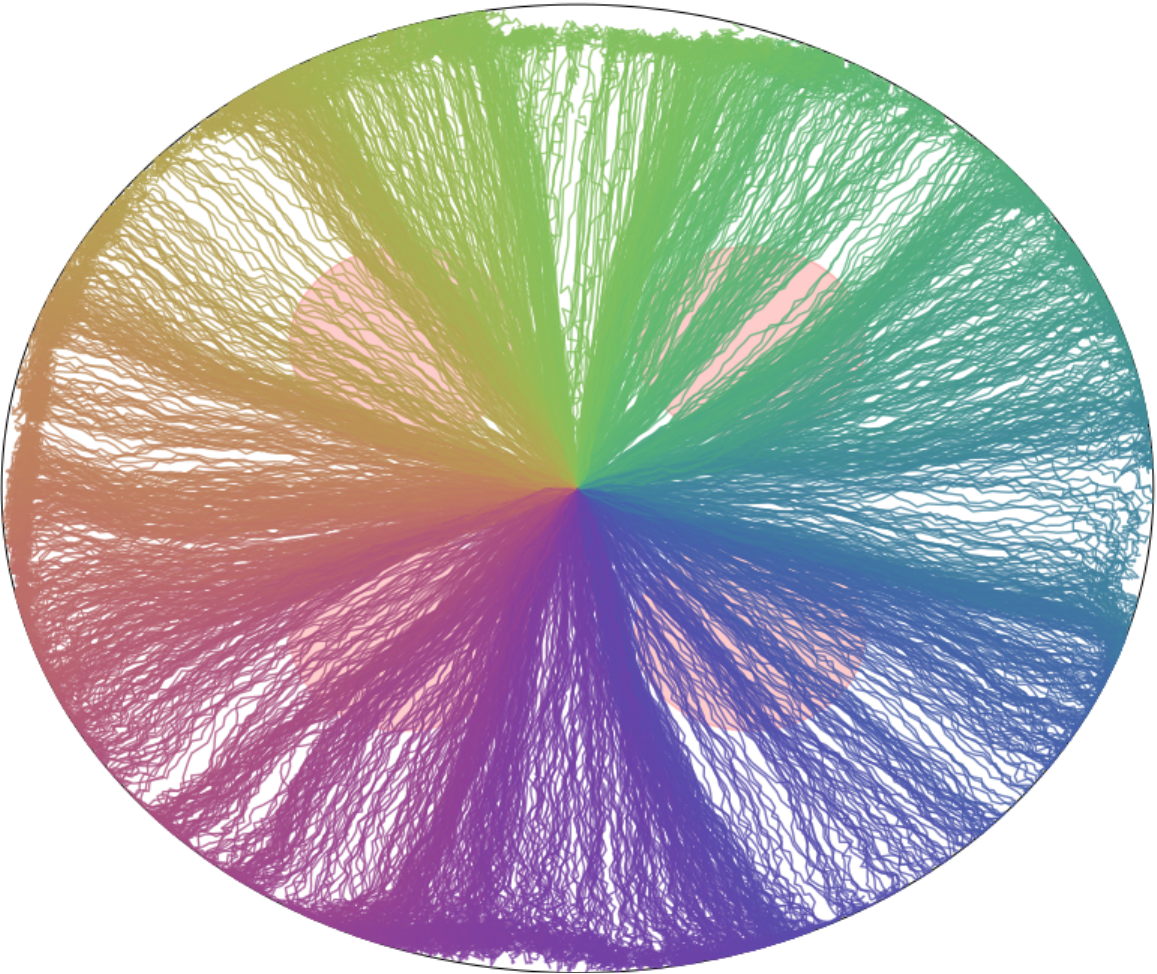} 
        \caption{0-HaSD}
        \label{fig: 0-HaSD}
    \end{subfigure}
    \hfill
    \begin{subfigure}[b]{0.15\textwidth}
        \centering
        \includegraphics[width=\textwidth]{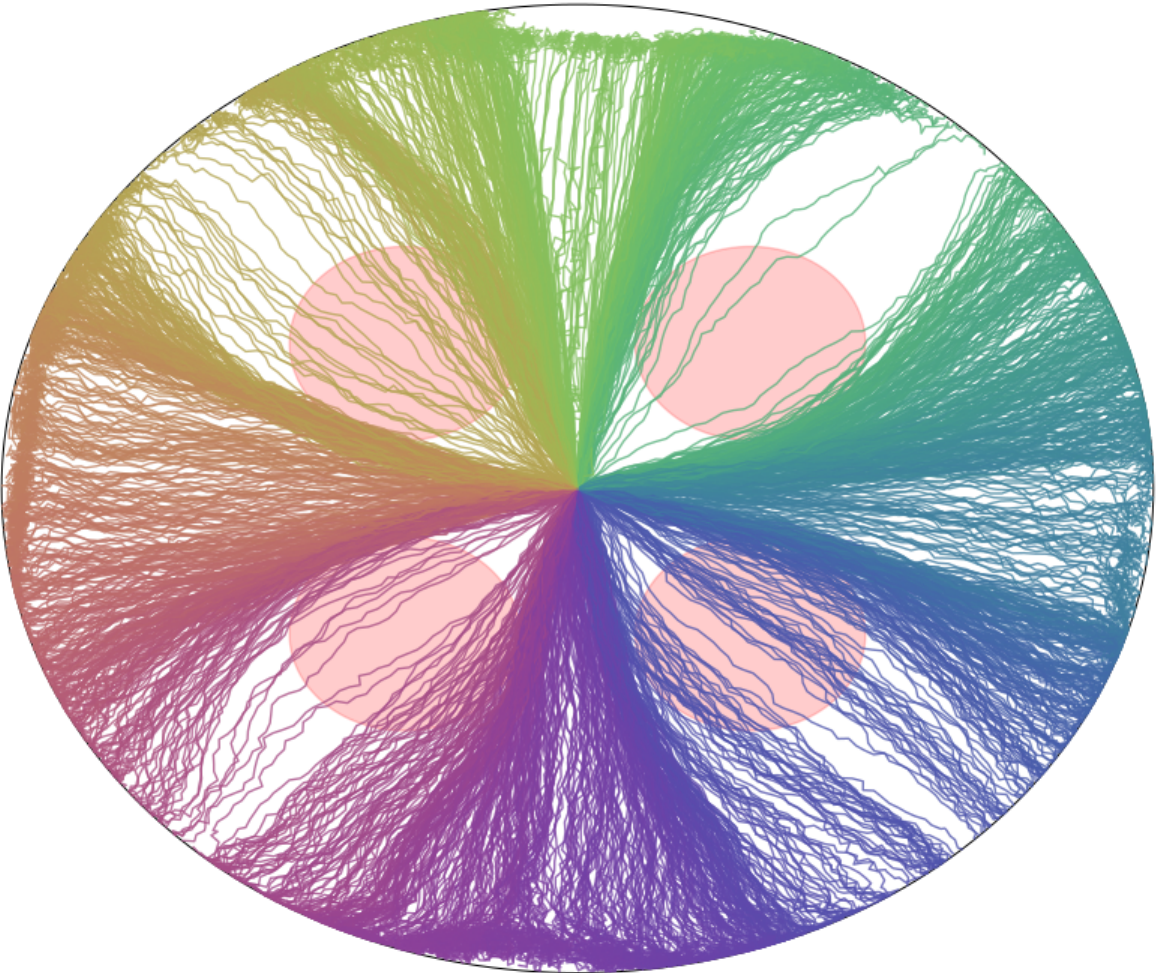} 
        \caption{0.1-HaSD}
        \label{fig: 0.1-HaSD}
    \end{subfigure}
    \hfill
        \begin{subfigure}[b]{0.15\textwidth}
        \centering
        \includegraphics[width=\textwidth]{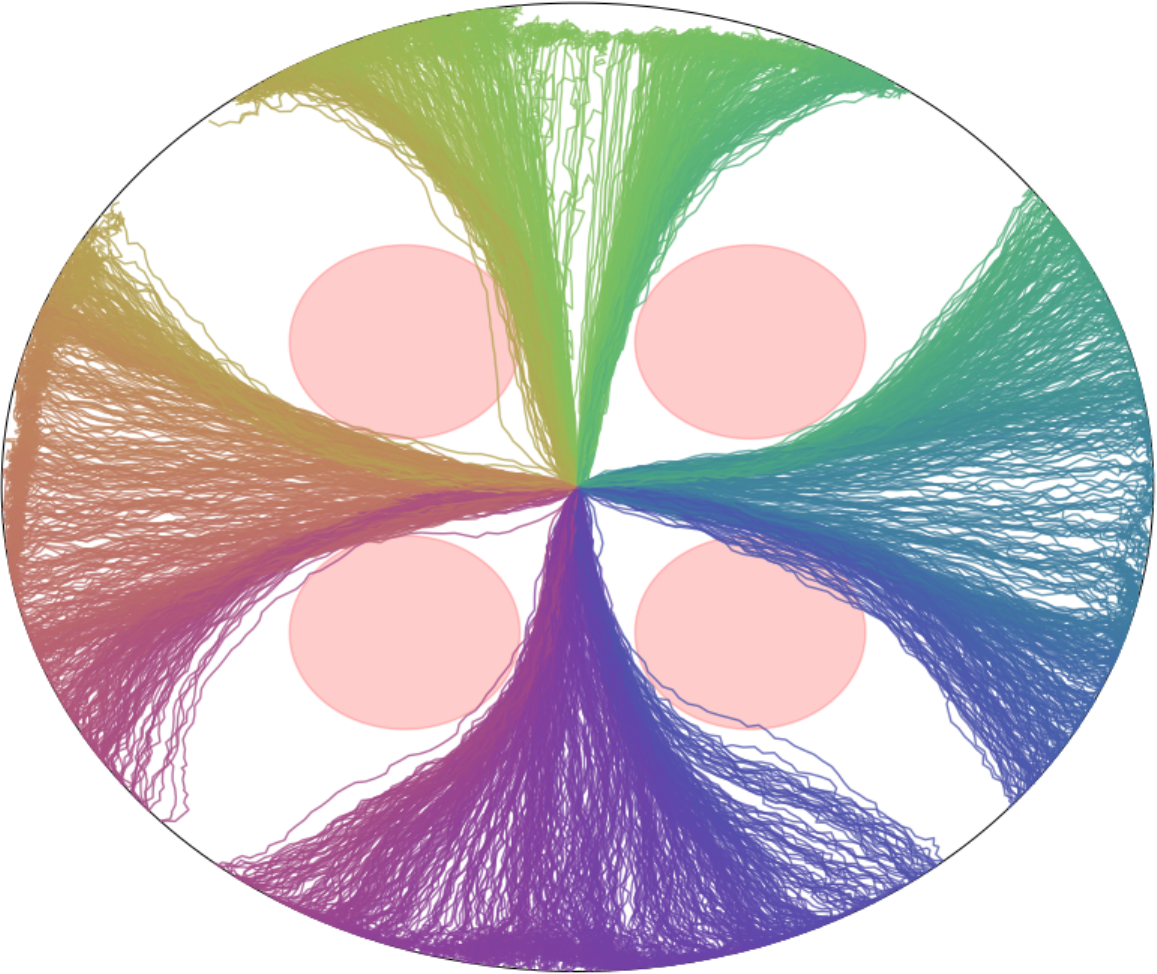} 
        \caption{0.2-HaSD}
        \label{fig: 0.2-HaSD}
    \end{subfigure}
    \hfill
    \begin{subfigure}[b]{0.15\textwidth}
        \centering
        \includegraphics[width=\textwidth]{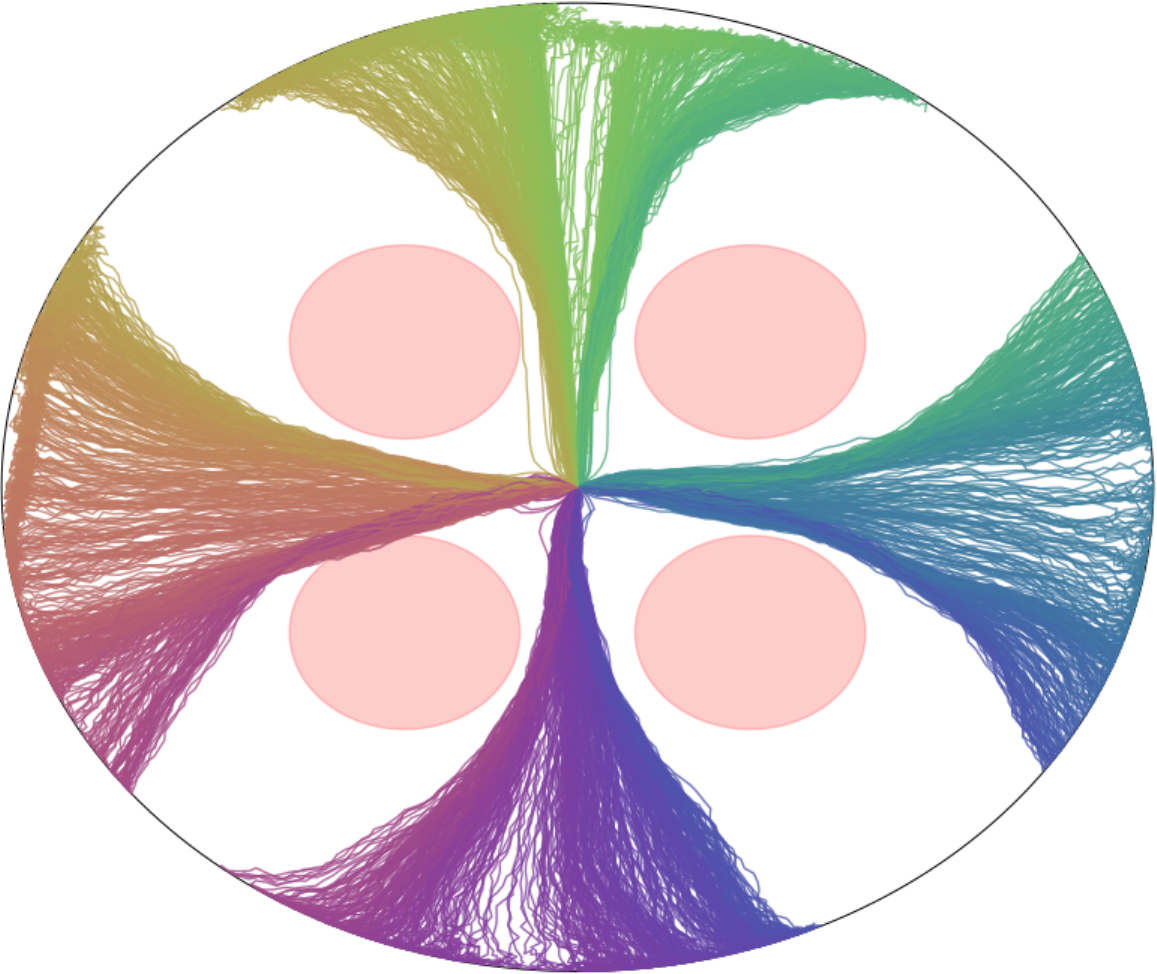} 
        \caption{0.5-HaSD}
        \label{fig: 0.5-HaSD}
    \end{subfigure}
    \hfill
    \begin{subfigure}[b]{0.15\textwidth}
        \centering
        \includegraphics[width=\textwidth]{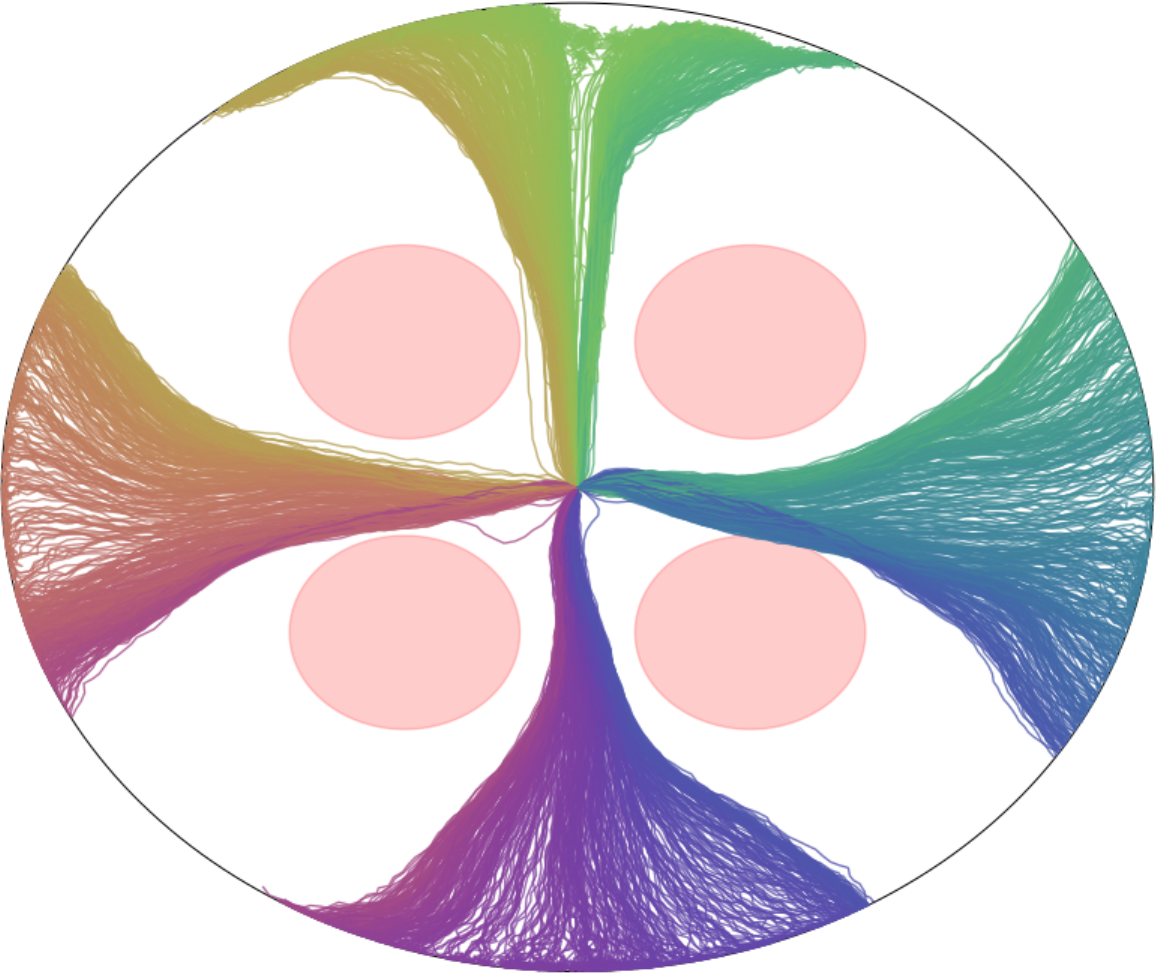} 
        \caption{1-HaSD}
        \label{fig: 1-HaSD}
    \end{subfigure}
    \caption{Comparing visually skill set obtained with \begin{math}\alpha\end{math}-HaSD by changing \begin{math} \alpha\end{math} in the 2D navigation environment. When alpha is set to 0 the skill set is similar to LSD. The higher the \begin{math}\alpha\end{math}, the lower the diversity and coverage is.}
    \label{fig: Experiment|Nav2d|a_hasd_skills}
    \Description{A visual presentation of the skills learned with \begin{math}\alpha\end{math}-HaSD for different value of \begin{math}\alpha\end{math} in a 2D navigation environment.}
\end{figure*}

%Skill sets learned from all baselines in the 2D navigation environment. LSD covers a larger region of the environment than DIAYN, while HaSD avoids hazardous areas while maintaining good coverage in safe places. SMERL methods are well aligned, but the coverage is not optimal. Finally, \begin{math}\alpha\end{math}-HaSD produce a range of aligned skills, where the higher the \begin{math}\alpha\end{math}, the lower the diversity and coverage is.
    %{Skill sets learned with DIAYN, LSD and HaSD in the 2D navigation environment. LSD covers a larger region of the environment than DIAYN, while HaSD avoids hazardous areas while maintaining good coverage in safe places.}

%% file: paper/experiments/sections/environments.tex
\subsection{Environment Descriptions}
\label{subsec: Experiment|Environments_Description}
\subsubsection{Nav2D}
\label{subsubsec: Experiment|Environments_Description|nav2d}
The 2D navigation environment consists of a two-dimensional circular room enclosed by walls that confine the agent to the circular area. The environment contains 4 fixed hazardous areas (red regions in Figure \ref{fig: Experiment|Nav2d|lsd_hasd_skills}) associated with a safety cost. The agent begins each episode in the center of the room, until episode termination, which occurs after $75$ steps. The agent only accesses its horizontal and vertical coordinates (X, Y). It can deterministically change its direction and amplitude of steps to freely move in the environment. State and action spaces are continuous. The ground truth reward function is designed to mimic the following preference: `\textit{Trajectories that travel as far as possible from the initial position while avoiding unsafe regions should be preferred}'. Details of the corresponding precise ground truth reward function used to simulate this objective can be found in Appendix \ref{subsec: appendix|gt}.

\subsubsection{Safety Gymnasium Environments}
In the Safety Gymnasium environment, we evaluated our method with five different agents (point, car, racecar, doggo and ant) shown in Figure \ref{fig: safety_gymnasium agents} across two environments. Each agent has its own state and action space, and learning the corresponding agent controller increases in complexity - i.e., point (least complex) to ant (most complex).
\subsubsection{Hazardous-Room}
\label{subsubsec: Experiment|Environments_Description|H1}
The Hazardous-Room environment as presented in Figure \ref{fig: Hazardous-Room environment} consist of a two-dimensional square room enclosed by walls that doesn't restrain the agent but is associated with some safety (alignment) cost. The environment contains four fixed hazardous areas (purple regions in Figure \ref{fig: Hazardous-Room environment}) associated with some safety (alignment) cost. The agent begins each episode in the center of the room until episode termination, which occurs after $200$ steps. The agent accesses internal state and lidar information about the hazardous areas. The ground truth reward function is designed to mimic the following preference `\textit{Trajectories that travel as far as possible from the initial position while avoiding unsafe regions and staying in the enclosed area  should be preferred }' - details can be found in Appendix \ref{subsec: appendix|gt}.

\subsubsection{Push-Room}
\label{subsubsec: Experiment|Environments_Description|HR3}
The Push-Room environment as presented in Figure \ref{fig: Push-Room environment} consists of a two-dimensional square room enclosed by walls that do not restrain the agent, but is associated with some safety (alignment) cost. The environment contains a box in the center of the room that can be pushed. The agent begins each episode in the northeast corner of the room until episode termination, which occurs after $400$ steps. The agent accesses internal state and lidar information about the box. The ground truth reward function is designed to mimic the following preference `\textit{Trajectories that make the box travel as far as possible from its initial position should be preferred }' details can be found in Appendix \ref{subsec: appendix|gt}.

%% file: paper/experiments/sections/nav2d/QandQ.tex
\subsection{Qualitative and Quantitative Comparisons in Nav2D}
\label{subsec: exp|nav2d|hasd}

In this section, we demonstrate that HaSD and \begin{math}\alpha\end{math}-HaSD can acquire a continuous set of skills that cover the environment while avoiding unsafe regions according to human preference. To this end, we sampled 2000 skills from policies trained across 5 seeds with DIAYN, LSD, SMERL, SMERL+PbRL, CDP HaSD and \begin{math}\alpha\end{math}-HaSD, and compared them qualitatively and quantitatively. 

Qualitatively, Figure \ref{fig: lsd_skills_nav2d} shows how LSD's skills cover the entire room, but ignore the unsafe region. In contrast, HaSD's skills as shown in Figure \ref{fig: hasd_skills_nav2d} adequately cover the environment while avoiding unsafe areas.  On the other hand, $\alpha$-HaSD can learn to generate a range of diversity-alignment trade-offs. \Cref{fig: 0-HaSD,fig: 0.1-HaSD,fig: 0.2-HaSD,fig: 0.5-HaSD,fig: 1-HaSD} illustrates how \begin{math}\alpha\end{math} can be adjusted to zero to produce a skills set similar to LSD as in Figure \ref{fig: 0-HaSD}, corresponding to a skill set that does not consider human preferences. By increasing \begin{math}\alpha\end{math}, we gradually lose diversity in favour of alignment as seen in the figures \ref{fig: 0.2-HaSD} and \ref{fig: 1-HaSD}. 
Figures \ref{fig: smerl_skills_nav2d} and \ref{fig: smerl_pbrl_skills_nav2d} illustrates that SMERL and SMERL+PbRL are also able to learn diverse skills that avoid unsafe areas however they will generally cover less than HaSD as they focus on optimising the alignment reward first. CDP successfully uses preferences to discover diverse safe goals however it learns to reach some of the desired goals in an unaligned manner as illustrated in Figure \ref{fig: cdp_skills_nav2d}.
\begin{figure}[htbp]
\centering
\includegraphics[width=1.0\linewidth]{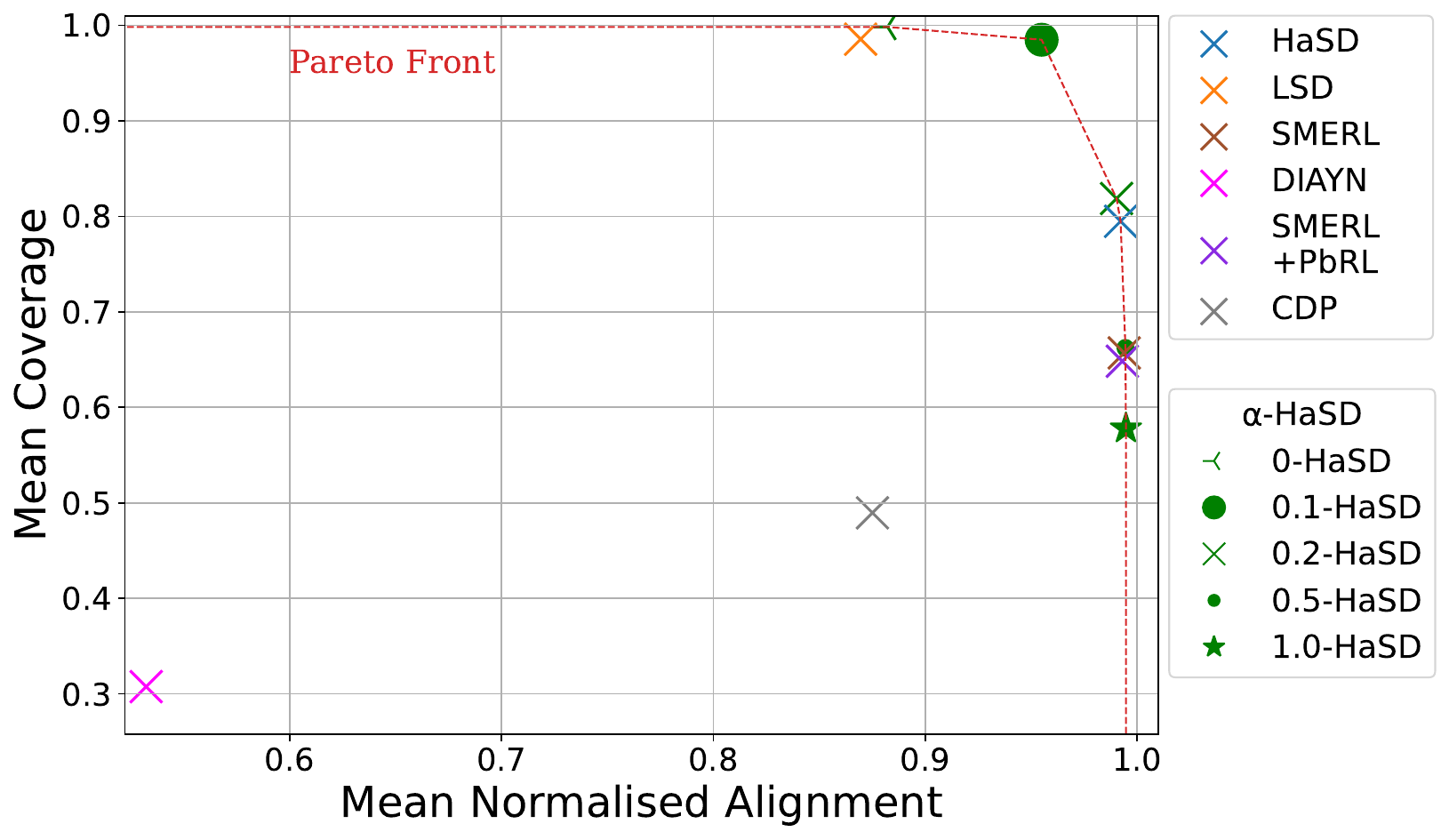} 
\caption{The approximated Pareto front shows that in the 2D navigation environment, LSD solutions achieve high coverage with low alignment on the left side. HaSD solutions are more to the right, offering high alignment while maintaining coverage. $\alpha$-HaSD covers more areas with diverse diversity-alignment trade-offs. SMERL methods attain high alignment but have lower coverage compared to HaSD.}
\label{fig: experiments|nav2d|QandQ|pf}
\Description{The approximated Pareto front shows that in the 2D navigation environment, LSD solutions achieve high coverage with low alignment on the left side. HaSD solutions are more to the right, offering high alignment while maintaining coverage. $\alpha$-HaSD covers more areas with diverse diversity-alignment trade-offs. SMERL methods attain high alignment but have lower coverage compared to HaSD.}
\end{figure}
We report in Figure \ref{fig: experiments|nav2d|QandQ|pf} each skill set (solutions) obtained with LSD or HaSD the normalised total coverage of a skill set (y-axis) and the normalised mean alignment value generated by skills in a skill set (x-axis). Coverage is measured by the number of \begin{math} 0.1 \times 0.1 \end{math} square bins occupied by the agent at least once, while we used the ground truth reward as the alignment value. We highlight the approximate Pareto frontier obtained. Quantitatively, we observe that LSD solutions achieve high coverage but low alignment. In contrast, HaSD solutions deliver high alignment while maintaining high coverage. Meanwhile, as expected SMERL and SMERL+PbRL have marginally better alignment than HaSD, but with significantly less coverage. Finally, DIAYN and CDP demonstrates a lower performance than LSD, HaSD and SMERL methods. Generally, HaSD offers a better balance between exploration and alignment than SMERL or LSD. Lastly, we can observe that \begin{math}\alpha\end{math}-HaSD solutions lie on/close to the approximated Pareto frontier meaning that we learn qualitative solutions over both objectives.

%% file: paper/experiments/sections/nav2d/feedback_sensitivity.tex
\subsection{Sensitivity to Human Feedback Budget}
\label{subsex: experiment|budget|nav2d}
\subsubsection{HaSD and SMERL+PbRL Sensitivity:}
\begin{figure}[htbp]
\centering
\includegraphics[width=0.9\linewidth]{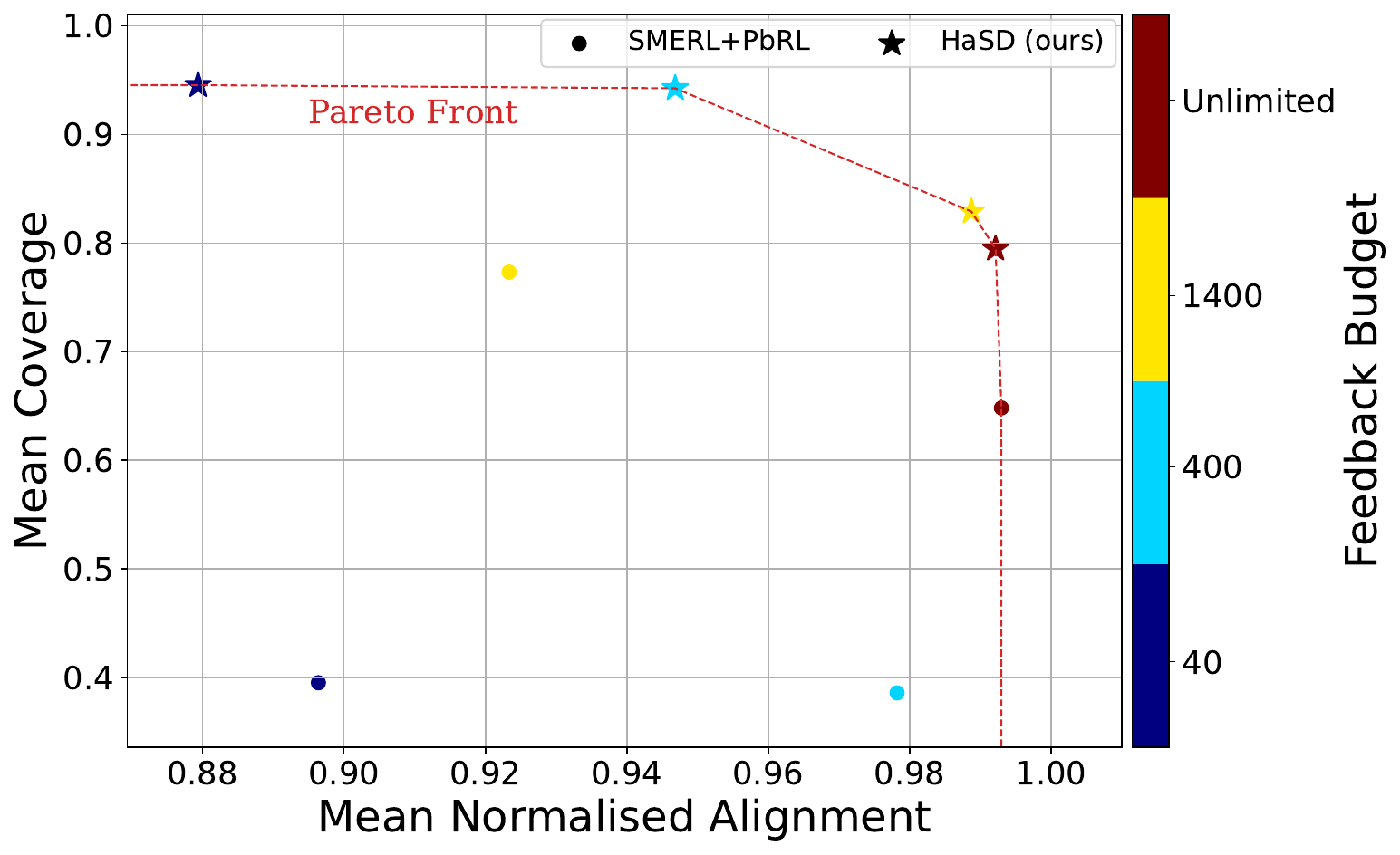} 
\caption{Approximated Pareto front considering SMERL+PbRL and HaSD with different human feedback budget in Nav2d environment. HaSD is more robust to reduction feedback than SMERL+PbRL.}
\label{fig: exp|budget|nav2d|pf}
\Description{Approximated Pareto front considering SMERL+PbRL and HaSD with different human feedback budget in Nav2d environment. HaSD is more robust to reduction feedback than SMERL+PbRL.}
\end{figure}

In this section we analyse the sensitivity of our methods to available budget feedback. The degree to which skills are aligned depends on how well the reward model captures human values. This naturally depends on the availability of feedback labels. As illustrated in Figure \ref{fig: exp|budget|nav2d|pf}, we found that as we decreased the number of feedback received, HaSD's performance over the alignment objective decreased. At the lowest ($40$ feedback instances), its performance was comparable to that of LSD. Additionally, we found that HaSD was more robust to a reduction of human feedback instances than SMERL+PbRL. This is because SMERL's Objective first seeks to optimise performance over task reward, which here is poorly approximated. This illustrates the limitations of SMERL's reward in our situation, where both rewards and a skill discovery objective are required.
\subsubsection{\begin{math} \alpha \end{math}-HaSD Sensitivity:}
\begin{figure}[htbp]
\centering
\includegraphics[width=1\linewidth]{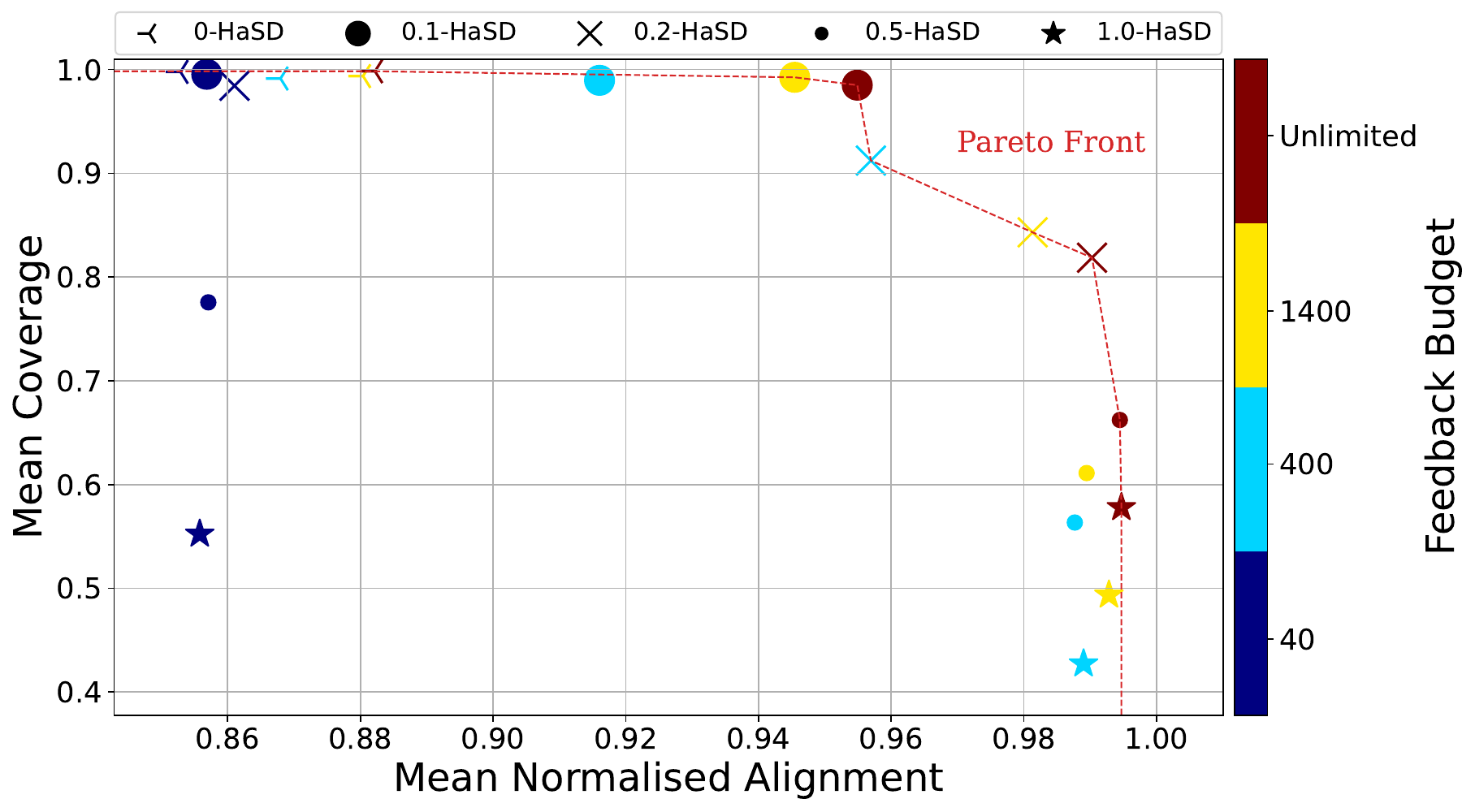} 
\caption{Approximated Pareto front considering \begin{math} \alpha \end{math}-HaSD with different human feedback budget in Nav2d environment.}
\label{fig: exp|budget|nav2d|pf_a_hasd}
\Description{Approximated Pareto front considering \begin{math} \alpha \end{math}-HaSD with different human feedback budget in Nav2d environment.}
\end{figure}

We also analysed the sensitivity of \begin{math} \alpha \end{math}-HaSD to human feedback budgets. 
Figure \ref{fig: exp|budget|nav2d|pf_a_hasd} illustrates the approximate Pareto frontier obtained with all solutions. \begin{math} \alpha \end{math}-HaSD exhibits similar behaviour as HaSD when the budget is restricted, resulting in less accurate alignment rewards. Nevertheless, \begin{math} \alpha \end{math}-HaSD is still able to learn a range of alignment values even on very restrictive budgets. % \textcolor{orange}{highlight reasons if any}
\subsubsection{Hypervolume:}
\begin{figure}
\centering
\includegraphics[width=1.0\linewidth]{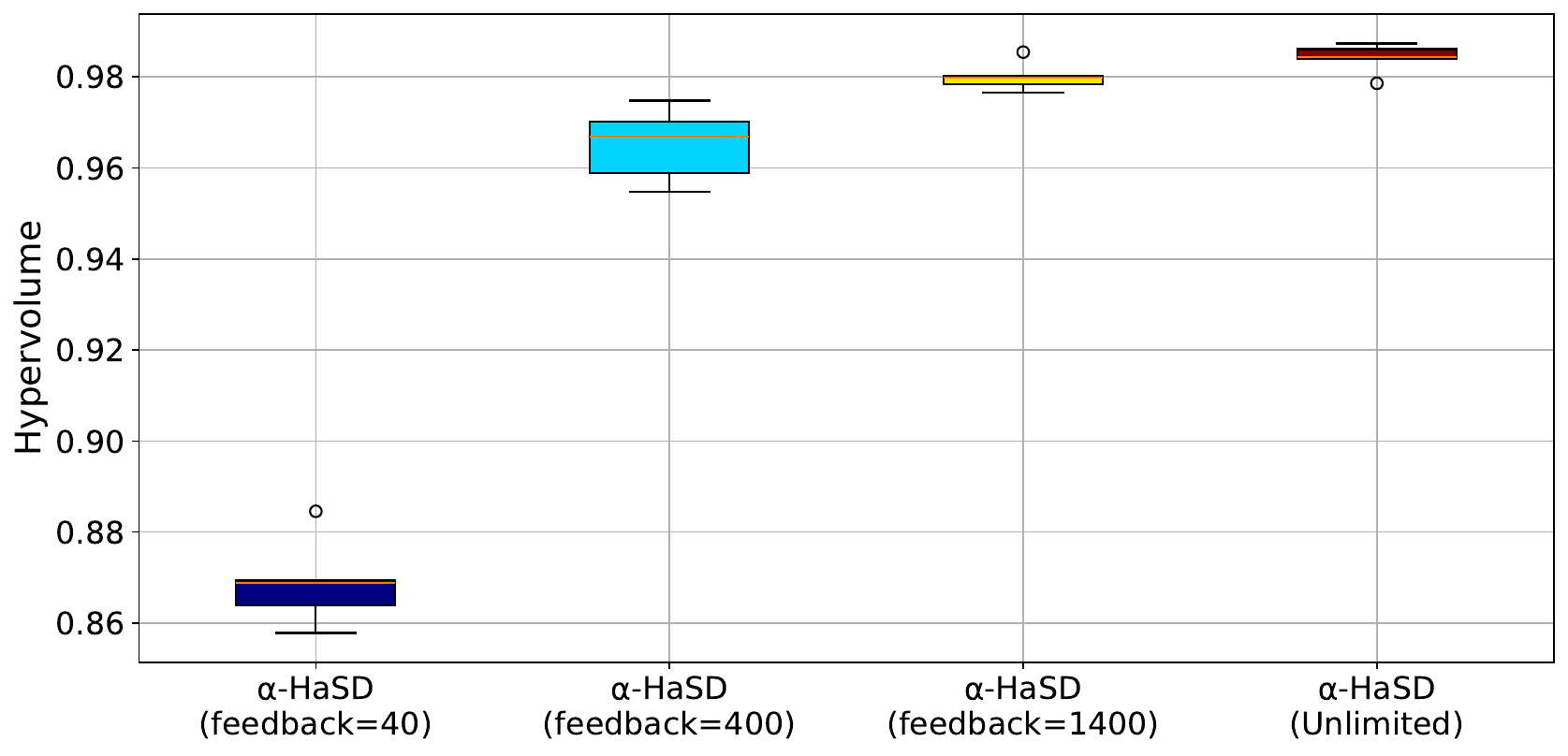} 
\caption{Hypervolume computed for each set of solutions produced by \begin{math} \alpha \end{math}-HaSD with different human feedback budget in Nav2d environment.}
\label{fig: exp|budget|nav2d|hpv}
\Description{Hypervolume computed for each set of solutions produced by \begin{math} \alpha \end{math}-HaSD with different human feedback budget in Nav2d environment.}
\end{figure}

To more accurately quantify the impact of restricted human feedback on \begin{math} \alpha \end{math}-HaSD, we showcase in Figure \ref{fig: exp|budget|nav2d|hpv} the Hypervolume\citep{Zitzler1999EvolutionaryAF} computed for each set of solutions. Hypervolume is a popular metric in multi-objective problems\citep{felten2023a} to measure how a set of solutions is diverse but simultaneously performant as well.  The Hypervolume corresponds to the set of points "contained" within an n-dimensional space. As expected we can see that as the amount of feedback diminishes, the Hypervolume metrics also decline, indicating a reduction in the quality of the solution set. Additionally, the variability of the solution set as feedback decreases.

%% file: paper/experiments/sections/nav2d/downstreamTask.tex
\subsection{Nav2D: Downstream Task}
\label{subsec: exp|nav2d|ft}
\begin{SCfigure}
    \begin{subfigure}[b]{0.125\textwidth}
        \centering
        \includegraphics[width=\textwidth]{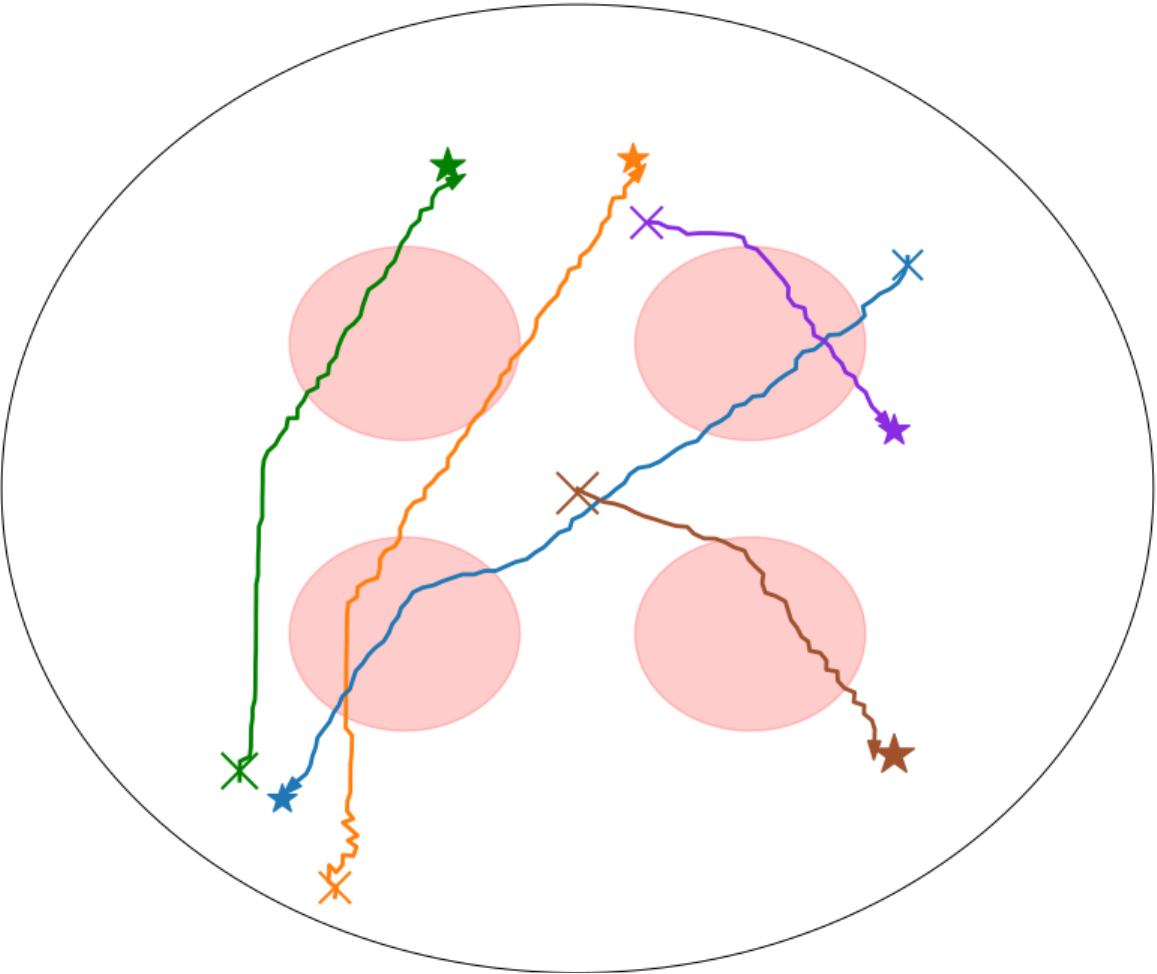} 
        \caption{LSD}
        \label{fig: rebuttal|ft|LSD}
    \end{subfigure}
    \hfill
    \begin{subfigure}[b]{0.125\textwidth}
        \centering
        \includegraphics[width=\textwidth]{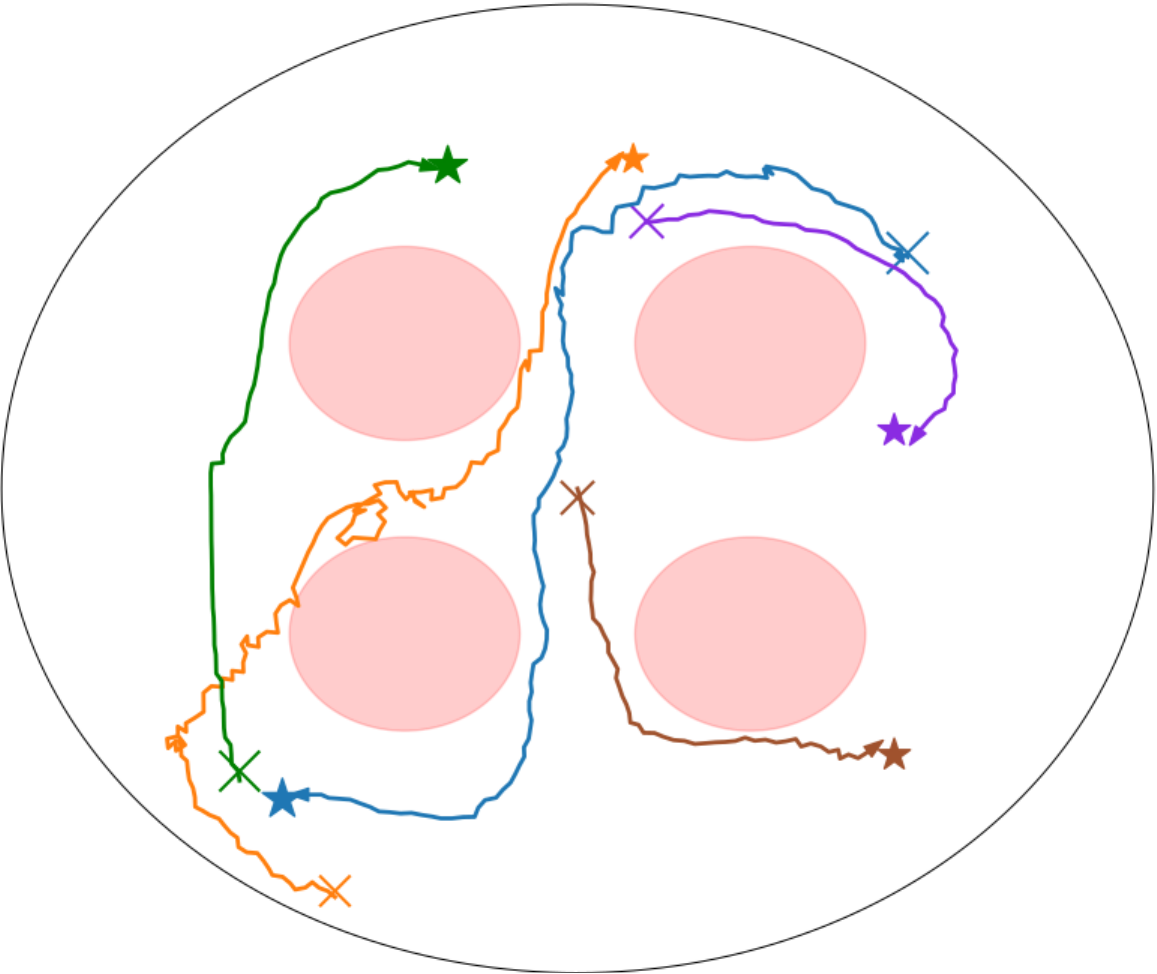} 
        \caption{HaSD (Ours)}
        \label{fig: rebuttal|ft|a_haSD}
    \end{subfigure}
    \caption{Goal-reaching trajectories produced with LSD skills and HaSD skills in the Nav2d environment.}
    \label{fig: nav2|ft|qualitative|goal-reaching}
    \Description{Goal-reaching trajectories produced with LSD skills and HaSD skills in the 2D Navigation environment.}
\end{SCfigure}

In this section, we demonstrate how we can use skills learned with HaSD to achieve downstream tasks in more challenging settings. To this end we implemented a meta-controller in our 2D navigation environment to demonstrate the utility of the discovered skills. 
In this setup:
\begin{itemize}
  \item The agent starts at a random location in the environment, excluding unsafe areas, and aims to reach a goal at another random location, also excluding unsafe areas. 
  \item The environment provides sparse rewards. Specifically, the agent receives a reward signal only upon reaching the goal; otherwise, it receives no reward.
  \item The meta-controller selects a skill which is then executed for one environment-step, replacing the traditional action space with the learned skill space.
  \item Importantly, no explicit information about unsafe areas is provided during this downstream task.
\end{itemize}
We provide more details on the settings in Appendix \ref{subsec: apdx|implementation_details}. We evaluated the meta-controller over 1000 sample goals and reported (across 9 seeds) both the average goal achievement rate and the mean cost generated to achieve them in Table \ref{nav2d|ft-table}. In addition, we compute a ranking score for each method based on their separate ranking on scores and costs. Our results demonstrate that HaSD attains a goal achievement rate comparable to LSD while maintaining significantly lower costs. Compared to HaSD, SMERL and SMERL+PbRL achieve significantly lower goal achievement. This may be due to the lower coverage of these methods, which highlights the importance of our approach to first discovery skill and then aligning with human preferences. The qualitative results in Figure \ref{fig: nav2|ft|qualitative|goal-reaching} demonstrate that the skill space acquired with HaSD retains alignment information, which allows our meta-controller to avoid unsafe areas more  than with LSD, despite no information regarding safety being provided during the downstream task. This evaluation substantiates the fact that skills discovered through our method are not only safe but also useful in downstream tasks.
\begin{table}[htbp]
  \centering
  \caption{Comparison of baselines (DIAYN,LSD, HaSD, SMERL, and SMERL+PbRL) based on their performance scores and associated costs}
  \label{nav2d|ft-table}
  \begin{tabular}{llll }              \\
    \toprule
    Method     & Score & Cost  & Rank  \\
    \midrule
    \textbf{HaSD(Ours)}  & \begin{math} 98.5\% \pm 1.0 \end{math} &  \begin{math}6.7 \pm 0.5 \end{math} & \textbf{ 4 (2,2)}  \\
    LSD       & \begin{math} 99.5\% \pm 0.6 \end{math} & \begin{math} 12.01 \pm 0.3  \end{math} & 5 (1,4) \\
    SMERL+PbRL  & \begin{math} 68.9\% \pm 4.1 \end{math} &  \begin{math}3.86 \pm 1.23 \end{math} & 5 (4,1) \\
    SMERL  & \begin{math} 75.6\% \pm 4.0 \end{math} &  \begin{math}9.49 \pm 2.0 \end{math} & 6 (3,3) \\
    DIAYN       & \begin{math} 23.9\% \pm 3.3 \end{math} & \begin{math} 23.9 \pm 7.2  \end{math} & 10 (5,5) \\
    \bottomrule
  \end{tabular}
\end{table}

%% file: paper/experiments/sections/safety_gymnasium/hazard.tex
\subsection{Safety Gymnasium: Hazard-Room}
\label{subsec: Experiment|HR1}
\begin{figure*}[htbp]
    \centering
    \includegraphics[width=1\textwidth]{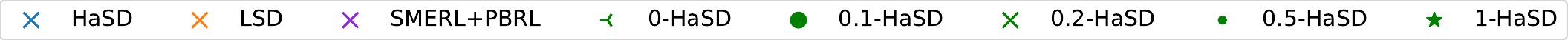}\\
    %\centering
    \begin{subfigure}[b]{0.19\textwidth}
        \centering
        \includegraphics[width=\textwidth]{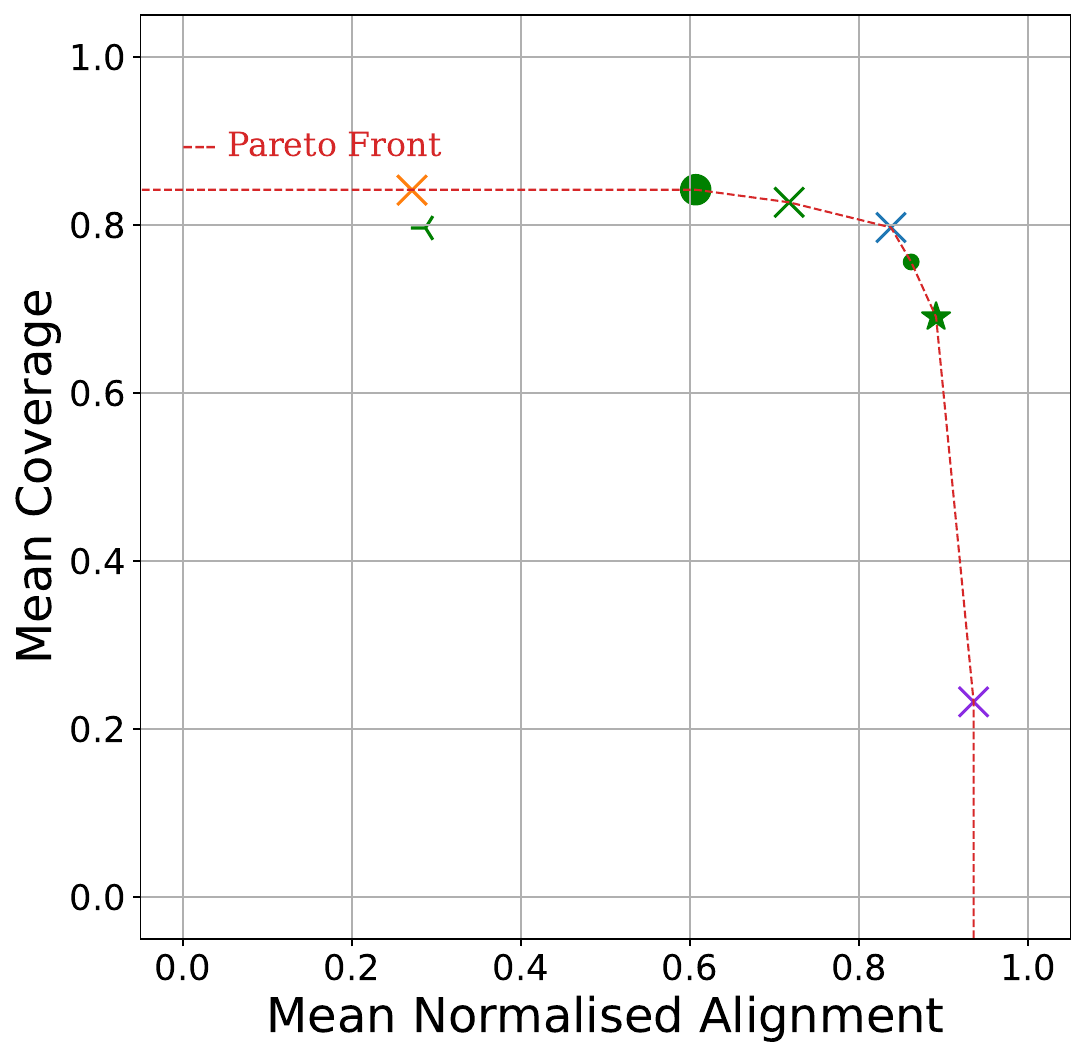} 
        \caption{Point}
        \label{fig: hr1|point|pt}
    \end{subfigure}
    \hfill
    \begin{subfigure}[b]{0.19\textwidth}
        \centering
        \includegraphics[width=\textwidth]{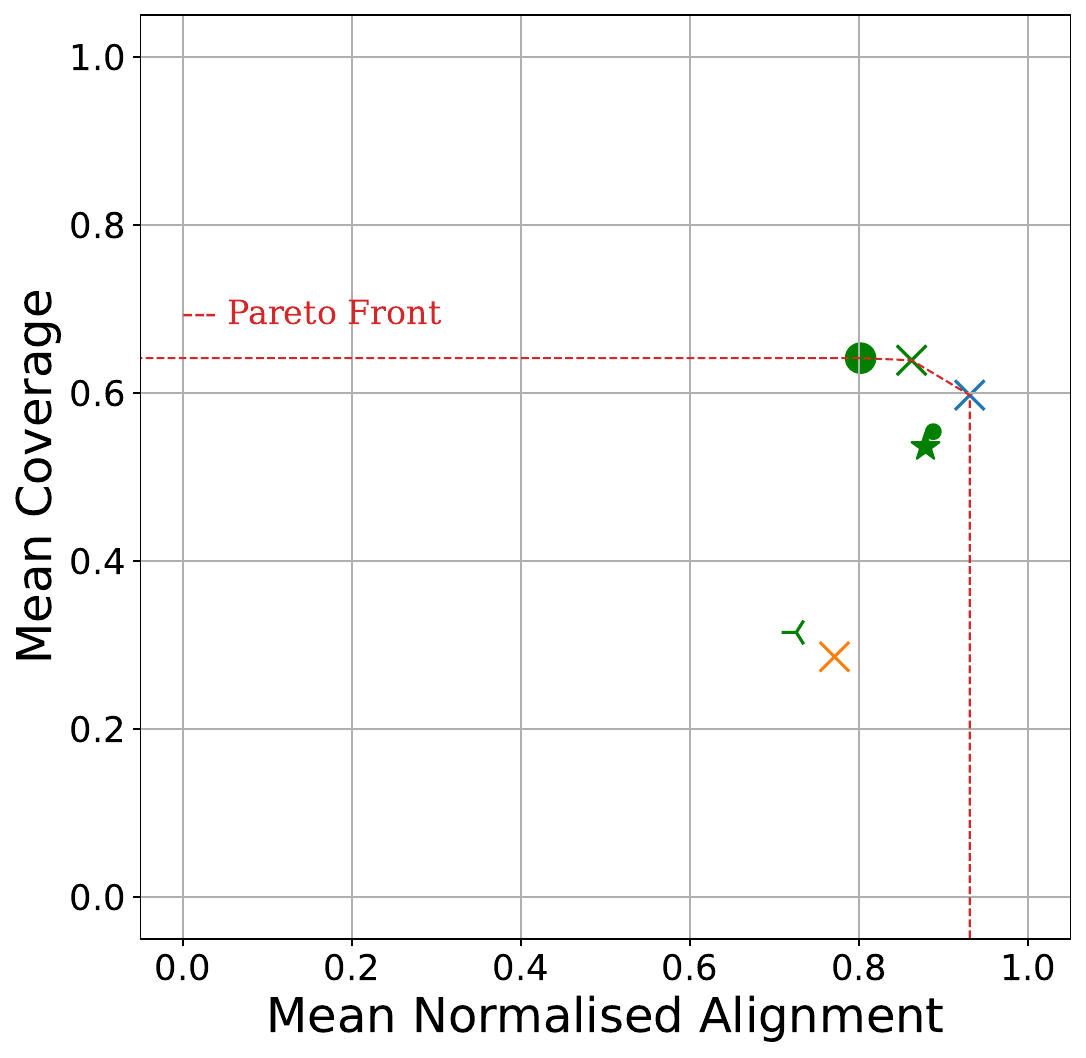}
        \caption{Car}
        \label{fig: hr1|car|pt}
    \end{subfigure}
    \hfill
    \begin{subfigure}[b]{0.19\textwidth}
        \centering
        \includegraphics[width=\textwidth]{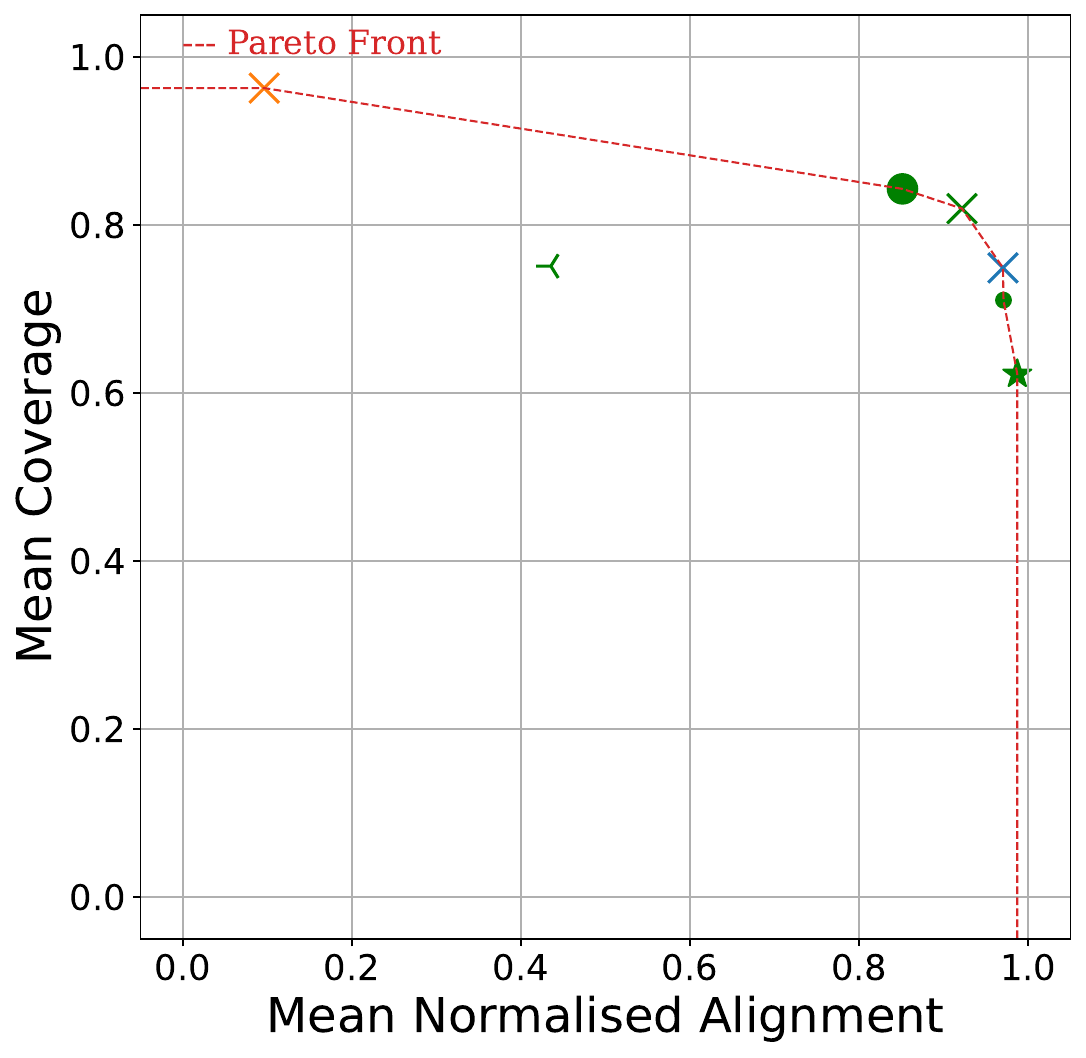}
        \caption{Racecar}
        \label{fig: hr1|racecar|pt}
    \end{subfigure}
    \hfill
    \begin{subfigure}[b]{0.19\textwidth}
        \centering
        \includegraphics[width=\textwidth]{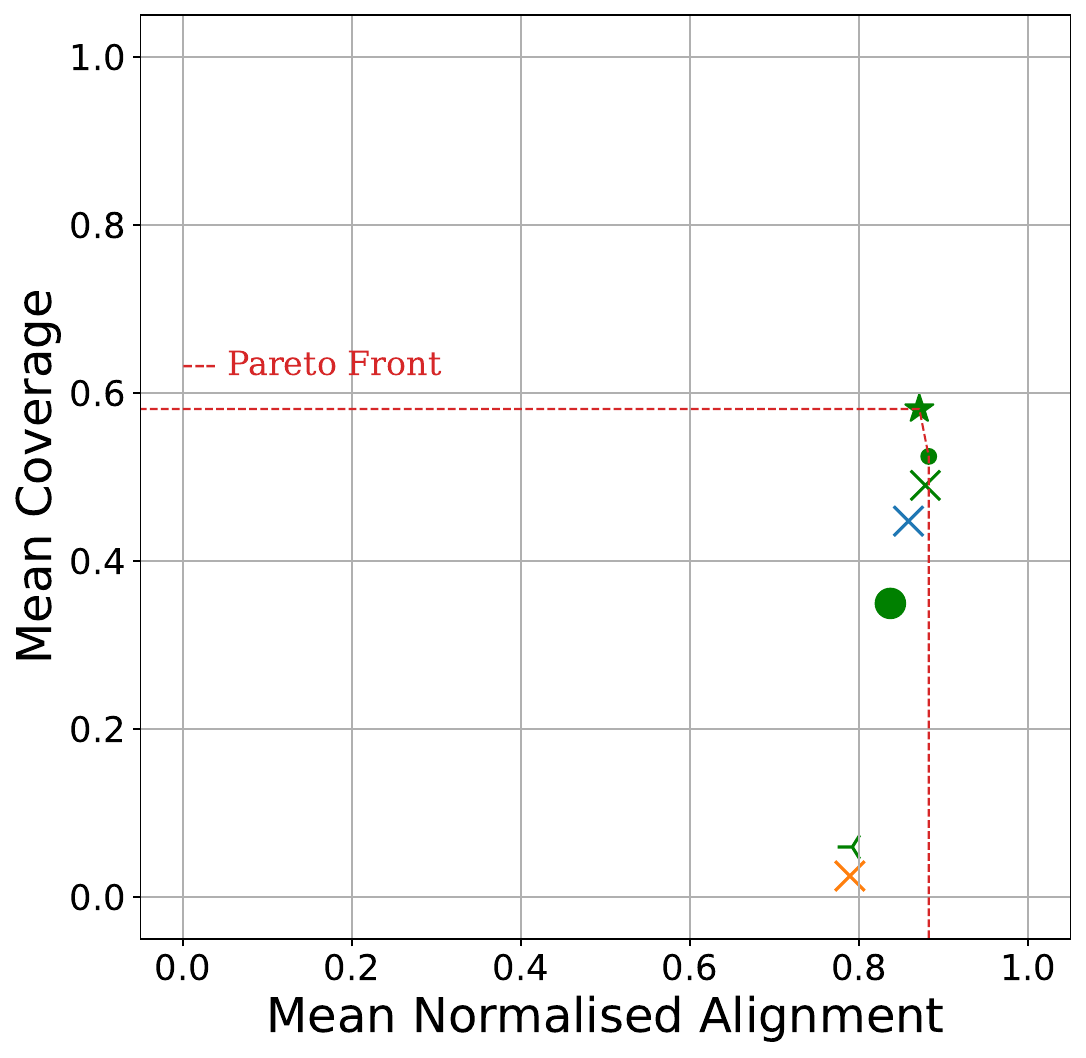} 
        \caption{Doggo}
        \label{fig: hr1|doggo|pt}
    \end{subfigure}
    \hfill
    \begin{subfigure}[b]{0.19\textwidth}
        \centering
        \includegraphics[width=\textwidth]{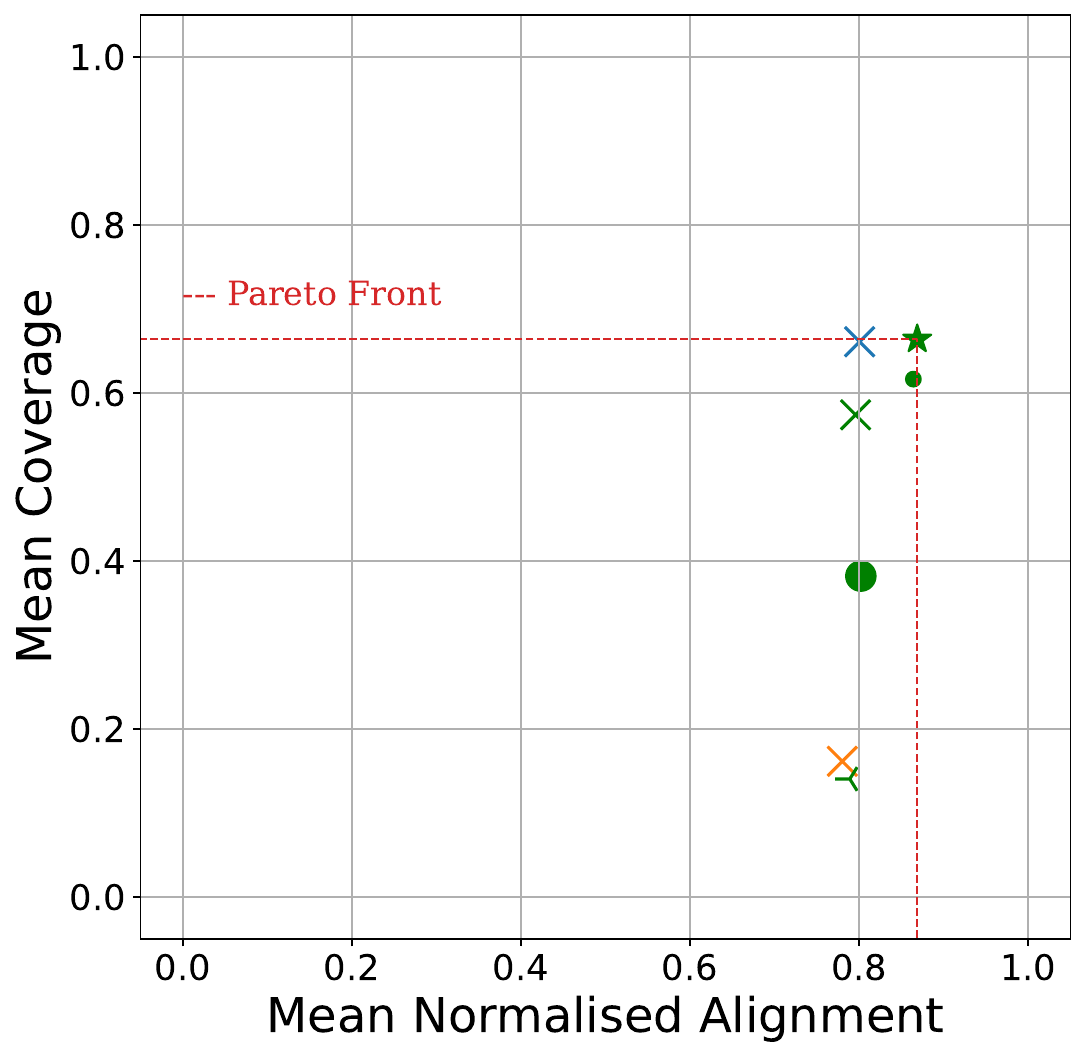} 
        \caption{Ant}
        \label{fig: hr1|ant|pt}
    \end{subfigure}
    \caption{Approximated Pareto Front obtained with LSD, HaSD and \begin{math}\alpha\end{math}-HaSD for each agent in the Hazardous-Room environment. In general, solutions are arranged from left to right, as alignment objectives are more influential.}% In Doggo and Ant, the LSD objective fails to discover skills, leading to HaSD methods relying mostly on the alignment objective.}
    \label{fig: pf_hasard_room1}
    \Description{Approximated Pareto Front obtained with LSD, SMERL+PBRL HaSD and \begin{math}\alpha\end{math}-HaSD for each agent in the Hazardous-Room environment. In general, solutions are arranged from left to right, as alignment objectives are more influential.}
\end{figure*}
In this section we present results obtain in the Hazard-Room environment described in Section \ref{subsubsec: Experiment|Environments_Description|HR3}. All results are presented after sampling 1000 skills from the policy and averaging across 5 seeds. We report a similar pareto front as in Section \ref{subsec: exp|nav2d|hasd} and highlight the approximate Pareto frontier obtained. We trained SMERL+PbRL only on the point agent to minimise experimental workload. We observe similar trends as in nav2d, where SMERL+PbRL achieves high alignment but very low coverage as shown in Figure \ref{fig: hr1|point|pt}. Although we have not run SMERL+PbRL on other agents, we expect similar characteristics (ie., high alignment and low coverage) for these agents as well. Quantitatively, In Figures \ref{fig: hr1|point|pt} and \ref{fig: hr1|racecar|pt}, Point and RaceCar display a similar pattern as Section \ref{subsec: exp|nav2d|hasd} where LSD achieves high coverage but low alignment while HaSD provides a better trade-off between metrics and \begin{math}\alpha\end{math}-HaSD generates diverse favourable trade-offs. In \Cref{fig: hr1|car|pt,fig: hr1|doggo|pt,fig: hr1|ant|pt} Car, Doggo, and Ant LSD fails to learn relevant skills, resulting in a low-coverage skill set that does not incur much cost. It is also important to note that the LSD's objective also fails to drive exploration in our method, leading to HaSD methods relying mostly on the alignment objective. We report qualitative results in the Appendix \ref{sec: Apdx|qr}.

%% file: paper/experiments/sections/safety_gymnasium/push.tex
\subsection{Safety Gymnasium: Push-Room}
\label{subsec: Experiment|HR3}
\begin{figure}
\centering
\includegraphics[width=1.0\linewidth]{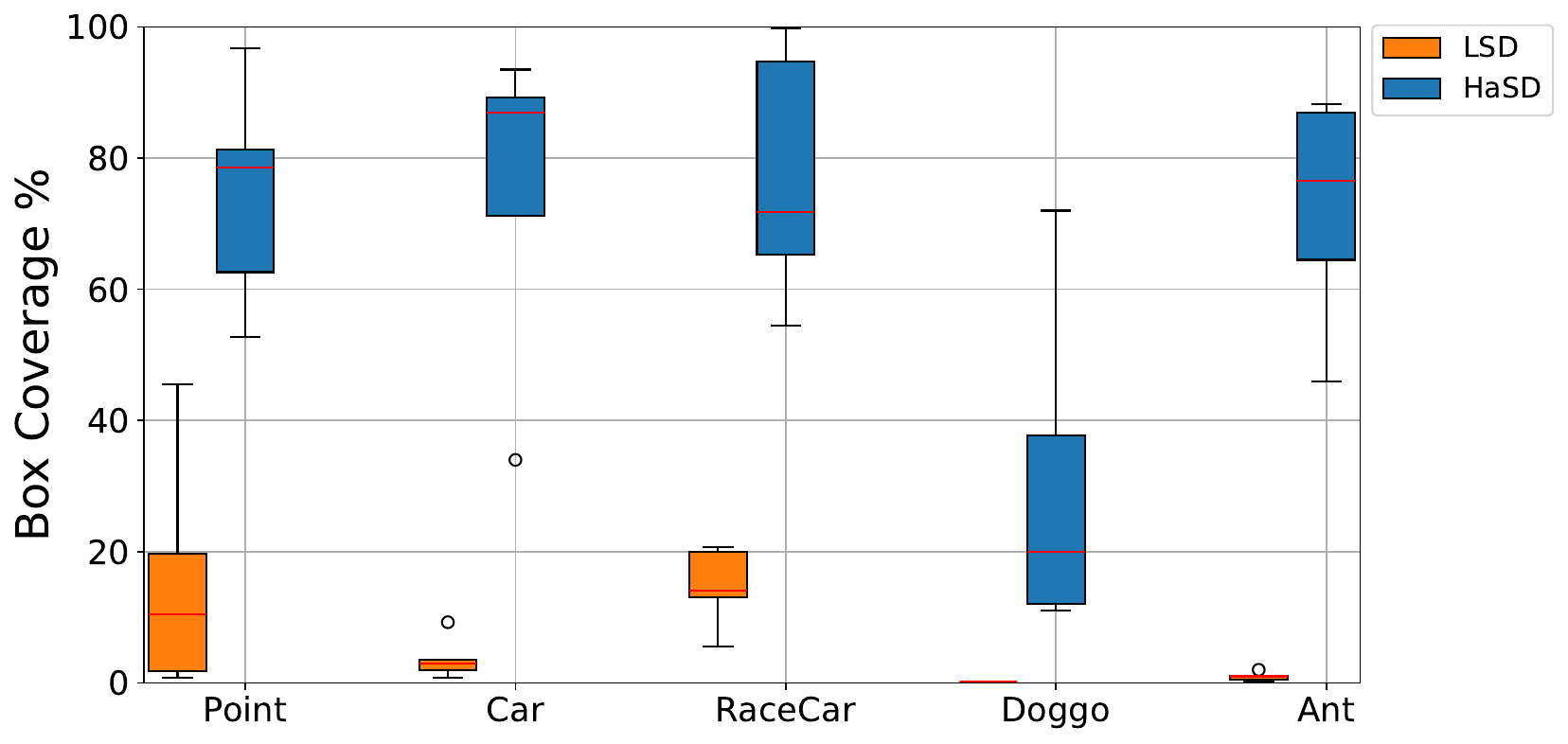} 
\caption{Comparing box-pushing agent's coverage of skill sets discovered by LSD and HaSD with the different robots in the Push-Room environment.}
    \label{fig: exp|hr3|push_box_coverage}
\Description{Comparing box-pushing agent's coverage of skill sets discovered by LSD and HaSD with the different robots in the Push-Room environment.}
\end{figure}

In this section we present results obtain in the Push-Room environment described in Section \ref{subsubsec: Experiment|Environments_Description|HR3}. All results are presented after sampling 1000 skills from the policy across 5 seeds. We report results regarding the box's coverage of the room in Figure \ref{fig: exp|hr3|push_box_coverage} This coverage is measured by the number of 0.1 × 0.1 square bins occupied by the box at least once. Having a high coverage by the box indicates that the agent interacts with the box in a wide range of ways. HaSD obtain better coverage with every agent than LSD. This means that the alignment objective helps indicate desired behaviour while allowing the diversity objective to discover diverse skills. We report qualitative results in the Appendix \ref{sec: Apdx|qr}.

%% file: paper/experiments/sections/Human_Feedback/hyp_sec.tex
\subsection{Sensitivity to Human Feedback}
\label{subsec: experiment|Human|nav2d}
\begin{SCfigure}
\centering
\includegraphics[width=0.5\linewidth]{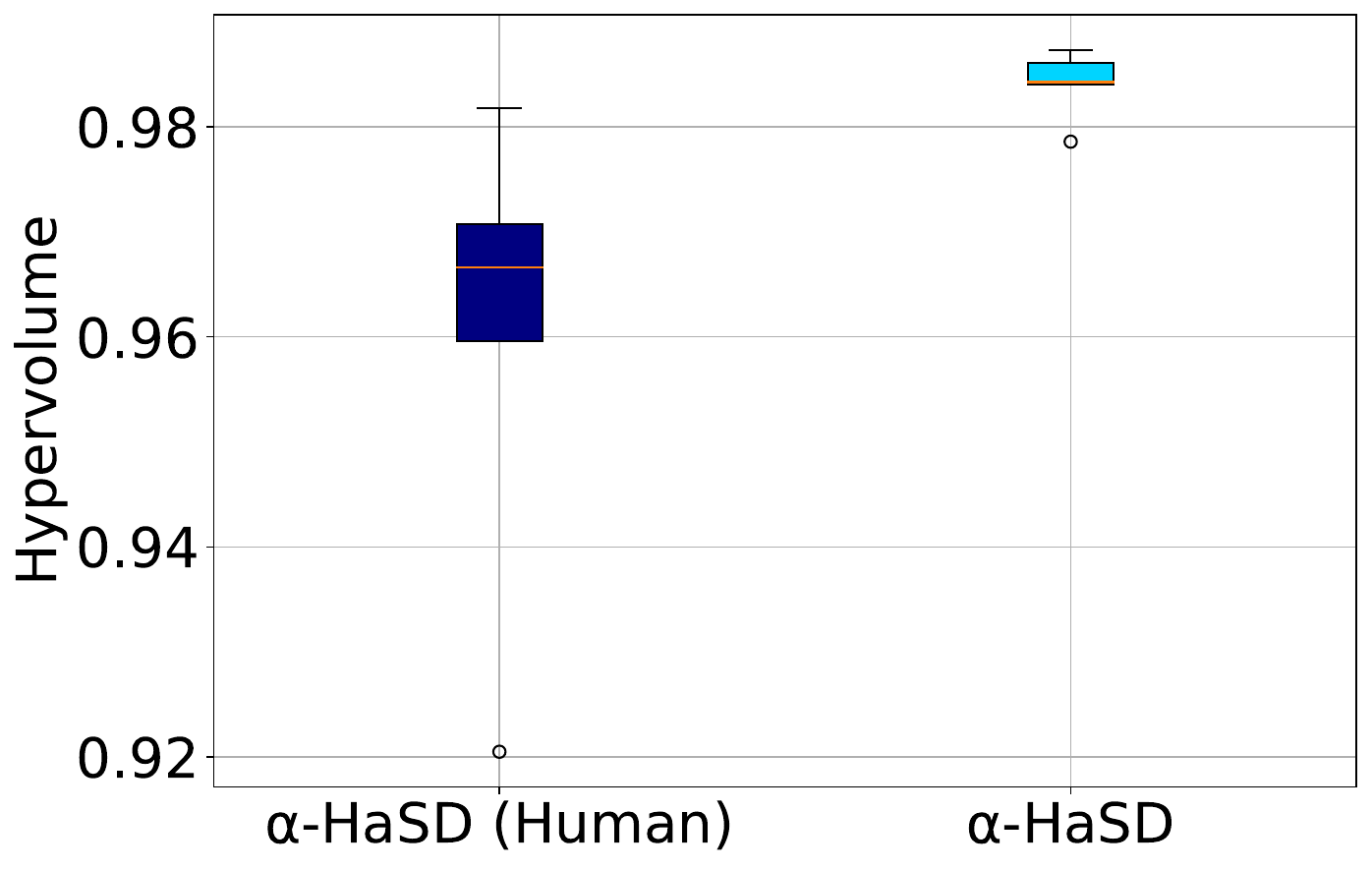} 
\caption{Comparing hypervolume computed for each set of solutions produced by \begin{math} \alpha \end{math}-HaSD with actual human feedback budget and ground truth reward in Nav2d environment.}
\label{fig: exp|real|nav2d|hpv}
\Description{Comparing hypervolume computed for each set of solutions produced by \begin{math} \alpha \end{math}-HaSD with actual human feedback budget and ground truth reward in Nav2d environment.}
\end{SCfigure}

In this section, we analysed the sensitivity of \begin{math} \alpha \end{math}-HaSD to real human feedback.  For practical reasons, we train offline a reward model from actual human feedback subjects (the authors) familiar with the task. This reward is then used to simulate human preferences during training.  The reward model was trained with 1400 feedback which took 1h to collect. We provide more detail in Appendix \ref{subsec: apdx|collecting_human_preferences}. We show in Figure \ref{fig: exp|real|nav2d|hpv} the hypervolume obtained with \begin{math} \alpha \end{math}-HaSD(Human) which used the real human process presented. The hypervolume indicates that \begin{math} \alpha \end{math}-HaSD(Human) can obtain similar results with actual human than ground truth reward. However we can observe that \begin{math} \alpha \end{math}-HaSD(Human). This is largely due to the noise inherent in the process. This method is inherently noisier than scenarios where preferences are directly provided during training. The accumulation of approximation errors while learning the reward offline can contribute to this noise, and these errors may carry over to the use of this reward as simulated human input during \begin{math} \alpha \end{math}-HaSD training. Other than that we are able to obtain similar results from actual human feedback.

%% file: paper/sections/conclusion.tex
In this work, we introduced a novel approach to address under-constrained skill discovery. Our proposed approach, Human aligned skill discovery, is based on the idea of optimising both a skill discovery objective and an alignment objective. Using these objectives, we explicitly designed HaSD, an approach to learn diverse skills that align with human preference over their entire trajectory. With different alignment objectives, we empirically demonstrated that HaSD learned diverse skills that align with human preferences in various navigation and box-pushing environments. Additionally, we showed that we can condition skills on the diversity alignment trade-off variable to produce a range of skills relevant to different diversity-alignment trade-offs.
One of the inherent limitations of this work is that the degree of alignment depends on the accuracy of the reward model. This can heavily depend on the number of feedback available.  Although this is a common limitation of HIL-methods, it is exacerbated in USD settings due to their inherent long training time, sample inefficiency and the size of the behaviour space. In these circumstances, the questions of when to seek human feedback and what queries to select to maximise information gain require more sophisticated methods. In future work, we will investigate how to leverage the diversity and amount of interactions generated by the skill discovery objective to increase reward model accuracy with low feedback budgets in USD settings.

%% file: paper/appendix/sections/appendix.tex
\appendix
\section{Appendix}
\input{paper/appendix/sections/implementation}

\input{paper/appendix/sections/conflicting_reward}
\input{paper/appendix/sections/inter_extra-polation}
\input{paper/appendix/sections/collecting_human_preferences}
\input{paper/appendix/sections/societal_impact}

\subsection{Safety Gymnasium Tasks}
\label{subsec: apdx|sg_tasks}
\begin{figure}[htbp]
    \begin{subfigure}[b]{0.15\textwidth}
        \centering
        \includegraphics[width=\textwidth]{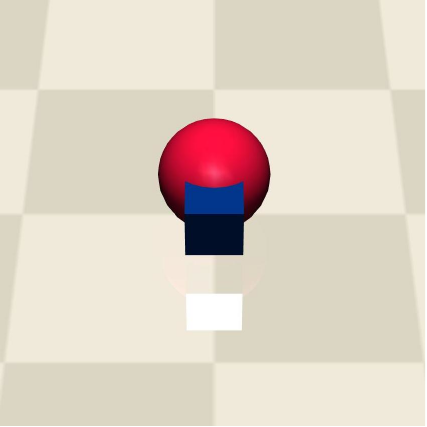} 
        \caption{Point}
    \end{subfigure}
    \hfill
    \begin{subfigure}[b]{0.15\textwidth}
        \centering
        \includegraphics[width=\textwidth]{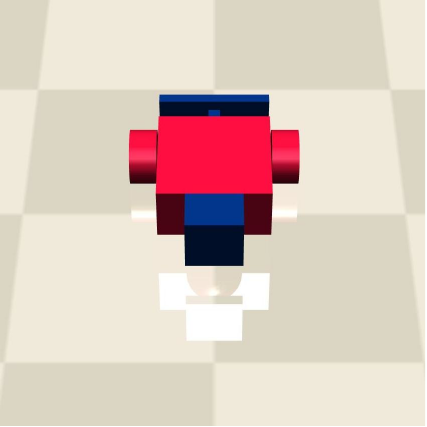} 
        \caption{Car}
    \end{subfigure}
    \hfill
        \begin{subfigure}[b]{0.15\textwidth}
        \centering
        \includegraphics[width=\textwidth]{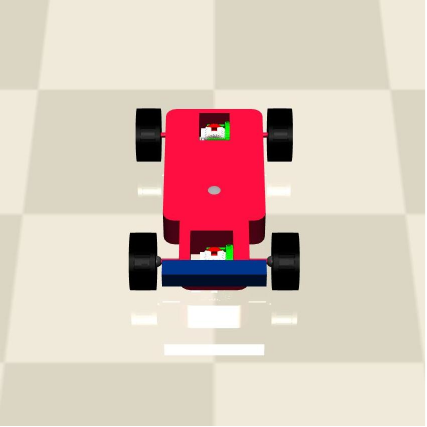} 
        \caption{Racecar}
    \end{subfigure} 
    \hfill 
    \begin{subfigure}[b]{0.15\textwidth}
        \centering
        \includegraphics[width=\textwidth]{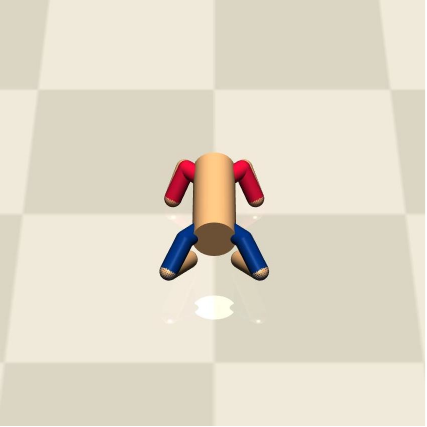} 
        \caption{Doggo}
    \end{subfigure}
    \hfill
    \begin{subfigure}[b]{0.15\textwidth}
        \centering
        \includegraphics[width=\textwidth]{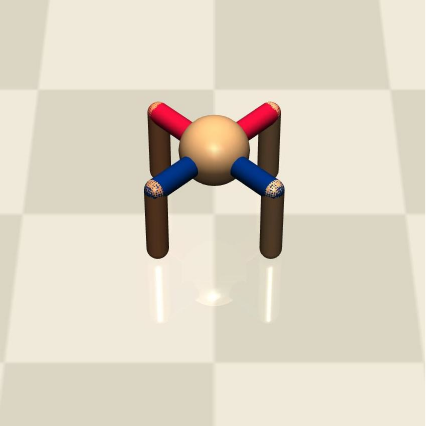} 
        \caption{Ant}
    \end{subfigure}
    \hfill
    \caption{Range of agents used to evaluate our method on safety gymnasium.}
    \label{fig: safety_gymnasium agents}
    \Description{Range of agents used to evaluate our method on safety gymnasium.}
\end{figure}
\begin{figure}[htbp]
    \begin{subfigure}[b]{0.45\textwidth}
        \centering
       \includegraphics[width=\linewidth]{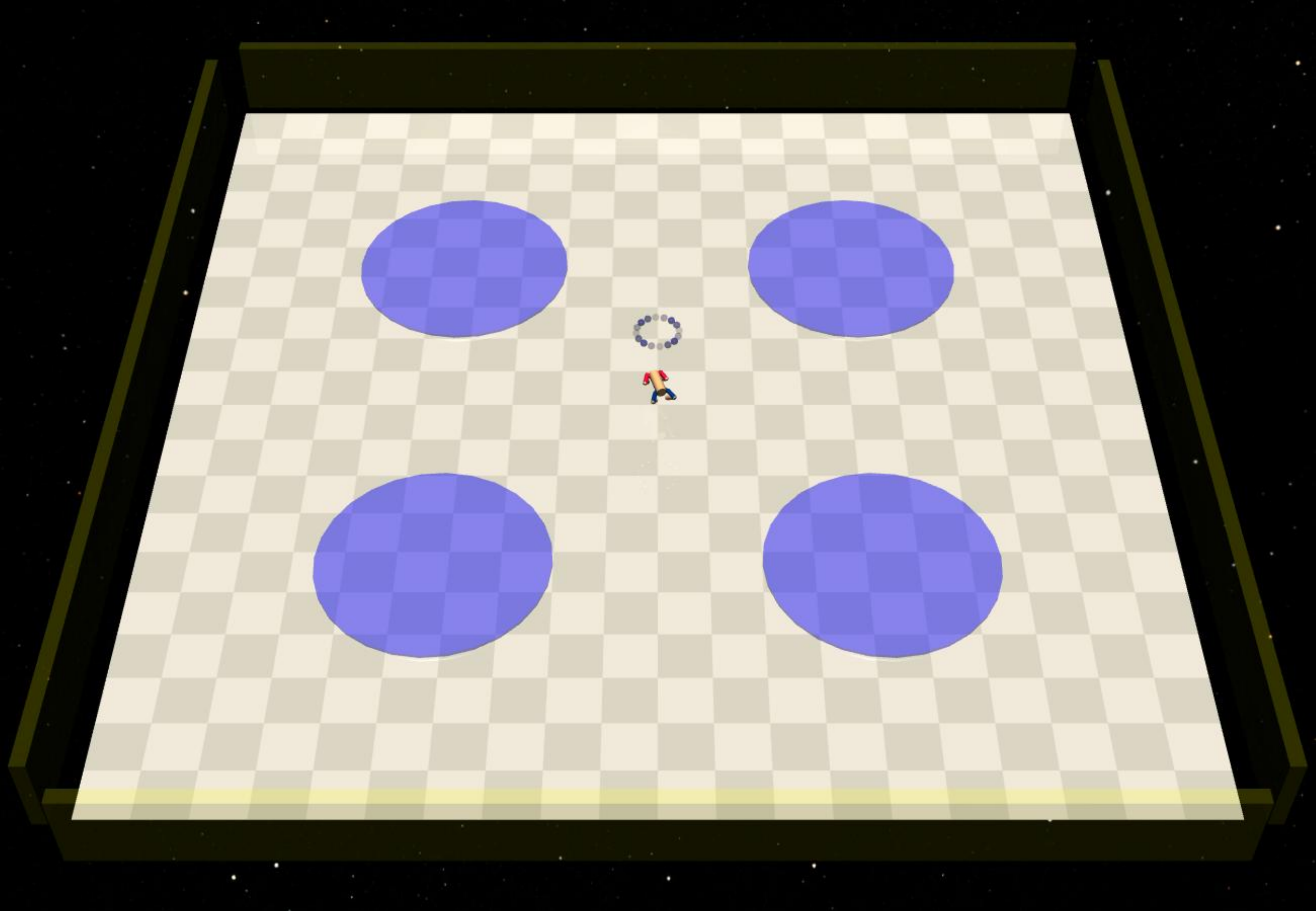} 
        \caption{Hazardous-Room}
        \label{fig: Hazardous-Room environment}
    \end{subfigure}
    \hfill
    \begin{subfigure}[b]{0.45\textwidth}
        \centering
        \includegraphics[width=\linewidth]{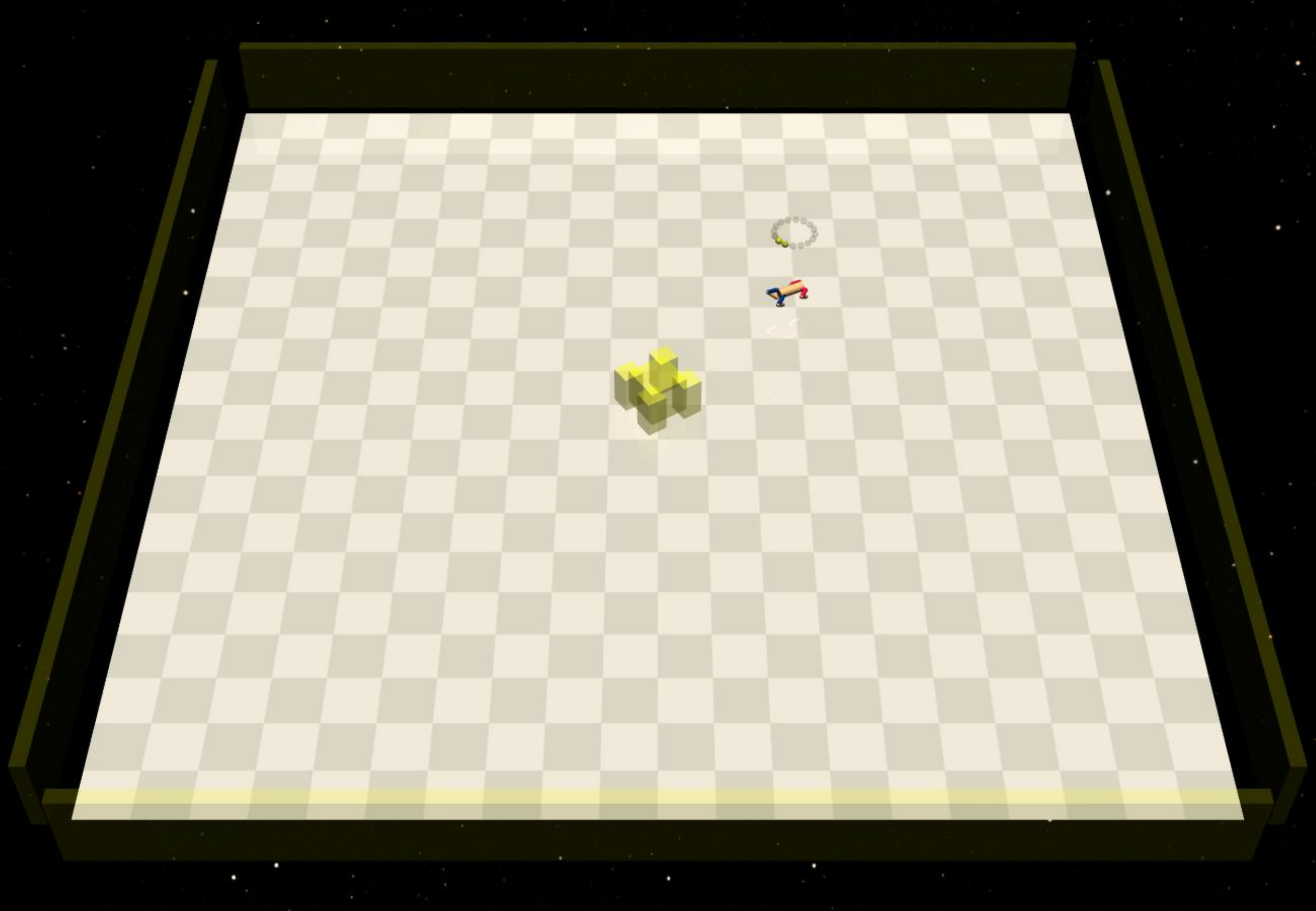} 
        \caption{Push-Room}
        \label{fig: Push-Room environment}
    \end{subfigure}
    \caption{(a) In the Hazardous-Room environment, the agent should ideally avoid the purple circles while navigating. (b) In the Push-Room environment, the agent should ideally move the yellow box around the room.}
    \label{fig: safety_gymnasium  environment}
    \Description{(a) In the Hazardous-Room environment, the agent should ideally avoid the purple circles while navigating. (b) In the Push-Room environment, the agent should ideally move the yellow box around the room.}
\end{figure}

\input{paper/appendix/sections/qualitative_results}

%% file: paper/appendix/sections/implementation.tex
\subsection{Implementation details}
\label{subsec: apdx|implementation_details}

\paragraph{Environments.} We used a (\href{https://github.com/shufflebyte/gym-nav2d}{custom 2D Navigation}) publicly released repository to experiment in simpler navigation environments. We used the customisable capabilities that Safety-Gymnasium \citep{ji2023safety} offers to design the Hazard-Room and Push-Room environments. 

\paragraph{Hazard-Room environment.}
In the Hazard-Room environment, lidar information disrupts the LSD objective, resulting in non-dynamic and non-diverse skills. To improve performance, we mask lidar information from the \begin{math}\phi\end{math} function.

\paragraph{DownStream Task}
In the experiment in section \ref{subsec: exp|nav2d|ft}, the skills discovered are trained in a slightly modified nav2d where the agent starts in a random position within the environment. This adjustment aims to mitigate potential issues arising from changes in the initial state distribution of the skills. By ensuring that the agent experiences a variety of initial conditions, we can more accurately combines skills. We then train a SAC-TQC on top of the skills discover, we report hyperparameter in Table \ref{apdx: ft-table}.

\paragraph{Algorithm.} We implement HaSD based on the publicly available LSD codebase \cite{park2021lipschitz} and PEBBLE\cite{2021pebble}. As our backbone reinforcement learning algorithm, we use the SAC implementation from the publicly available codebase cleanRL\cite{huang2022cleanrl} and a publicly available codebase for (\href{https://github.com/SamsungLabs/tqc_pytorch?tab=readme-ov-file}{TQC}). For all experiment with HaSD we select \begin{math} \alpha=0.2\end{math} and for \begin{math} \alpha\end{math}-HaSD we sample \begin{math} \alpha\end{math} uniformly from \begin{math} \mathrm{A}=\{1,0.5,0.2,0.1,0\}\end{math}. We report other hyperparameters used in Tables \ref{sac-table}, \ref{lsd-table} and  \ref{rlhf-table}:
\paragraph{Resources.} All experiments are conducted on an Ubuntu 20.4 server with 36 cores CPU, 767GB RAM, and a V100 32GB GPU with CUDA version 12.0, each run in Sections \ref{subsec: Experiment|HR1} and \ref{subsec: Experiment|HR3} took  24hours (point, Car, Racecar) and 72hours (Doggo, Ant).

\begin{table}[htbp]
  \caption{Hyperparameter SAC-TQC}
  \label{sac-table}
  \centering
  \begin{tabular}{ll}              \\
    \toprule
    Hyperparameter     & Value    \\
    \midrule
    Training iteration          & 1M(Nav2d), 5M(Point, Car, Racecar), 10M(Doggo, Ant)       \\
    Learning rate critic       & \begin{math}3.0 \times 10^{-04}\end{math} (Nav2d), \begin{math}1.0\times 10^{-04}\end{math} (Safety-Gymnasium)       \\
    Learning rate actor         & \begin{math}1.0 \times 10^{-04}\end{math} (Nav2d), \begin{math}3.0\times 10^{-05}\end{math} (Safety-Gymnasium)      \\
    Update policy frequency     & 2       \\
    Update-to-data              & 4       \\
    Optimiser                   &  Adam \\
    Minibatch size              &  256         \\
    Discount factor $\gamma$    &  0.99         \\
    Replay buffer size          &  \begin{math}10^6\end{math}         \\
    Hidden layers               &  2        \\
    Hidden units per layers     & 256(Nav2d), 512(Safety-Gymnasium)         \\
    Target network smoothing coefficient      &  0.995        \\
    Entropy coefficent          &  auto-adjust \cite{pmlr-v80-haarnoja18b}         \\
    Number of quantiles: & 25\\
    Number of networks: & 3\\
    Number of top quantiles to drop: &2\\
    \bottomrule
  \end{tabular}
\end{table}

\begin{table}[htbp]
  \caption{Hyperparameter LSD}
  \label{lsd-table}
  \centering
  \begin{tabular}{ll}              \\
    \toprule
    Hyperparameter     & Value    \\
    \midrule
    Learning rate critic       & \begin{math}3.0 \times 10^{-04}\end{math} (Nav2d), \begin{math}1.0\times 10^{-04}\end{math} (Safety-Gymnasium)       \\
    Hidden layers      &  2        \\
    Hidden units per layers      & 256(Nav2d), 512(Safety-Gymnasium)         \\
    Z dim  & 2 \\
    LSD $\epsilon $     &  \begin{math}1.0\times 10^{-06}\end{math}         \\
    LSD initial \begin{math}\lambda\end{math}      &  3000         \\
    \bottomrule
  \end{tabular}
\end{table}

\begin{table}[htbp]
  \caption{Hyperparameter RLHF}
  \label{rlhf-table}
  \centering
  \begin{tabular}{ll}              \\
    \toprule
    Hyperparameter     & Value    \\
    \midrule
    Learning rate       & \begin{math}3.0\times 10^{-4}\end{math}      \\
    Optimizer           &  Adam \\
    Minibatch size      &  128(Nav2d), 256 (Safety-Gymnasium)         \\
    Ensemble size     &  3         \\
    Size segment     &   25(Nav2d), 50 (Safety-Gymnasium)         \\
    Sampling mode     &  Uniform       \\
    Queries per feedback session       &  128(Nav2d), 280(Safety-Gymnasium)        \\
    Number of feedback session       &  10        \\
    Frequency of feedback session       &  12K (Nav2d), 50K(Safety-Gymnasium)        \\
    Start feedback      &  30K (Nav2d), 150K(Safety-Gymnasium)    \\
    \bottomrule
  \end{tabular}
\end{table}

\begin{table}[htbp]
  \caption{Hyperparameter SAC-TQC Downstream-Task}
  \label{apdx: ft-table}
  \centering
  \begin{tabular}{ll}              \\
    \toprule
    Hyperparameter     & Value    \\
    \midrule
    Training iteration          & 500k(Nav2d)     \\
    Learning rate critic       & \begin{math}3.0 \times 10^{-03}\end{math} (Nav2d)      \\
    Learning rate actor         & \begin{math}1.0 \times 10^{-03}\end{math} (Nav2d)  \\
    Update policy frequency     & 2       \\
    Update-to-data              & 1       \\
    Optimiser                   &  Adam \\
    Minibatch size              &  256         \\
    Discount factor $\gamma$    &  0.99         \\
    Replay buffer size          &  \begin{math}10^6\end{math}         \\
    Hidden layers               &  2        \\
    Hidden units per layers     & 256(Nav2d) \\
    Target network smoothing coefficient      &  0.995        \\
    Entropy coefficent          &  auto-adjust \cite{pmlr-v80-haarnoja18b}         \\
    Number of quantiles: & 25\\
    Number of networks: & 3\\
    Number of top quantiles to drop: &2\\
    \bottomrule
  \end{tabular}
\end{table}

\subsection{Ground Truth rewards}
\label{subsec: appendix|gt}
\subsubsection{Nav2d}
The following ground truth reward function is designed to mimic the following preference `\textit{Trajectories that travel as far as possible from the initial position while avoiding unsafe regions should be preferred}':
\begin{equation}
    r^{\text{Ground Truth}} = \left\lVert a^t_{xy} - a^0_{xy}\right\rVert +\left\lVert a^t_{xy} - a^{t-1}_{xy}\right\rVert -\mathds {1}[a^t_{xy} \in \textit{hazardous areas}] 
\end{equation} 
where \begin{math}a^t_{xy}\end{math} is the agent position in the Cartesian coordinates at time $t$.

\subsubsection{Hazard-Room}
The following ground truth reward function is designed to mimic the following preference `\textit{Trajectories that travel as far as possible from the initial position while avoiding unsafe regions and staying in the enclosed area  should be preferred }':
\begin{equation}
  r^{\text{Ground Truth}} = \left\lVert a^t_{xy} - a^0_{xy}\right\rVert  -100\times\mathds {1}[a^t_{xy} \in \textit{hazardous areas}]  -10\times\mathds {1}[a^t_{xy} \in \textit{Passing through a wall}] 
\end{equation}
where \begin{math}a^t_{xy}\end{math} is the agent position in the Cartesian coordinates at time $t$.

\subsubsection{Push-Room}
The following ground truth reward function is designed to mimic the following preference `\textit{Trajectories that make the box travel as far as possible from its initial position should be preferred }':
\begin{equation}
\begin{aligned}
   r^{\text{Ground Truth}} &= r^{\text{to box}} + r^{\text{from box}} - \mathds {1}[a^t_{xy} \in \textit{Passing through a wall}] 
\end{aligned}
\end{equation}
where :
\begin{equation}
\begin{aligned}
\label{eq: r_to_box}
r^{\text{to box}} = \left\lVert b^t_{xy} - a^t_{xy}\right\rVert  \quad\text{if}\quad   \left\lVert b^{t-1}_{xy} - a^{t-1}_{xy}\right\rVert \geq \alpha_{ba} 
\end{aligned}
\end{equation}

\begin{equation}
\begin{aligned}
\label{eq: r_from_box}
r^{\text{from box}} = \left\lVert b^t_{xy} - b^0_{xy}\right\rVert   \quad\text{if}\quad  \left\lVert b^{t}_{xy} - b^{t-1}_{xy}\right\rVert \geq \alpha_{bb'}
\end{aligned}
\end{equation}
where \begin{math}a^t_{xy}\end{math} and  \begin{math}b^t_{xy}\end{math} is respectively the agent position and the box position in the Cartesian coordinates at time \begin{math}t\end{math}.

%% file: paper/appendix/sections/conflicting_reward.tex
\subsection{Conflicting Objectives}
\label{subsec: apdx|conflicting_objectives}
In settings where both rewards conflict, the agent might give up a certain degree of diversity to follow human preferences, or give up a degree of human preferences in order to discover novel skills. This trade-off is inherent to multi-objective problems, which is why we propose to learn the trade-off with \begin{math}\alpha\end{math}-HaSD, enabling the user to choose from multiple solutions at the end. Both rewards are non-orthogonal in the safety experiments, since the skill discovery reward encourages crossing unsafe regions while the preferences reward penalises it. In these settings \Cref{fig: 0-HaSD,fig: 0.1-HaSD,fig: 0.2-HaSD,fig: 0.5-HaSD,fig: 1-HaSD} from the paper shows how \begin{math}\alpha\end{math} can deal with this situation.

\begin{figure}[htbp]
    \centering
    \includegraphics[width=0.23\linewidth]{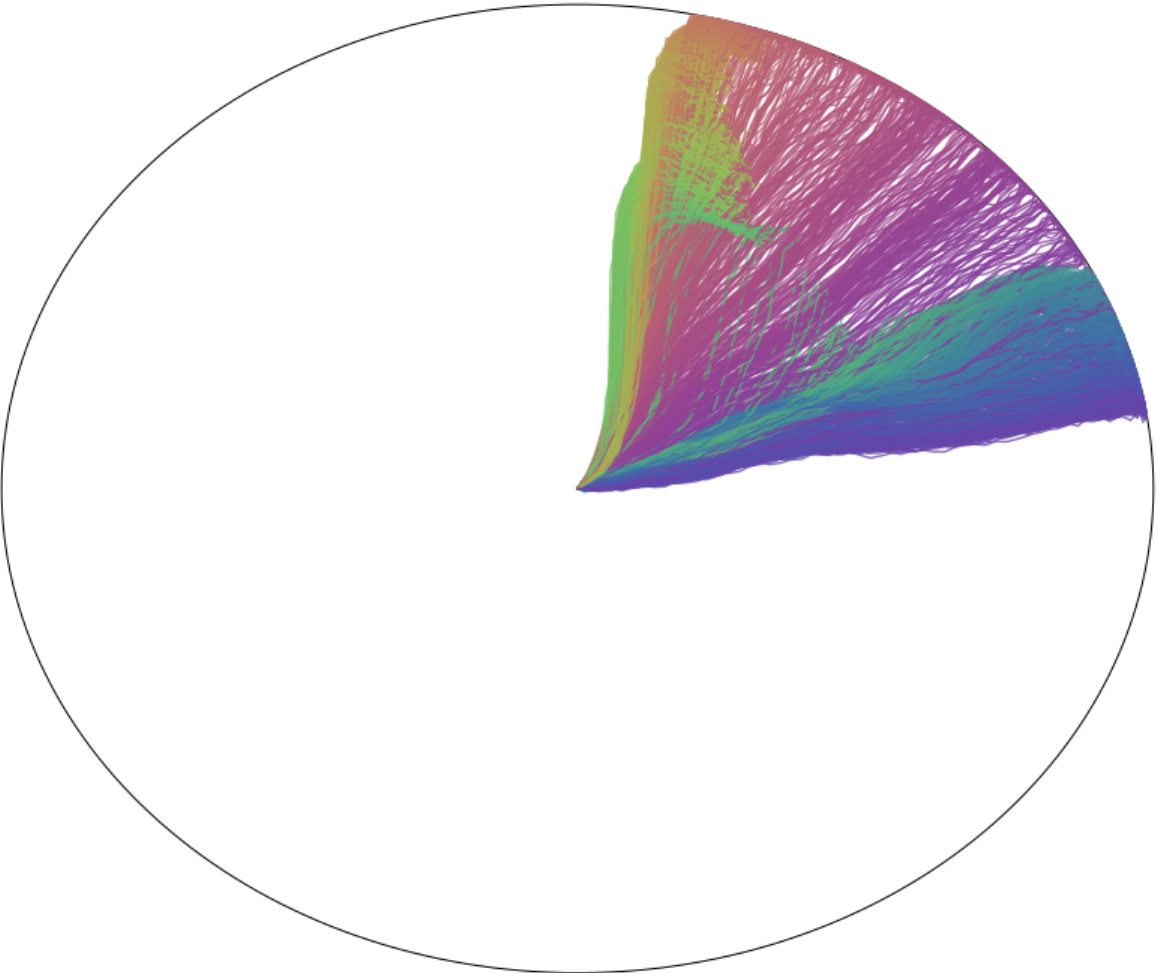} 
    \caption{Skill sets obtained with preference for covering the north-east regions.}
    \label{fig: apdx|conflicting_objectives|N-E}
\end{figure}

\begin{figure}[htbp]
    \centering
    \begin{subfigure}[b]{0.23\textwidth}
        \centering
        \includegraphics[width=\textwidth]{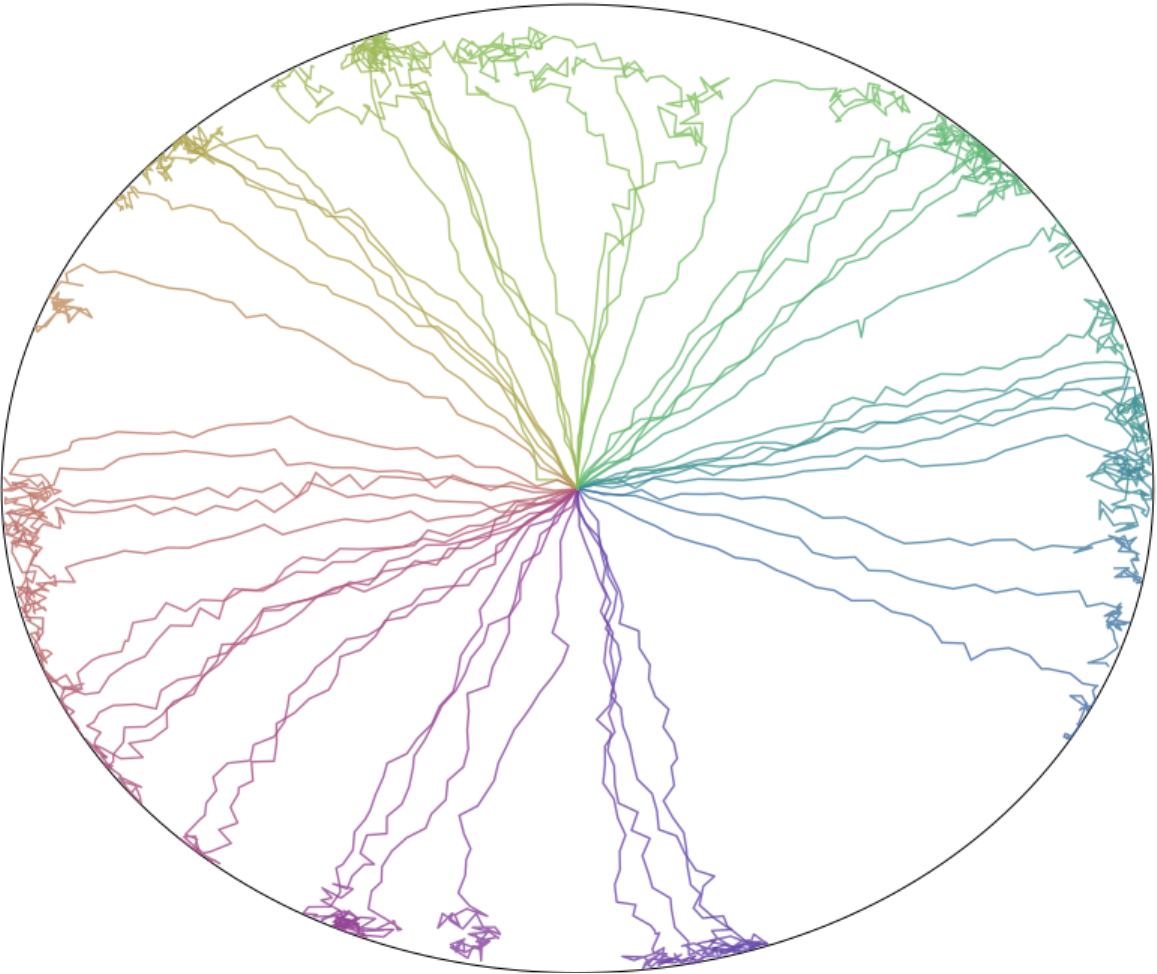} 
        \caption{0.0-HaSD}
         \label{fig:apdx|conflicting_objectives|L|0.0}
    \end{subfigure}
    \hfill
    \begin{subfigure}[b]{0.23\textwidth}
        \centering
        \includegraphics[width=\textwidth]{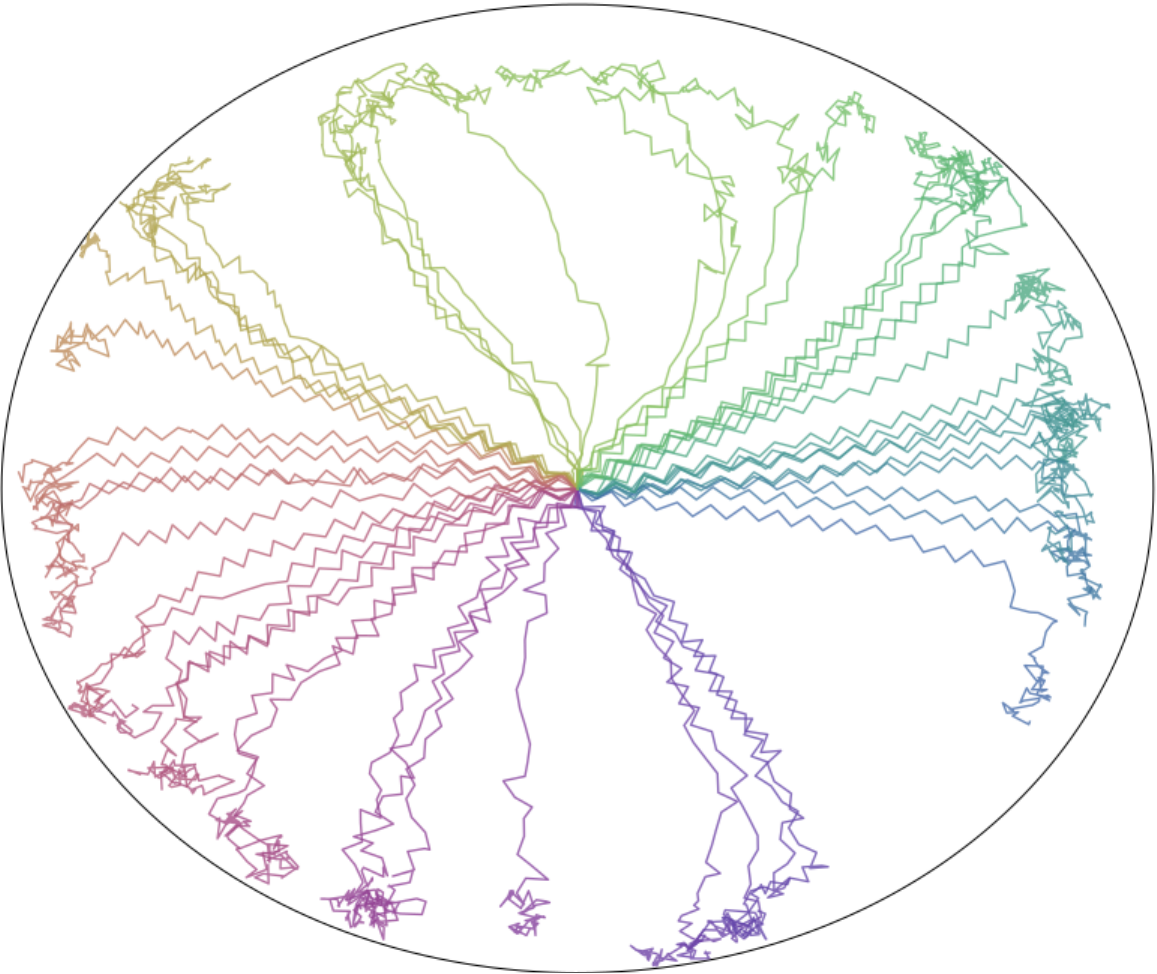} 
        \caption{0.2-HaSD}
         \label{fig: apdx|conflicting_objectives|L|0.2}
    \end{subfigure}
    \hfill
    \begin{subfigure}[b]{0.23\textwidth}
        \centering
        \includegraphics[width=\textwidth]{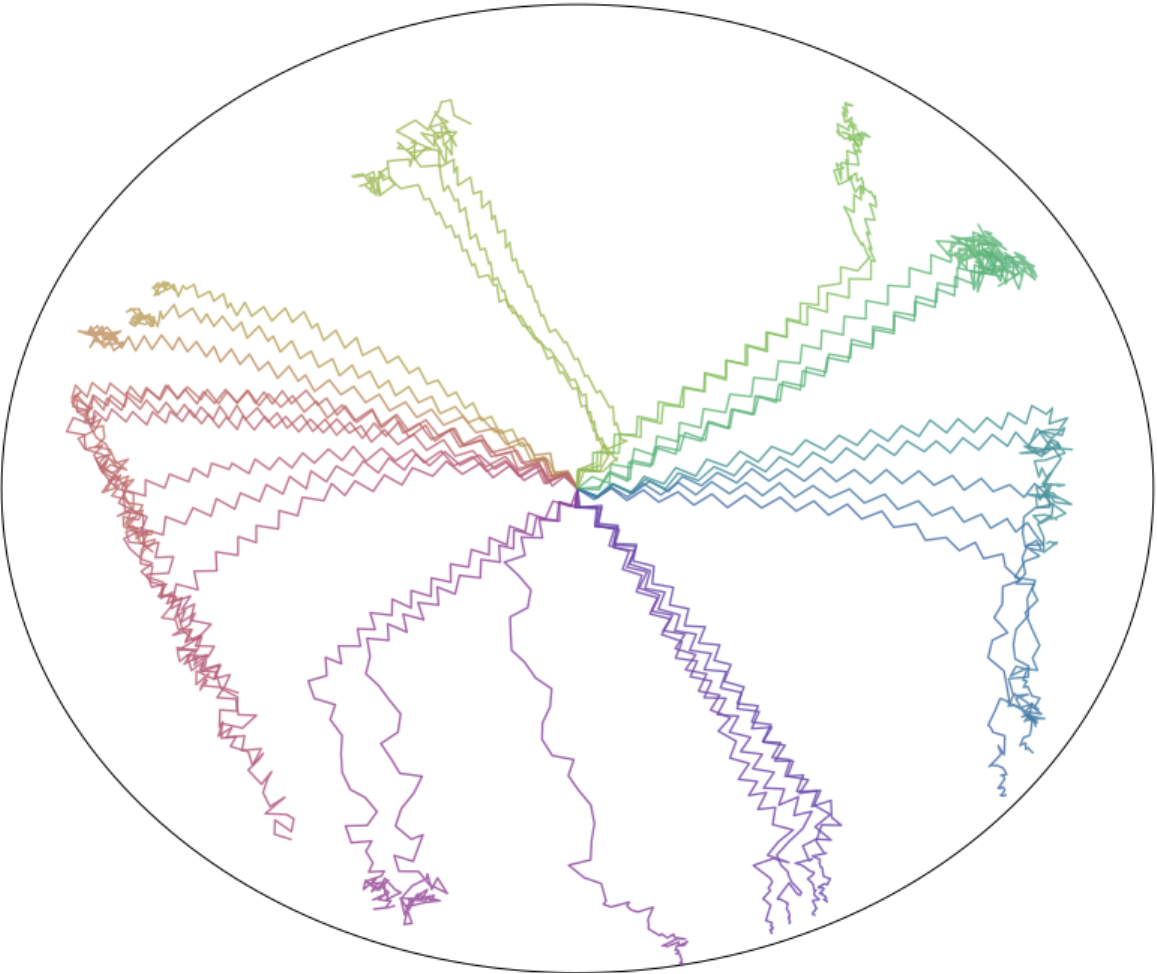} 
        \caption{0.5-HaSD}
         \label{fig: apdx|conflicting_objectives|L|0.5}
    \end{subfigure}
    \caption{Skill sets obtained with preference for L-shape trajectories.}
    \label{fig: apdx|conflicting_objectives|L}
\end{figure}

In this section we provide additional experiments to show that our method can manage conflicting objectives in different settings. To this end, we introduce two conflicting settings by changing the preferences to 'Trajectories travelling in the North-East region as far as possible from the initial position should be preferred' and 'Trajectories that travel in an L-shaped should be preferred'. Both preferences are mimicked by the rewards in Equation \ref{fig: apdx|conflicting_objectives|N-E|reward} and in Equation \ref{fig: apdx|conflicting_objectives|L|reward}. For the L-shaped, reward we specifically tried to enforce a 90 degree angle between the last 3 agent positions. As the state space in the 2D navigation is the agent position, we had to stack the 3 last states for the policy to learn behaviour and the reward model to learn the preference. Figure \ref{fig: apdx|conflicting_objectives|N-E} and \ref{fig: apdx|conflicting_objectives|L} shows that even in those settings, we are able to learn a skill set covering only the North-East region in the first settings and to learn L-shaped looking skills in the second setting.

The following ground truth reward function is designed to mimic the following preference `\textit{Trajectories that travel in the North-East as far as possible from the initial}':

\begin{eqnarray}
\label{fig: apdx|conflicting_objectives|N-E|reward}
    r =\begin{cases}
			 \left\lVert a^t_{xy} - a^0_{xy}\right\rVert , & \text{if $a^t_{xy} \in \textit{North-East} $}\\
            -1, & \text{otherwise}
	\end{cases}
\end{eqnarray}
where \begin{math}a^t_{xy}\end{math} is the agent position in the Cartesian coordinates at time $t$.

The following ground truth reward function is designed to mimic the following preference `\textit{Trajectories that travel in the North-East as far as possible from the initial}':
\begin{eqnarray}
\label{fig: apdx|conflicting_objectives|L|reward}
    r = 1 - \left\lVert \theta^t - 90\right\rVert \quad\textrm{}\quad \theta^t = \arccos(\frac{\Delta a^{t''t'}_{xy} \Delta a^{t't}_{xy}}{\left\lVert \Delta a^{t''t'}_{xy}\right\rVert \left\lVert  \Delta a^{t't}_{xy} \right\rVert  })  
\end{eqnarray}
where \begin{math}a^t_{xy}\end{math} is the agent position in the Cartesian coordinates at time $t$.

%% file: paper/appendix/sections/inter_extra-polation.tex
\subsection{$\alpha$-Generalisation}
\label{subsec: apdx|a_generalisation}
\begin{figure}[htbp]
    \centering
    \begin{subfigure}[b]{0.23\textwidth}
        \centering
        \includegraphics[width=\textwidth]{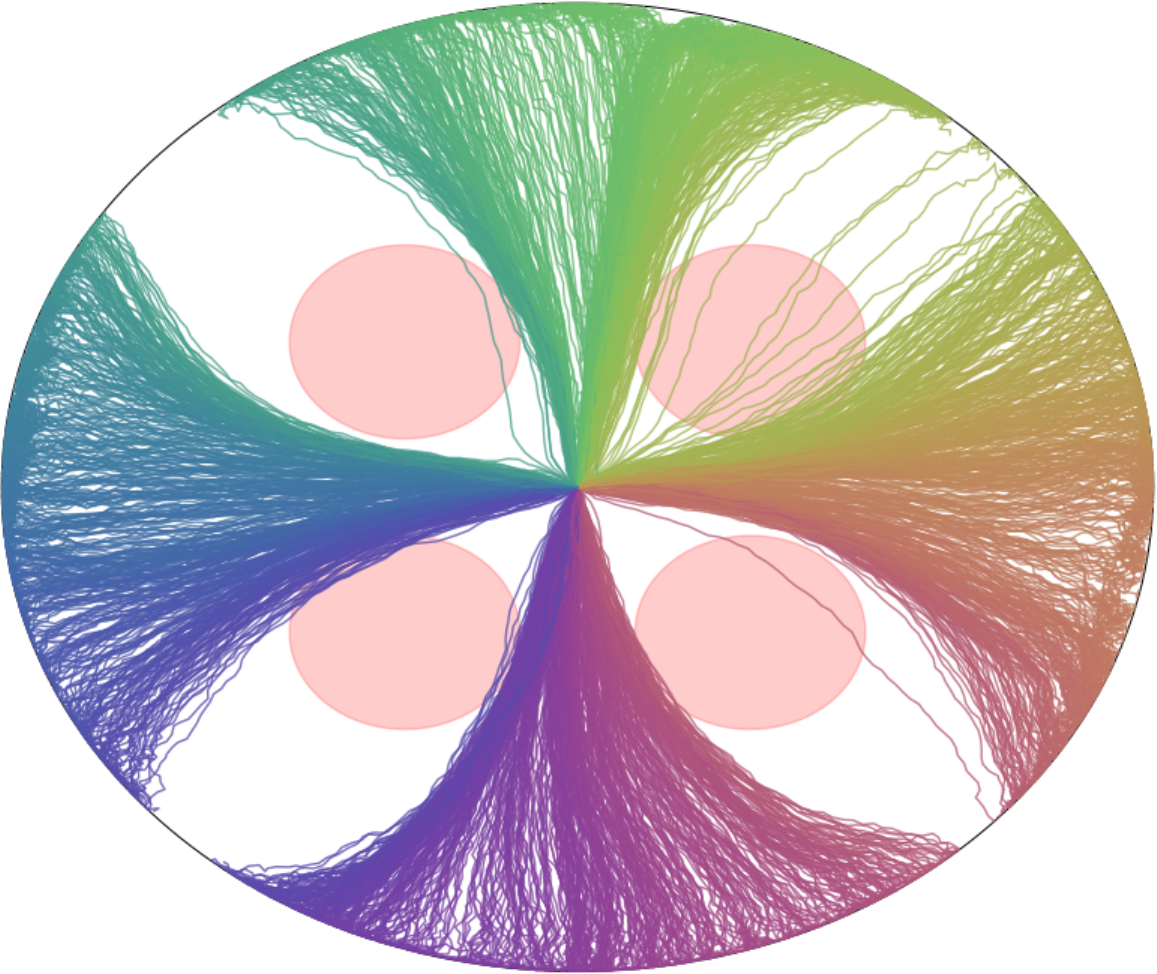} 
        \caption{0.4-HaSD}
         \label{fig:apdx|a_generalisation|0.4}
    \end{subfigure}
    \hfill
    \begin{subfigure}[b]{0.23\textwidth}
        \centering
        \includegraphics[width=\textwidth]{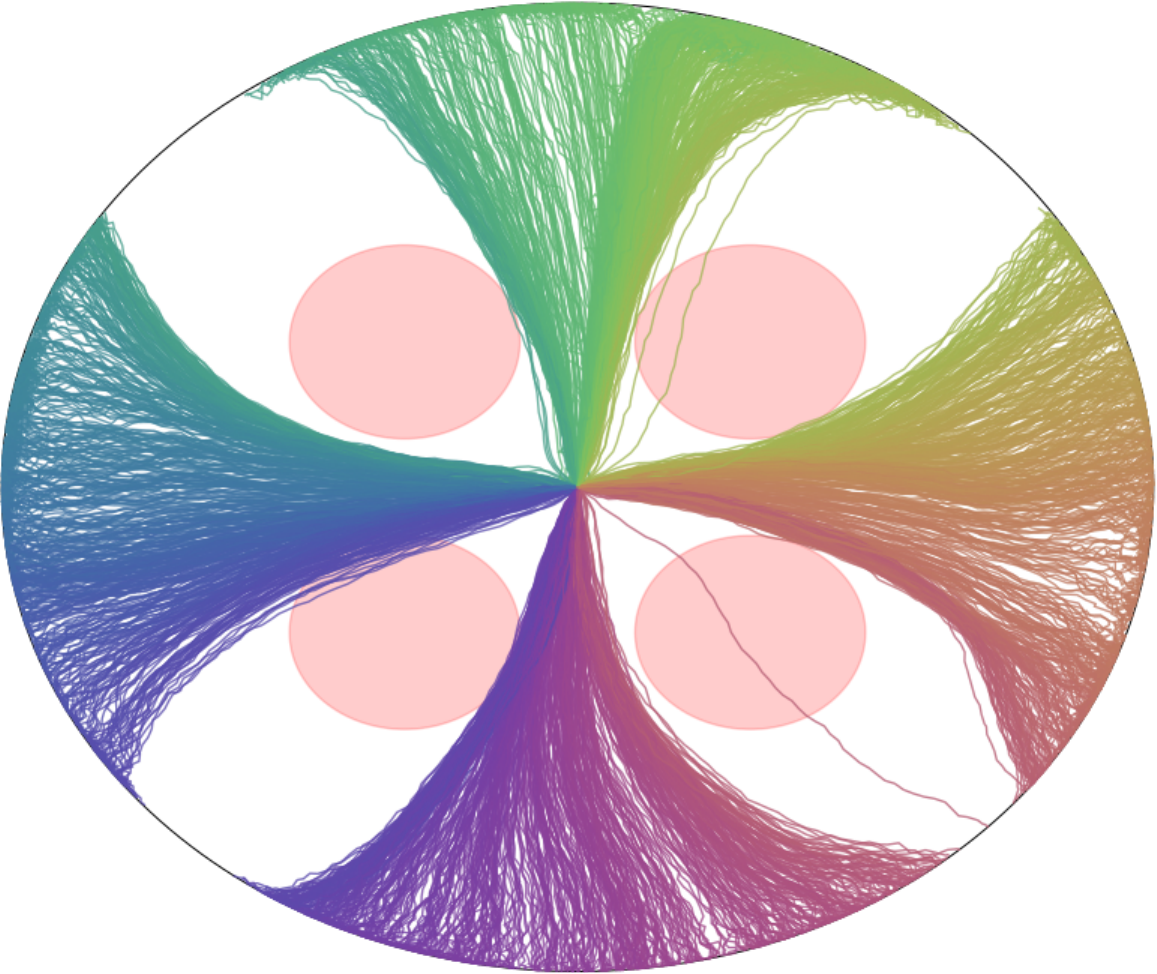} 
        \caption{0.7-HaSD}
         \label{fig: apdx|a_generalisation|0.7}
    \end{subfigure}
    \hfill
    \begin{subfigure}[b]{0.23\textwidth}
        \centering
        \includegraphics[width=\textwidth]{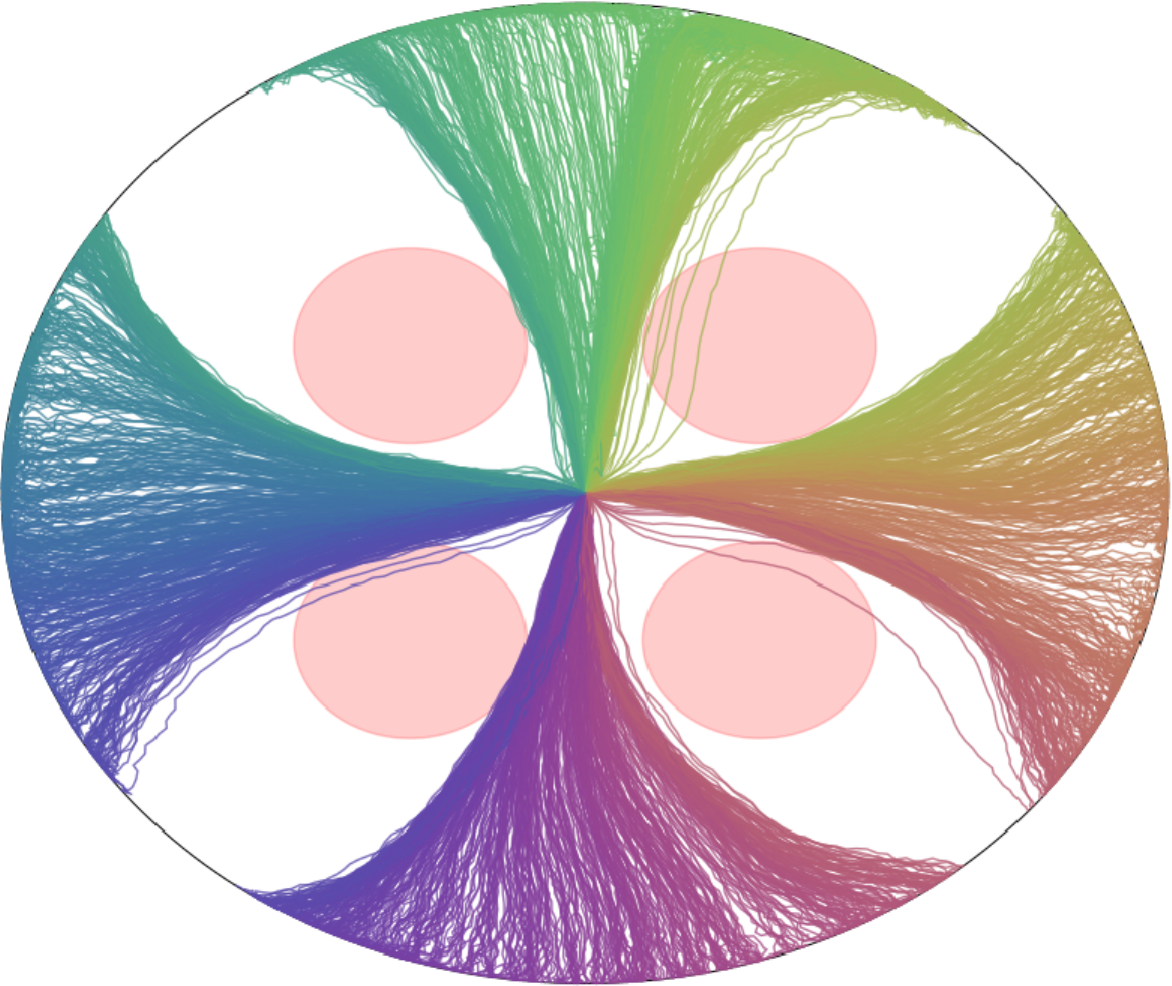} 
        \caption{0.8-HaSD}
         \label{fig: apdx|a_generalisation|0.8}
    \end{subfigure}
    \caption{ \begin{math}\alpha\end{math}-HaSD skills through interpolation.}
    \label{fig: apdx|a_generalisation|interpolation}
\end{figure}

\begin{figure}[htbp]
    \centering
    \begin{subfigure}[b]{0.23\textwidth}
        \centering
        \includegraphics[width=\textwidth]{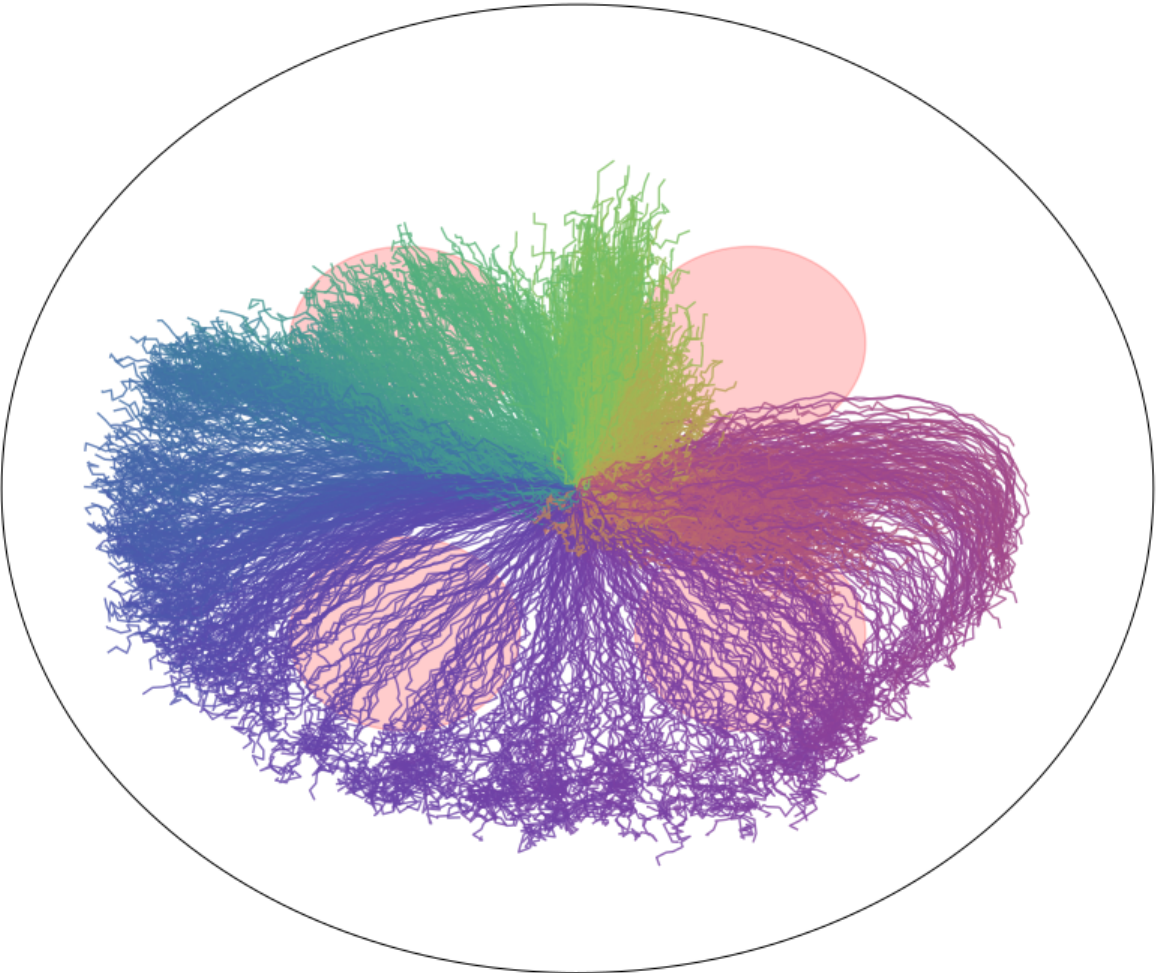} 
        \caption{-1.0-HaSD}
         \label{fig:apdx|a_generalisation|extrapolation|-1.0}
    \end{subfigure}
    \hfill
    \begin{subfigure}[b]{0.23\textwidth}
        \centering
        \includegraphics[width=\textwidth]{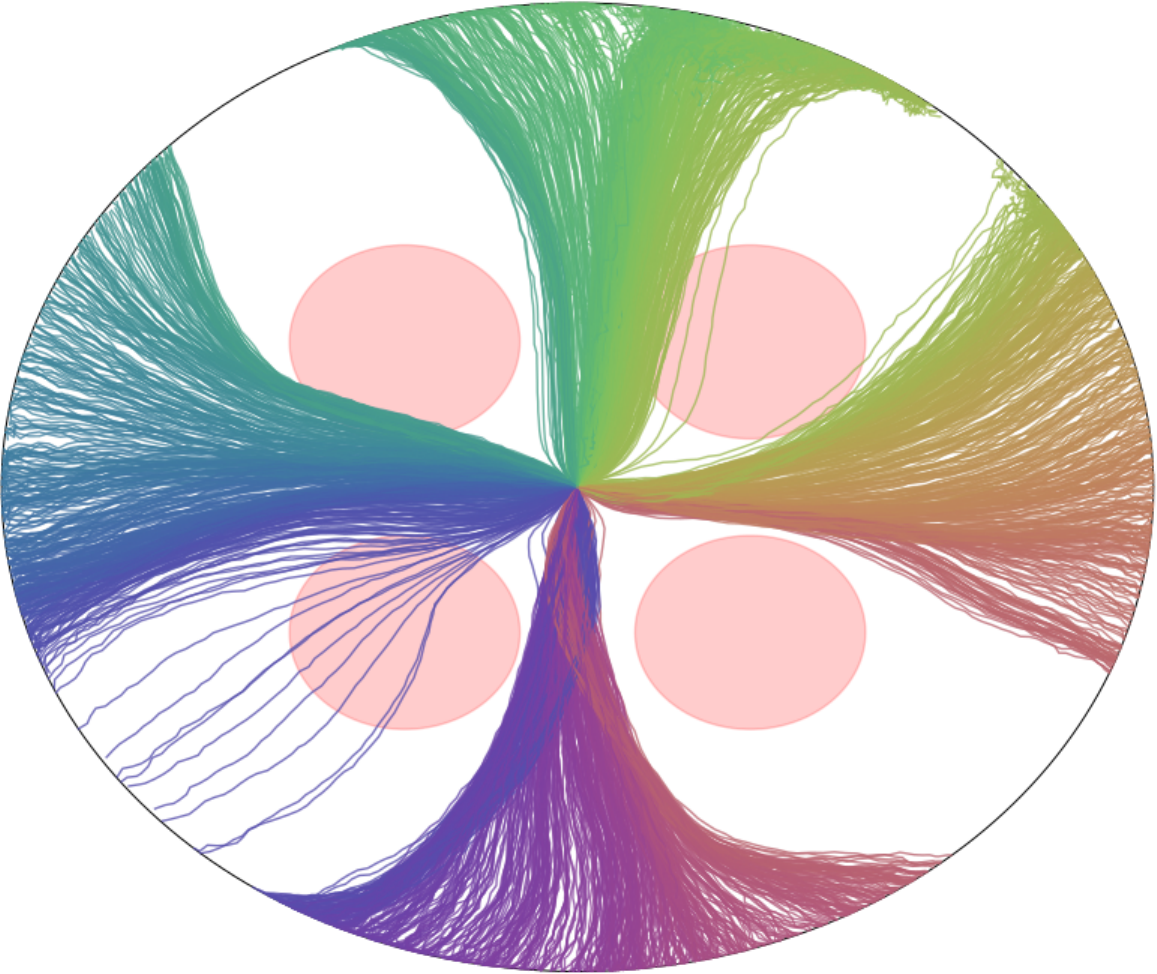} 
        \caption{1.5-HaSD}
         \label{fig: apdx|a_generalisation|extrapolation|1.5}
    \end{subfigure}
    \hfill
    \begin{subfigure}[b]{0.23\textwidth}
        \centering
        \includegraphics[width=\textwidth]{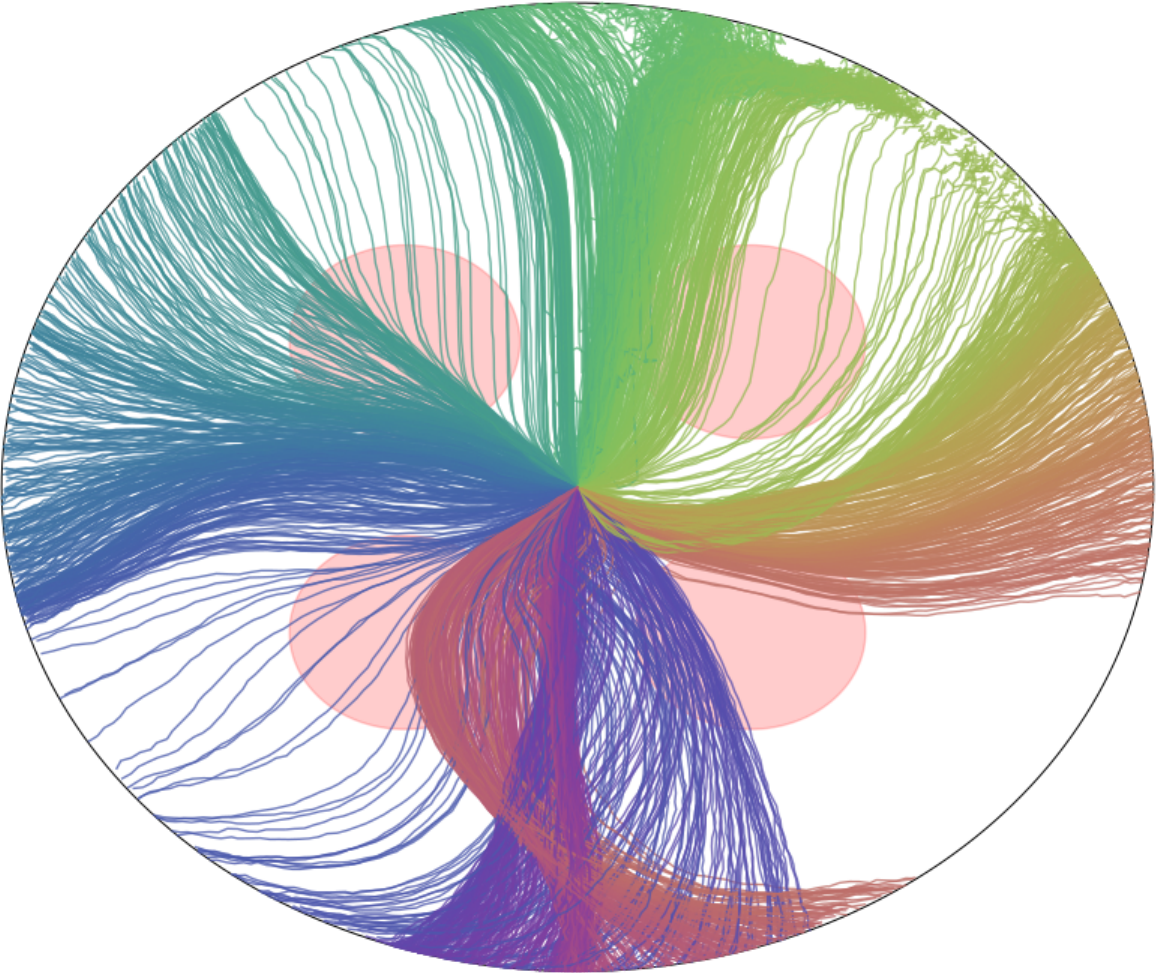} 
        \caption{2.0-HaSD}
         \label{fig: apdx|a_generalisation|extrapolation|2.0}
    \end{subfigure}
    \caption{ \begin{math}\alpha\end{math}-HaSD skills through extrapolation.}
    \label{fig: apdx|a_generalisation|extrapolation}
\end{figure}

In this section, we provide additional experiments on the generalisation of \begin{math}\alpha\end{math} in \begin{math}\alpha\end{math}-HaSD. \Cref{fig: 0-HaSD,fig: 0.1-HaSD,fig: 0.2-HaSD,fig: 0.5-HaSD,fig: 1-HaSD} shows the skills set learned from seen \begin{math}\alpha\end{math} values during training, \Cref{fig:apdx|a_generalisation|0.4,fig: apdx|a_generalisation|0.7,fig: apdx|a_generalisation|0.8} with unseen \begin{math}\alpha\end{math} values during training through interpolation and Figure \Cref{fig:apdx|a_generalisation|extrapolation|-1.0,fig: apdx|a_generalisation|extrapolation|1.5,fig: apdx|a_generalisation|extrapolation|2.0} with unseen \begin{math}\alpha\end{math} values during training through extrapolation. This result demonstrates that \begin{math}\alpha\end{math}-HaSD can interpolate unseen \begin{math}\alpha\end{math} values well but fails to extrapolate to unseen \begin{math}\alpha\end{math} values outside of the scope of the \begin{math}\alpha\end{math} used in training, this is a common challenge in machine learning which is not specific to our methods.

%% file: paper/appendix/sections/collecting_human_preferences.tex
\subsection{Collecting Human Preferences}
\label{subsec: apdx|collecting_human_preferences}
In this section, we detail how we collected human preferences for the experiment in Section \ref{subsec: experiment|Human|nav2d}. We first collect trajectories from previous runs by using \begin{math}\alpha\end{math}-HaSD which provides diverse trajectories. We then construct a dataset of queries by selecting segments of the trajectories collected and pairing them randomly with other created segments. 

\begin{figure}[htbp]
\centering
\includegraphics[width=0.3\linewidth]{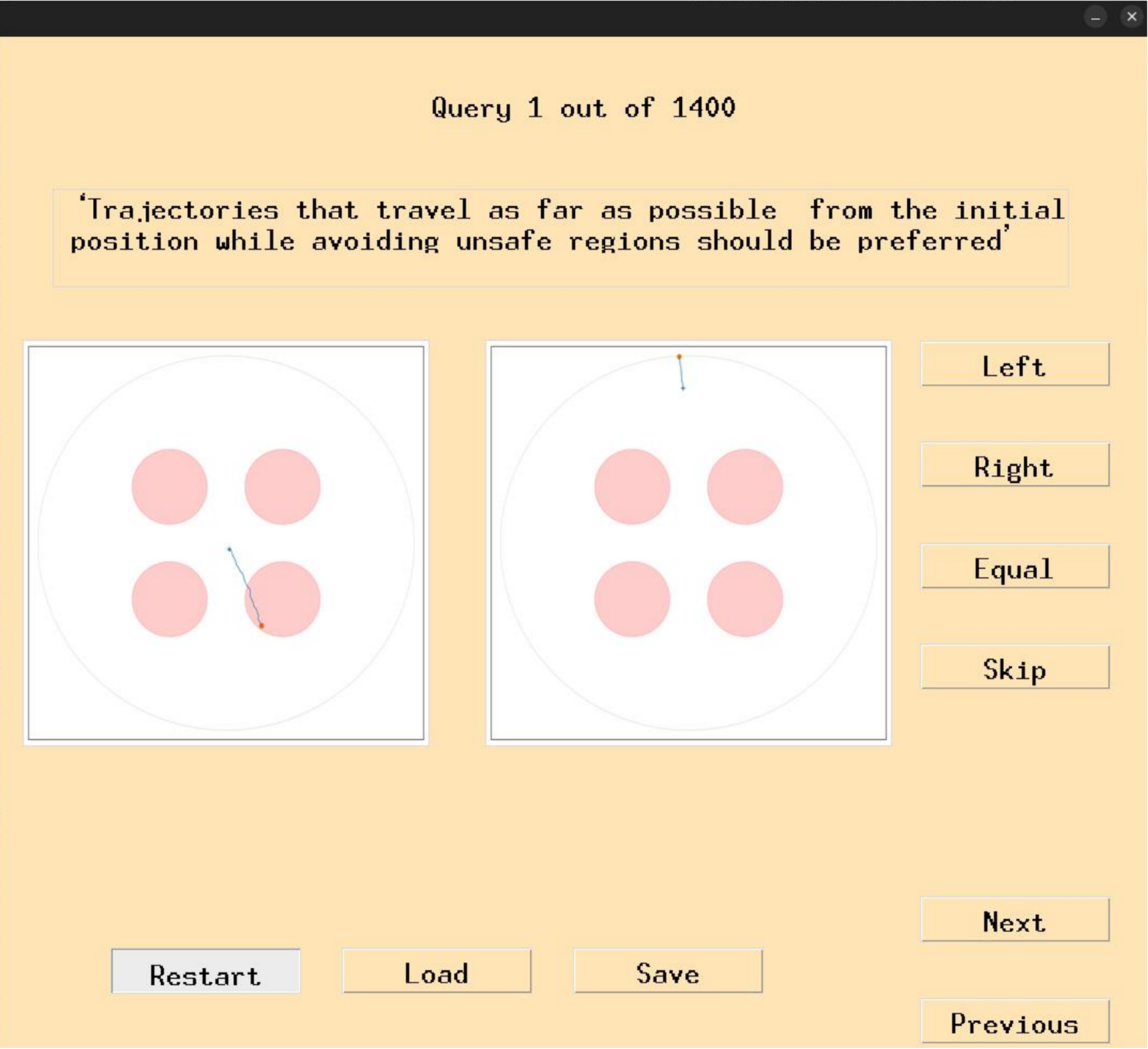} 
\caption{Application used to collect human preferences.}
    \label{fig: apdx|chp|gui}
\end{figure}

We show segments to humans in the form of an image representing the path followed by the agent during the segment selected. This is illustrated in Figure \ref{fig: apdx|chp|gui}. Each segment has its own image to easily distinguish its path. We found that showing trajectories in the form of an image allows us to provide our preferences quicker. Once we have collected our preferences dataset, we train a reward model with the hyperparameters specified in table \ref{rlhf-table} for 10000 epochs.

%% file: paper/appendix/sections/societal_impact.tex
\subsection{Broader Impacts: Potential Positive and Negative Societal Impacts }
\label{subsec: apdx|social_impact}
We believe that the unconstrained nature of Unsupervised Skill Discovery should be addressed to reduce the potential discovery of unsafe and undesirable skills. As such, our work on aligning skill discovery methods aims to reduce the potential negative societal impacts of unsupervised skill discovery methods. However, we also found that using a negative  \begin{math} \alpha\end{math} in Equation \ref{eq: hasd_reward}, can lead to the discovery of skills contrary to human value. This could have a negative impact, so we recommend that future applications ensure that \begin{math} \alpha\end{math} always remains positive.

%% file: paper/appendix/sections/qualitative_results.tex
\section{Full Qualitative Results}
\label{sec: Apdx|qr}
\begin{figure}[htb]
    \begin{subfigure}[b]{1\textwidth}
        \centering
        \includegraphics[width=0.14\textwidth]{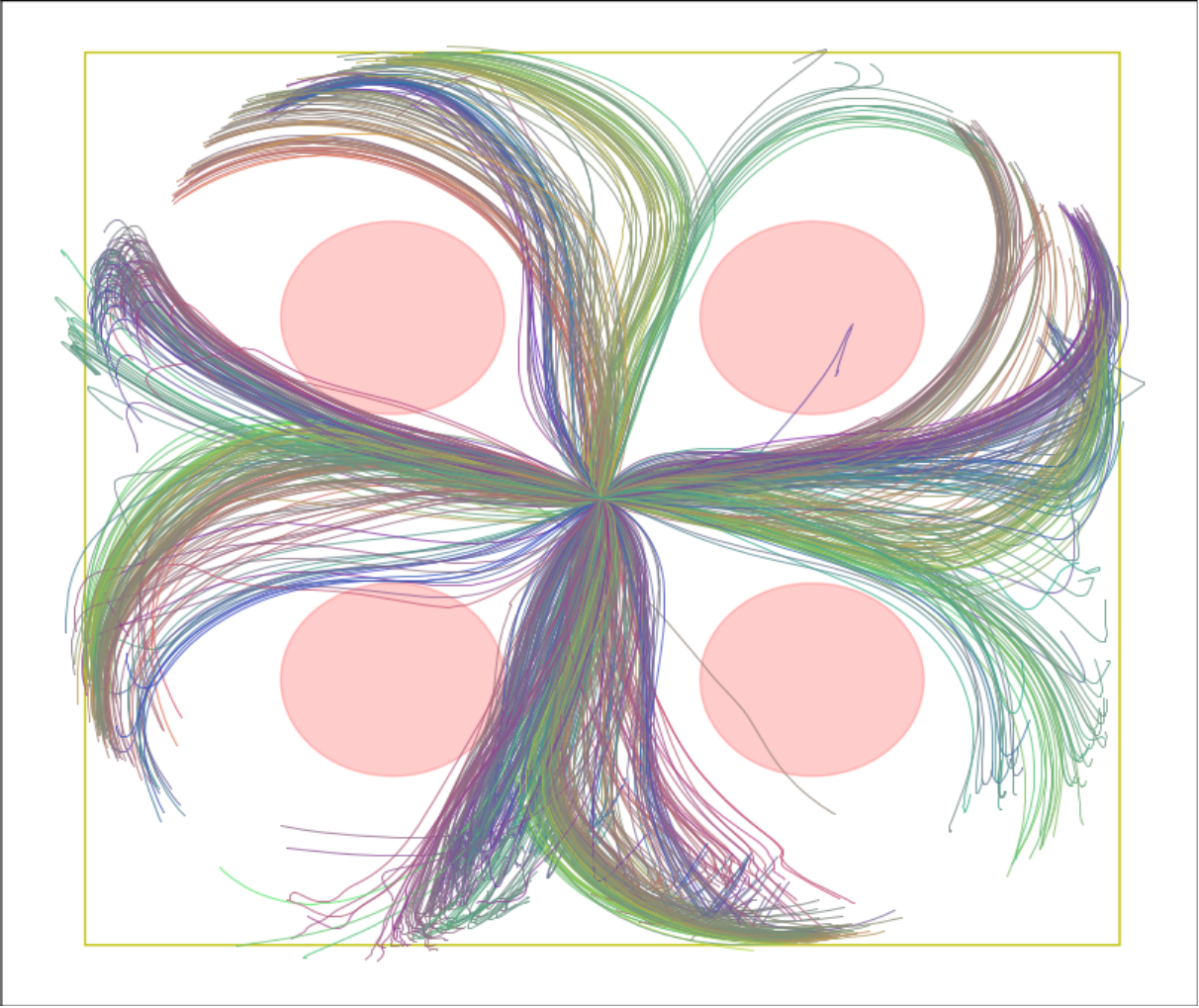} 
        \includegraphics[width=0.14\textwidth]{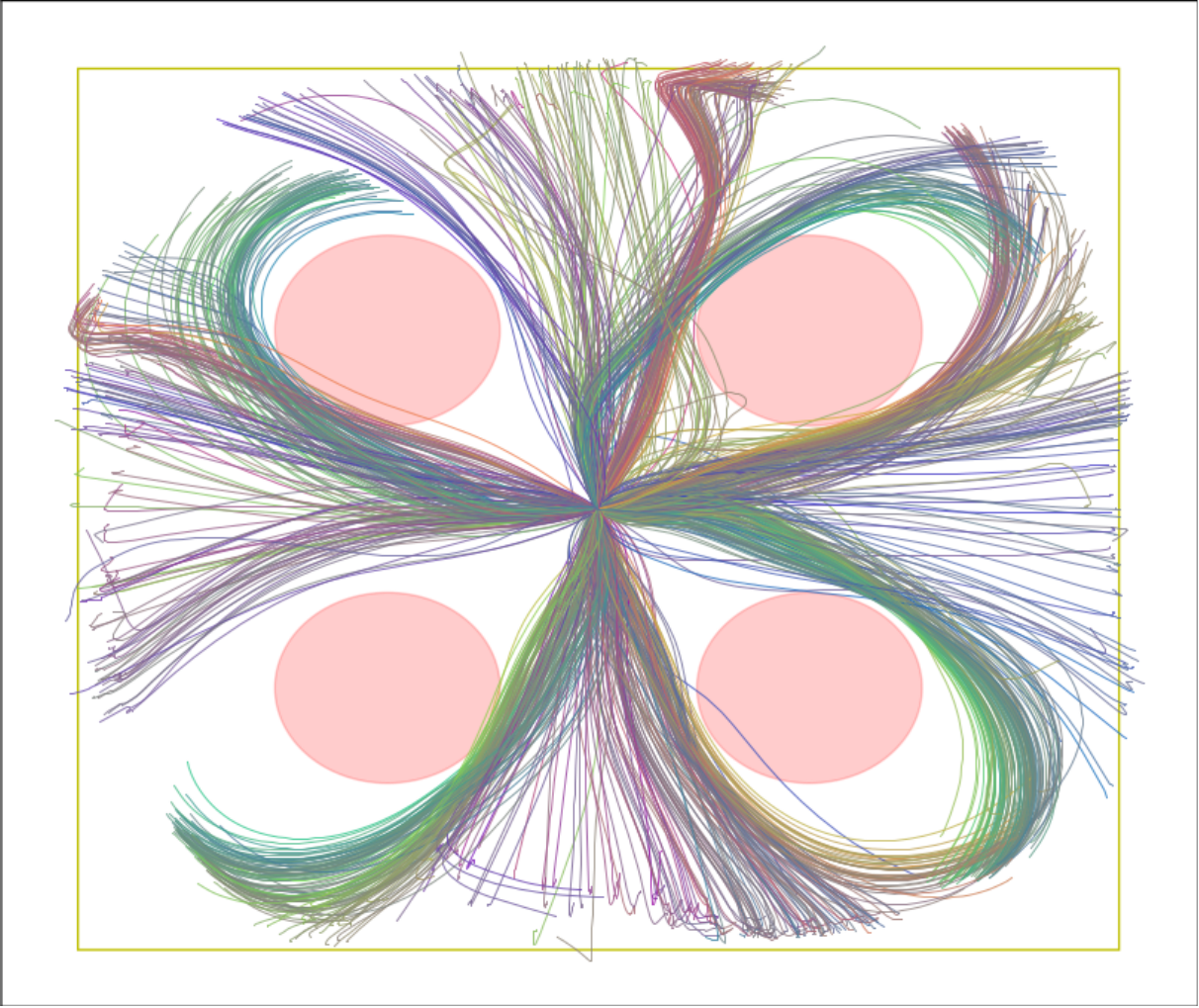} 
        \includegraphics[width=0.14\textwidth]{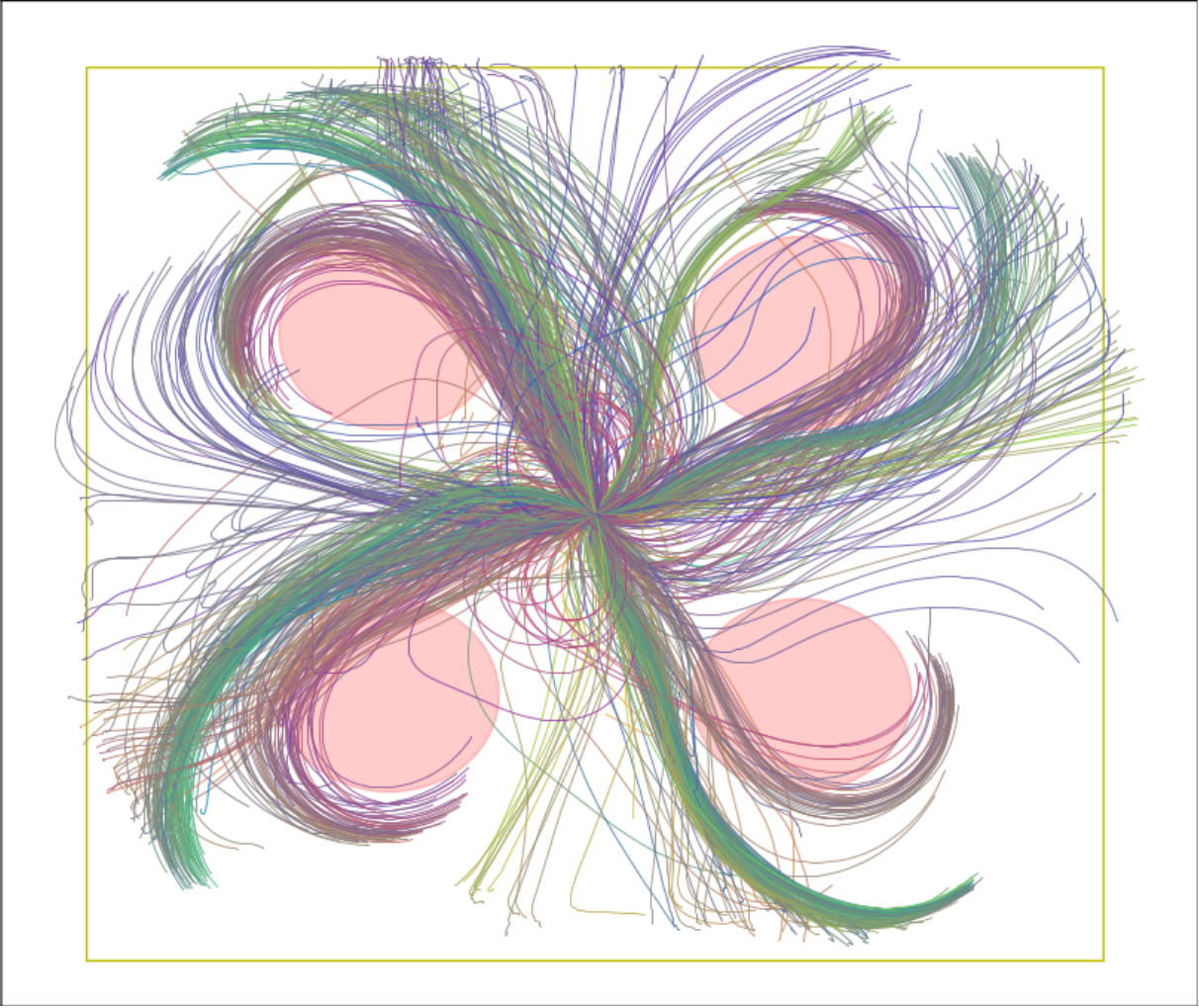} 
        \includegraphics[width=0.14\textwidth]{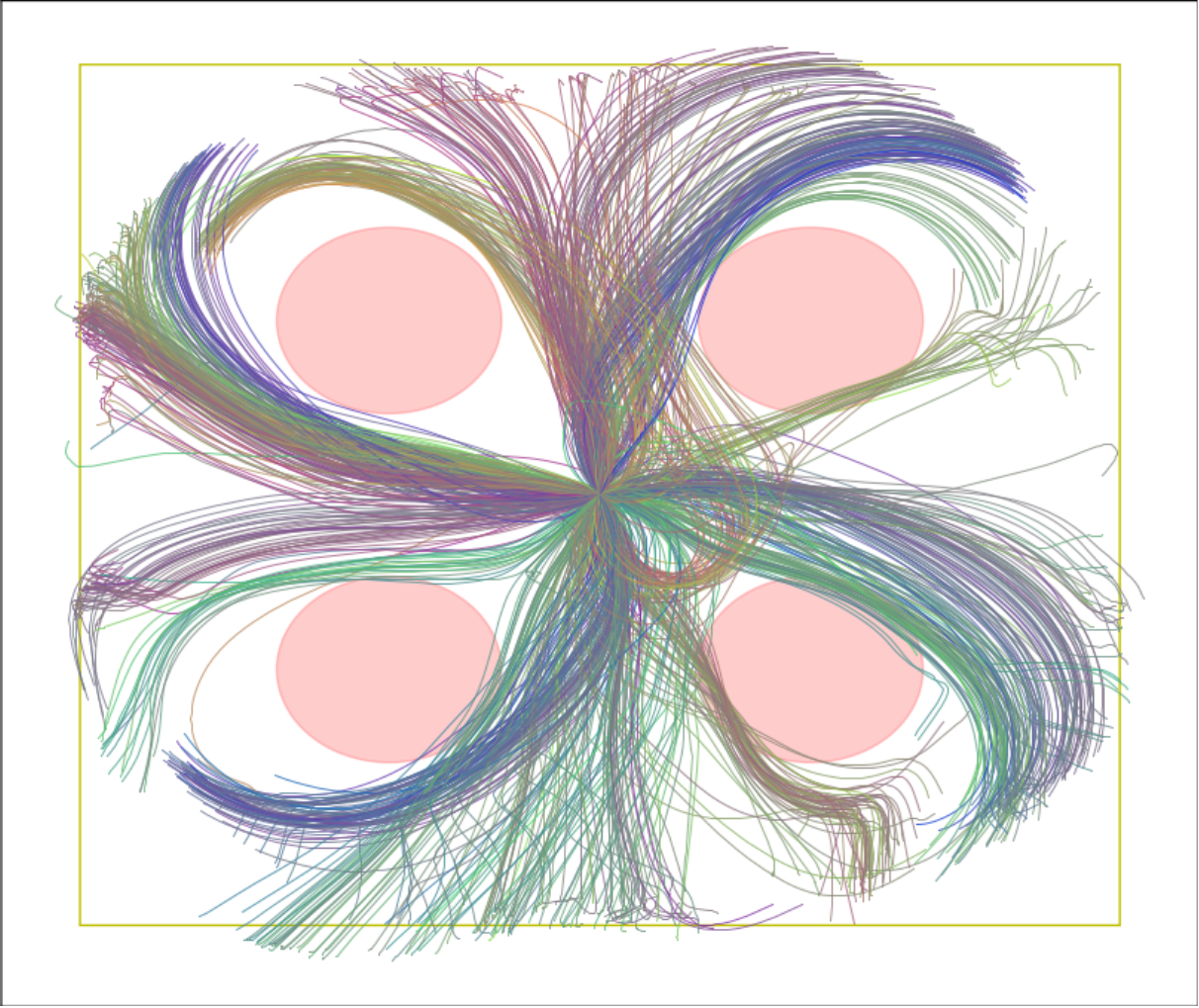} 
        \includegraphics[width=0.14\textwidth]{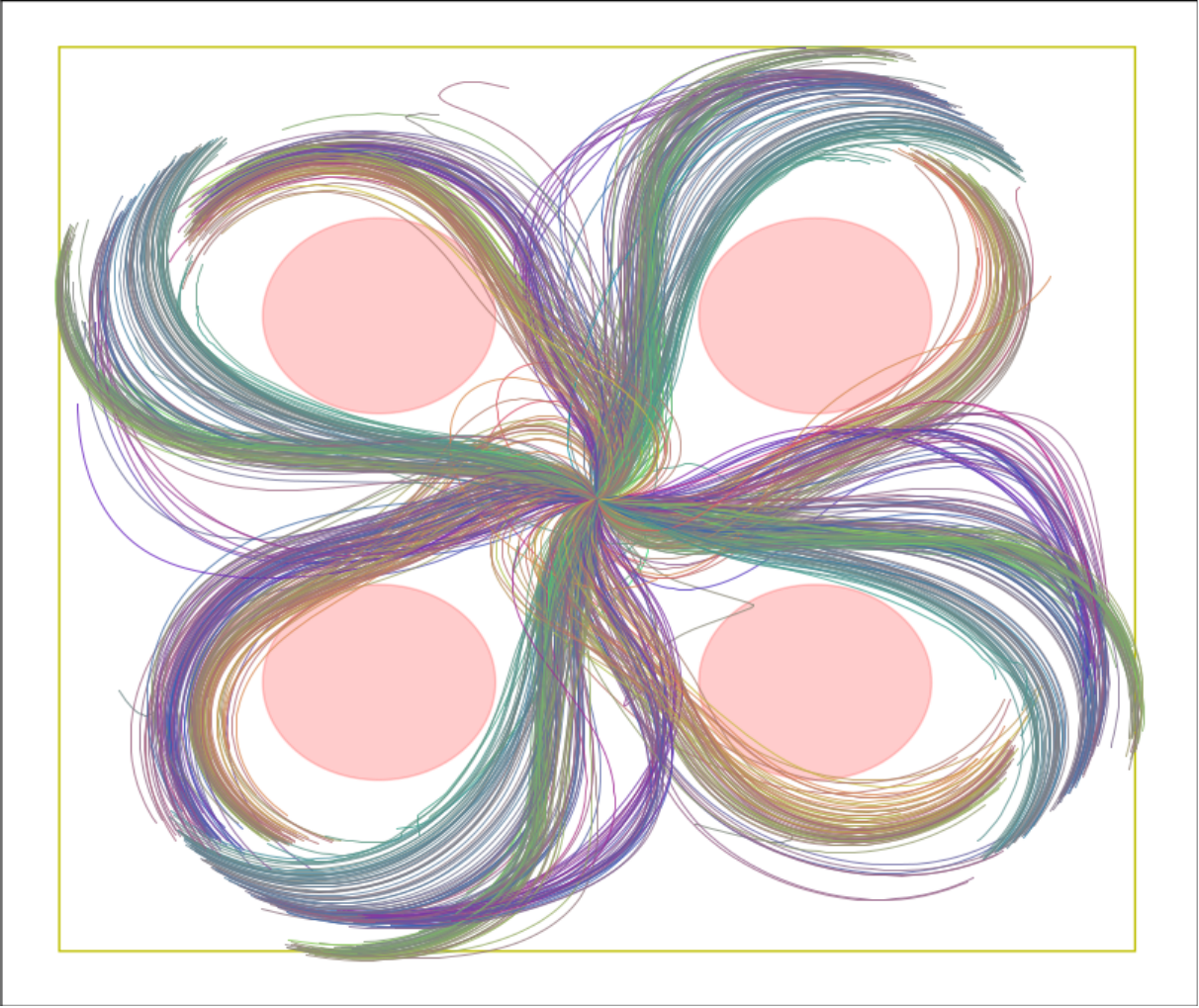} 
        \caption{Point}
        \label{fig: Apdx|hr1|qr|point}
    \end{subfigure}
    \begin{subfigure}[b]{1\textwidth}
        \centering
    \includegraphics[width=0.14\textwidth]{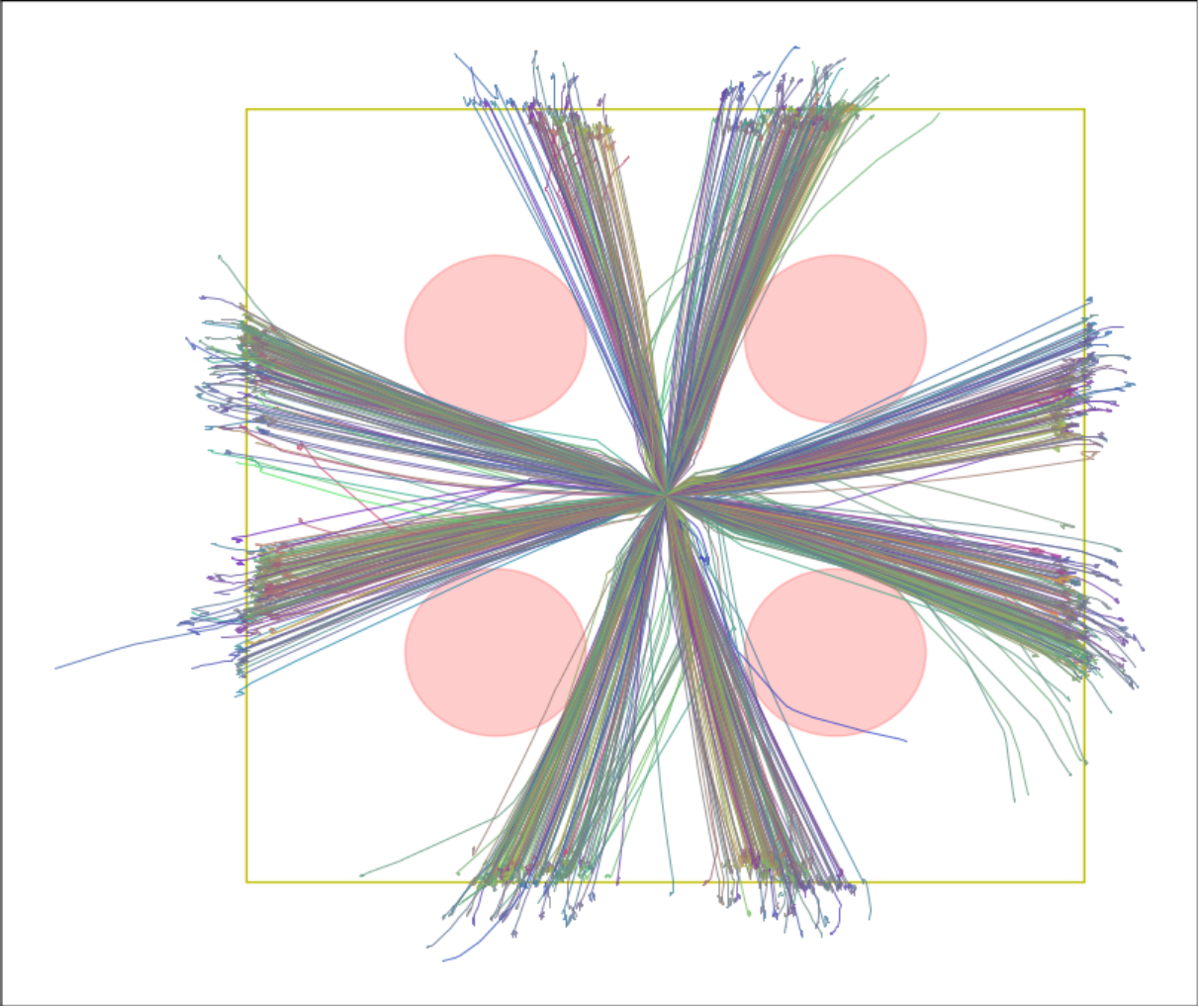} 
    \includegraphics[width=0.14\textwidth]{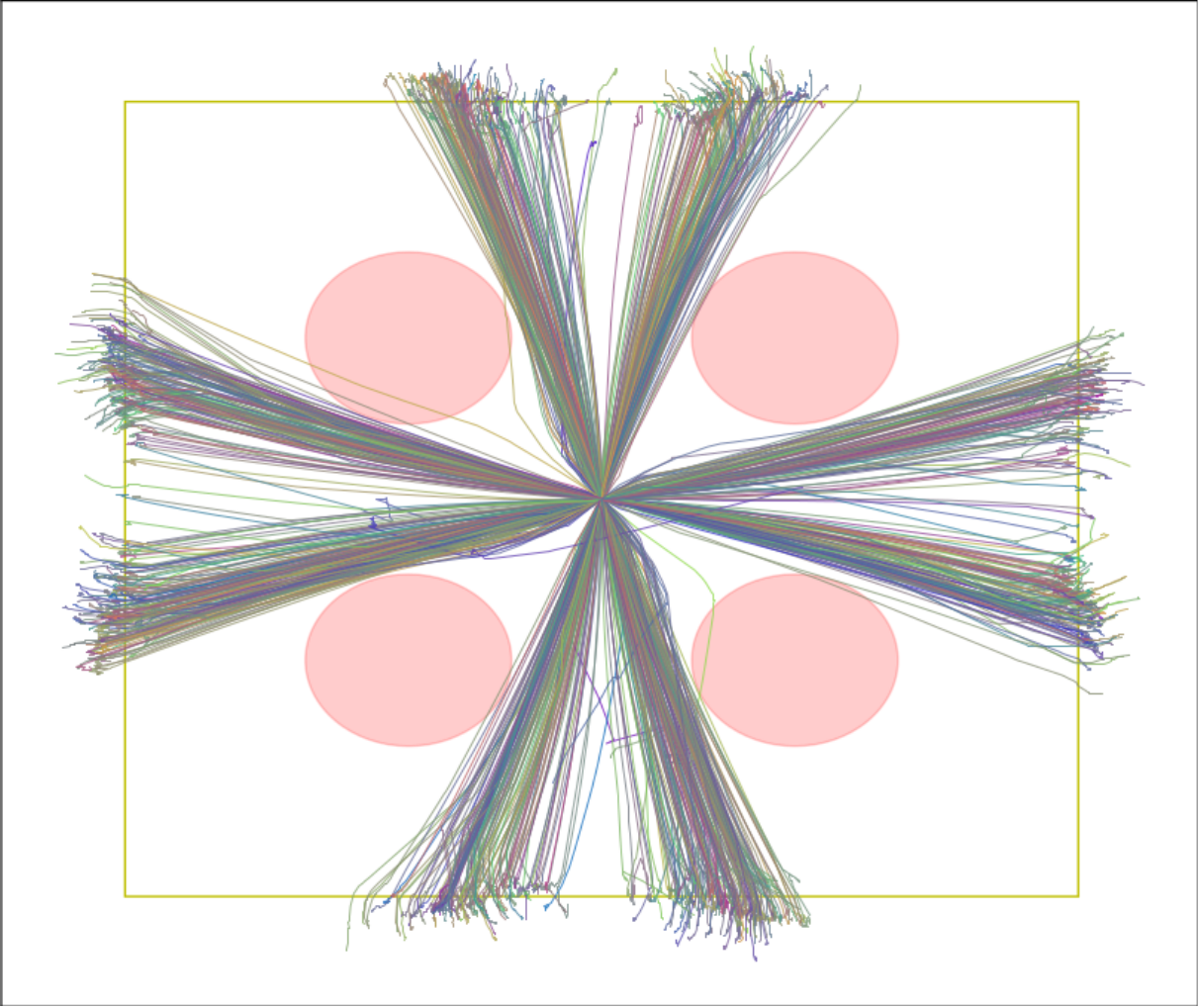} 
    \includegraphics[width=0.14\textwidth]{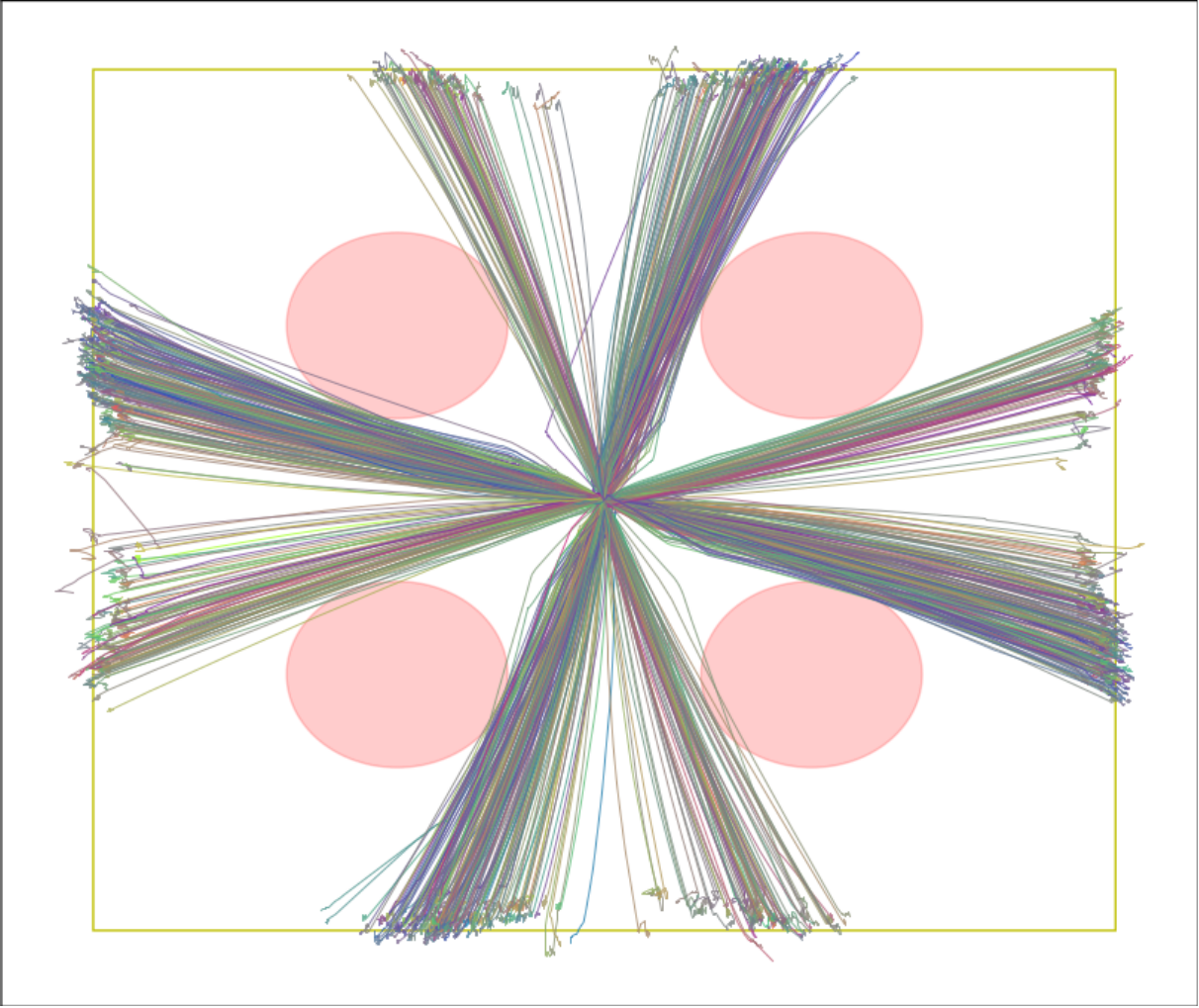} 
    \includegraphics[width=0.14\textwidth]{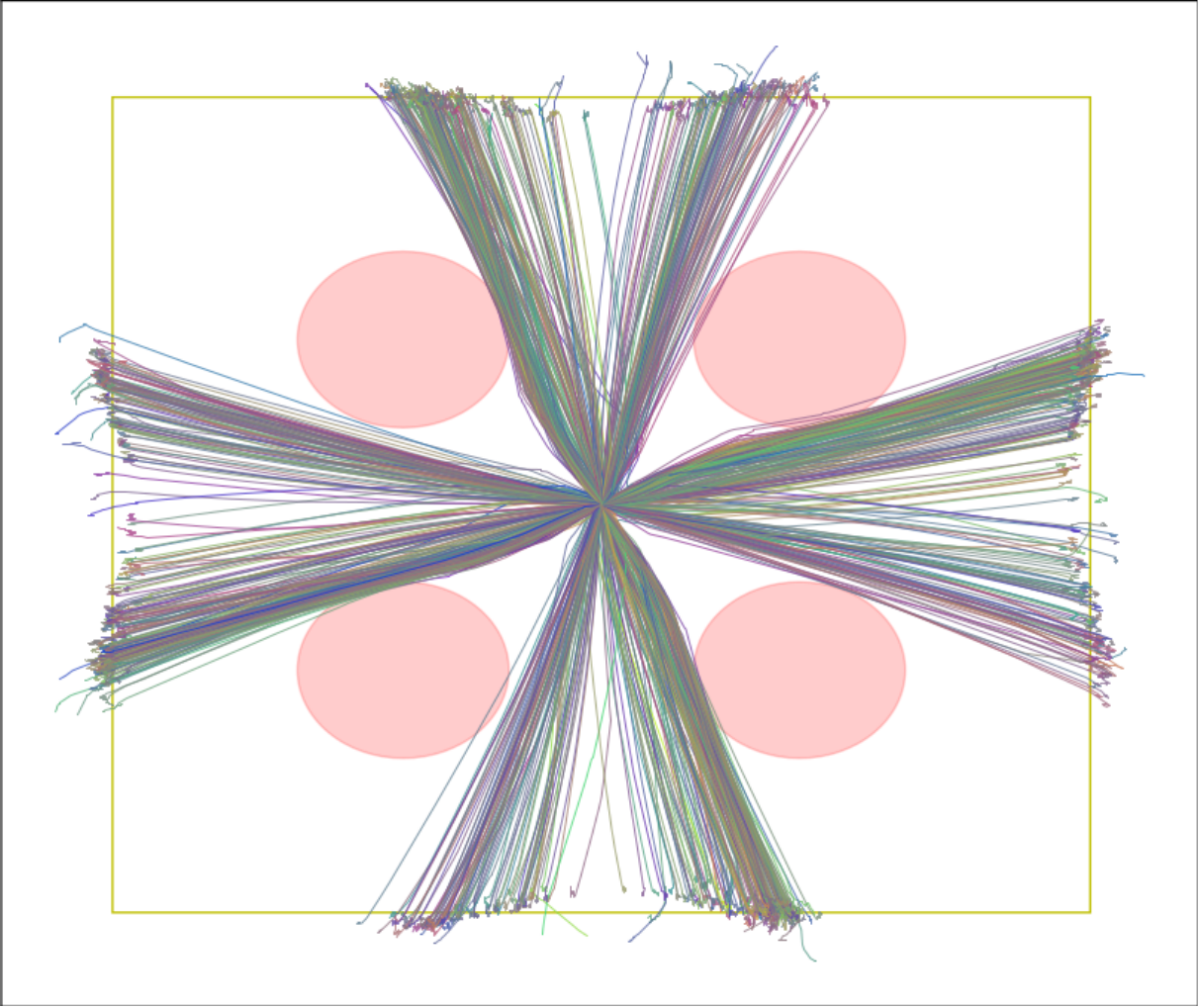} 
    \includegraphics[width=0.14\textwidth]{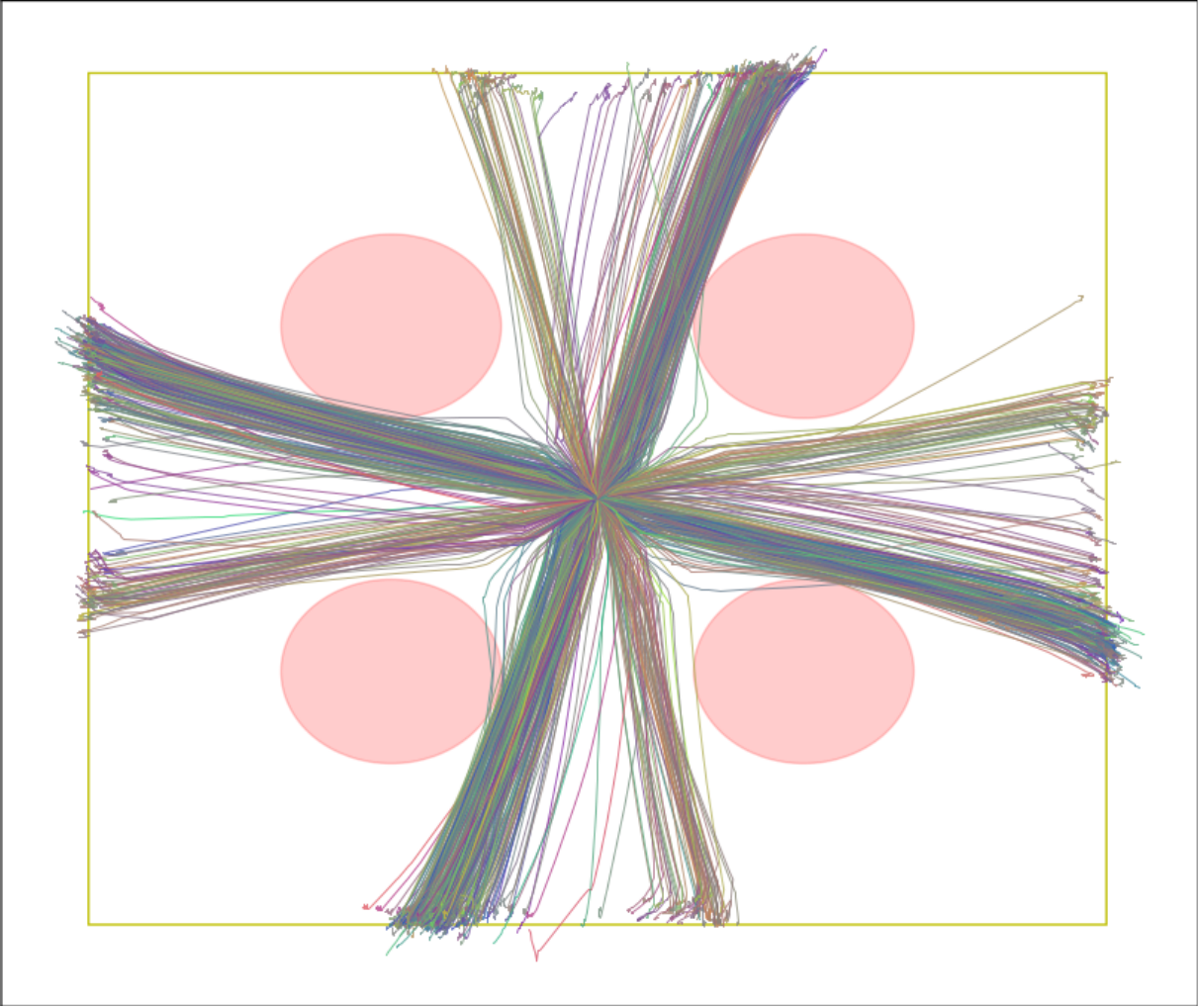} 
        \caption{Car}
        \label{fig: Apdx|hr1|qr|car}
    \end{subfigure}
     \begin{subfigure}[b]{1\textwidth}
        \centering
    \includegraphics[width=0.14\textwidth]{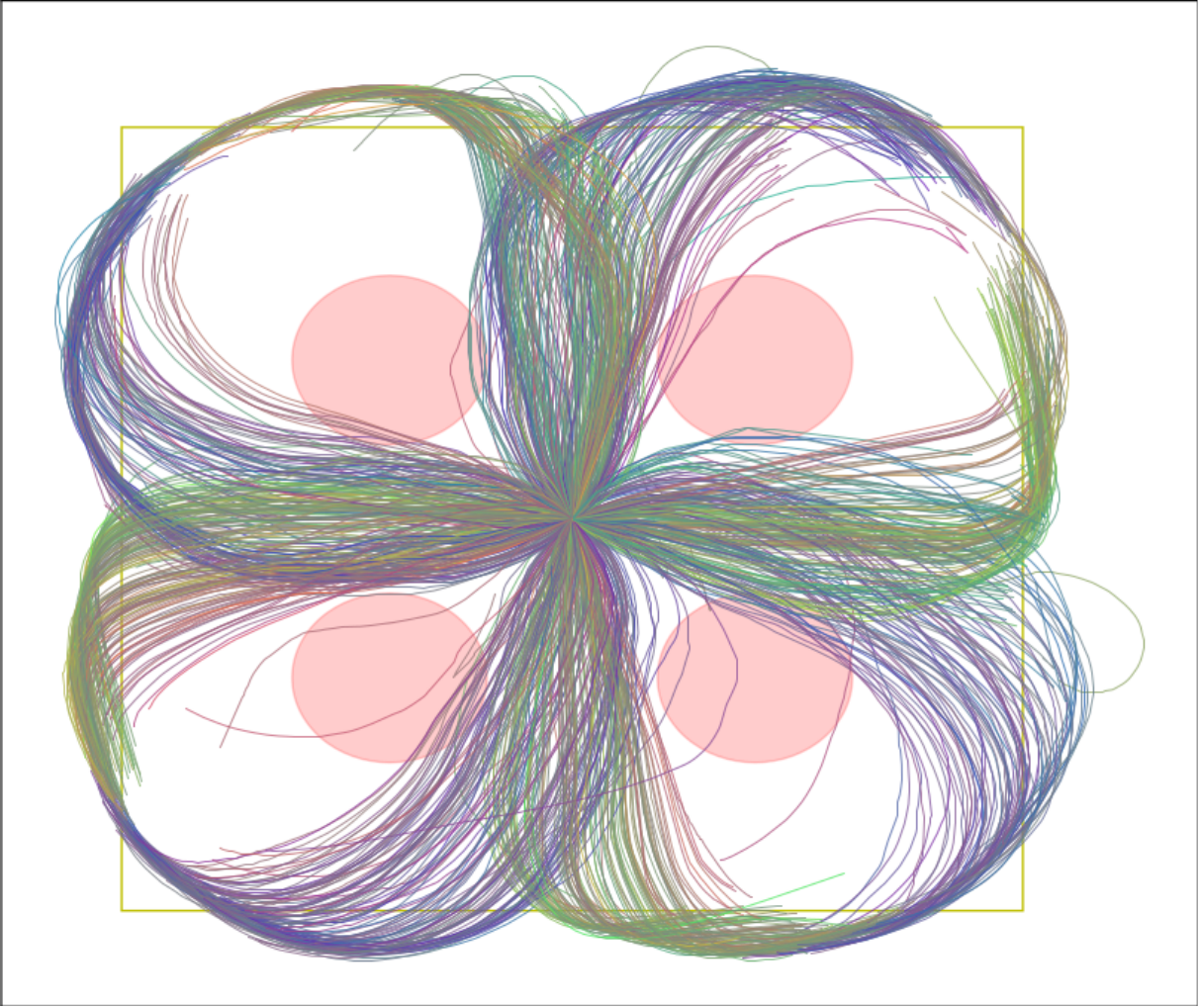} 
    \includegraphics[width=0.14\textwidth]{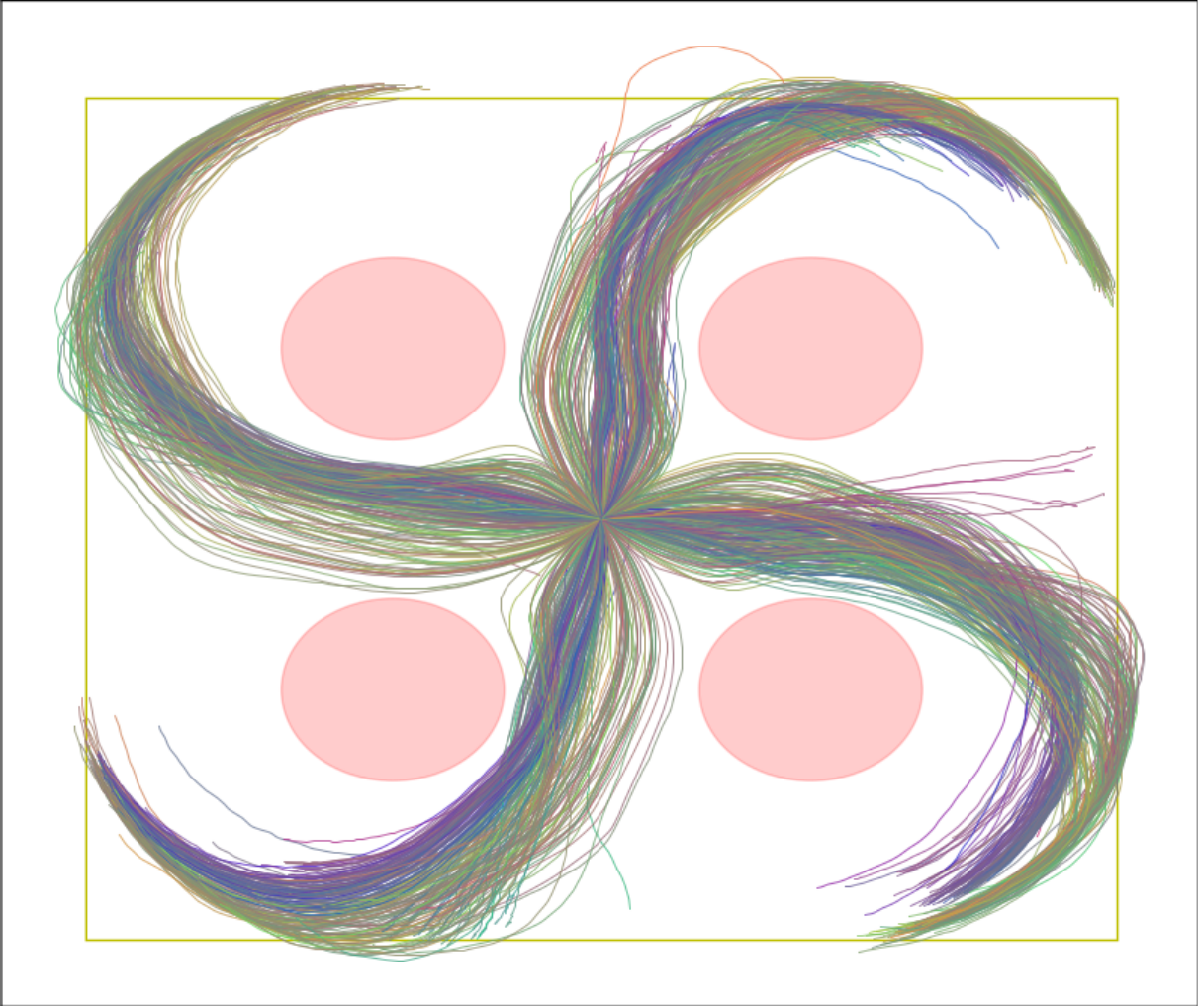} 
    \includegraphics[width=0.14\textwidth]{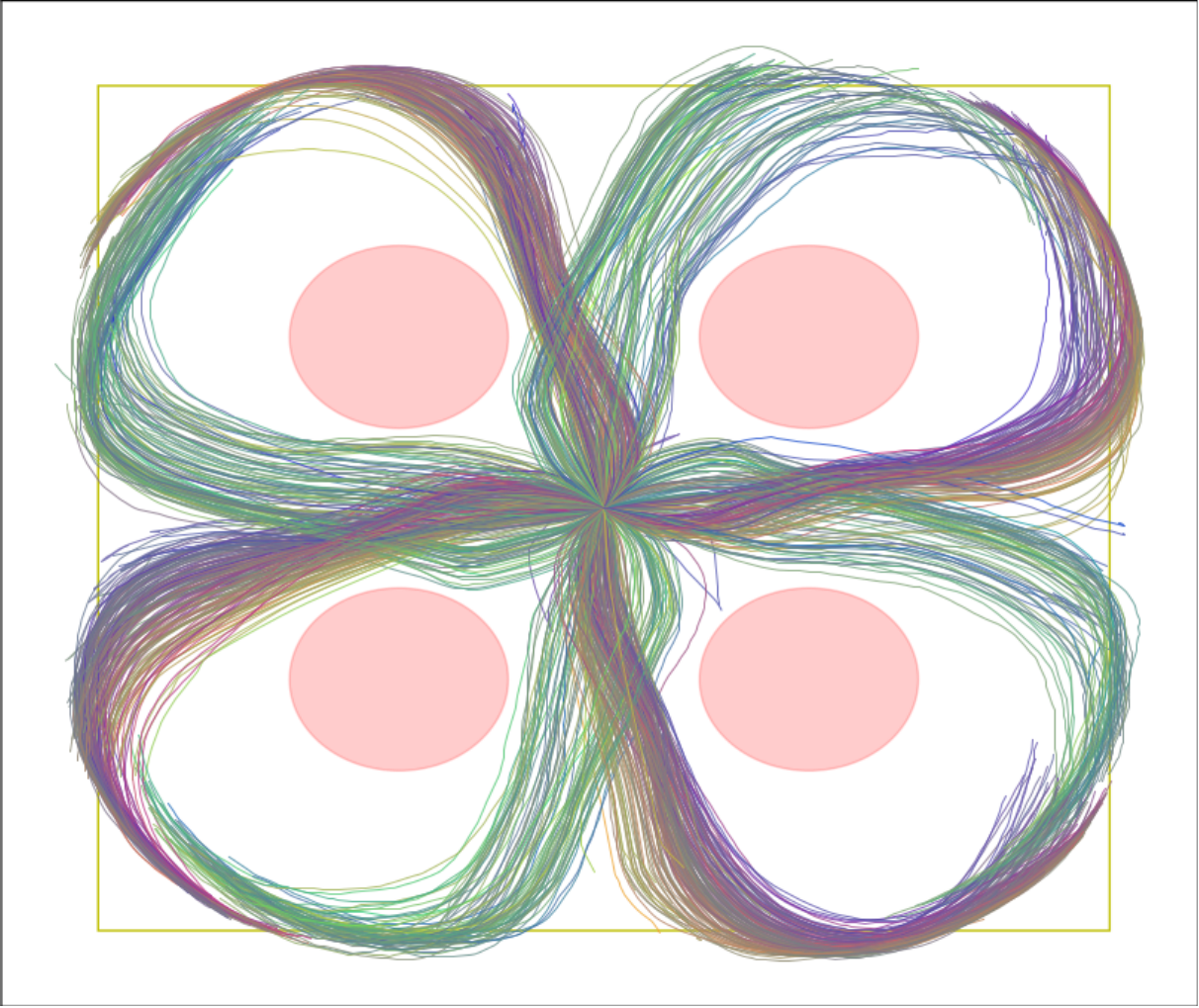} 
    \includegraphics[width=0.14\textwidth]{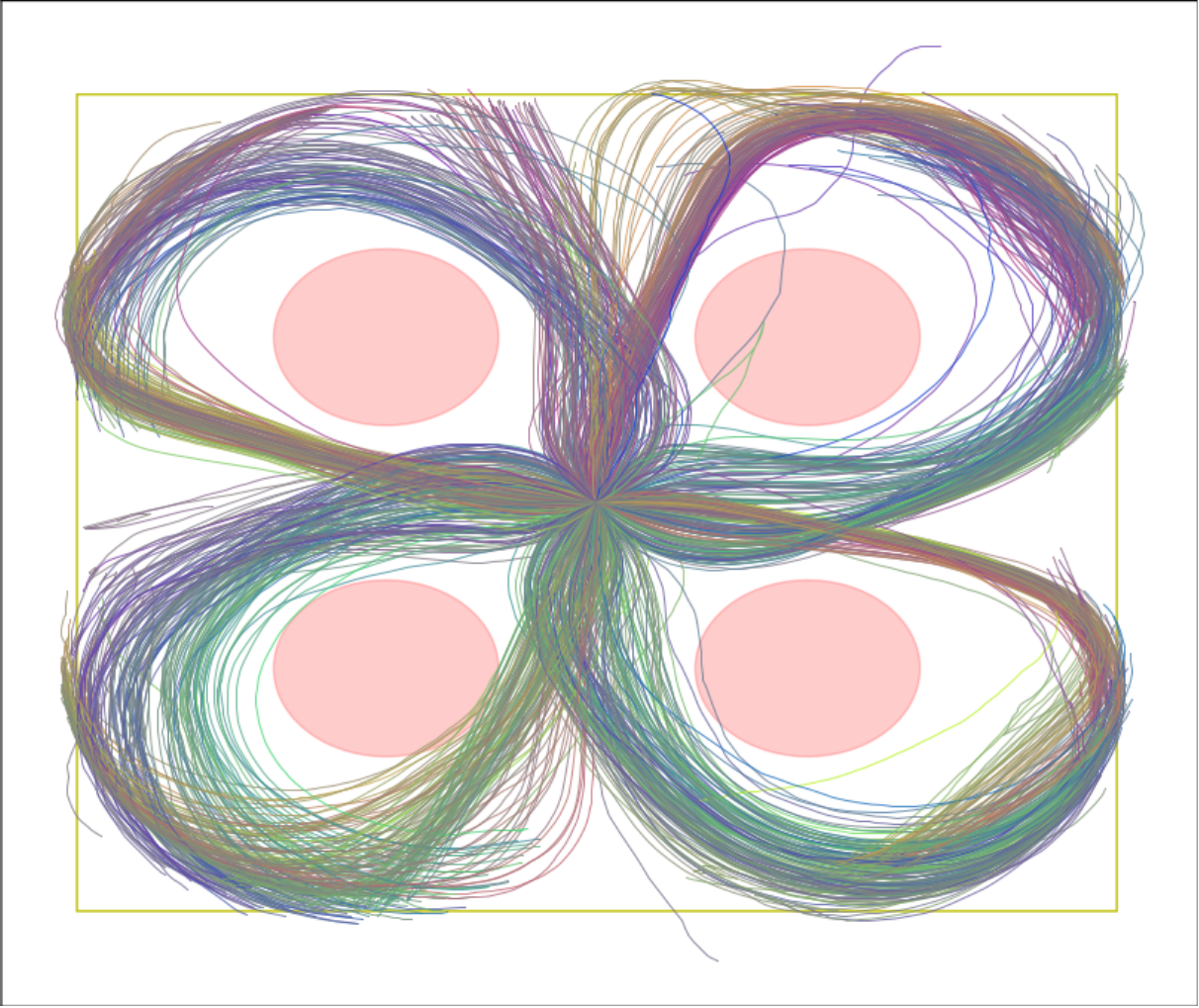} 
    \includegraphics[width=0.14\textwidth]{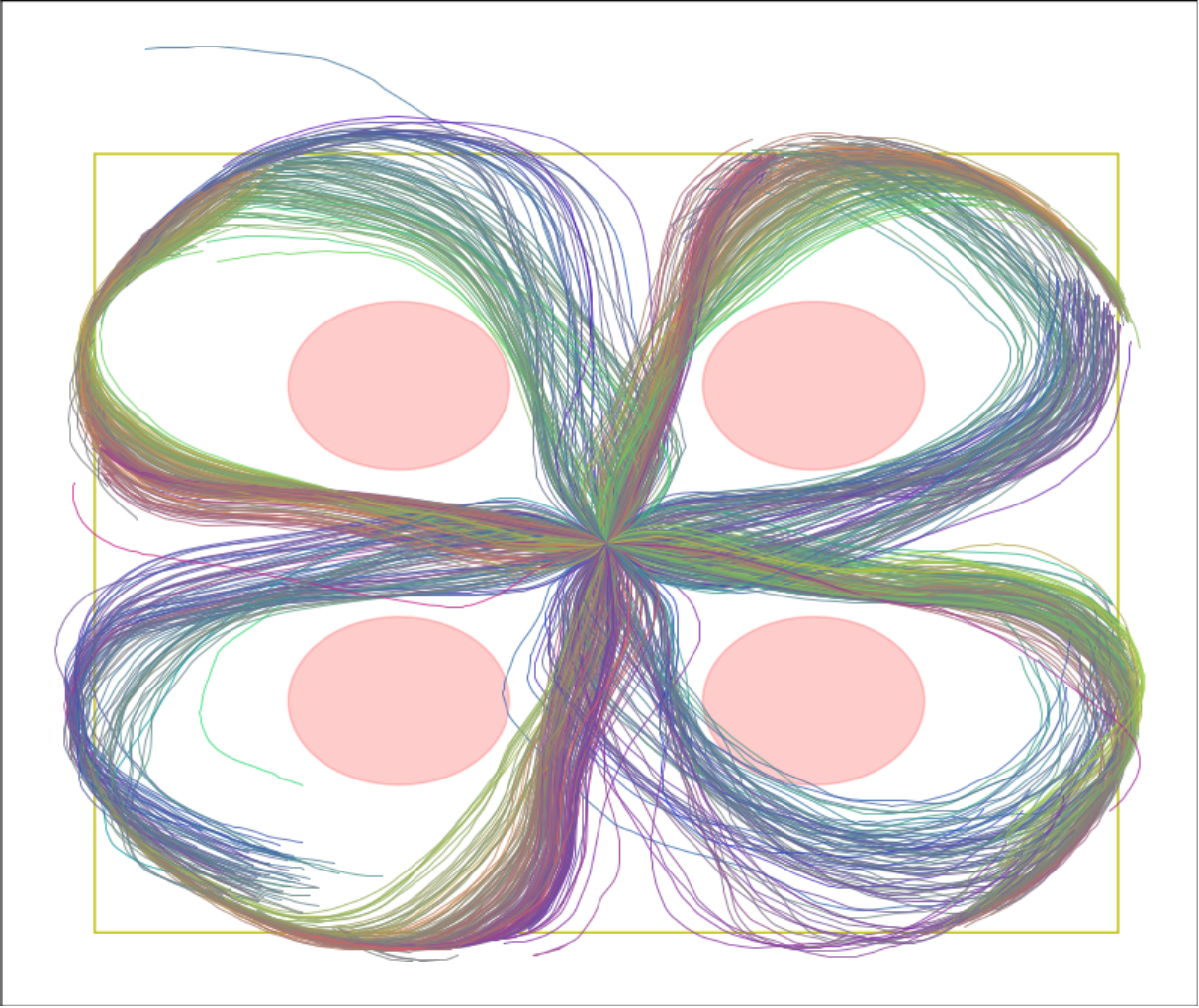} 
        \caption{Racecar}
        \label{fig: Apdx|hr1|qr|race}
    \end{subfigure}
    \begin{subfigure}[b]{1\textwidth}
        \centering
    \includegraphics[width=0.14\textwidth]{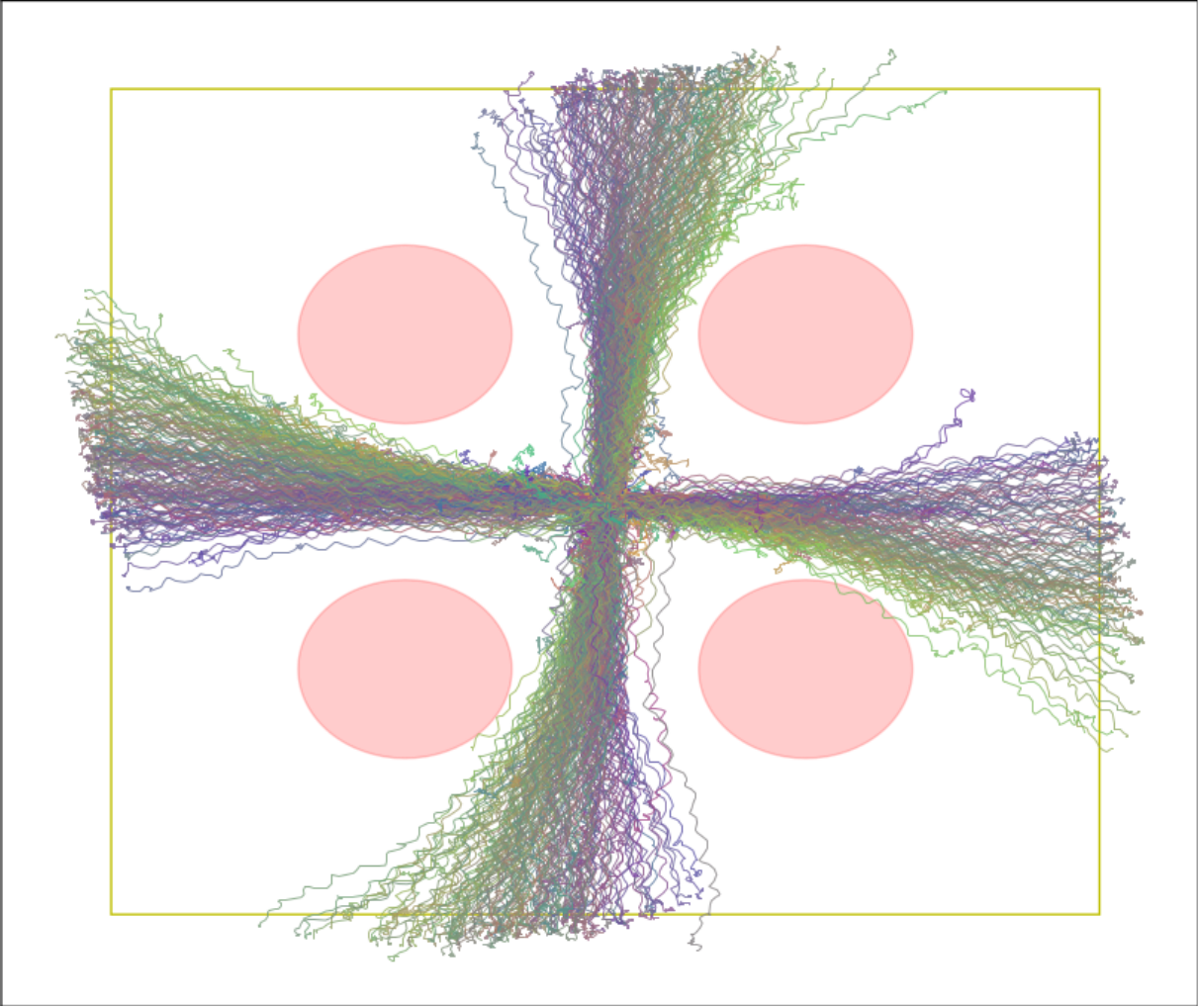} 
    \includegraphics[width=0.14\textwidth]{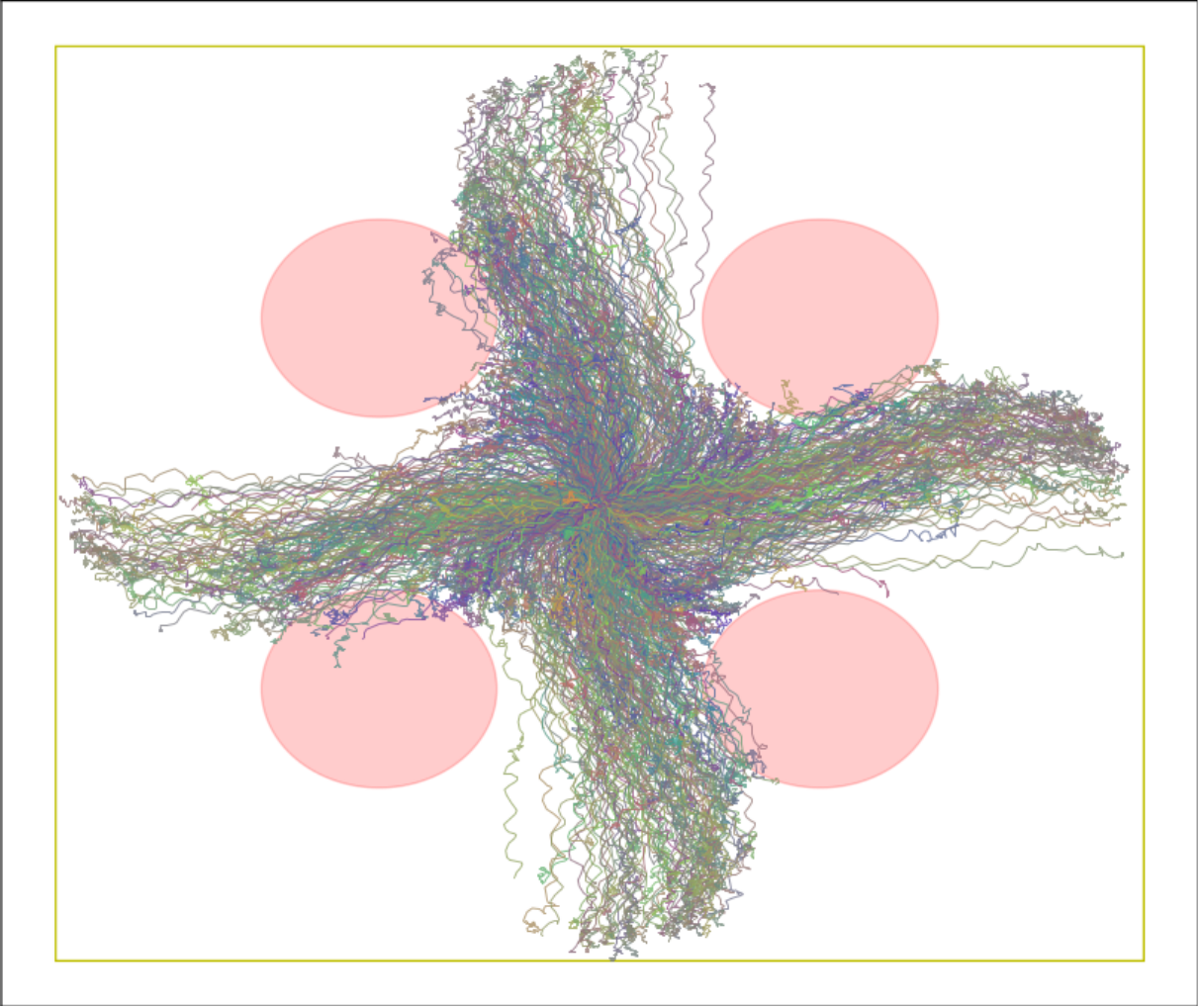} 
    \includegraphics[width=0.14\textwidth]{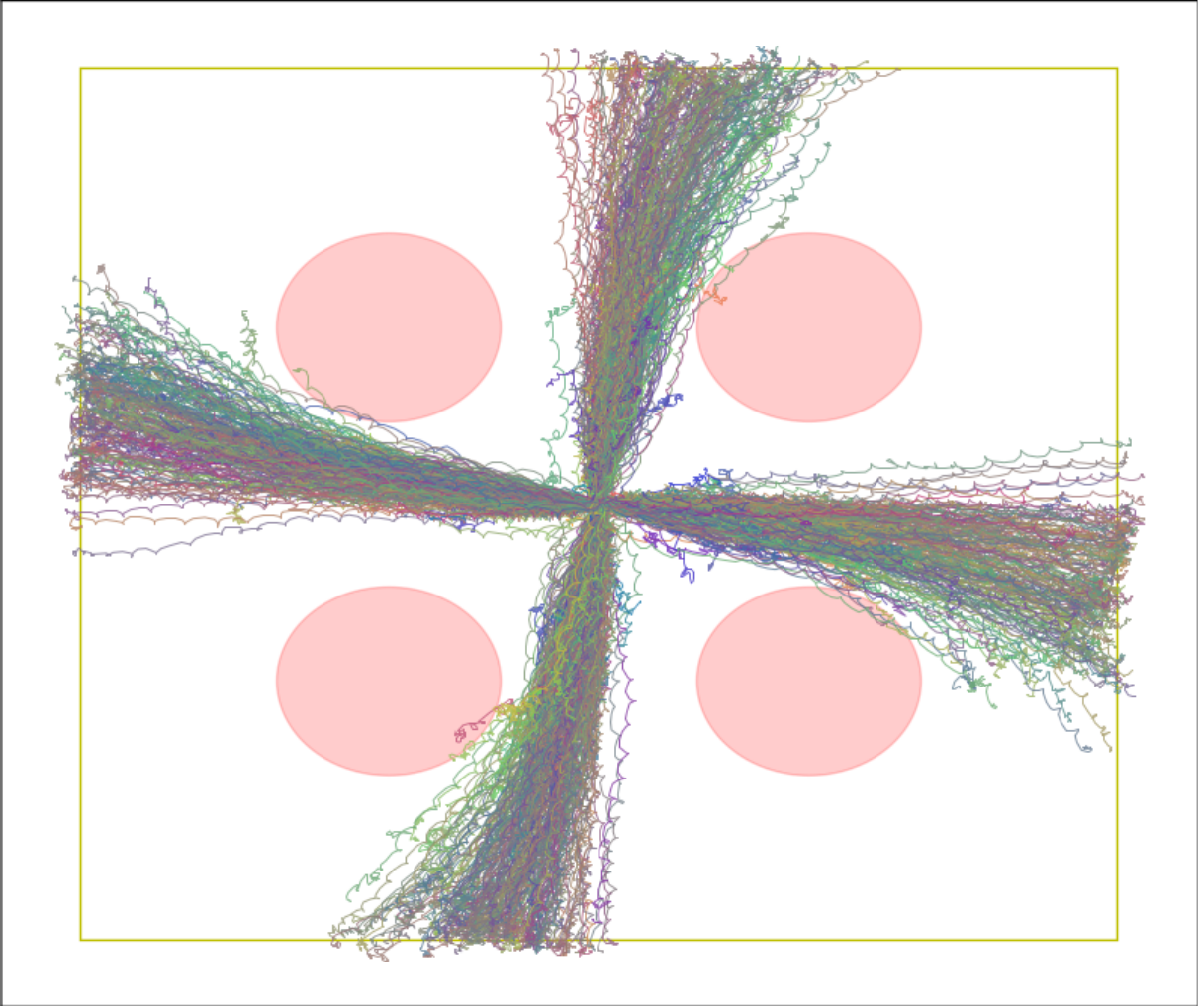} 
    \includegraphics[width=0.14\textwidth]{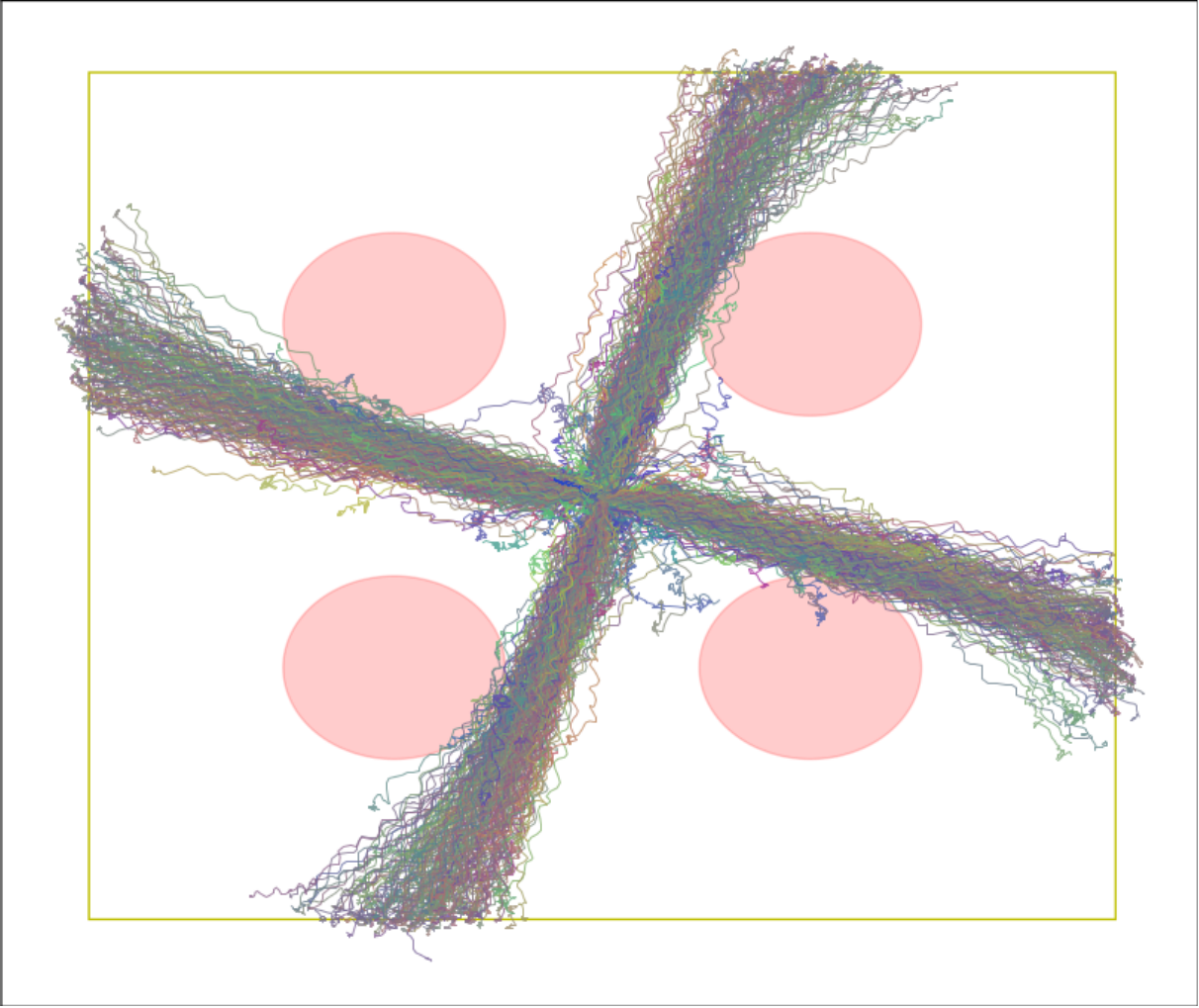} 
    \includegraphics[width=0.14\textwidth]{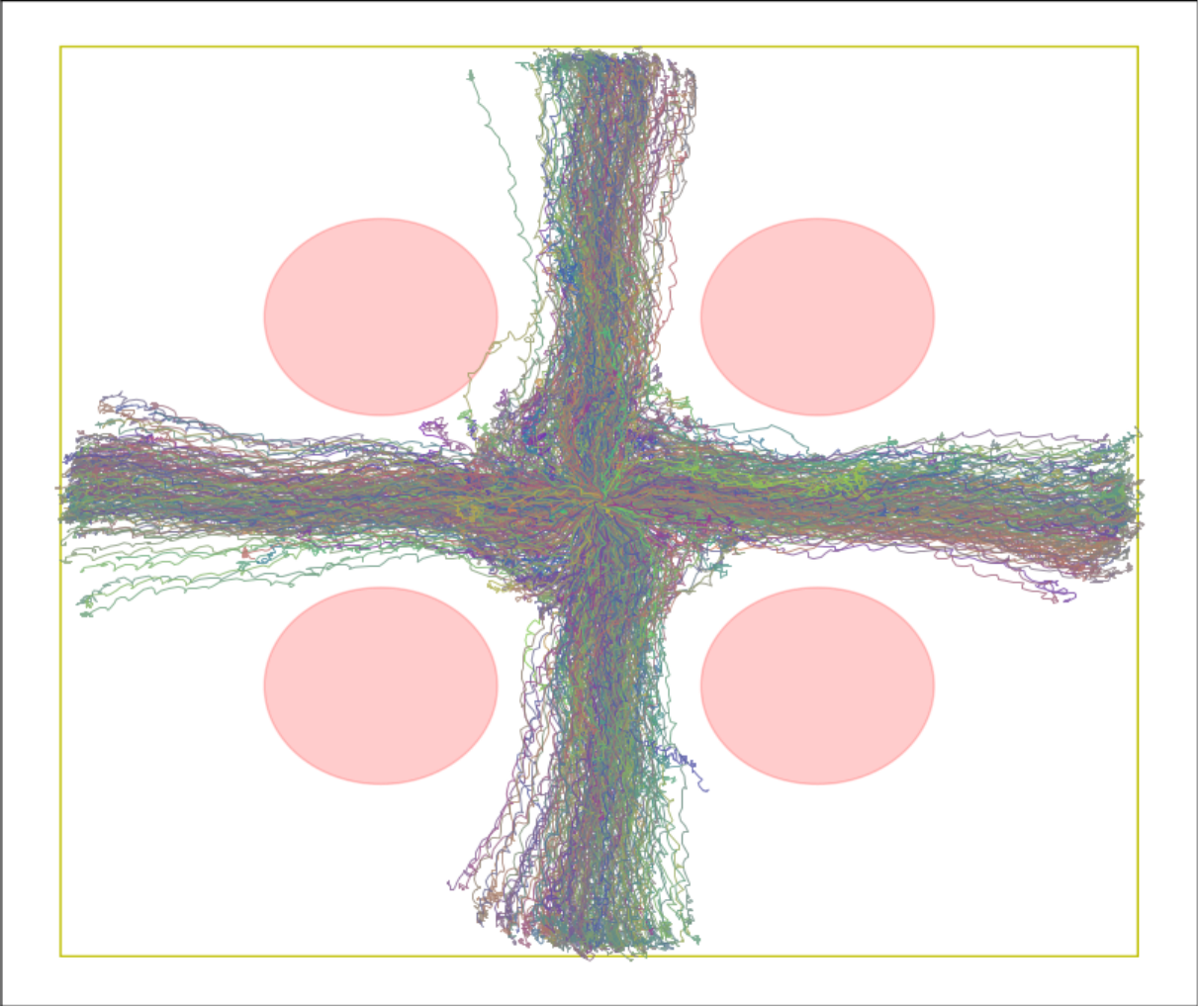} 
        \caption{Doggo}
        \label{fig: Apdx|hr1|qr|doggo}
    \end{subfigure}
    \begin{subfigure}[b]{1\textwidth}
        \centering
    \includegraphics[width=0.14\textwidth]{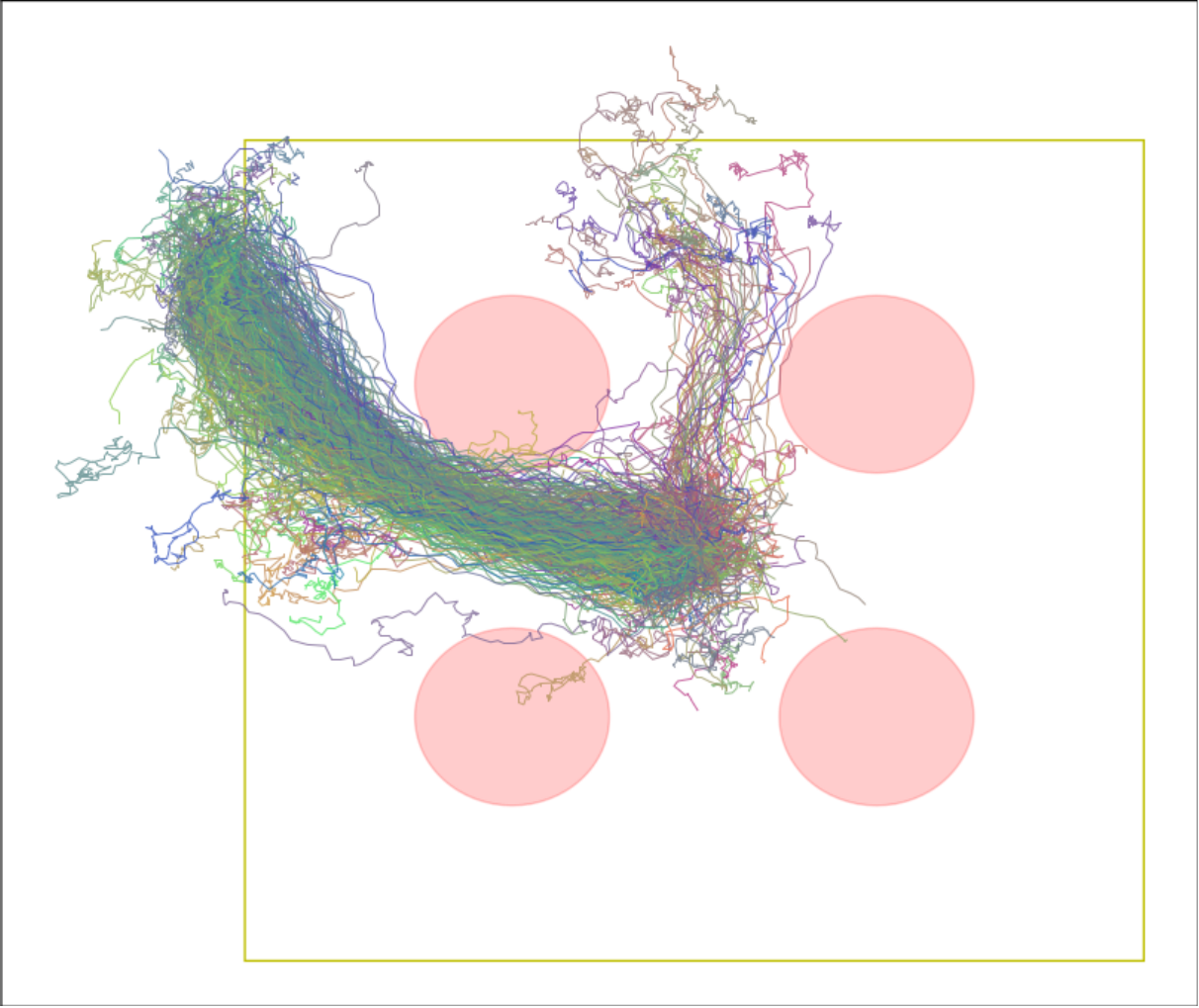} 
    \includegraphics[width=0.14\textwidth]{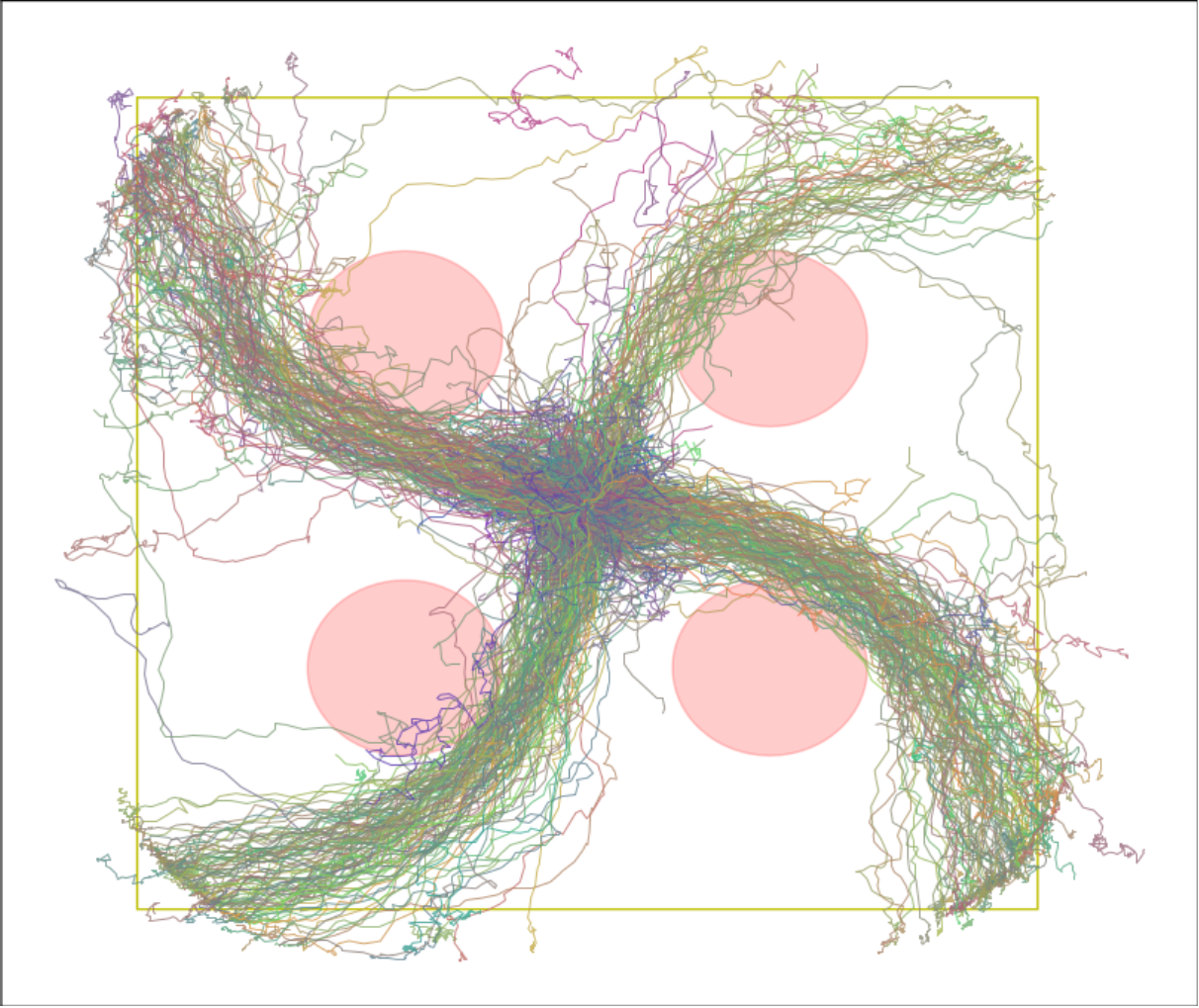} 
    \includegraphics[width=0.14\textwidth]{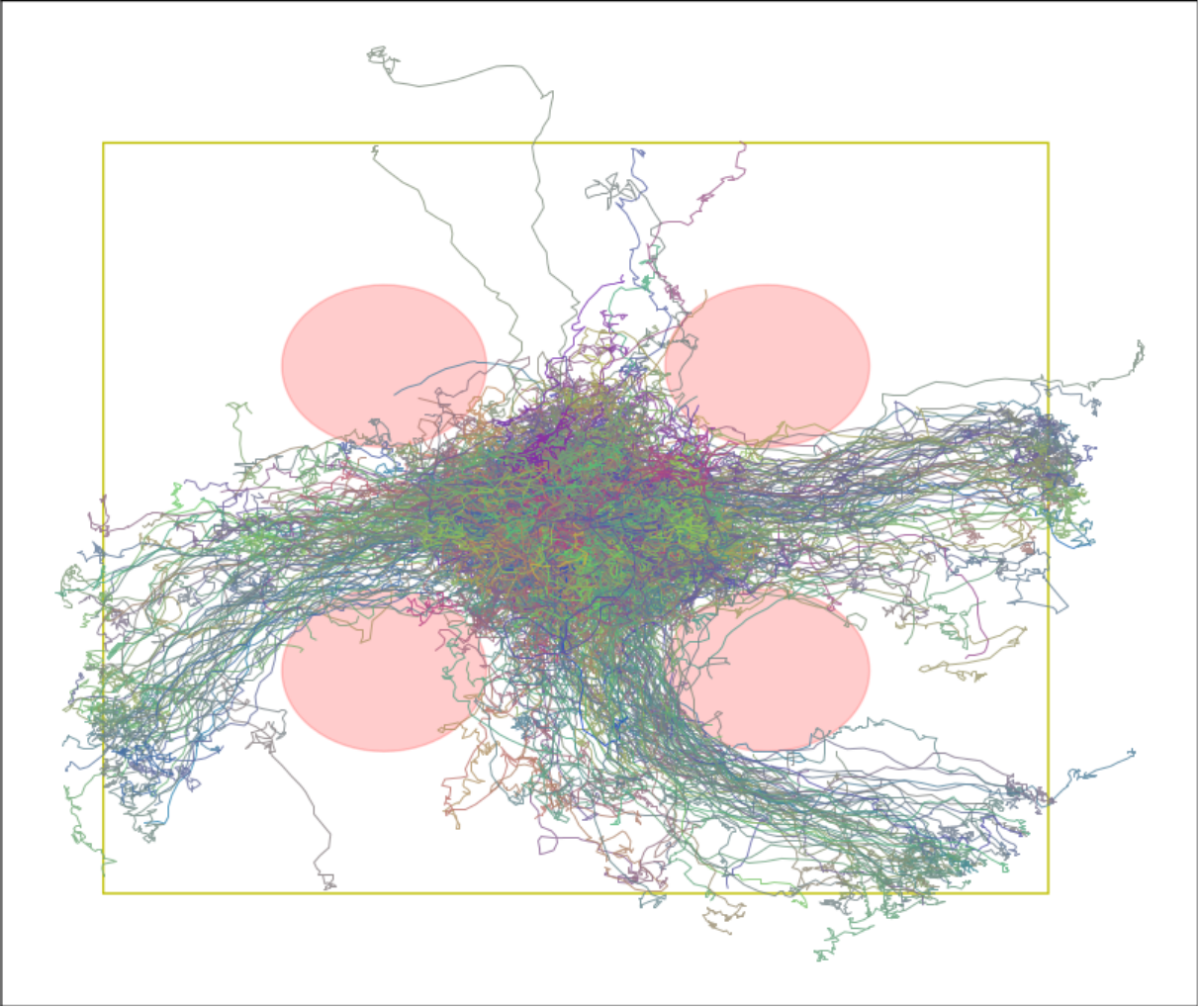} 
    \includegraphics[width=0.14\textwidth]{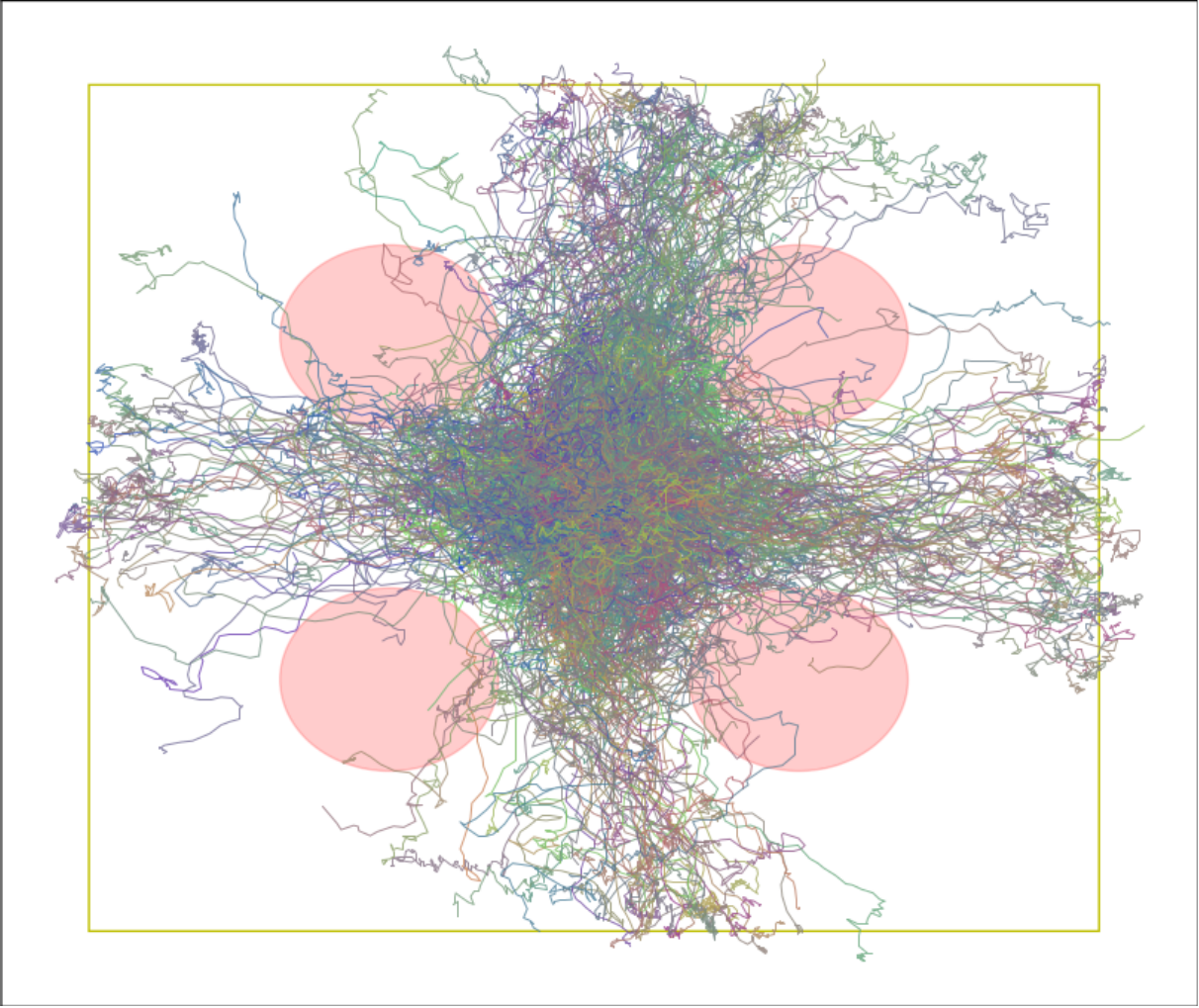} 
    \includegraphics[width=0.14\textwidth]{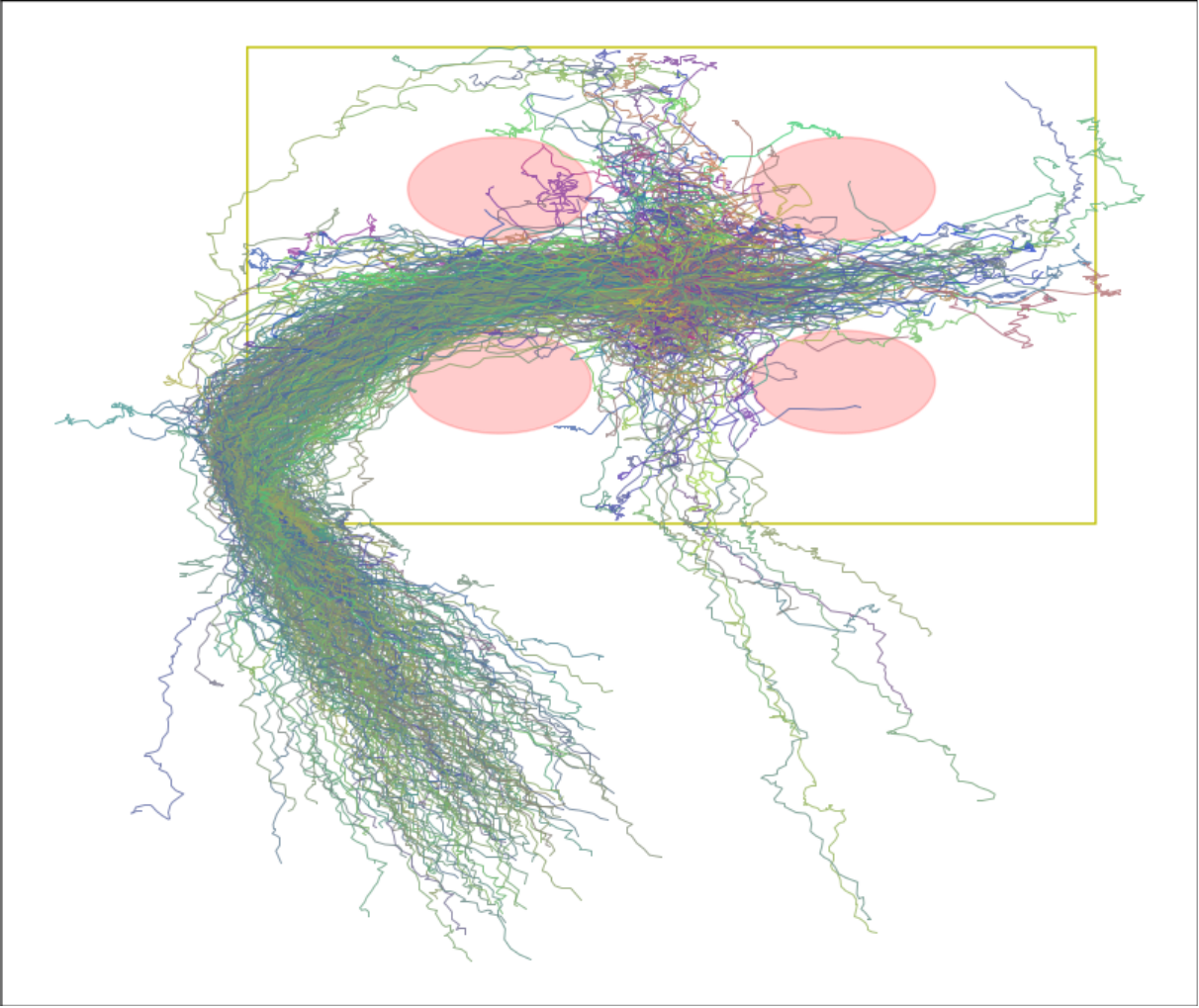} 
        \caption{Ant}
        \label{fig: Apdx|hr1|qr|ant}
    \end{subfigure}
    \caption{Qualitative results of HaSD (5seeds) in the Hazard Room environment with with each agents. After sampling 1000 skills, the first row shows the agent's trajectory.}
    \label{fig: Apdx|hr1|qr|all}
\end{figure}

\begin{figure}[htbp]
    \begin{subfigure}[b]{1\textwidth}
        \centering
        \includegraphics[width=0.13\textwidth]{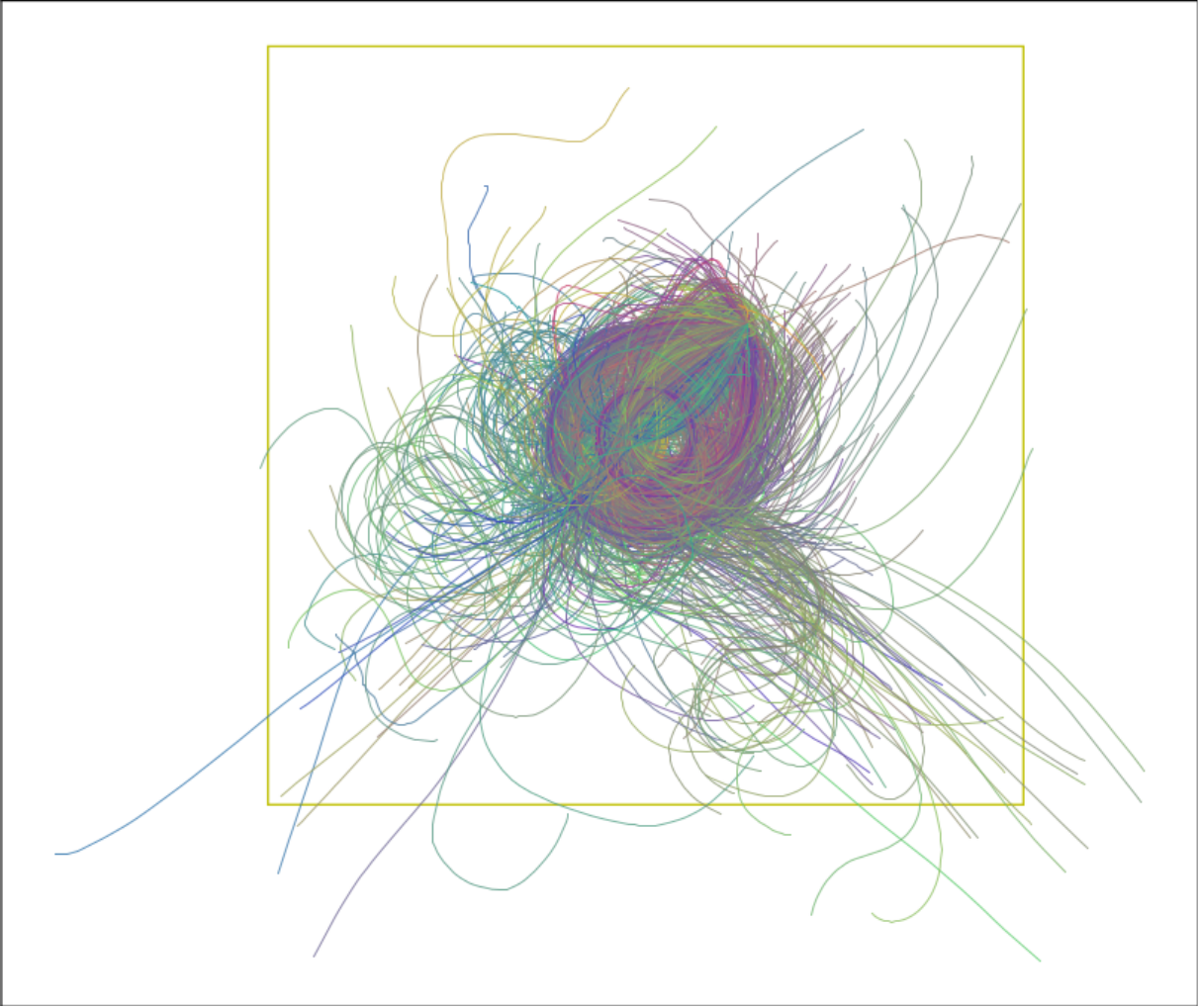} 
        \includegraphics[width=0.13\textwidth]{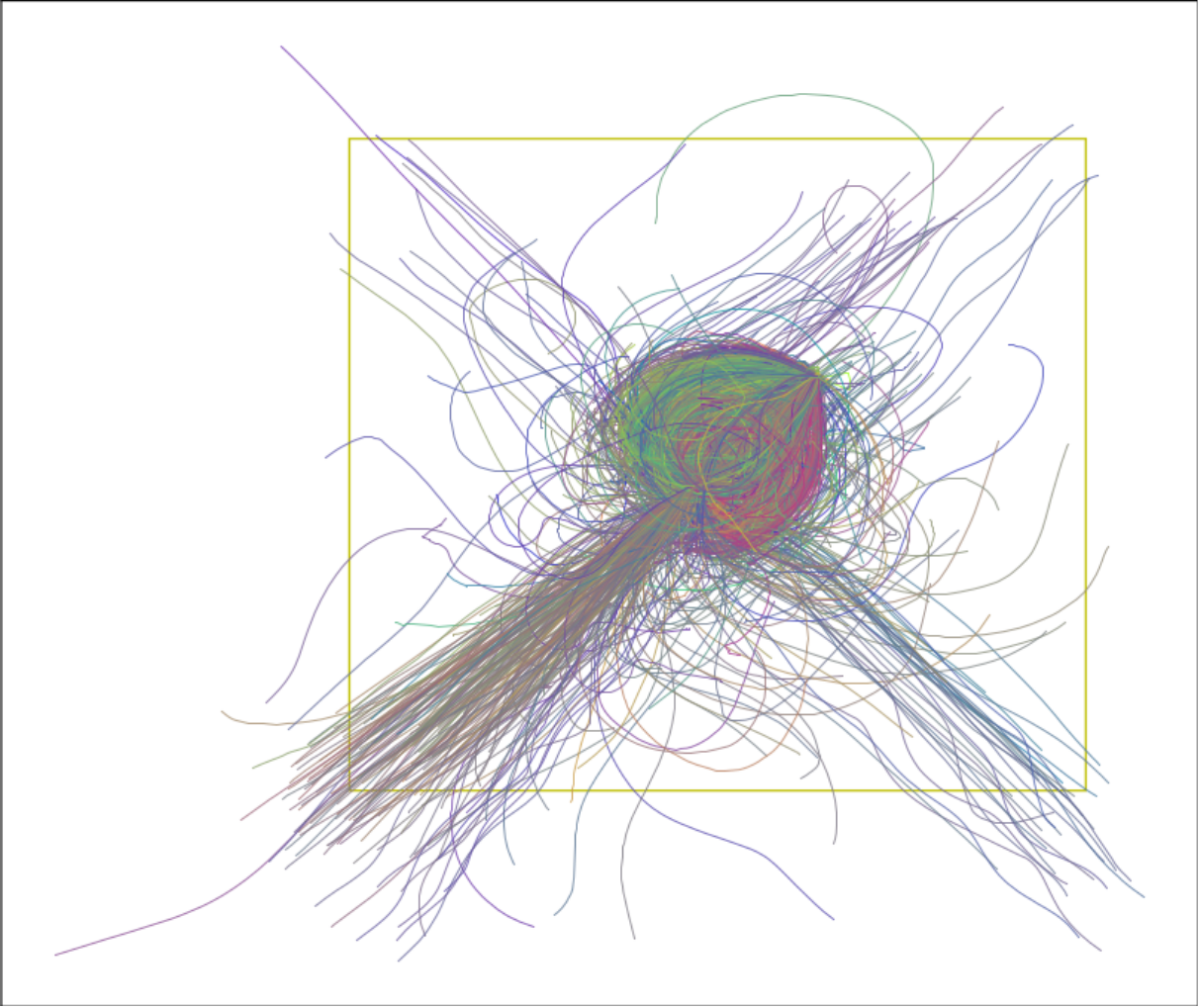}  
        \includegraphics[width=0.13\textwidth]{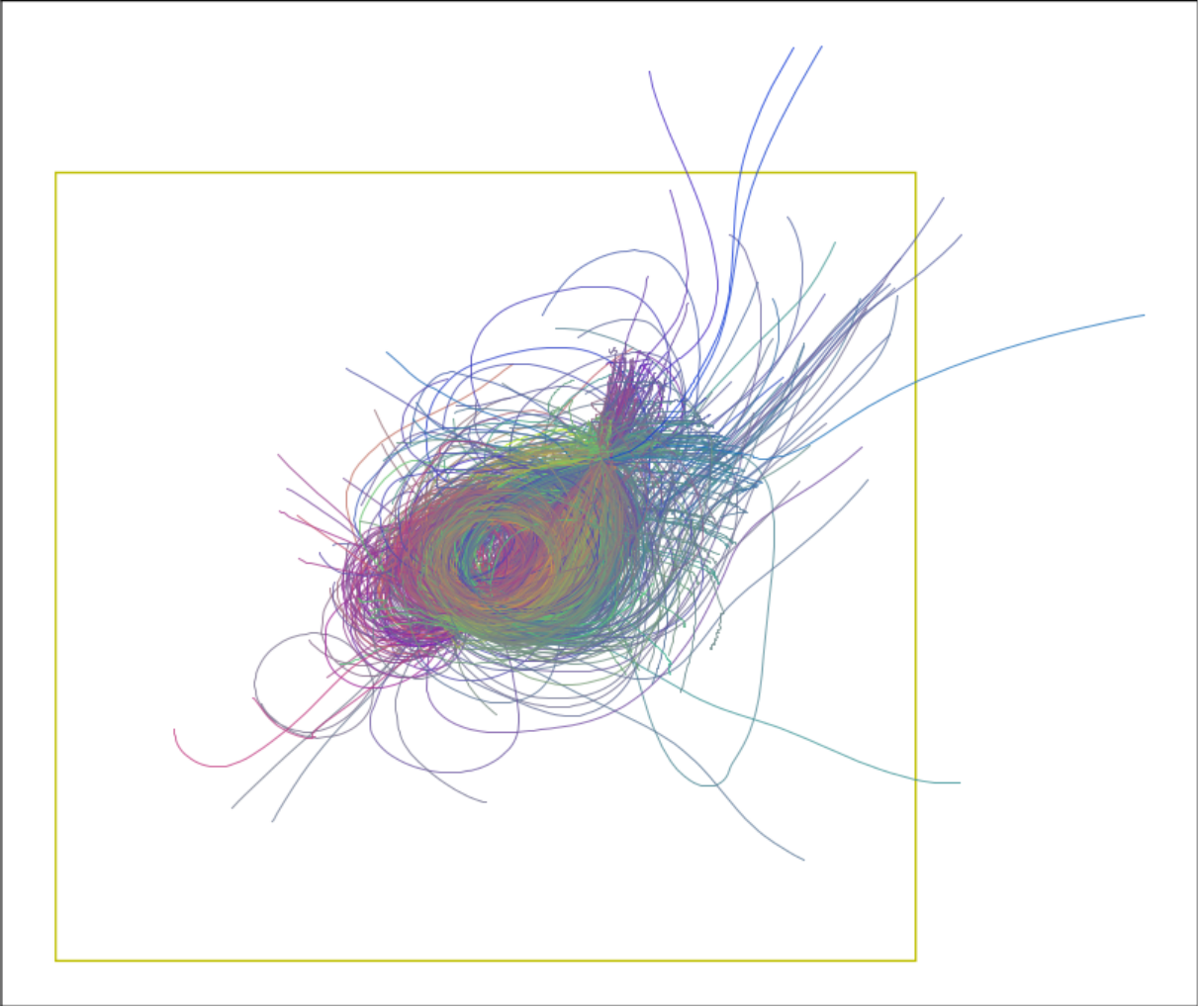} 
        \includegraphics[width=0.13\textwidth]{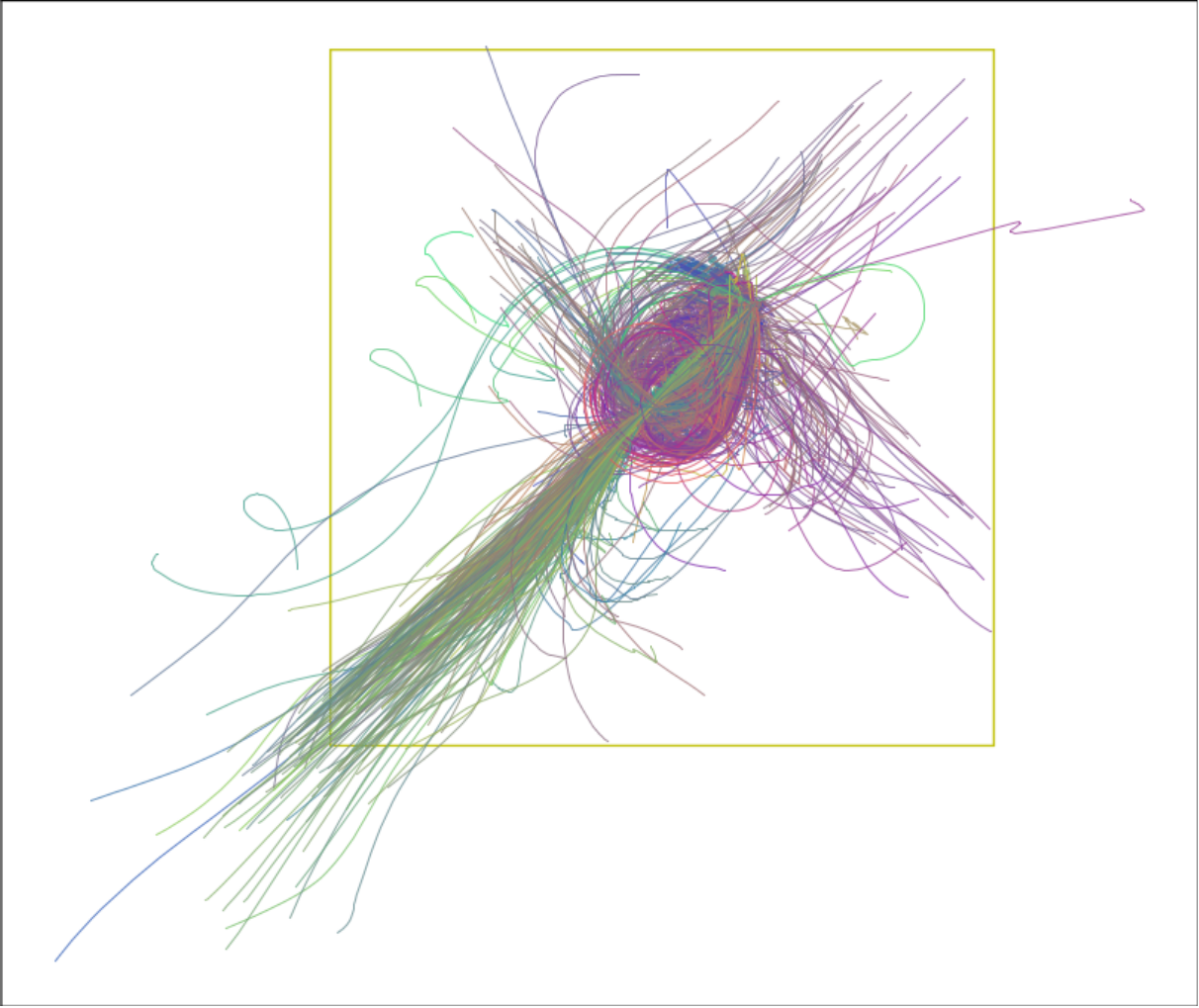}  \\
        \includegraphics[width=0.13\textwidth]{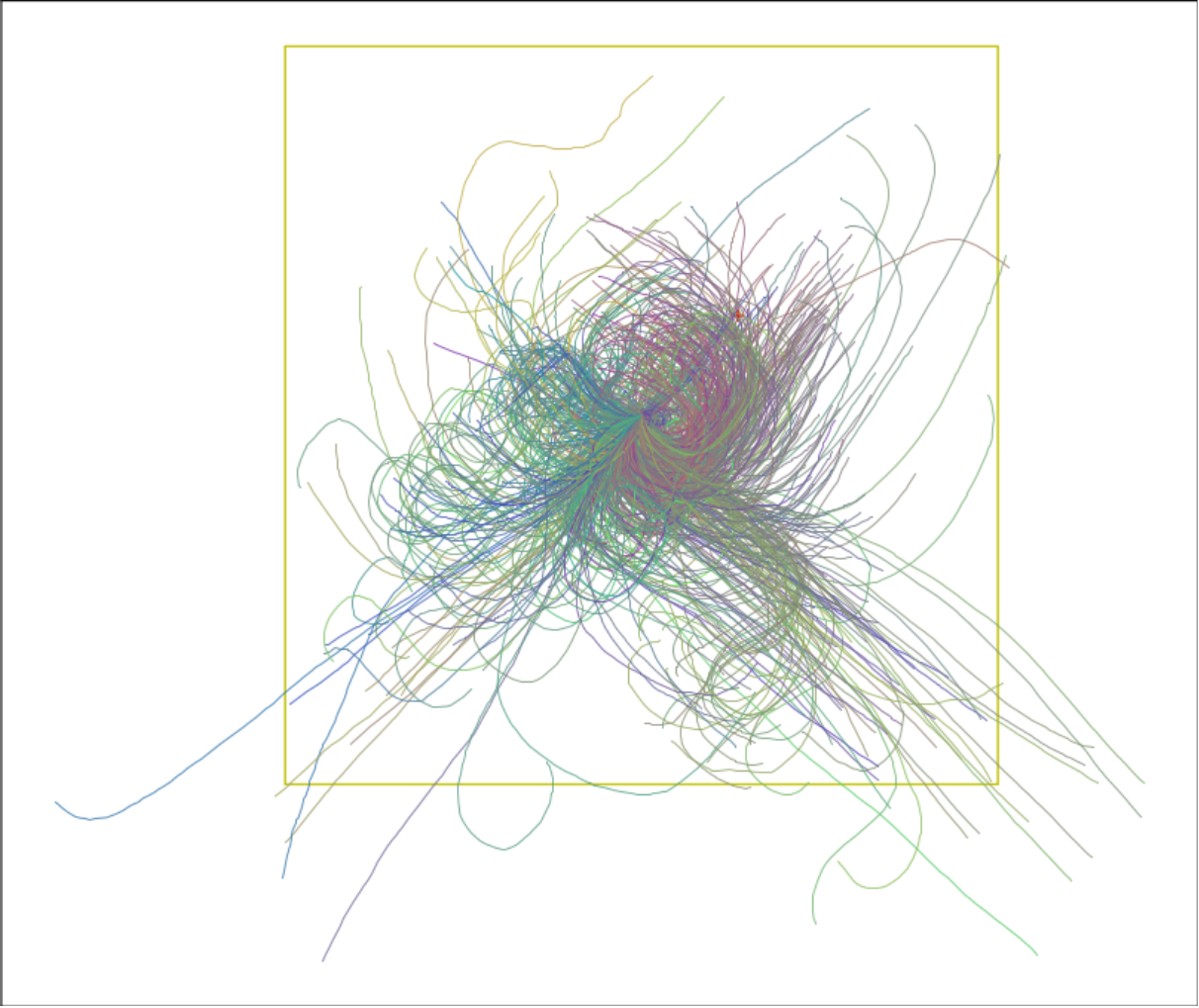} 
        \includegraphics[width=0.13\textwidth]{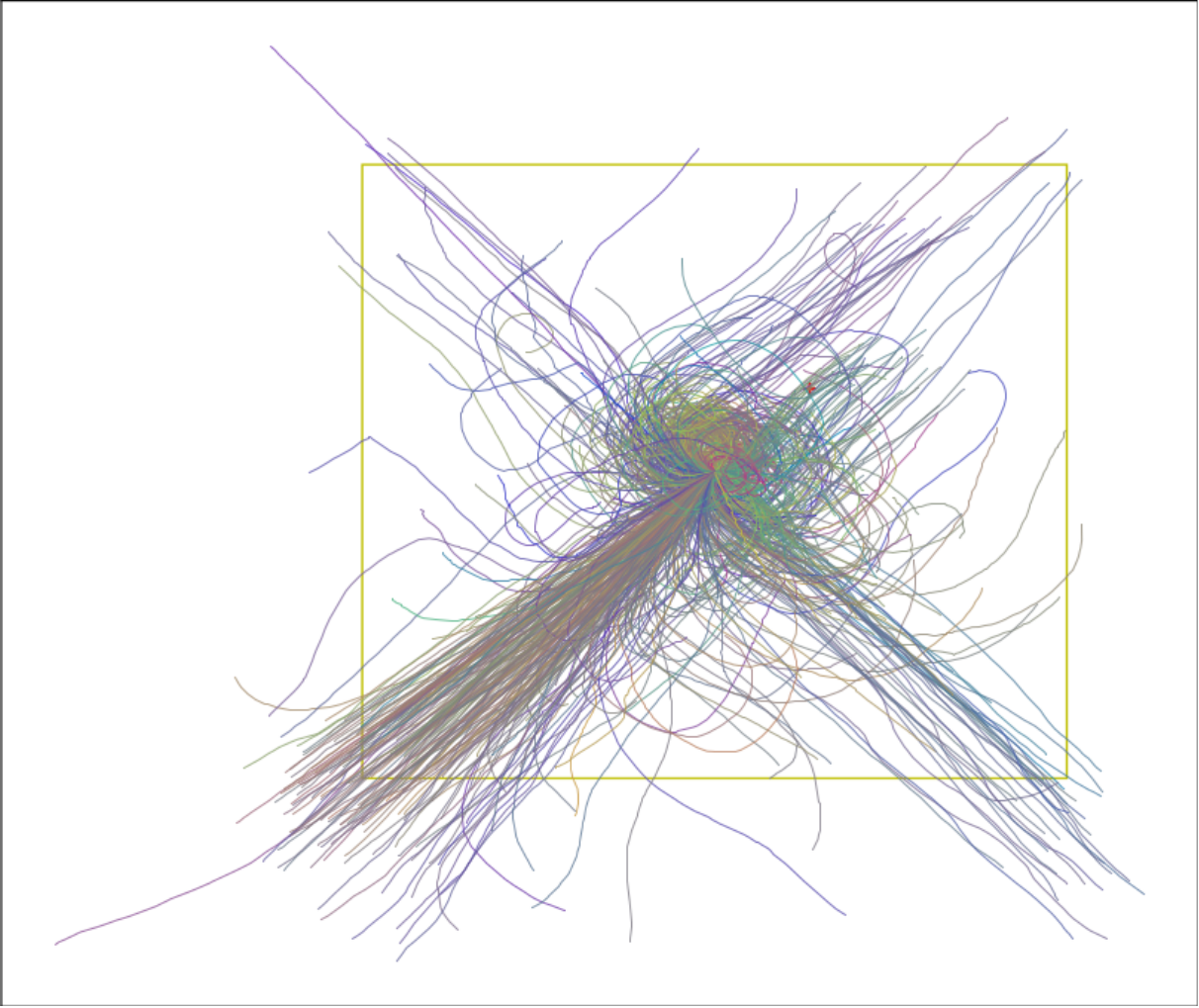}  
        \includegraphics[width=0.13\textwidth]{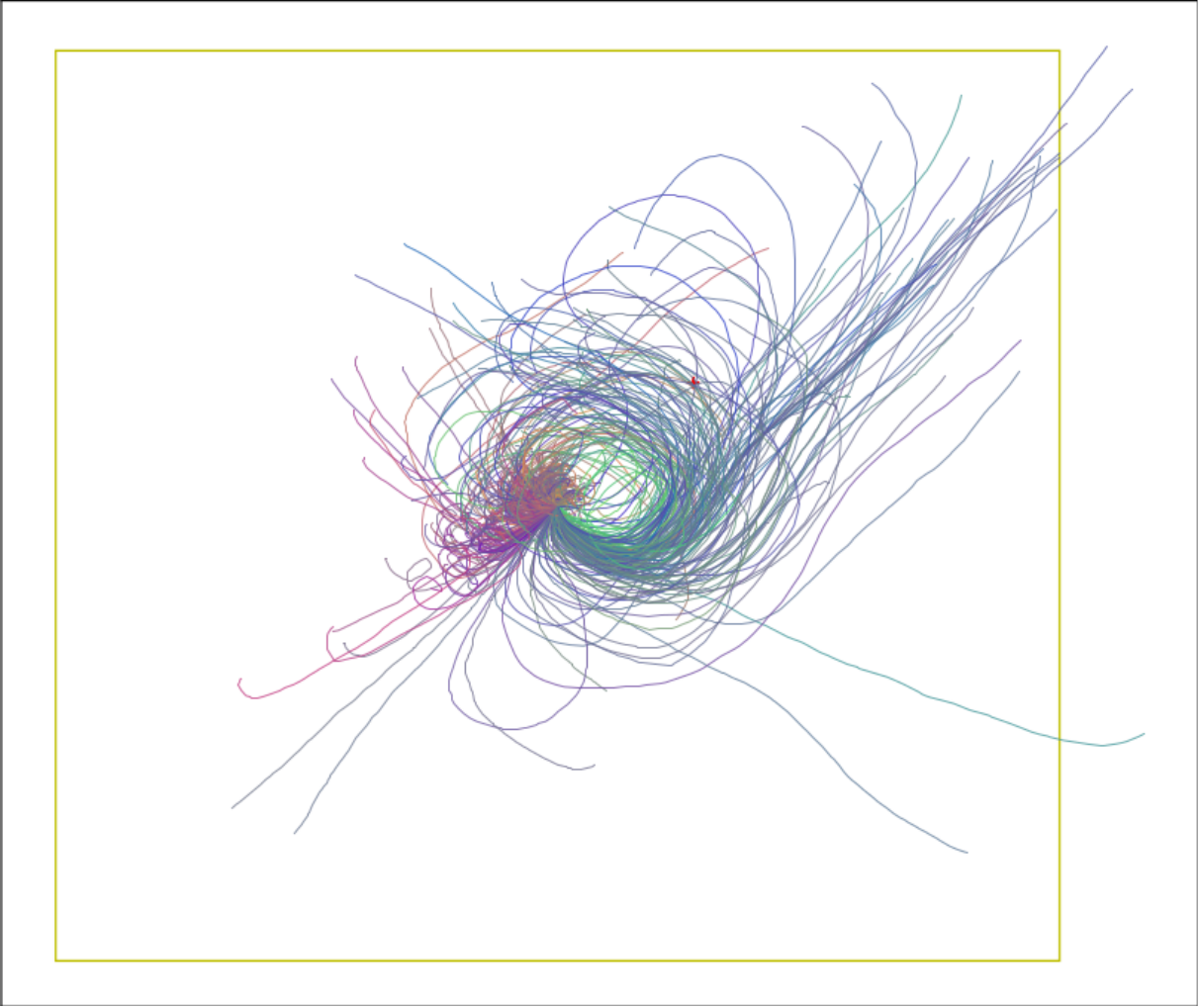} 
        \includegraphics[width=0.13\textwidth]{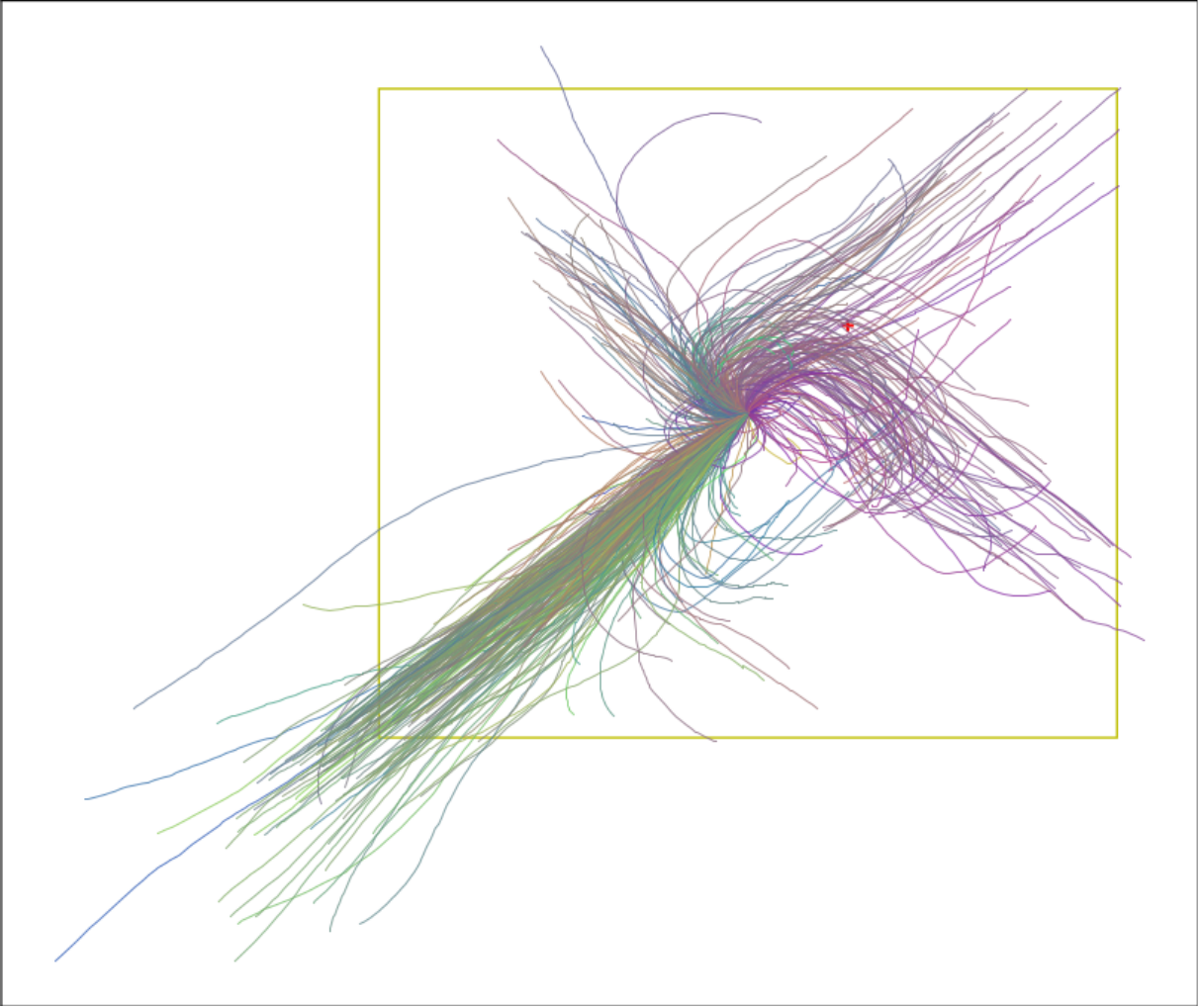}  
        \caption{Point}
        \label{fig: Apdx|hr3|qr|point}
    \end{subfigure}
    \begin{subfigure}[b]{1\textwidth}
        \centering
        \includegraphics[width=0.13\textwidth]{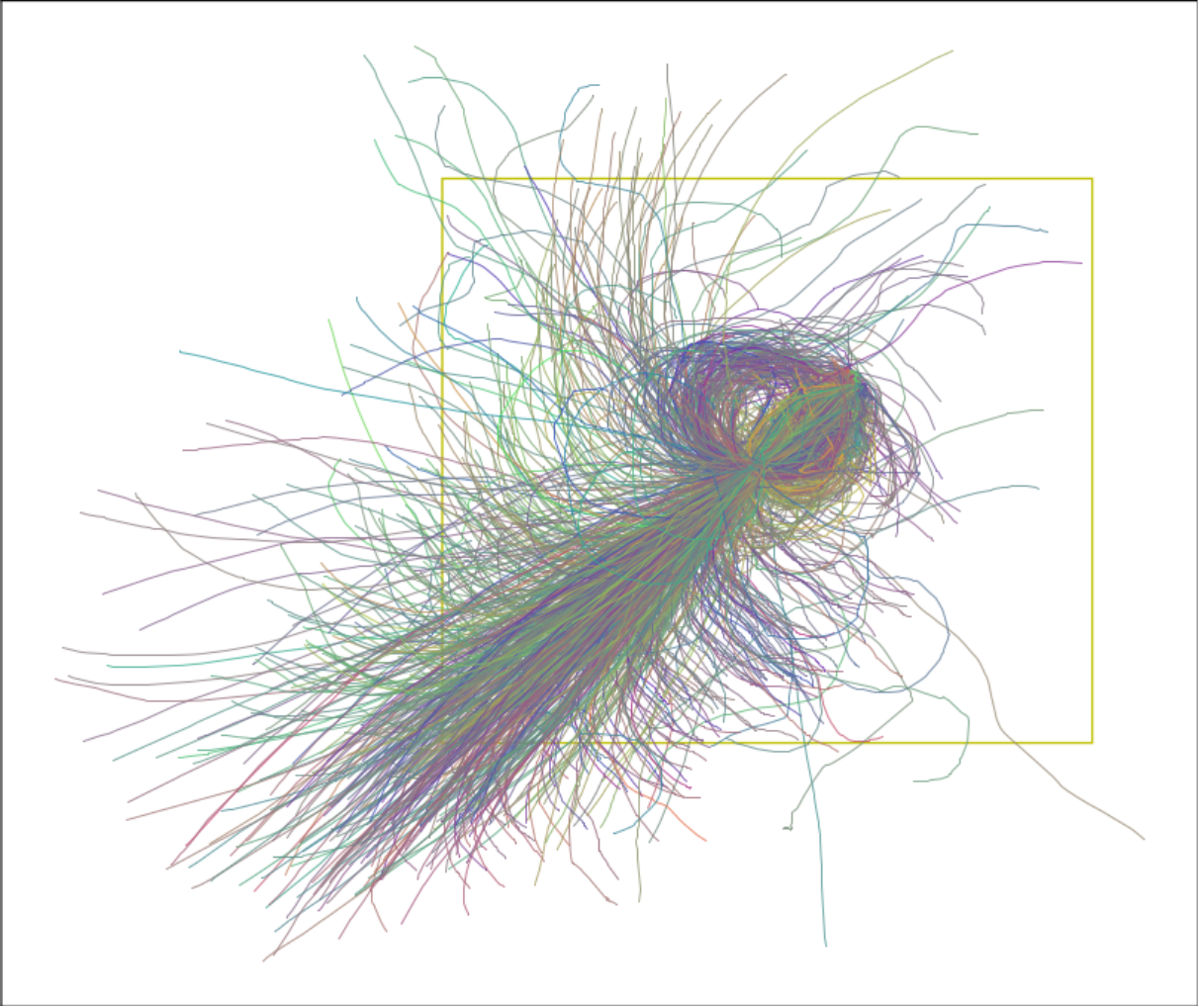} 
        \includegraphics[width=0.13\textwidth]{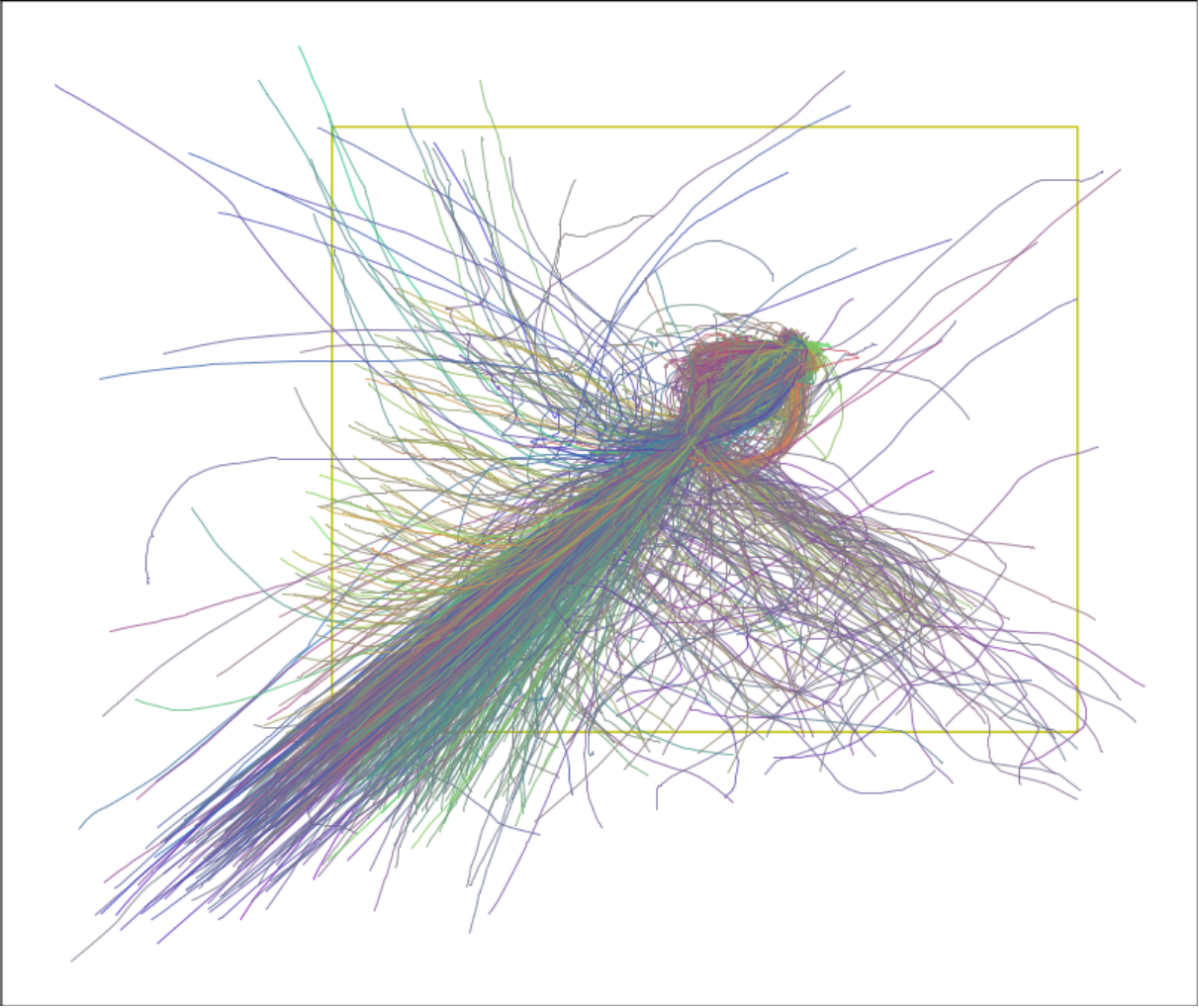} 
        \includegraphics[width=0.13\textwidth]{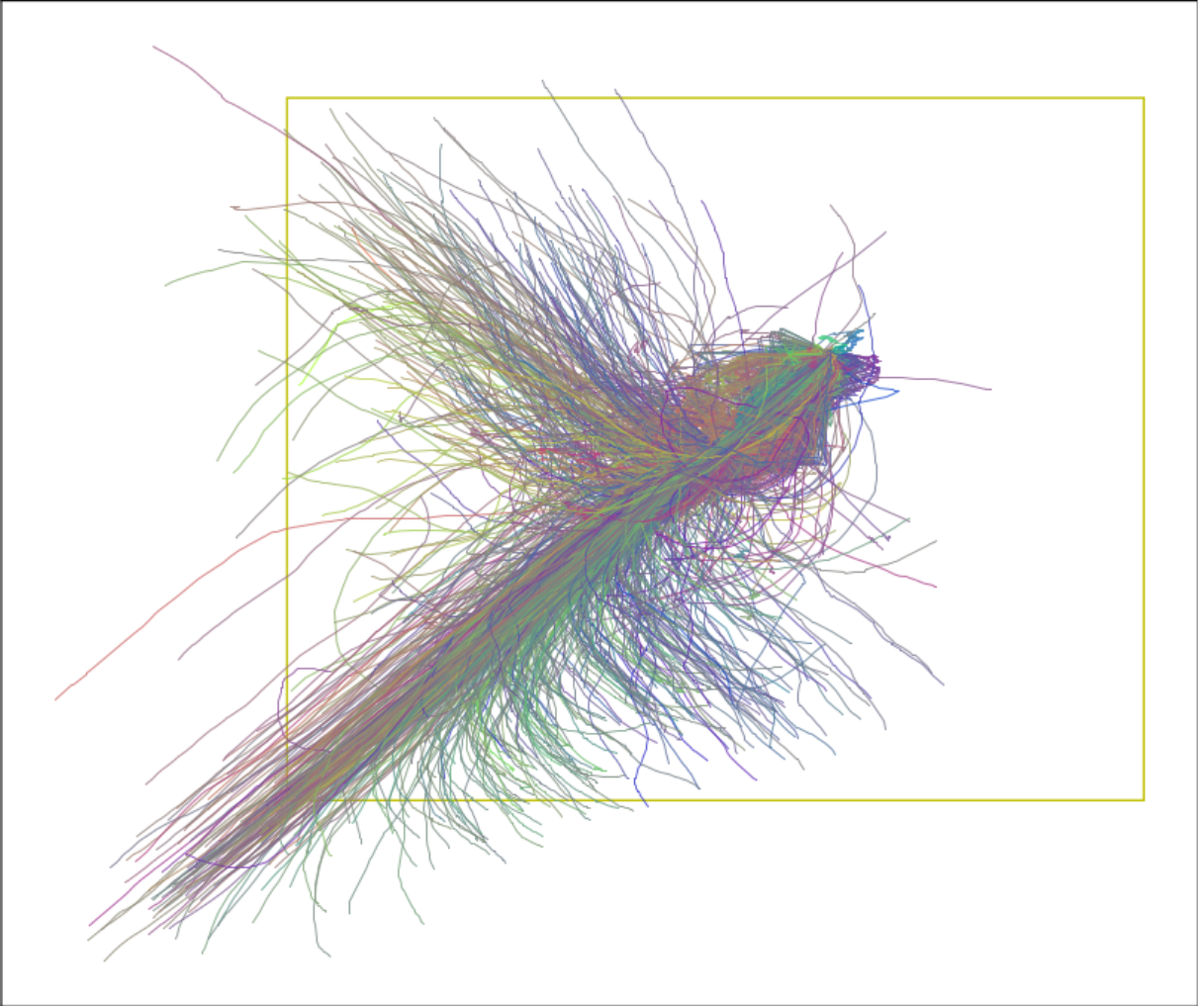}  
        \includegraphics[width=0.13\textwidth]{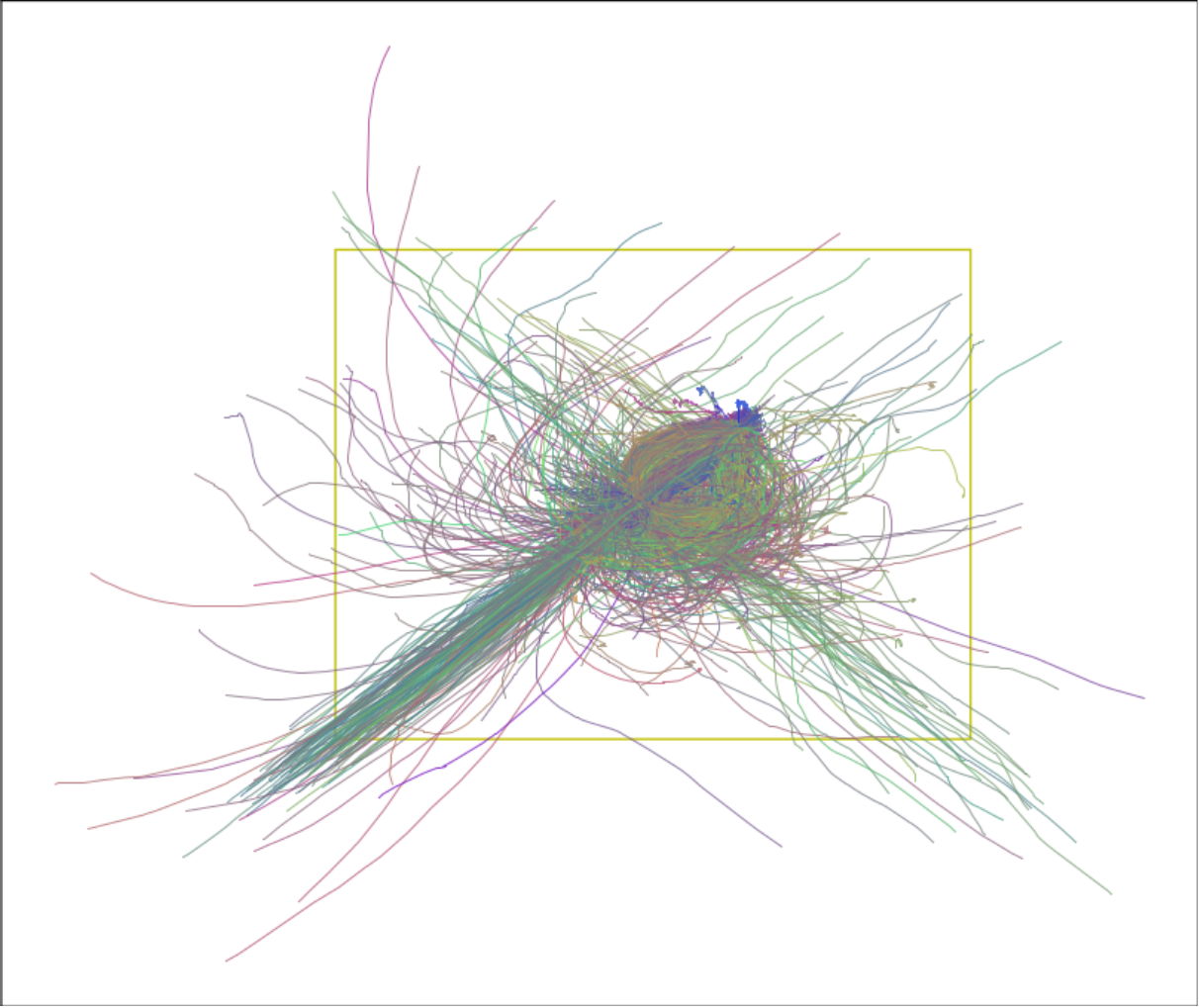}   \\
        \includegraphics[width=0.13\textwidth]{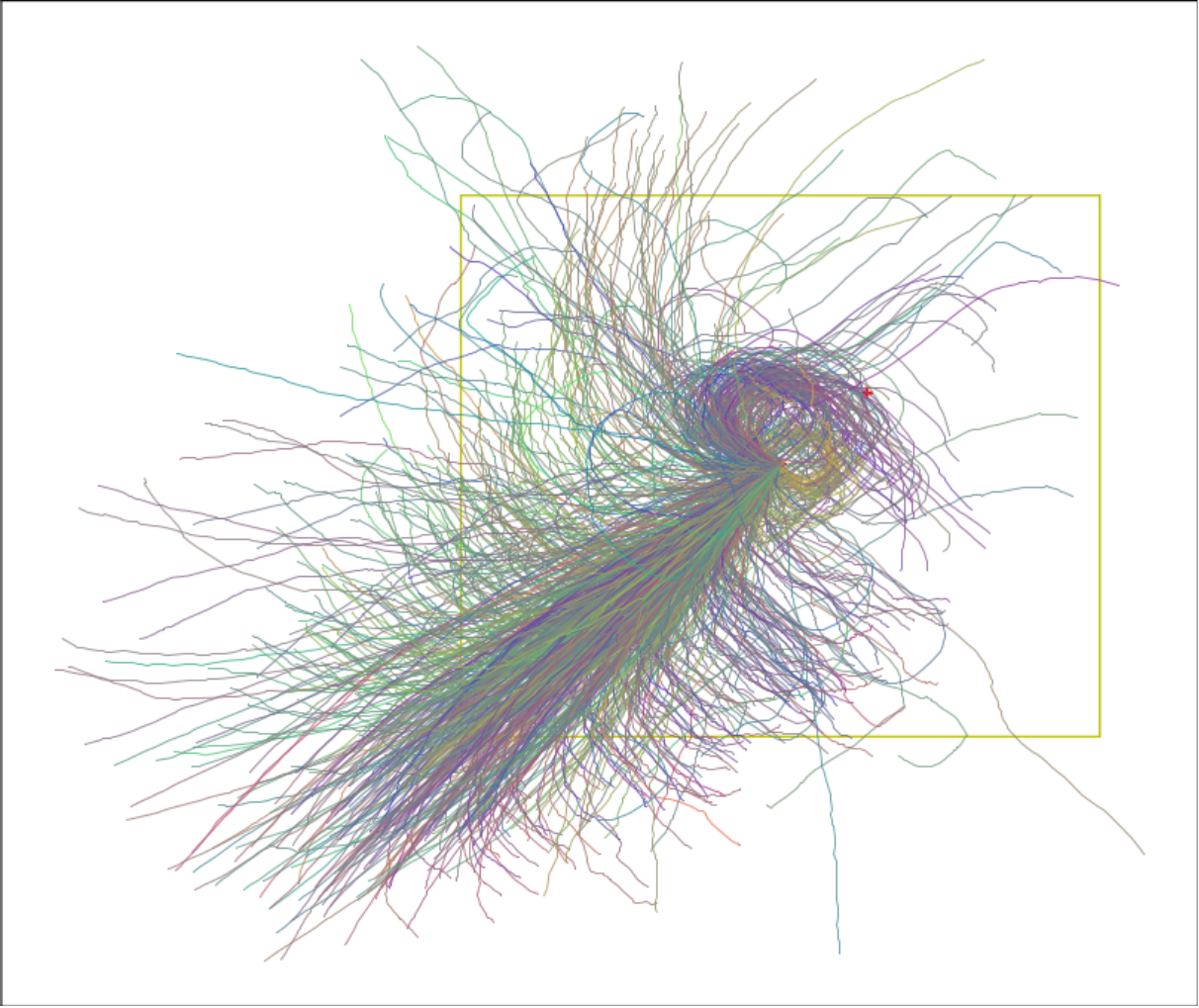} 
        \includegraphics[width=0.13\textwidth]{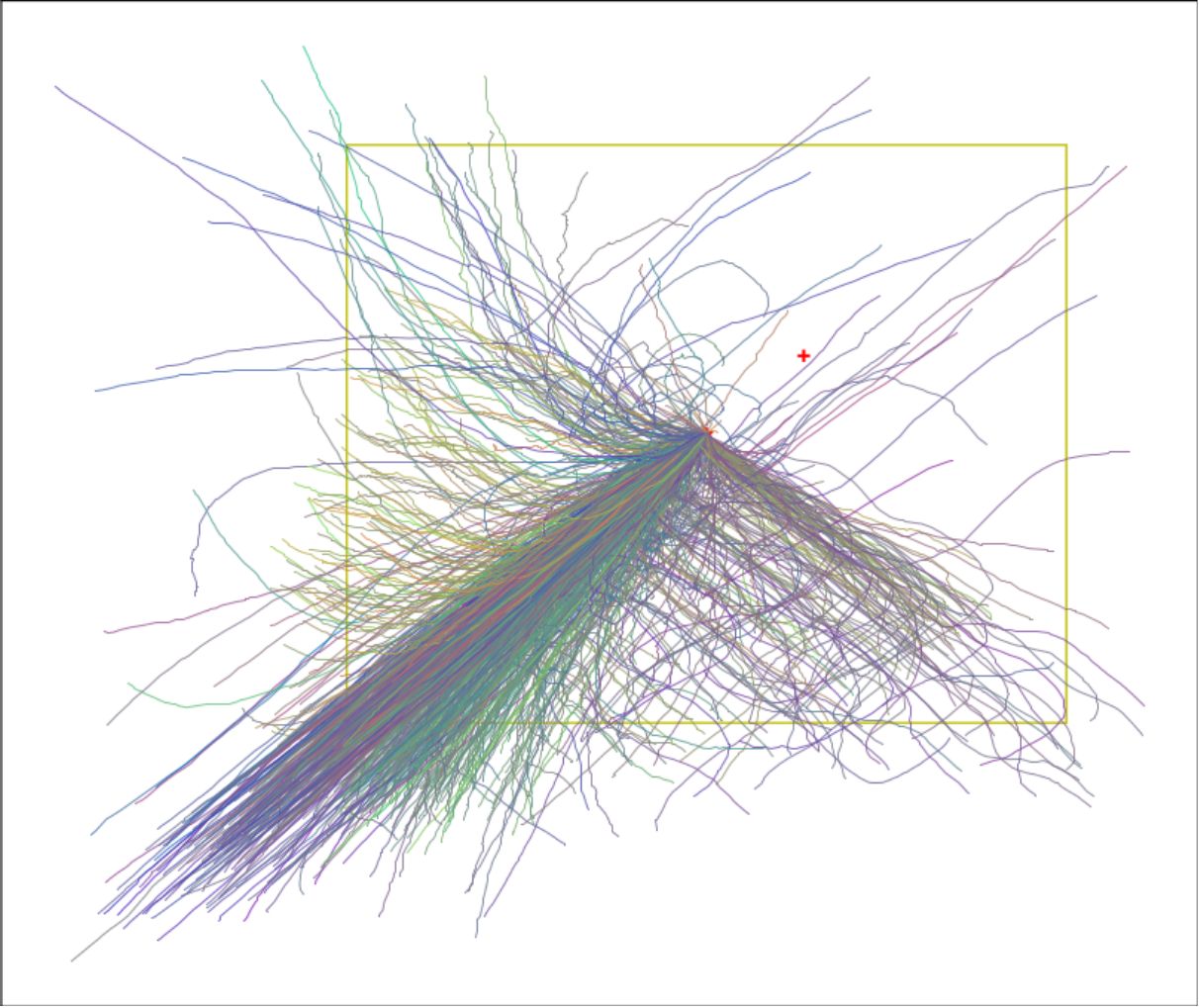} 
        \includegraphics[width=0.13\textwidth]{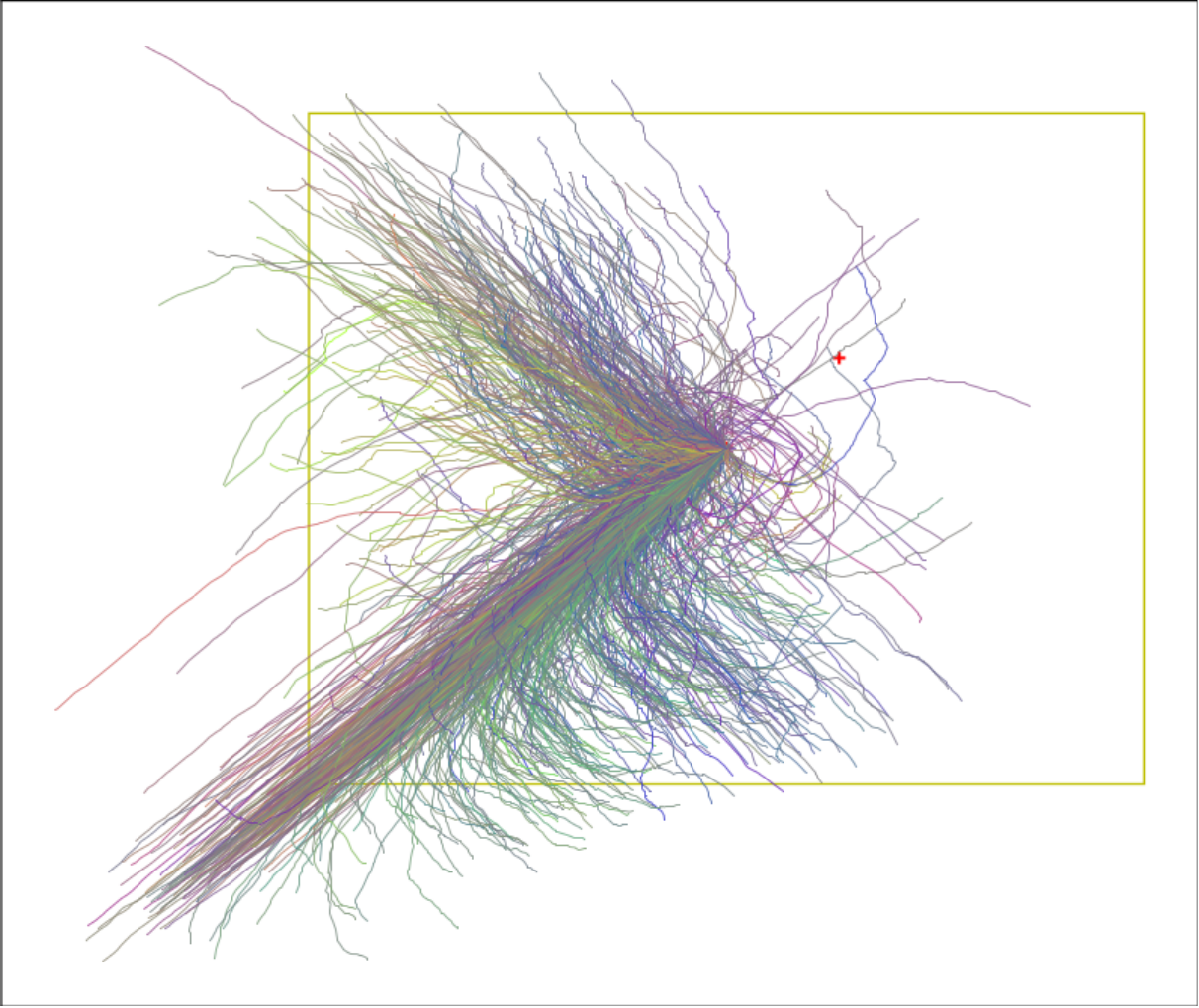}  
        \includegraphics[width=0.13\textwidth]{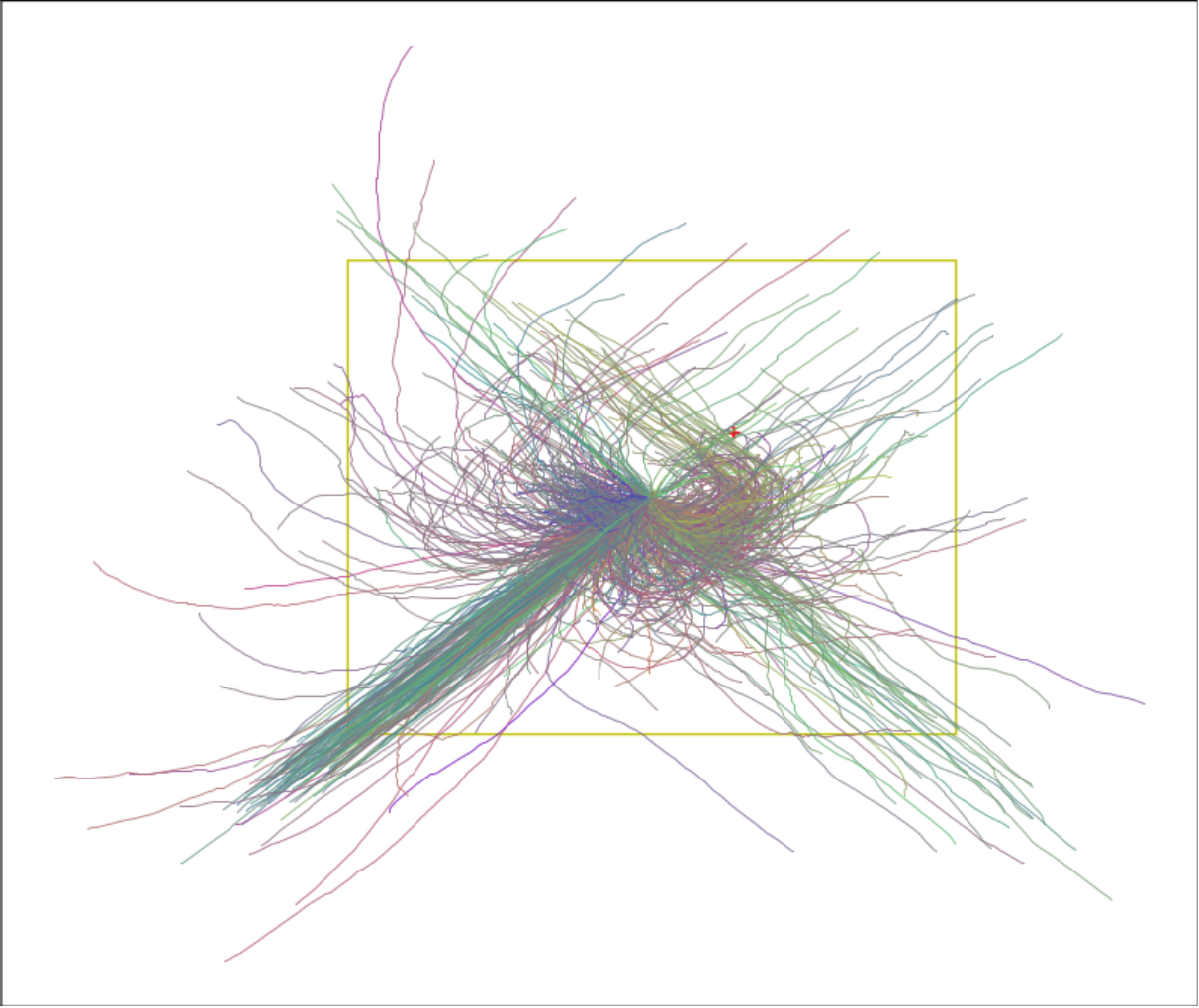} 
        \caption{Car}
        \label{fig: Apdx|hr3|qr|car}
    \end{subfigure}
     \begin{subfigure}[b]{1\textwidth}
        \centering
    \includegraphics[width=0.13\textwidth]{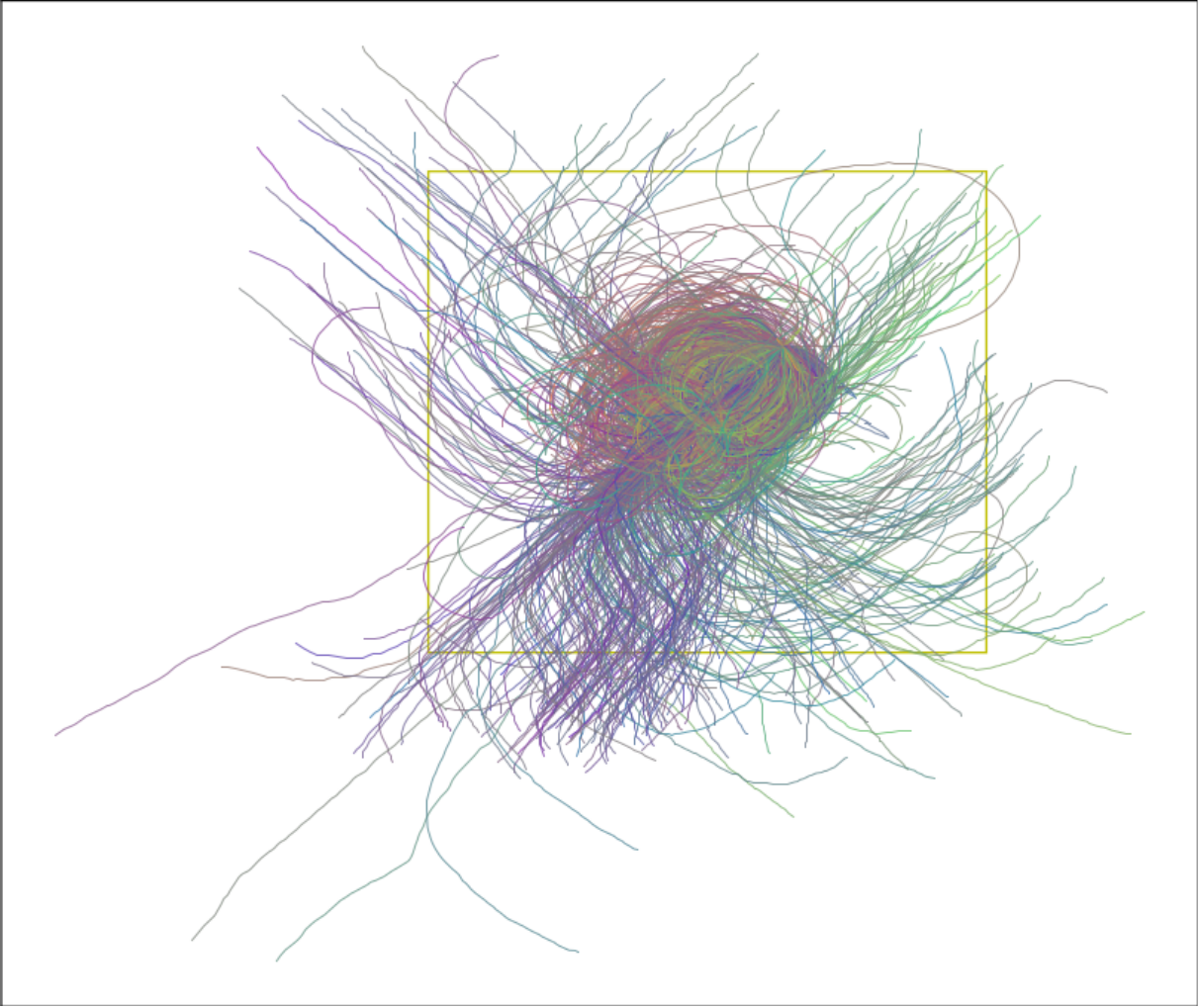} 
    \includegraphics[width=0.13\textwidth]{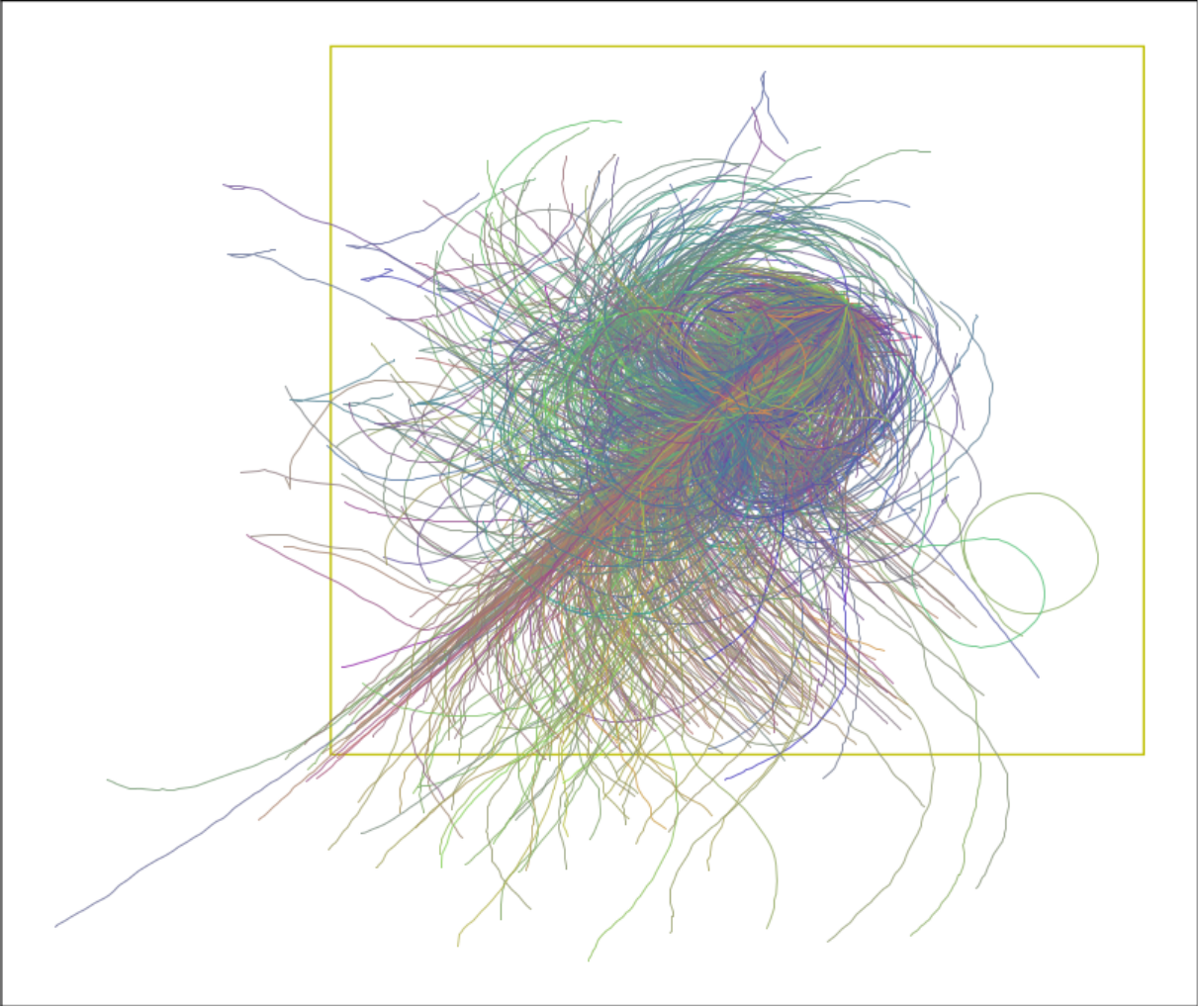} 
    \includegraphics[width=0.13\textwidth]{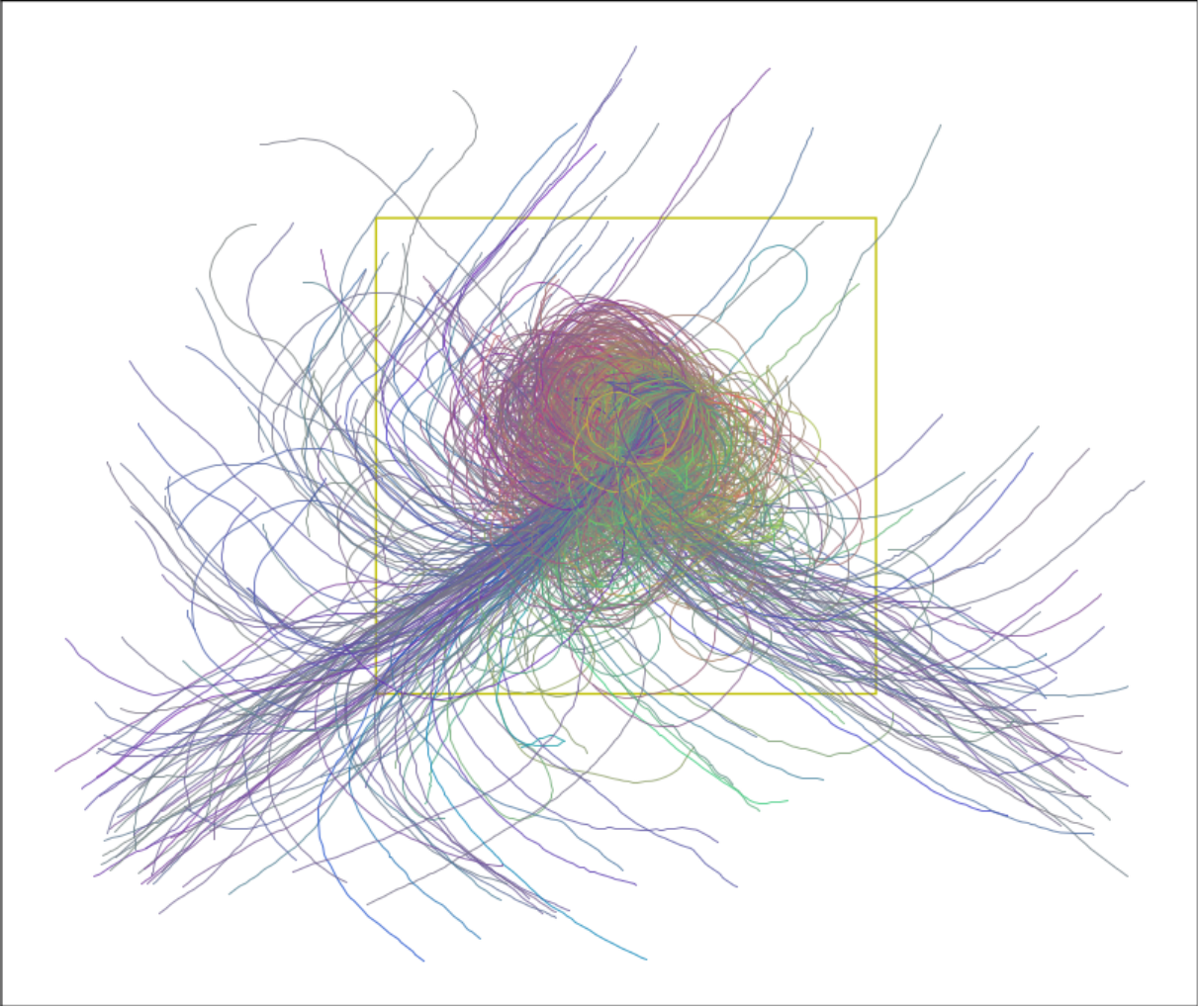}  
    \includegraphics[width=0.13\textwidth]{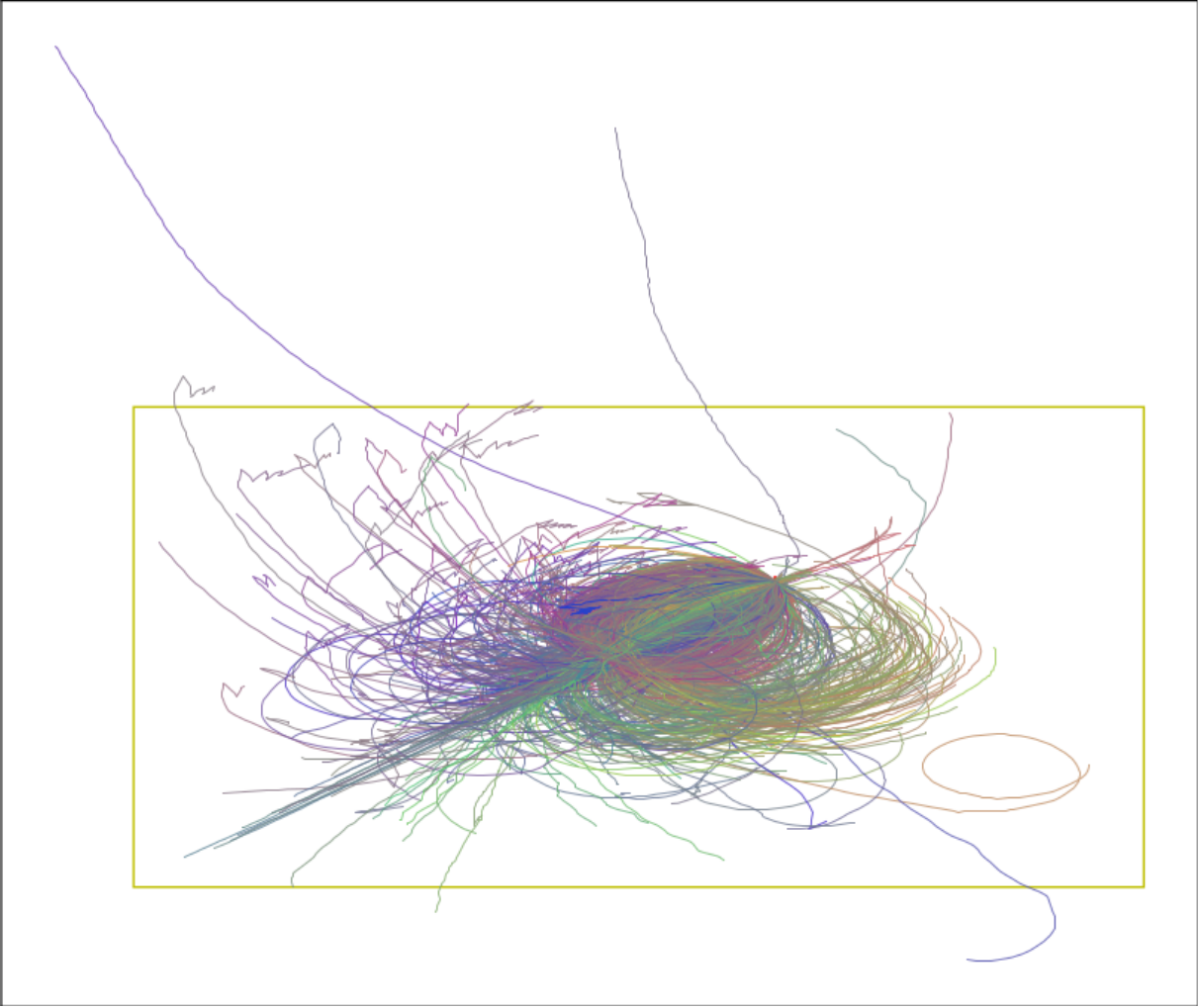}   \\
    \includegraphics[width=0.13\textwidth]{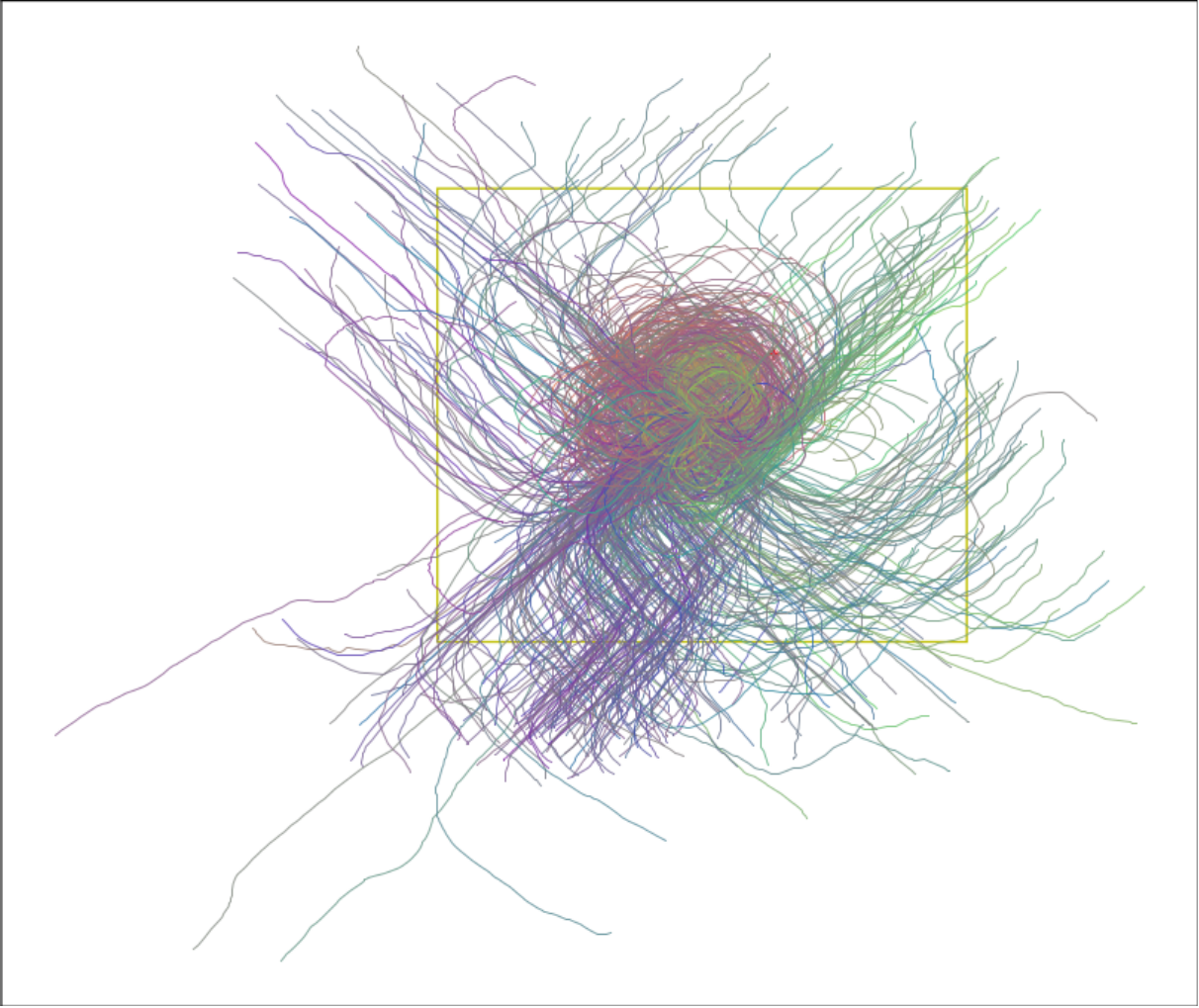} 
    \includegraphics[width=0.13\textwidth]{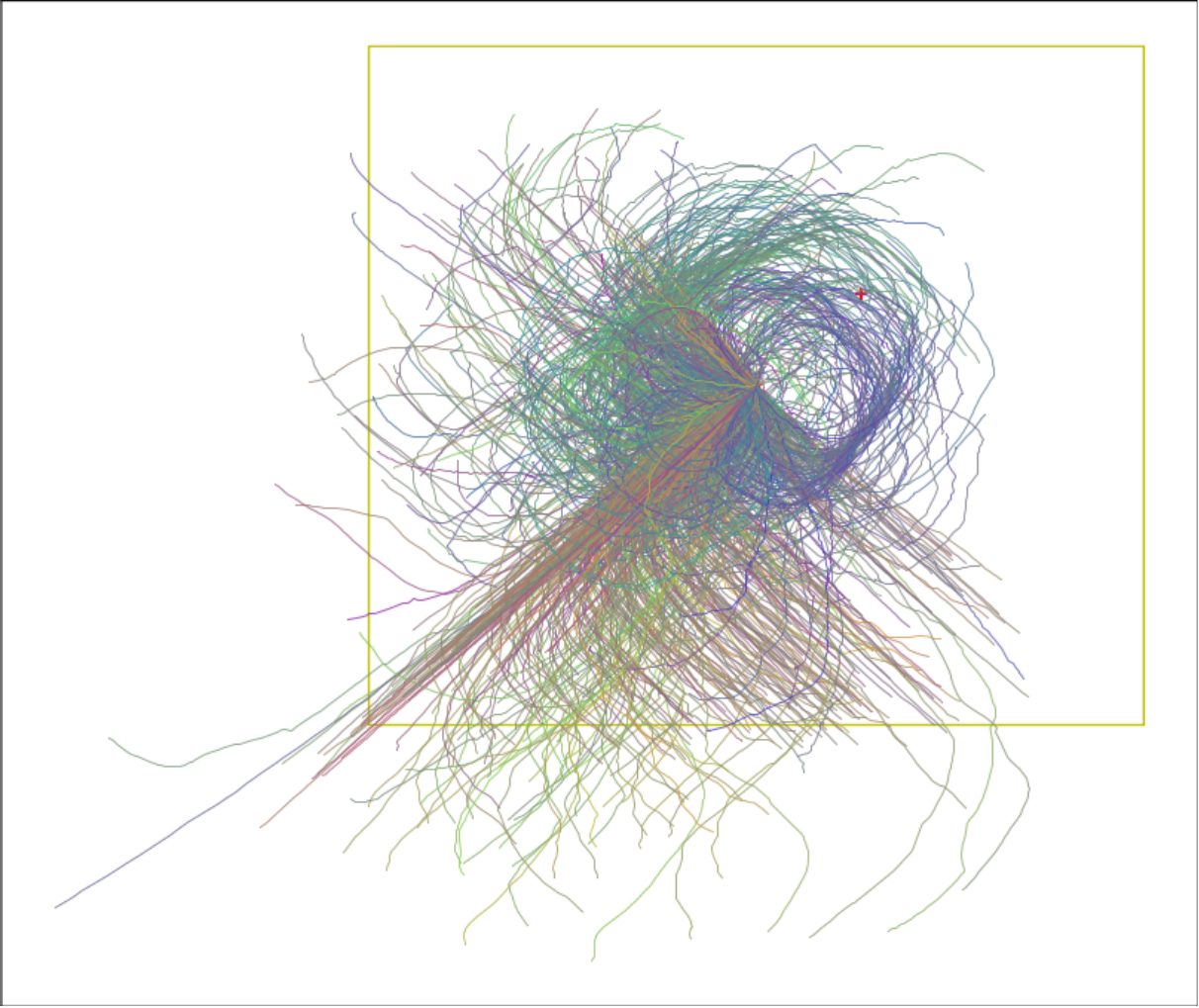} 
    \includegraphics[width=0.13\textwidth]{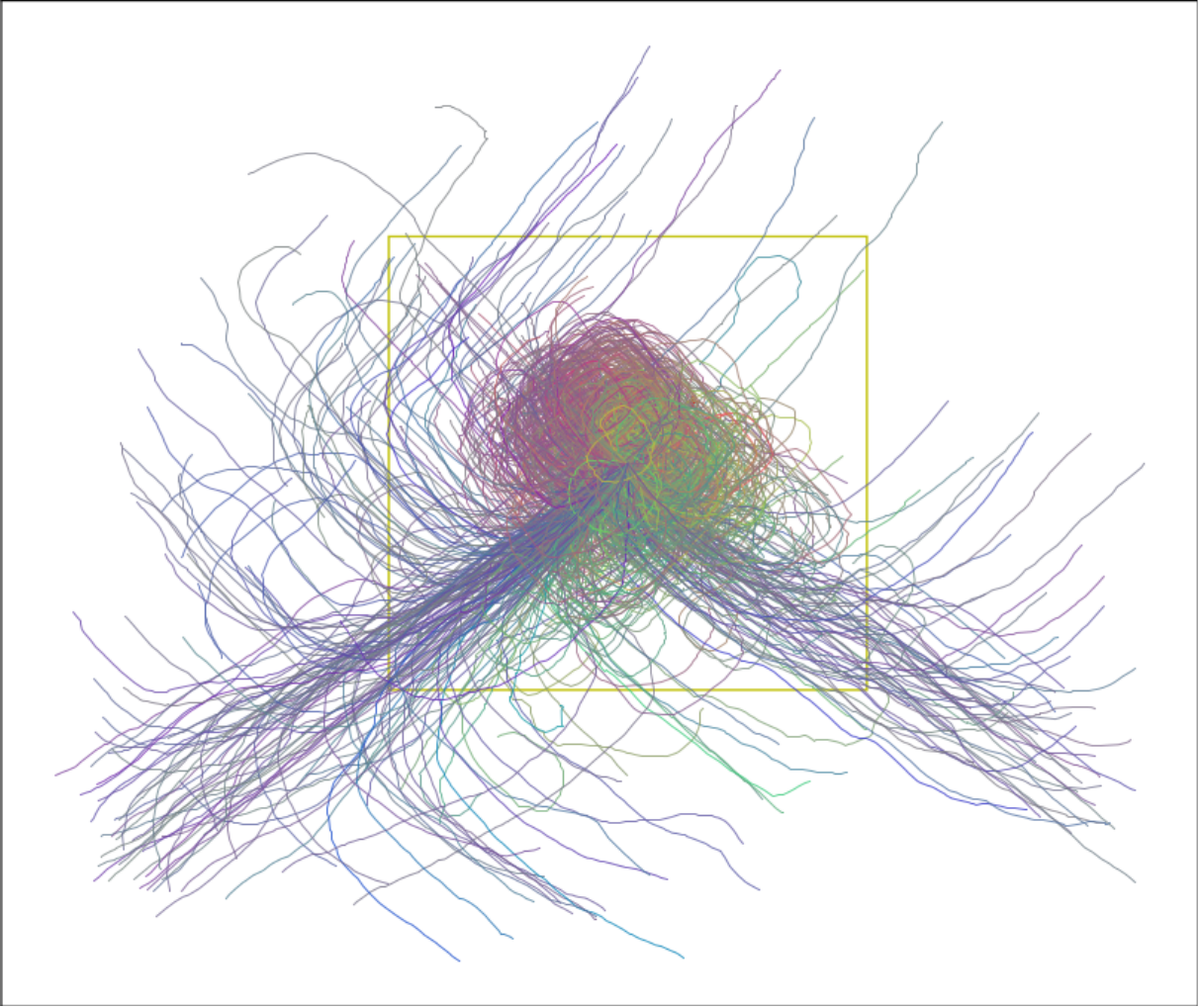}  
    \includegraphics[width=0.13\textwidth]{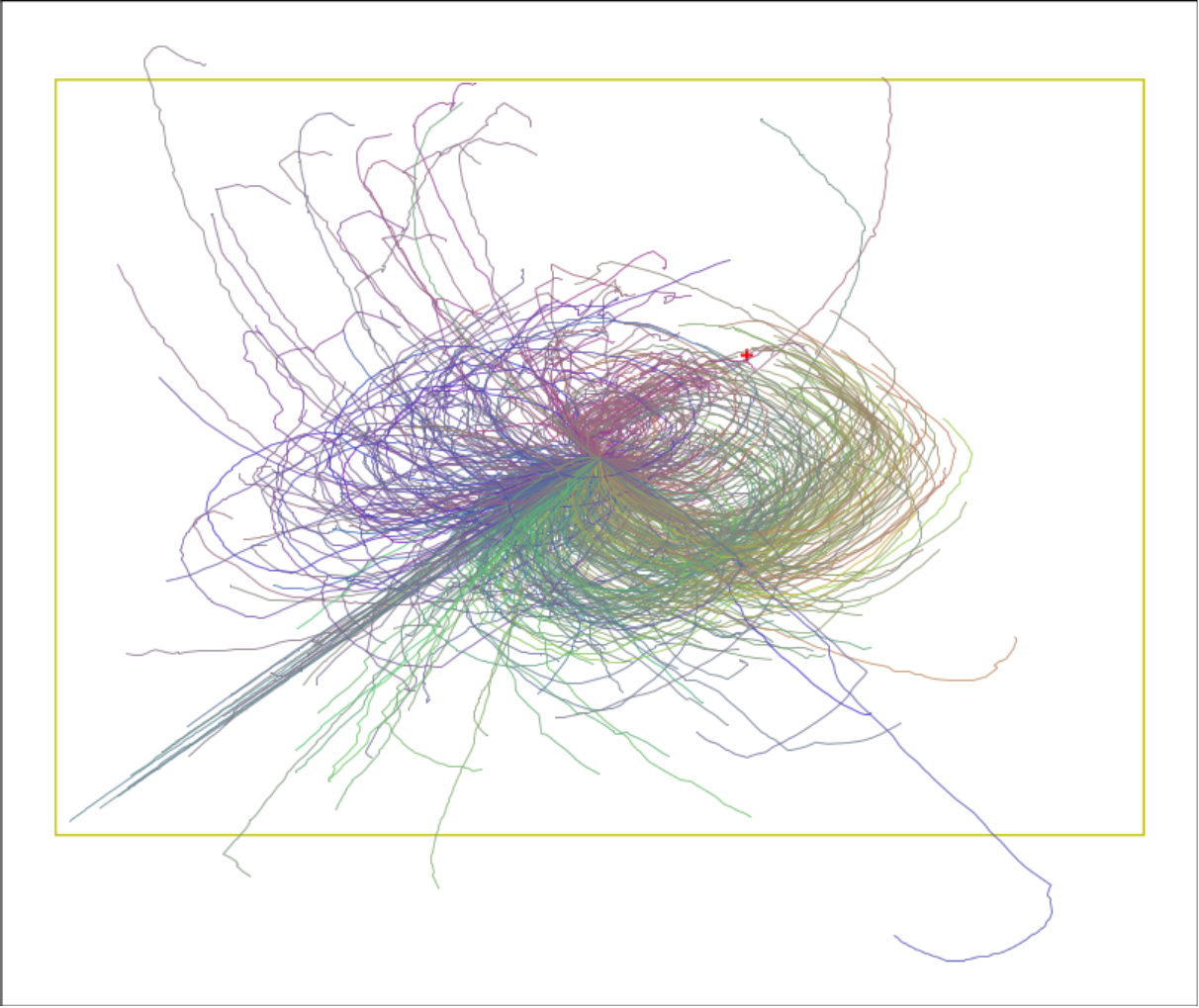} 
        \caption{Racecar}
        \label{fig: Apdx|hr3|qr|race}
    \end{subfigure}
    \begin{subfigure}[b]{1\textwidth}
        \centering
        \includegraphics[width=0.13\textwidth]{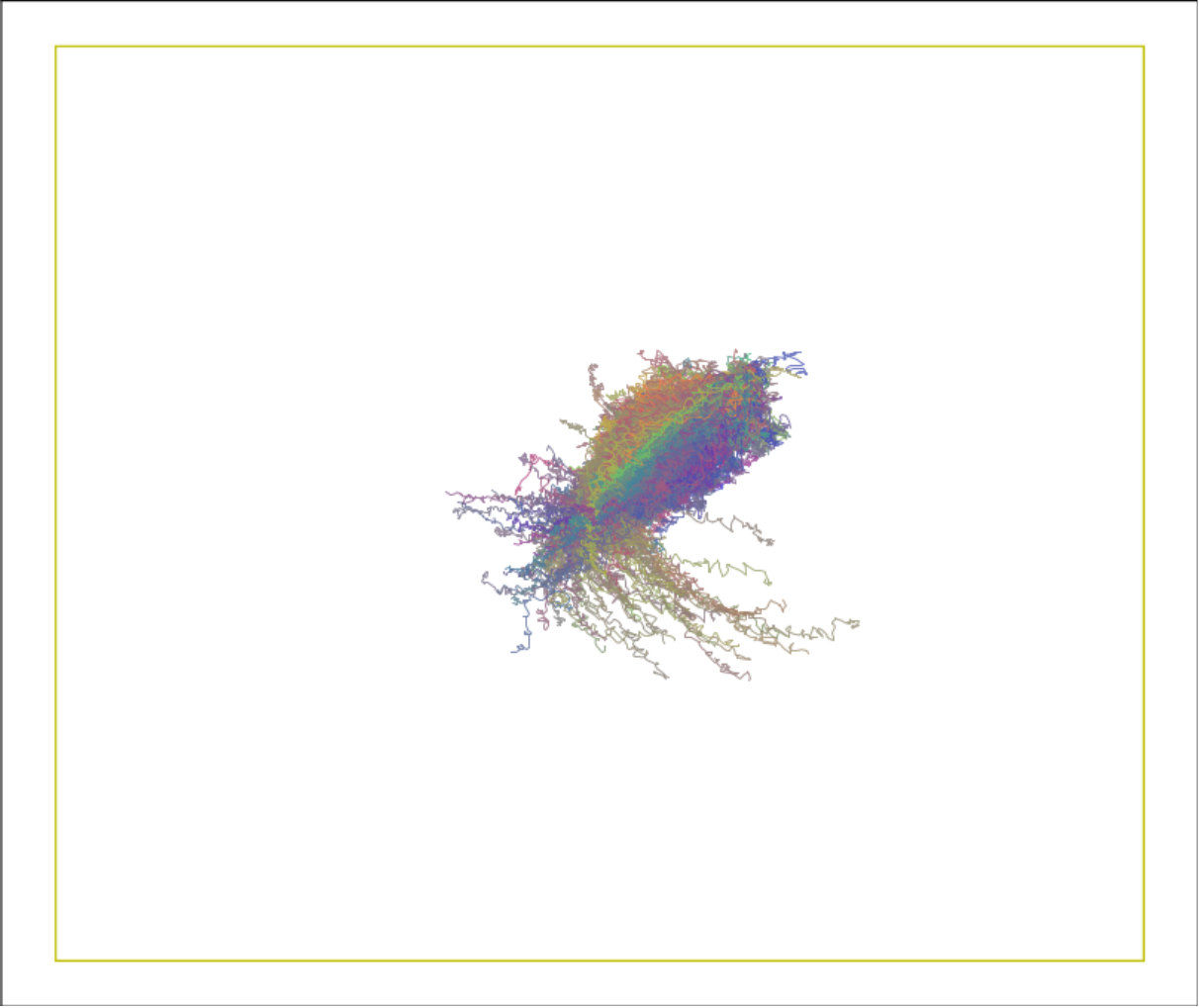} 
        \includegraphics[width=0.13\textwidth]{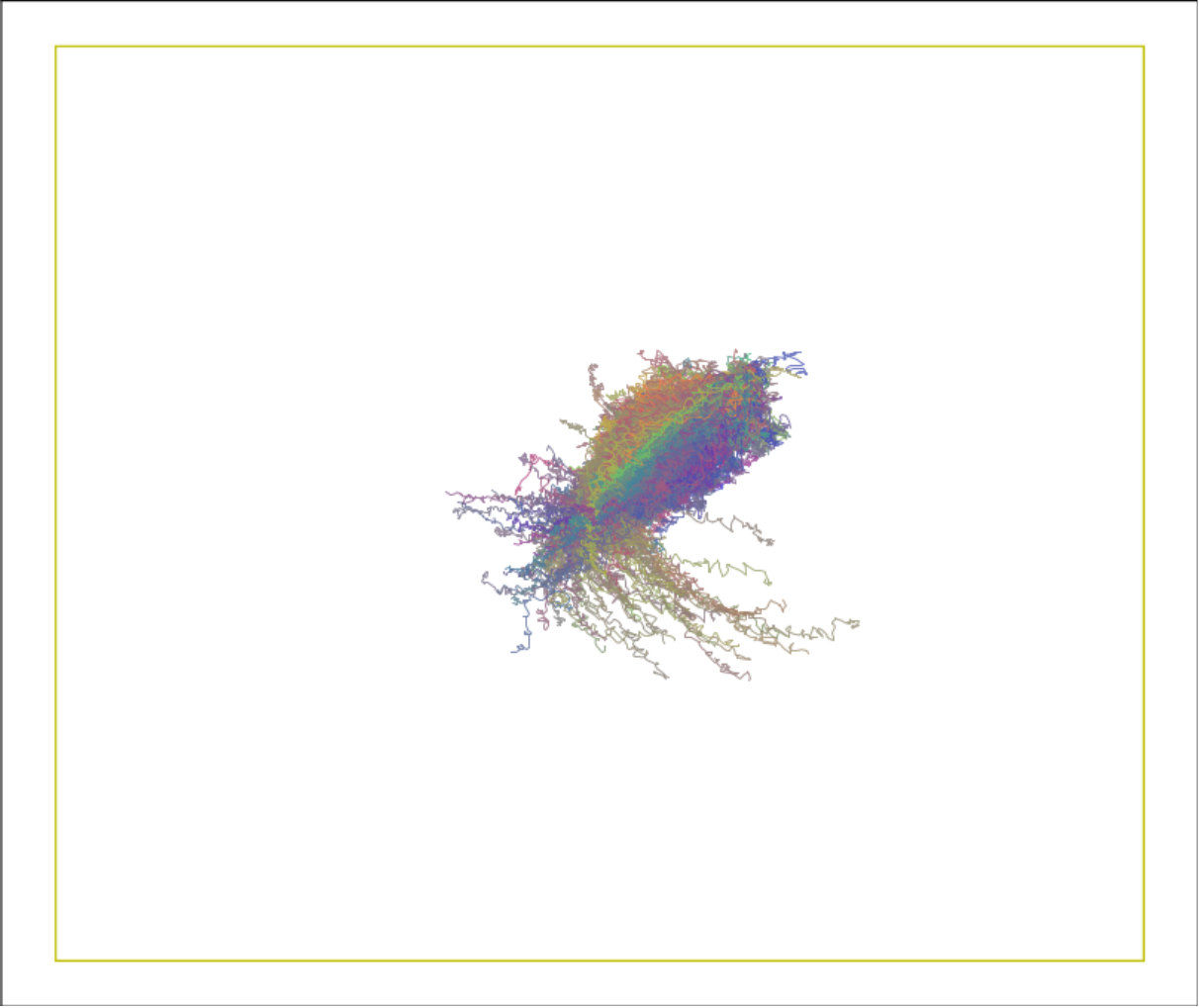} 
        \includegraphics[width=0.13\textwidth]{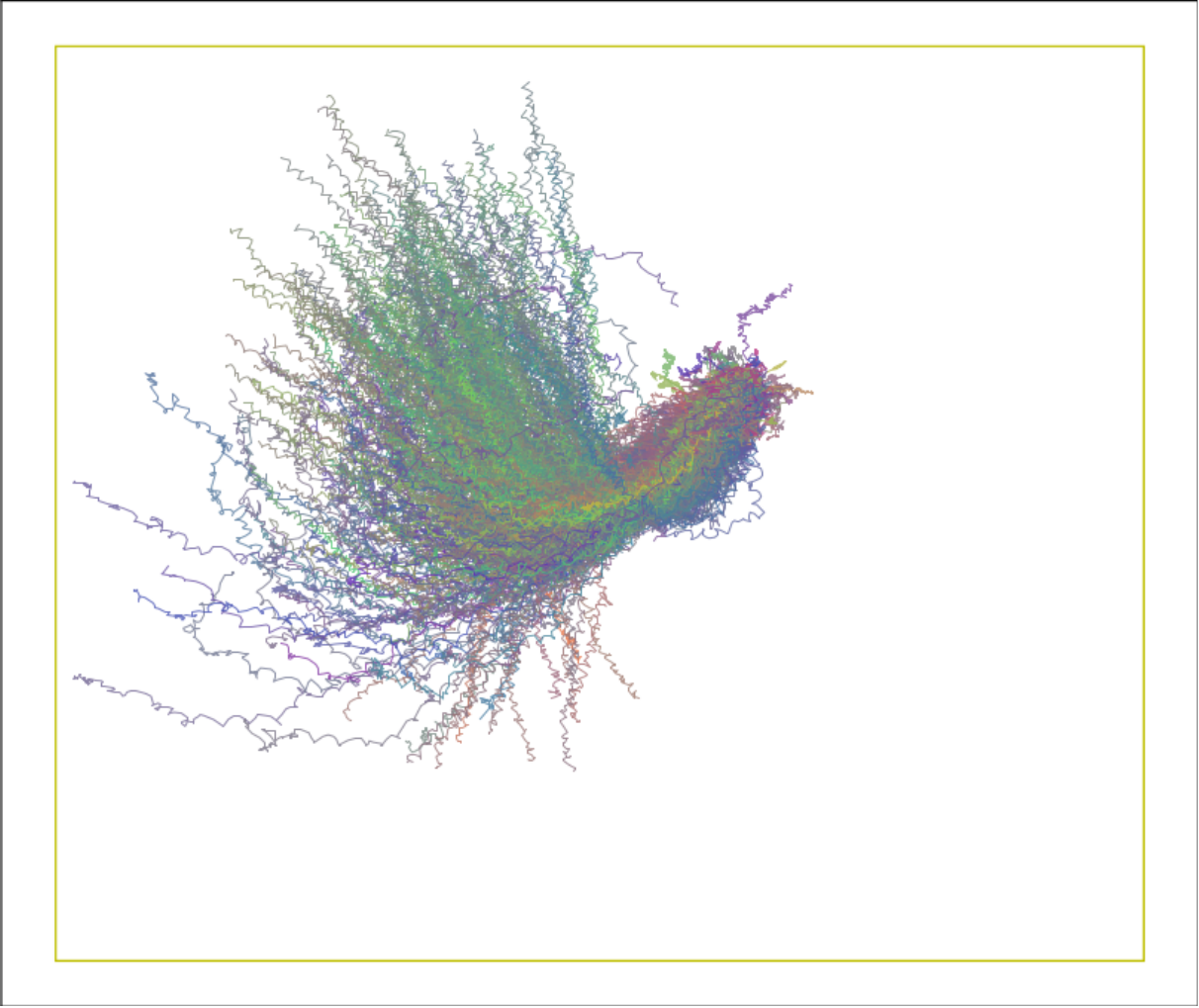}
        \includegraphics[width=0.13\textwidth]{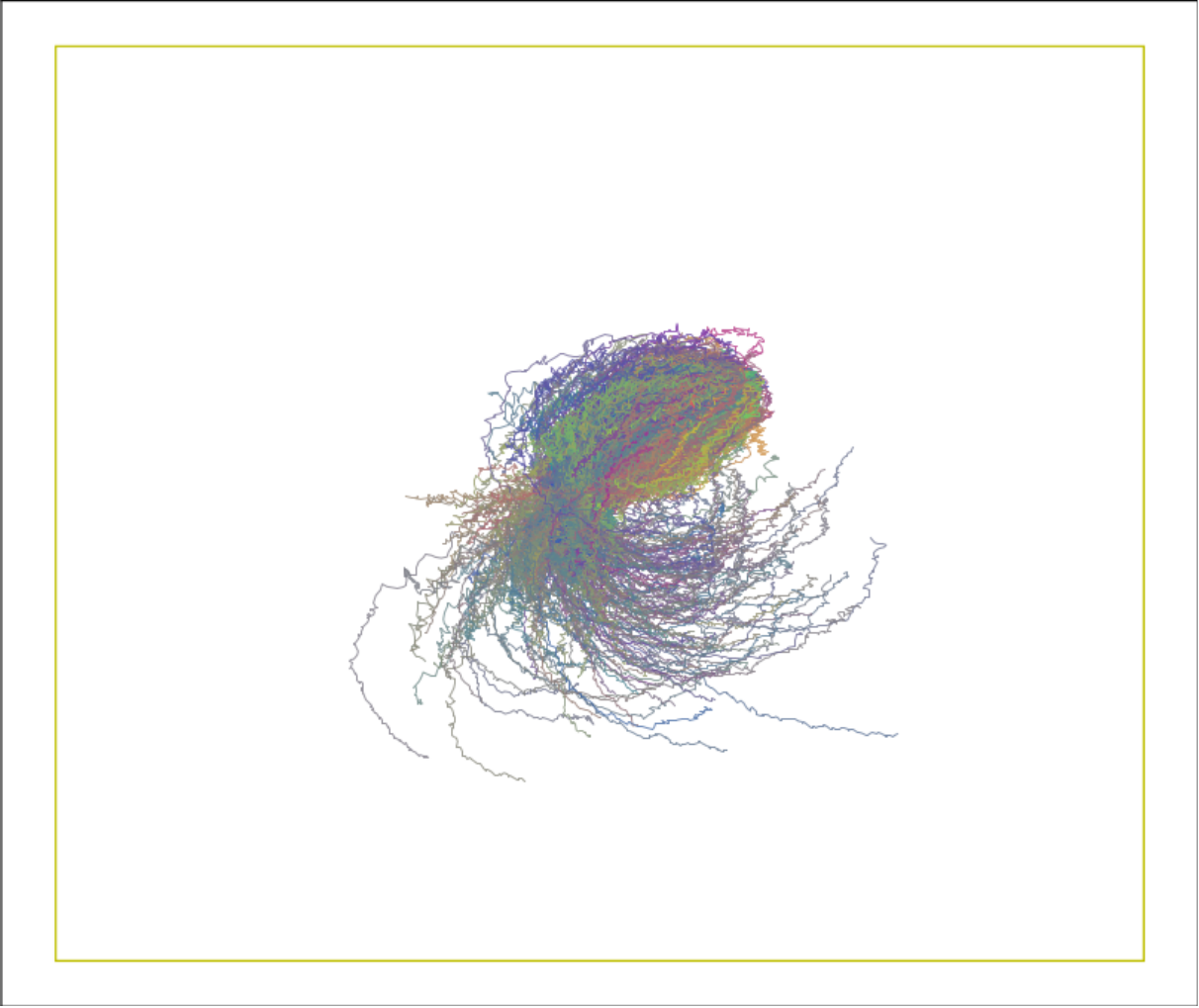}   \\
        \includegraphics[width=0.13\textwidth]{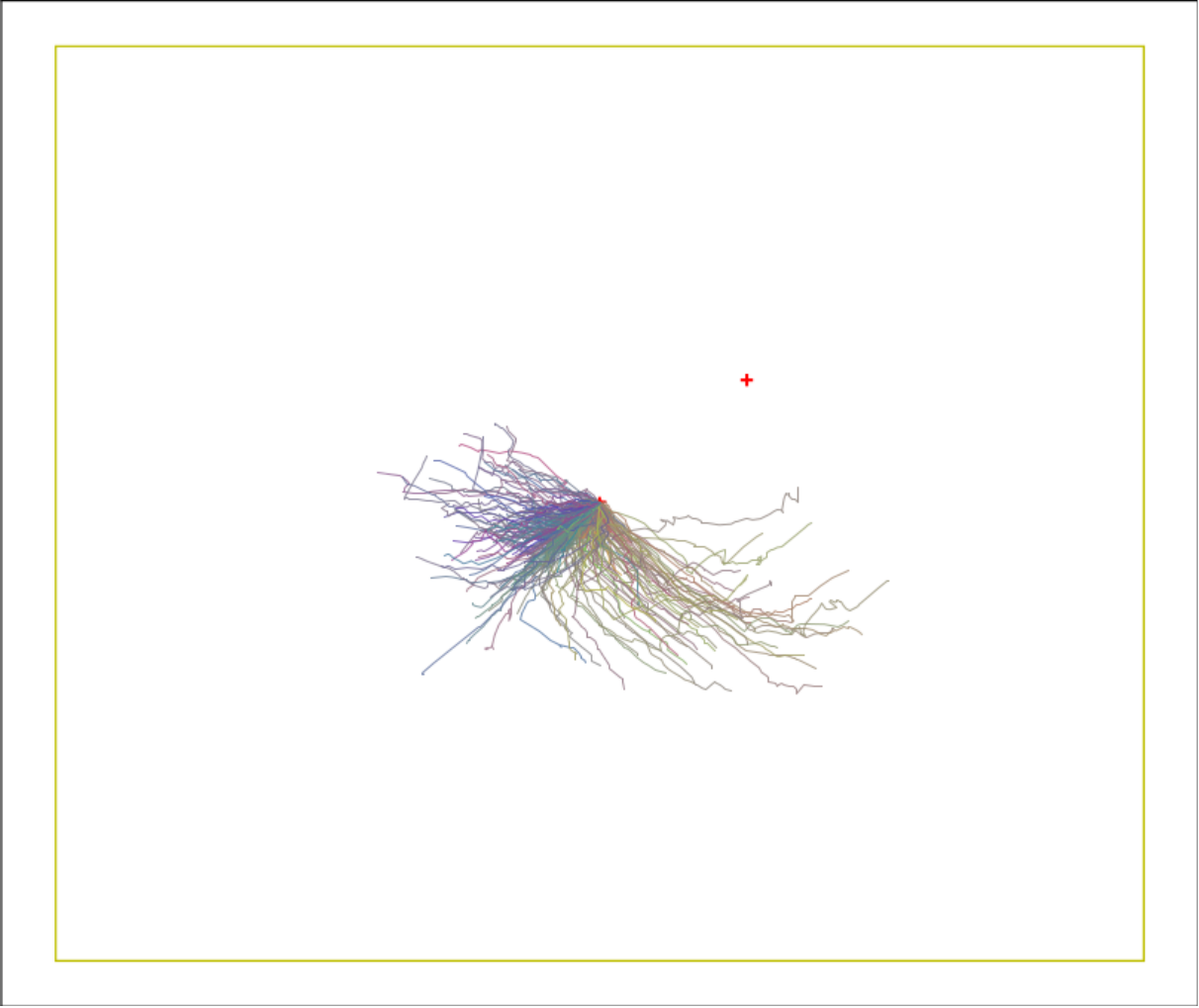} 
        \includegraphics[width=0.13\textwidth]{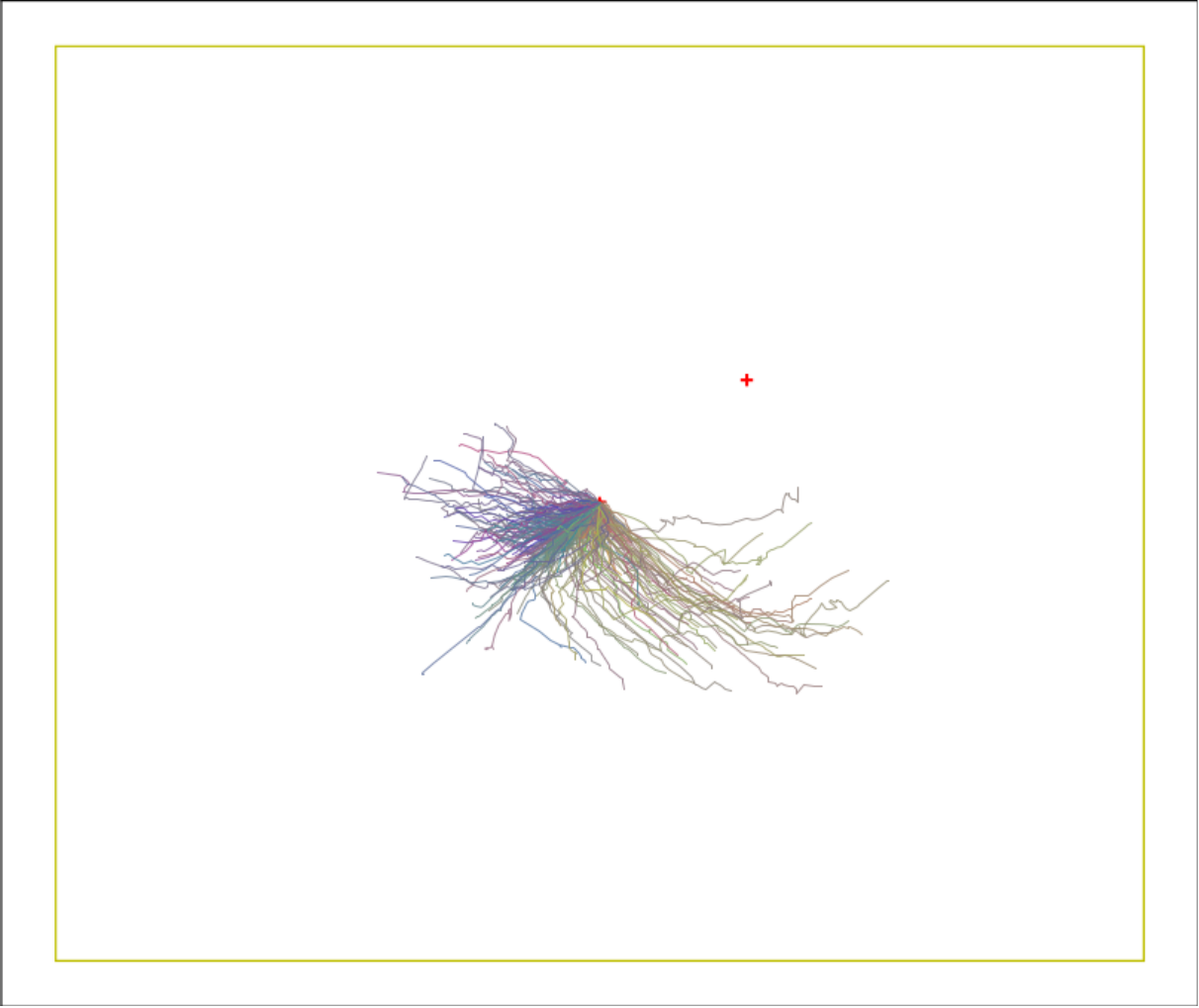} 
        \includegraphics[width=0.13\textwidth]{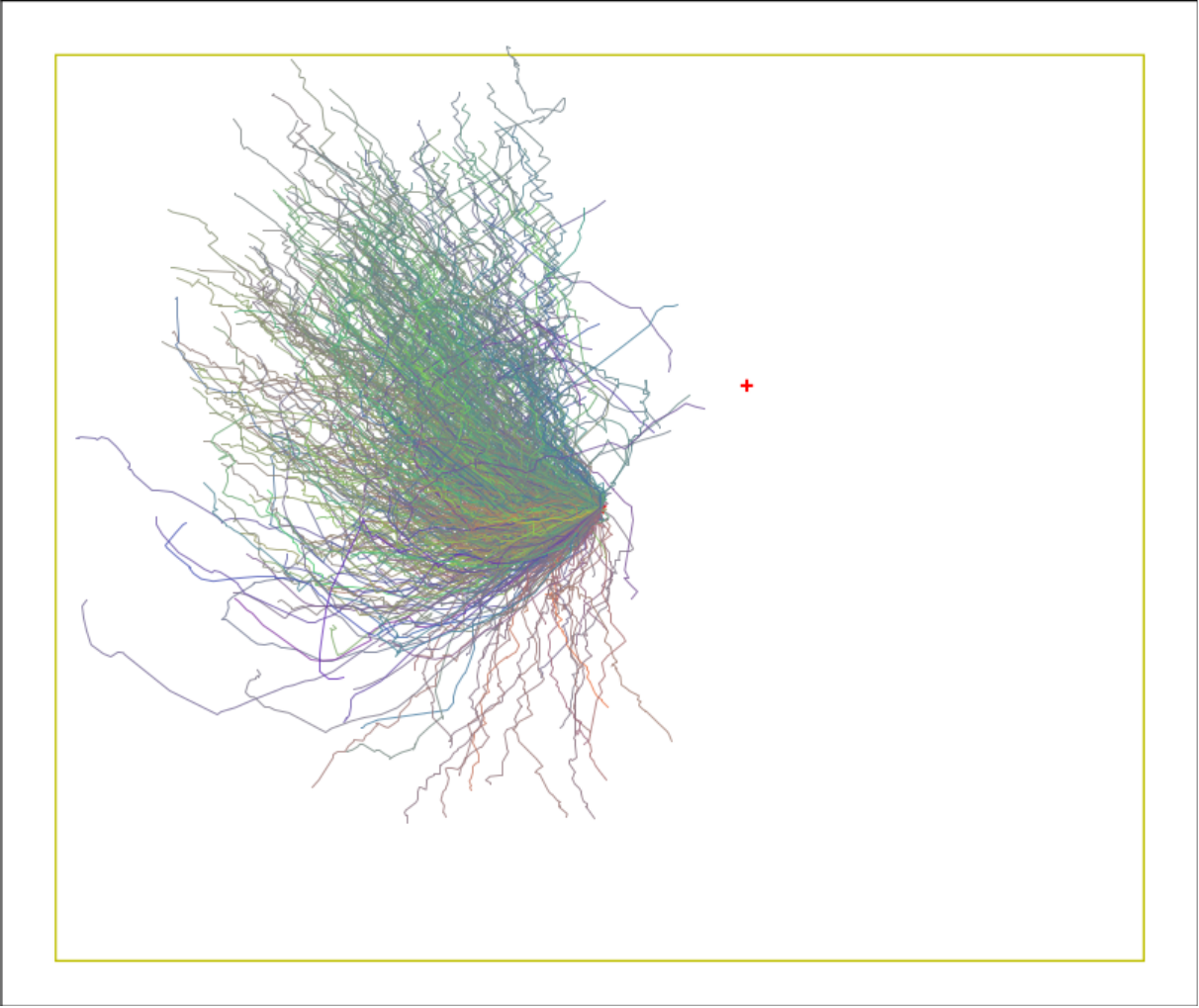}  
        \includegraphics[width=0.13\textwidth]{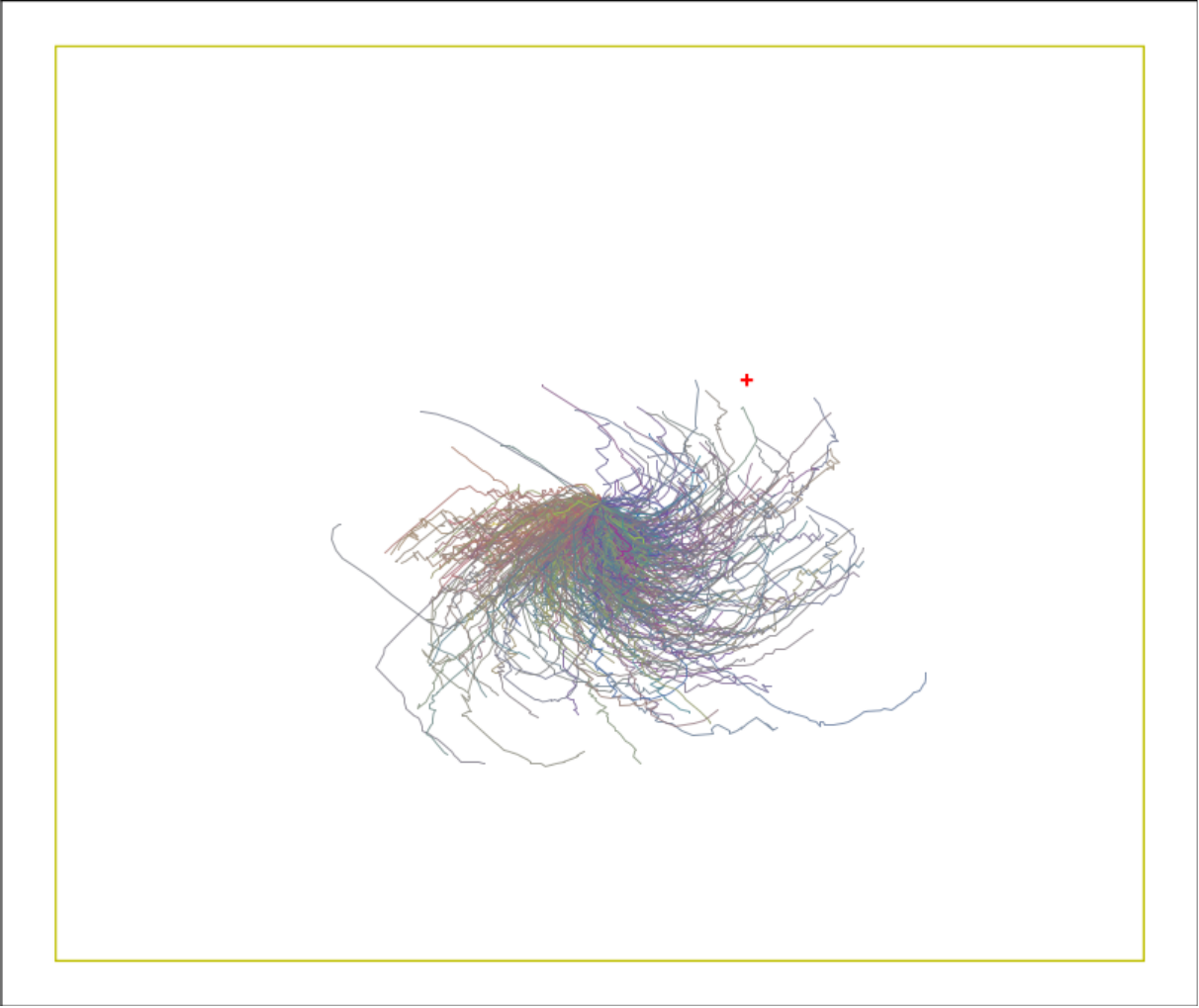}  
        \caption{Doggo}
        \label{fig: Apdx|hr3|qr|doggo}
    \end{subfigure}
    \begin{subfigure}[b]{1\textwidth}
        \centering
        \includegraphics[width=0.13\textwidth]{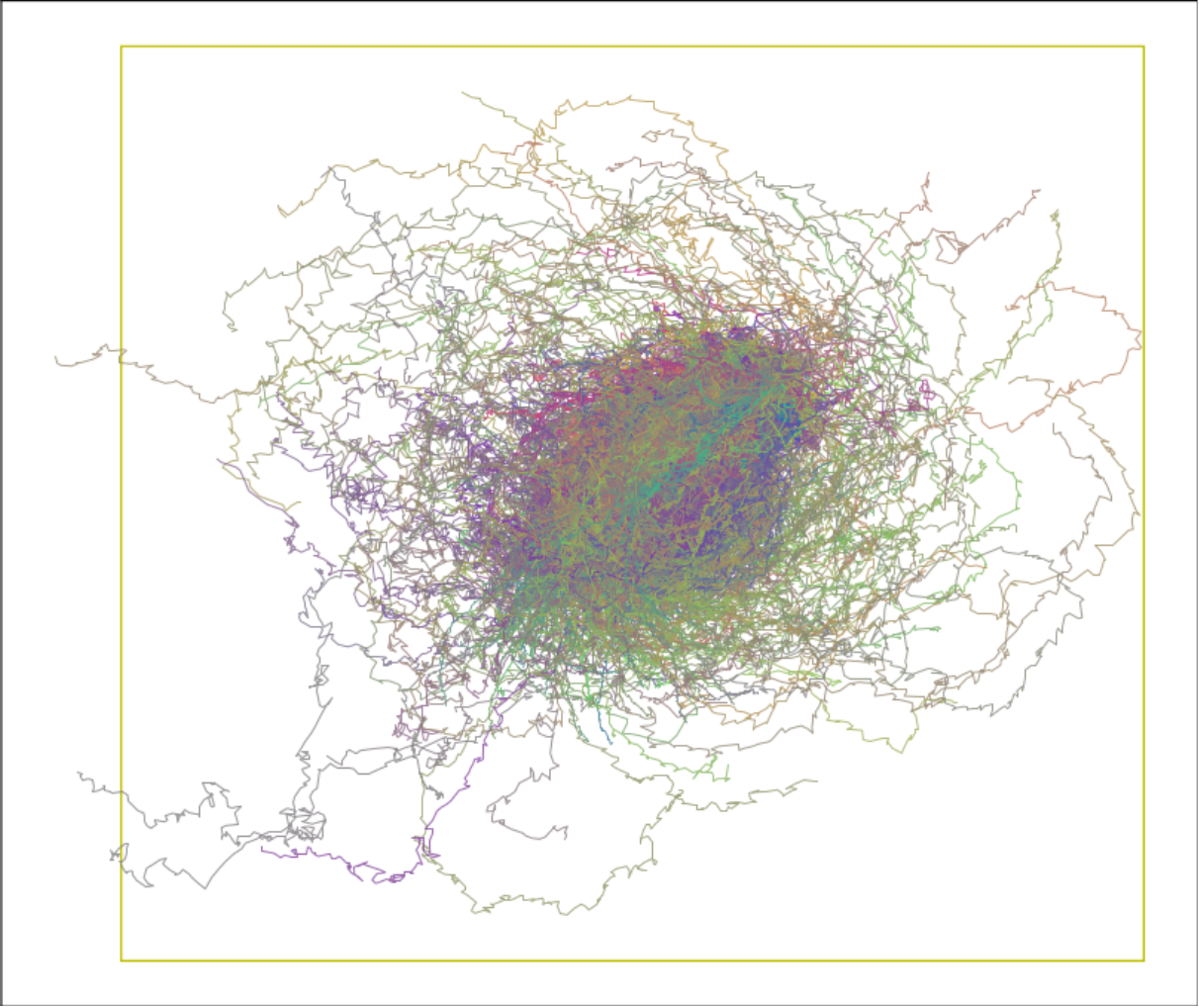} 
        \includegraphics[width=0.13\textwidth]{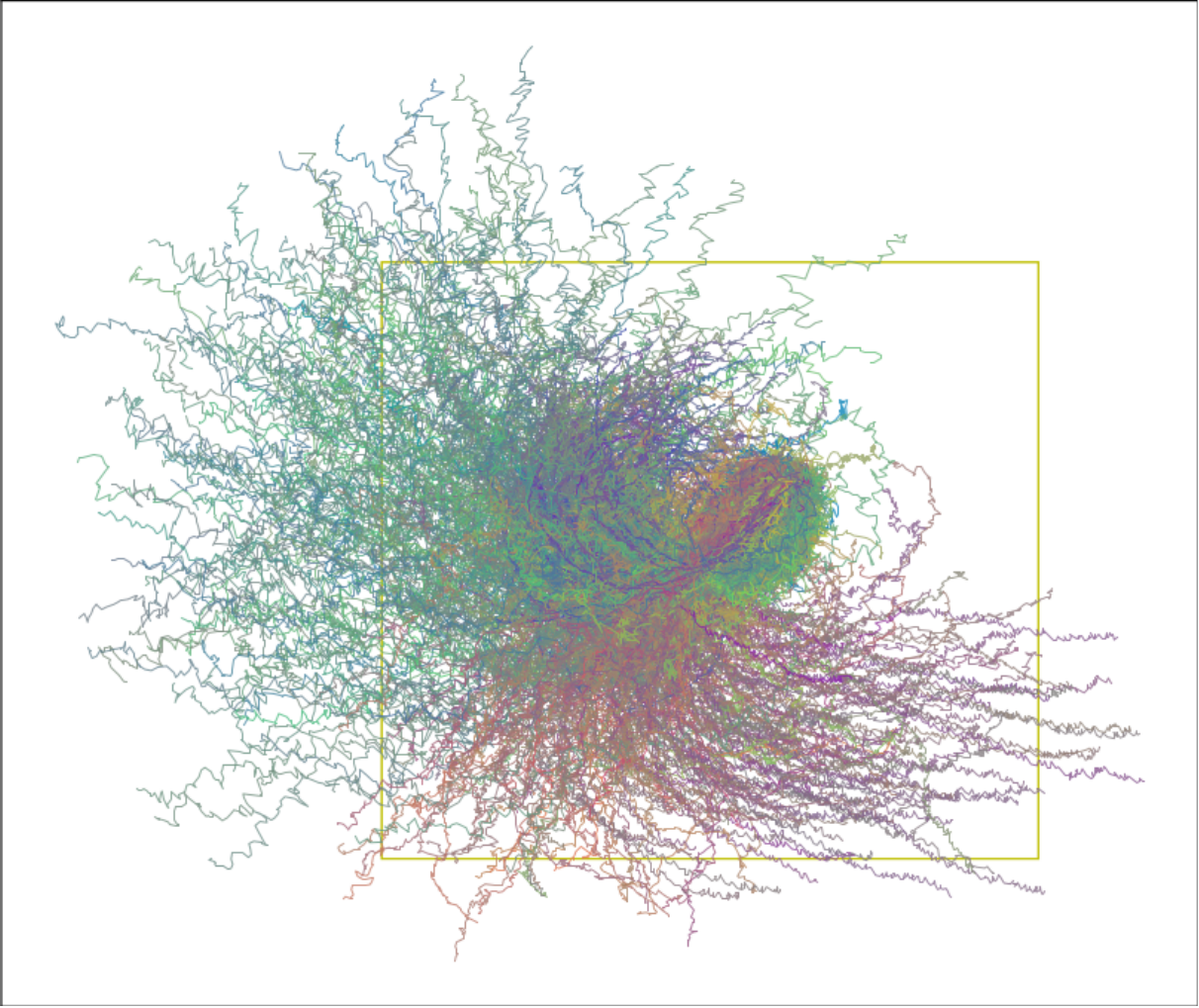}  
        \includegraphics[width=0.13\textwidth]{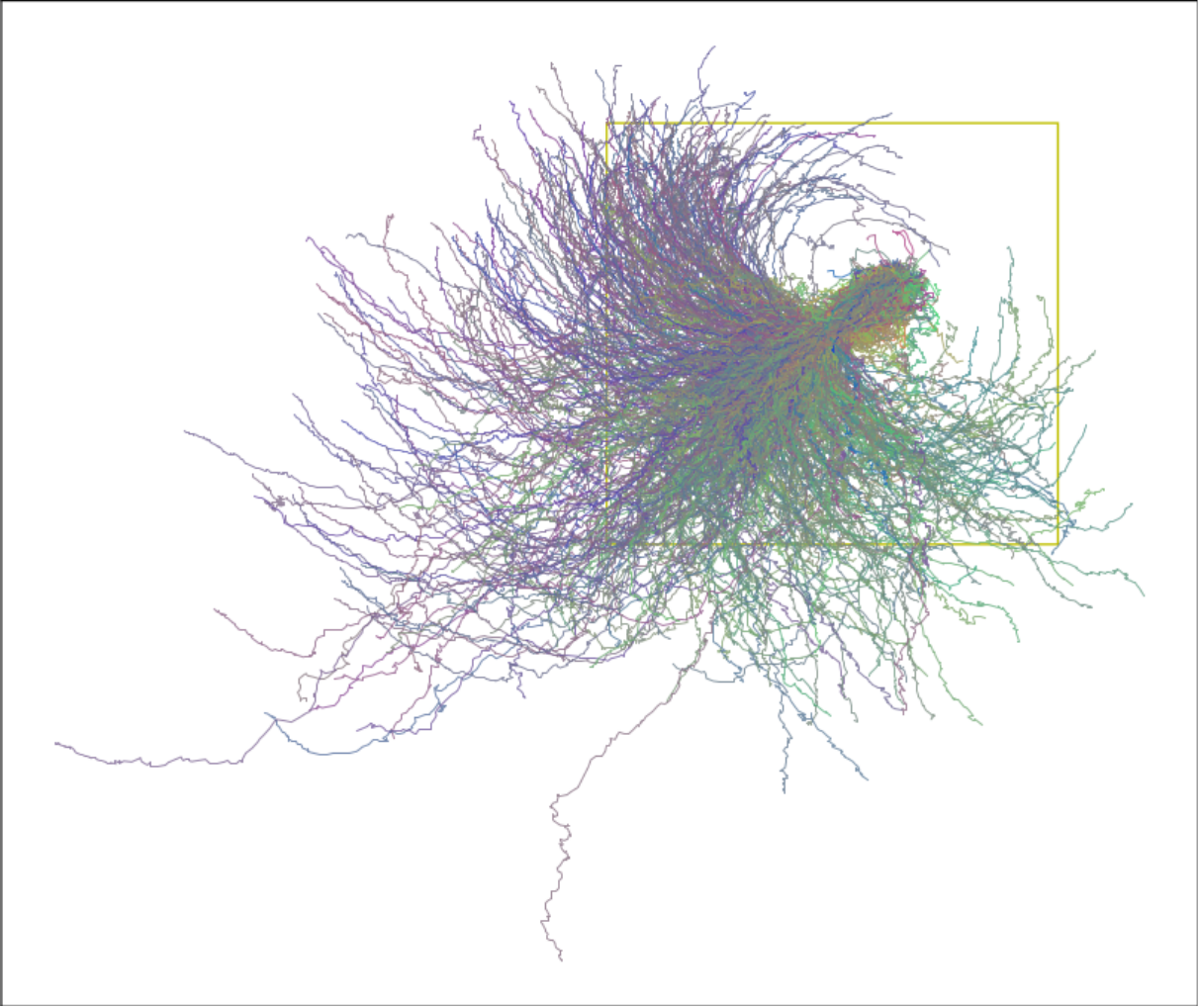} 
        \includegraphics[width=0.13\textwidth]{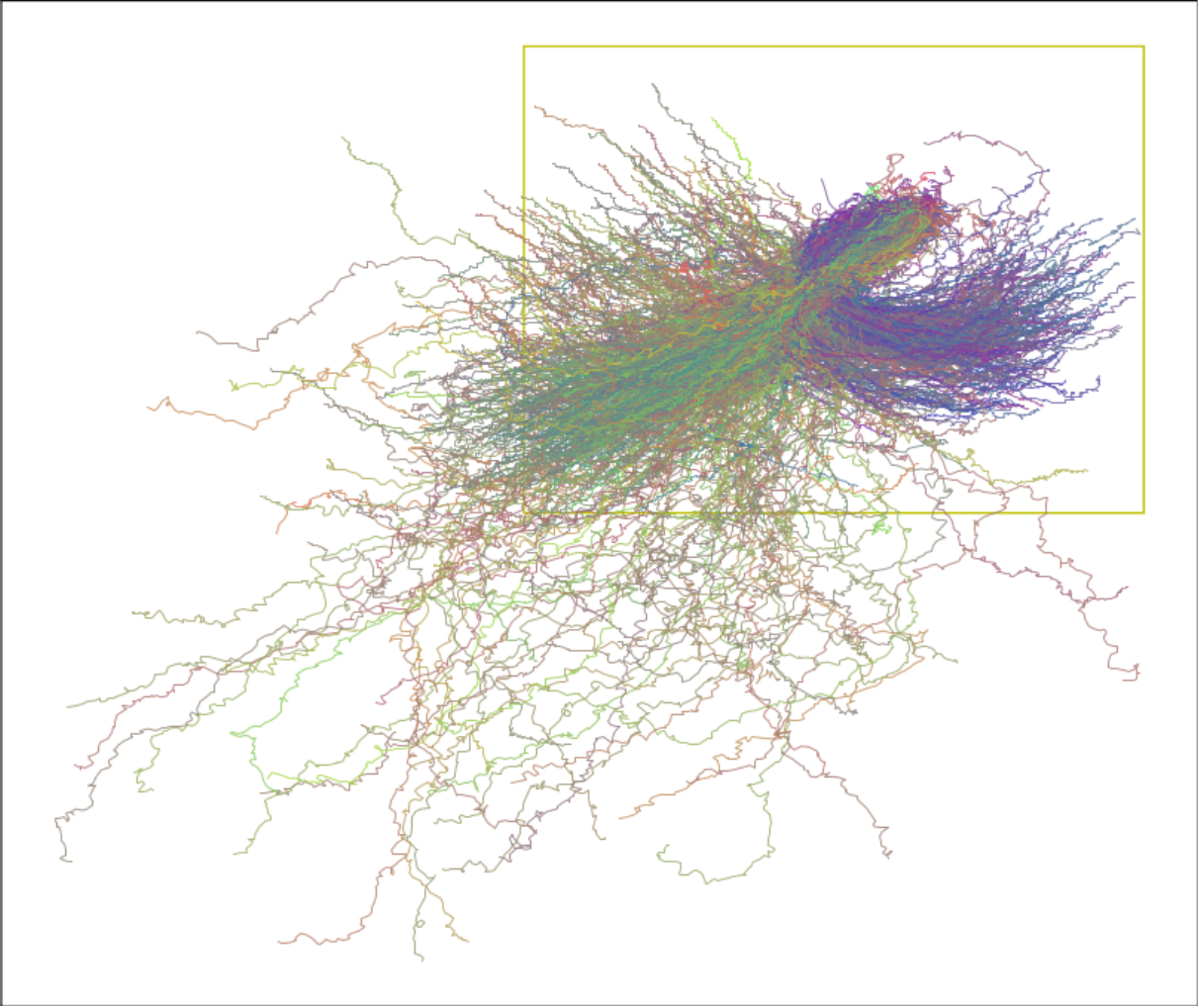} \\
        \includegraphics[width=0.13\textwidth]{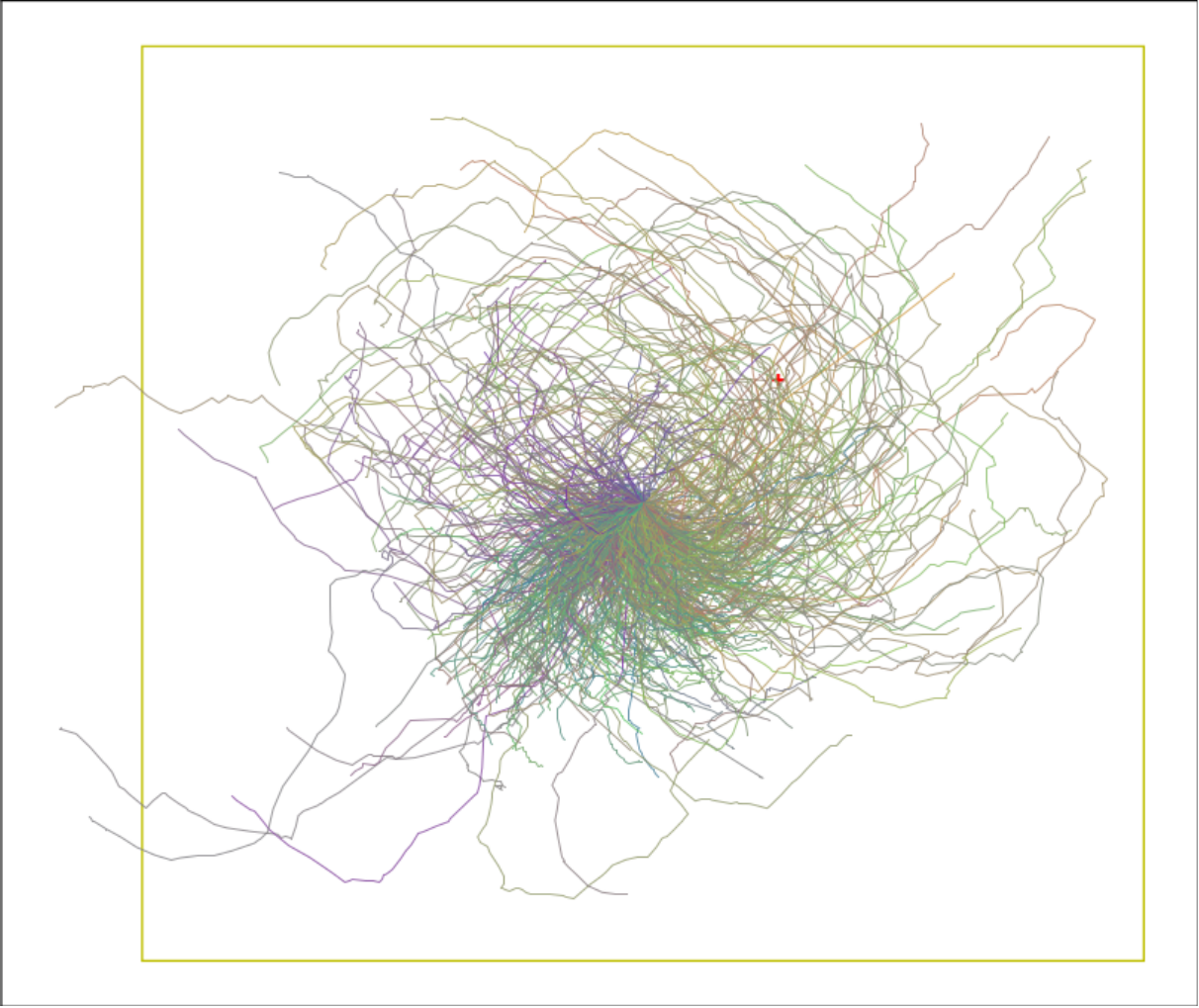} 
        \includegraphics[width=0.13\textwidth]{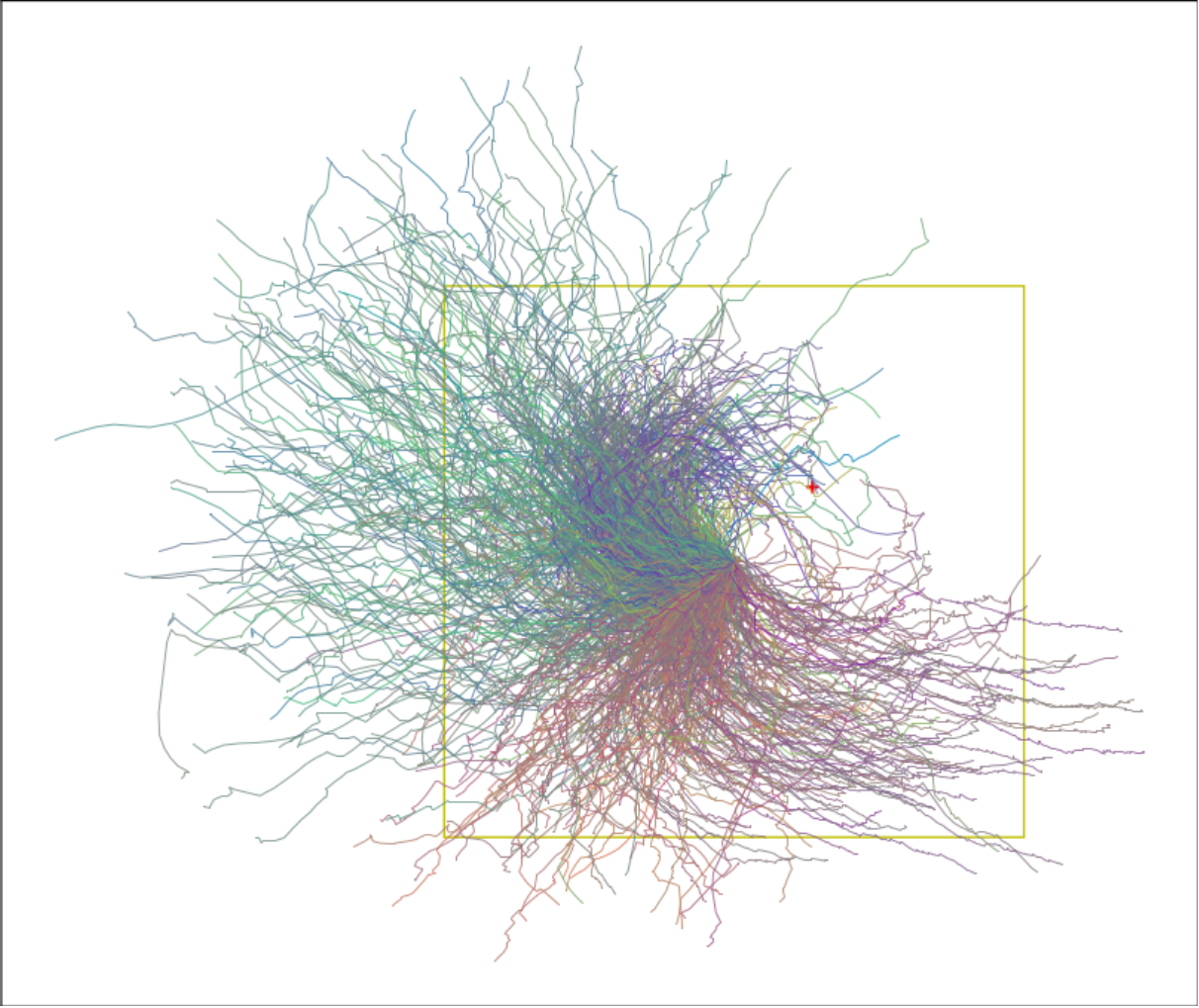}  
        \includegraphics[width=0.13\textwidth]{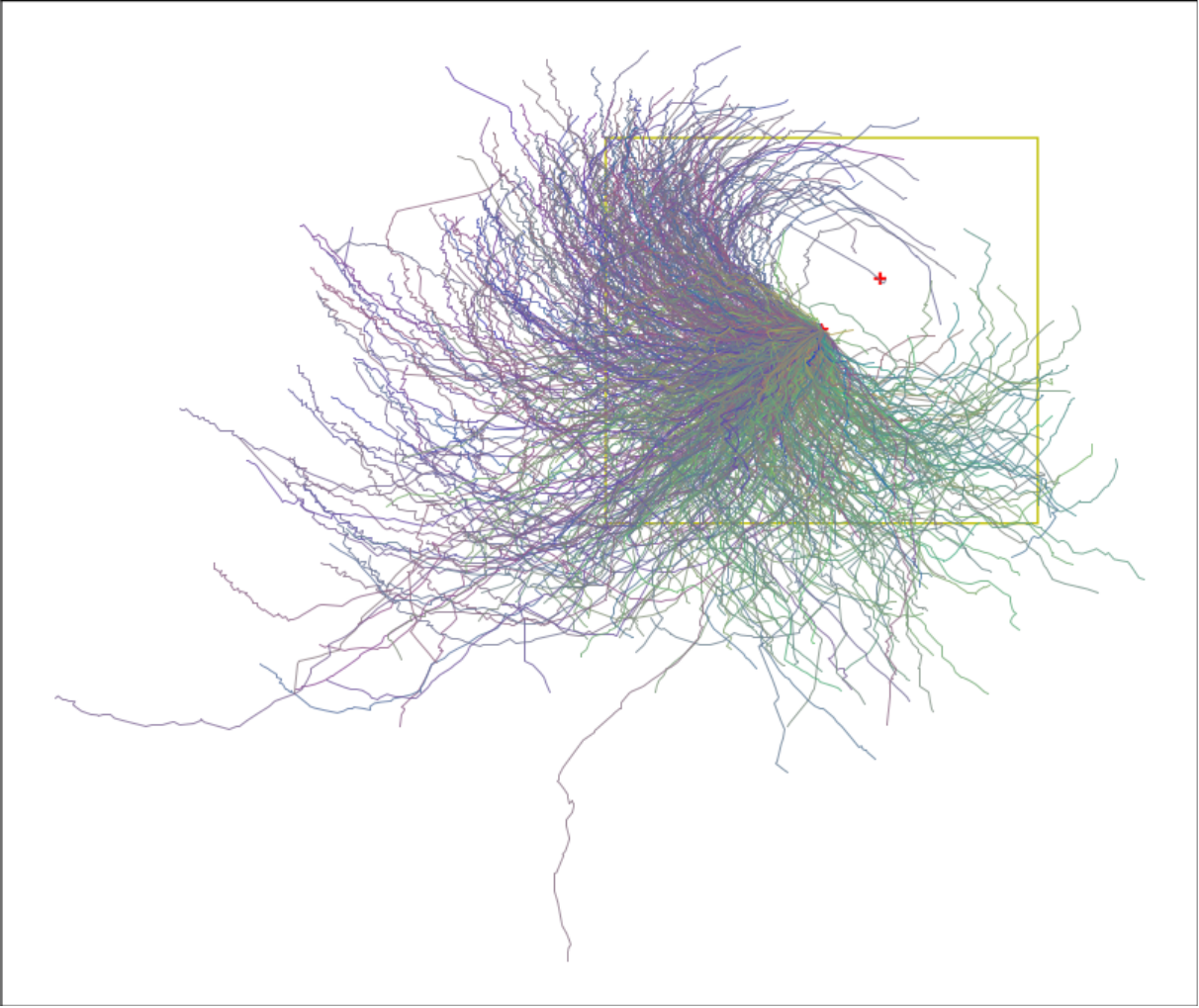} 
        \includegraphics[width=0.13\textwidth]{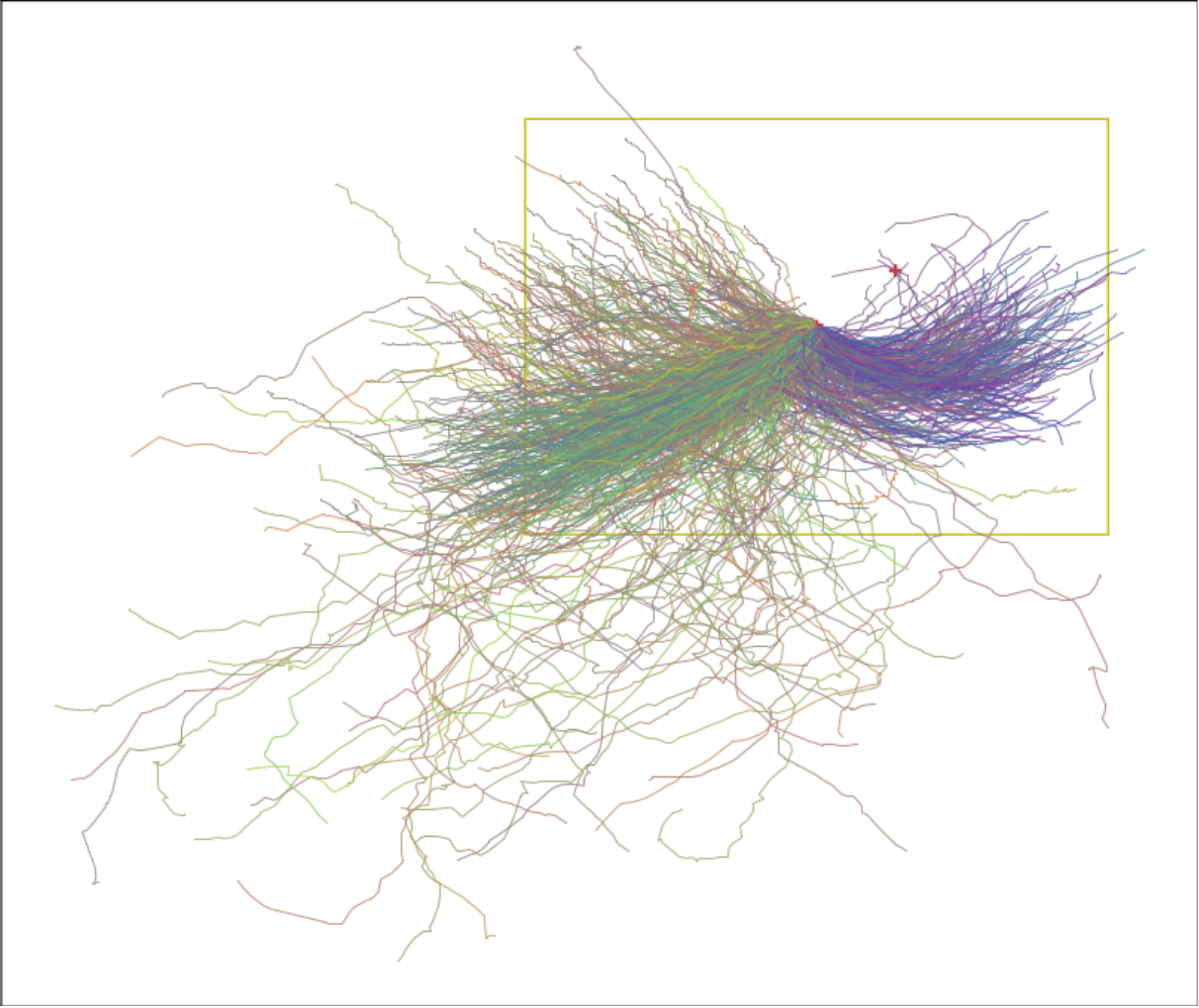} 
        \caption{Ant}
        \label{fig: Apdx|hr3|qr|ant}
    \end{subfigure}
    \caption{Qualitative results of HaSD (4seeds) in the Push Room environment with each agents. After sampling 1000 skills, the first row shows the agent's trajectory, while the second row shows the trajectory of the box.}
    \label{fig: Apdx|hr3|qr|all}
\end{figure}

Figures \ref{fig: Apdx|hr1|qr|all} and Figure \ref{fig: Apdx|hr3|qr|all} show the complete qualitative results of skills discovered by HaSD in the Hazard-Room and Push-Room environments across all agents. We use 2-D skills for all agents and environment. In most environments, HaSD discovers skills that align with human values regardless of the random seeds.

